\documentclass[journal]{IEEEtran}

\IEEEoverridecommandlockouts                              % This command is only needed if
\title{Game-theoretic Modeling of Traffic in Unsignalized Intersection Network for Autonomous Vehicle Control Verification and Validation}

 \author{Ran Tian, Nan Li, Ilya Kolmanovsky, Yildiray Yildiz, and Anouck Girard
 \thanks{This research has been supported by the National Science Foundation award CNS 1544844.}% <-this % stops a space
  \thanks{
         Ran Tian, Nan Li, Ilya Kolmanovsky, and Anouck Girard are with the Department of Aerospace Engineering, University of Michigan, Ann Arbor, MI 48109, USA.
          {\tt\small \{tianran, nanli, ilya, anouck\}@umich.edu}.
         Yildiray Yildiz is with the Department of Mechanical Engineering, Bilkent
        University, Ankara 06800, Turkey. {\tt\small yyildiz@bilkent.edu.tr}.
     }
  }
\usepackage{tabularx}
\usepackage{float}
\usepackage[caption=false]{subfig}
\usepackage{epsfig}
\usepackage{epstopdf}
\usepackage{bbm}
\usepackage{epsfig} % for postscript graphics files
\usepackage{amsmath} % assumes amsmath package installed
\usepackage{amssymb}  % assumes amsmath package installed
\usepackage{bm}
\usepackage{epstopdf}
\usepackage{color}
\usepackage{etoolbox}
\DeclareMathOperator*{\argmax}{arg\,max}
\DeclareMathOperator*{\argmin}{arg\,min}
\usepackage{comment}
\usepackage{diagbox}
\usepackage[ruled, lined, longend, linesnumbered]{algorithm2e}
\usepackage{booktabs,multirow}
\usepackage{bigstrut}
\usepackage{cite}
\newcommand{\di}[1]{{\color{black}#1}}

\begin{document}
\maketitle

% As a general rule, do not put math, special symbols or citations
% in the abstract or keywords.
\begin{abstract}
For a foreseeable future, autonomous vehicles (AVs) will operate in traffic together with human-driven vehicles. Their planning and control systems need extensive testing, including early-stage testing in simulations where the interactions among autonomous/human-driven vehicles are represented. Motivated by the need for such simulation tools, we propose a game-theoretic approach to modeling vehicle interactions, in particular, for urban traffic environments with unsignalized intersections. We develop traffic models with heterogeneous (in terms of their driving styles) and interactive vehicles based on our proposed approach, and use them for virtual testing, evaluation, and calibration of AV control systems. For illustration, we consider two AV control approaches, analyze their characteristics and performance based on the simulation results with our developed traffic models, and optimize the parameters of one of them.
\end{abstract}

% Note that keywords are not normally used for peerreview papers.
%\begin{IEEEkeywords}

%\end{IEEEkeywords}

% For peer review papers, you can put extra information on the cover
% page as needed:
% \ifCLASSOPTIONpeerreview
% \begin{center} \bfseries EDICS Category: 3-BBND \end{center}
% \fi
%
% For peerreview papers, this IEEEtran command inserts a page break and
% creates the second title. It will be ignored for other modes.
\IEEEpeerreviewmaketitle

\section{Introduction}

Autonomous driving technologies have greatly advanced in recent years with the promise of providing safer, more efficient, environment-friendly, and easily accessible transportation \cite{FAGNANT2015167, autonomous_safety, MEYER201780}. To fulfill such a commitment requires developing advanced planning and control algorithms to navigate autonomous vehicles (AVs), as well as comprehensive testing procedures to verify their safety and performance characteristics \cite{KALRA2016182,zhou2017reduced,waschl2019control}. It is estimated based on the collision fatalities rate that to confidently verify an AV control system, hundreds of millions of miles need to be driven \cite{KALRA2016182}, which can be highly time and resource consuming if these driving tests are all conducted in the physical world. Therefore, \di{an alternative solution is to use simulation tools to conduct early-stage testing and evaluation in a virtual world so that the overall verification process can be accelerated \cite{zhao2016accelerated,waschl2019virtual}}. The work of this paper is motivated by the need for virtual testing of AV control systems.

In the near to medium term, AVs are expected to operate in traffic together with human-driven vehicles. Therefore, accounting for the interactions among autonomous/human-driven vehicles is important to achieve safe and efficient driving behavior of an AV.

Control strategies for AVs that account for vehicle interactions include the ones based on Markov decision processes \cite{8248668, bandyopadhyay2013intention, 8569729, li2019stochastic}, model predictive control \cite{schwarting2017safe, 7990647}, game-theoretic models \cite{bahram2015game, sadigh2016planning, fisac2018hierarchical, yu2018human, dreves2018generalized,yu2018human}, as well as data-driven approaches \cite{vallon2017machine,xu2017end}. To evaluate the effectiveness of these algorithms requires simulation environments that can represent the interactions among autonomous/human-driven vehicles.

In our previous work \cite{li7993050}, we exploited a game-theoretic approach to modeling vehicle interactions in highway traffic. Compared to highway traffic, urban traffic environments with intersections are considered to be more challenging for both human drivers and AVs, as they involve more extensive and complex interactions among vehicles. For instance, almost $40\%$ of traffic accidents in the U.S. are intersection-related \cite{safety_intersection}.

In this paper, we extend the game-theoretic approach of \cite{li7993050} to modeling vehicle interactions in urban traffic. In particular, we consider urban traffic environments with unsignalized intersections. Firstly, unsignalized intersections may be even more challenging than signalized intersections because, due to the lack of guidance from traffic signals, a driver/automation needs to decide on its own, whether, when and how to enter and drive through the intersection. According to the U.S. Federal Highway Administration's report, almost $70 \%$ of fatalities due to intersection-related traffic accidents happened at unsignalized intersections \cite{Unsignalized_Intersection}. Thus, well-verified autonomous driving systems for unsignalized intersections may deliver significant safety benefits. Indeed, many previous works in the literature on AV control for intersections, including \cite{dreves2018generalized,7995818,8461233,pruekprasert2019game,pruekprasert2019decision}, deal with unsignalized intersections, although they do not always explicitly point this out.

Our approach formulates the decision-making processes of drivers/vehicles as a dynamic game, where each vehicle interacts with other vehicles by observing their states, predicting their future actions, and then planning its own actions. In addition to the difference in traffic scenarios being considered (i.e., urban traffic in this paper versus highway traffic in \cite{li7993050}), this paper contains the following methodological contribution compared to \cite{li7993050}: Due to the much larger state space for urban traffic environments with intersections compared to that for highway traffic, the reinforcement learning approach used in \cite{li7993050} to solve for control policies is computationally prohibitive. Therefore, we develop in this paper an alternative approach that uniquely integrates a game-theoretic formalism, receding-horizon optimization, and an imitation learning algorithm to obtain control policies. This new approach is shown to be computationally effective for the large state space of urban traffic.

\di{Our model representing vehicle interactions falls in-between macroscopic traffic models and microscopic driver behavior models. On the one hand, macroscopic traffic models typically assume a large number (e.g., hundreds to thousands) of vehicles and study the average or statistical properties of traffic flow, such as traffic flux (vehicles/hour) versus traffic density (vehicles/km) \cite{lighthill1955kinematic, may1990traffic, daganzo1995cell}. Individual vehicle behavior is usually not represented in such models. On the other hand, microscopic driver behavior models typically focus on modeling the decision-making as well as control processes of individual drivers \cite{gipps1981behavioural, brackstone1999car, salvucci2006modeling}, such as the responses of a human driver to various traffic situations. The interactions among multiple vehicles are usually not incorporated in such models. Consequently, neither models are particularly suitable for virtual testing of AV control systems. In contrast, our game-theoretic model represents the interactive decision-making processes of multiple drivers/vehicles, where each individual vehicle's behavior is represented using a kinematic vehicle model. This way, we can model traffic scenes with a medium number (e.g., dozens) of interacting vehicles, suitable as test scenarios for AV control systems.}

In \cite{li_leader_follower}, we modeled the interactions among vehicles at unsignalized intersections, but using a different game-theoretic approach from the one used in this paper: In \cite{li_leader_follower}, we model vehicle interactions based on a formulation of a leader-follower game; while in this paper, we consider the application of level-k game theory \cite{stahl1995players,costa2006cognition}. The control strategies of all interacting vehicles modeled using the framework of \cite{li_leader_follower} are homogeneous; while the control strategies of different vehicles modeled using the scheme of this paper are heterogeneous, differentiated by their level-$k$ control policies with different $k = 0,1,2,\dots$ This heterogeneity can be used to represent the different driving styles among different drivers, e.g., aggressive driving versus cautious/conservative driving. In addition, \cite{li_leader_follower} models a single intersection with up to $10$ interacting vehicles; while in this paper, thanks to the effective application of the aforementioned solution approach integrating game theory, receding-horizon optimization, and imitation learning to obtain control policies, the scheme of this paper can be used to model much larger road systems involving many intersections and many vehicles with manageable online computational effort. This enables the investigation into driving characteristics that are exhibited when a vehicle drives through multiple road segments, such as overall travel time, fuel consumption, etc. A road system with $15$ intersections and $30$ vehicles is illustrated as an example in Section~\ref{sec: traffic simulator}. Furthermore, application of the developed traffic models to verification and validation of AV control systems is comprehensively discussed in this paper, but not in \cite{li_leader_follower}.

Preliminary results of this paper have been reported in the conference papers \cite{Nanintersection} and \cite{8619275_tian}. The results modeling the interactions between two vehicles at a four-way intersection are reported in \cite{Nanintersection} and those for two vehicles at a roundabout intersection are in \cite{8619275_tian}. This paper generalizes the methodology to modeling the interactions among multiple (more than two) vehicles and to an additional intersection type -- T-shaped intersection. Constructing larger road systems based on the models of these three intersections is reported for the first time in this paper. This paper also demonstrates how the developed traffic models can be used for virtual testing, evaluation, and calibration of AV control systems, which is not provided in \cite{Nanintersection} and \cite{8619275_tian}.

In summary, the contributions of this paper are: 1) We describe an approach based on level-k game theory to modeling the interactions among vehicles in urban traffic environments with unsignalized intersections. 2) We propose an algorithm based on imitation learning to obtain level-$k$ control policies so that our approach to modeling vehicle interactions is scalable -- able to model traffic scenes with many intersections and many vehicles. \di{In particular, this new imitation learning approach is compared with the supervised learning approach used in our previous work \cite{8619275_tian} and is shown to provide better results}. 3) We demonstrate the use of the developed traffic models for virtual testing, evaluation, and calibration of AV control systems.  For illustration purposes, we consider two AV control approaches, analyze their characteristics and performance based on the simulation results with our traffic models, and optimize the parameters of one of them.

This paper is organized as follows: The models representing vehicle kinematics and driver decision-making processes are introduced in Section~\ref{sec: problem statement}. The game-theoretic model representing vehicle interactions and obtaining its explicit approximation via imitation learning are discussed in Section~\ref{sec:  game theorstic decision-making}. The procedure to construct traffic models of larger road systems based on the models of three basic intersection scenarios is described in Section~\ref{sec: traffic simulator}. We then propose two AV control approaches in Section~\ref{sec: VV}, used as case studies to illustrate the application of our developed traffic models to AV control verification and validation. Simulation results are reported in Section~\ref{sec: results}, and finally, the paper is concluded in Section~\ref{sec: conclusion}.

\section{Traffic Dynamics and Driver Decision-making Modeling}
\label{sec: problem statement}

In this section, we describe our models to represent the traffic dynamics and the decision-making processes of interacting drivers.

\subsection{Traffic dynamics}

Firstly, we describe the evolution of a traffic scenario using a discrete-time model as follows:
\begin{equation}
    \mathbf{s}_{t+1} = \mathcal{F}(\mathbf{s}_{t},\mathbf{u}_{t}),
    \label{equ:dynamic model}
\end{equation}
where $\mathbf{s} = (s^1,s^2,\dots,s^m)$ denotes the traffic state, composed of the states $s^i$, $i \in \mathcal{M} =\{1,2,\dots,m\}$, of all interacting vehicles in the scenario, $\mathbf{u} = (u^1,u^2,\dots,u^m)$ denotes the collection of all vehicles' actions $u^i$, and the subscript $t$ represents the discrete-time instant. In particular, the state of a vehicle is composed of two parts, $s^i = (s^{i,1},s^{i,2})$. The first part $s^{i,1} = (x^i, y^i, v^i, \theta^i)$ represents the state of vehicle kinematics, modeled using the following unicycle model:
\begin{align}\label{equ:kinematics}
    \left[\begin{matrix}
    x^i_{t+1} \\
    y^i_{t+1} \\
    v^i_{t+1} \\
    \theta^i_{t+1} \\
    \end{matrix}\right] = f(s^i_t,u_t^i) =\left[\begin{matrix}
    x_t^i + v_t^i\, \cos{\big(\theta_t^i\big)}\, \Delta t \\
    y_t^i + v_t^i\, \sin{\big(\theta_t^i\big)}\, \Delta t \\
    v_t^i + a_t^i\, \Delta t \\
    \theta_t^i + \omega_t^i\, \Delta t \\
    \end{matrix}\right],
\end{align}
where $(x^i,y^i)$, $v^i$, and $\theta^i$ represent, respectively, the vehicle's position in the ground-fixed frame, its speed, and its heading angle, the inputs $a^i$ and $\omega^i$ represent, respectively, the vehicle's acceleration and heading angle rate, while $\Delta t$ is the sampling period for decision-making. The second part $s^{i,2} = (r^i, \xi^i)$ contains additional information related to the vehicle's decision-making objective, including $r^i = (r_x^i, r_y^i)$, representing a target/reference position to go, and $\xi^i$, a feature vector containing key information about the road layout and geometry such as the road width, the angle of intersection, and etc \cite{li_leader_follower}. When vehicle $i$ is driving toward, in the middle of, or exiting a specific intersection, $s^{i,2}$ stays constant with $r^i$ being a point located in the center of the vehicle's target lane; $s^{i,2}$ gets updated after the vehicle has returned to straight road and is driving toward the next intersection.

\di{We remark that the unicycle model \eqref{equ:kinematics} is suitable for our purpose of modeling the interactive decision-making processes and the resulting dynamic behavior of multiple vehicles in intersection traffic scenarios. This model is simple while it can sufficiently accurately represent vehicle kinematics at low to medium vehicle speeds and involving turning behavior \cite{sadigh2016planning,Karimi2020}. In Section~\ref{sec: results} we show that 1) driving behaviors planned based on the model \eqref{equ:kinematics} can be accurately executed by vehicle systems with lower-level control, and 2) vehicle trajectories extracted from real-world traffic data can be satisfactorily reproduced by simulating the model \eqref{equ:kinematics} along with the decision logic developed following our approach.}

\subsection{Driver decision-making}

An action $u^i$ is a pair of values of the inputs $(a^i,\omega^i)$, i.e., $u^i = (a^i,\omega^i)$. We assume that the drivers of the vehicles make sequential decisions based on receding-horizon optimization as follows: At each discrete-time instant $t$, the driver of vehicle $i$ solves for
\begin{align}
   (\mathbf{u}^i_t)^* &= \big\{ (u_{0|t}^i)^*,(u_{1|t}^i)^*,\dots,(u_{N-1|t}^i)^*\big\} \label{equ:general_rhc} \\
   & \in \argmax_{\mathbf{u}^i_t \in \mathcal{U}^N }\sum_{\tau=0}^{N-1}\lambda^\tau R(s_{\tau|t}^i, \mathbf{s}^{-i}_{\tau|t},u_{\tau|t}^i,\mathbf{u}_{\tau|t}^{-i}),
    \nonumber
\end{align}
where $\mathbf{u}^i_t = \big\{u_{0|t}^i,u_{1|t}^i,\dots,u_{N-1|t}^i\big\}$ represents a sequence of predicted actions of vehicle $i$, with $u_{\tau|t}^i$ denoting the predicted action for time step $t+\tau$ and taking values in a finite action set $\mathcal{U}$; the notations $s_{\tau|t}^i$, $\mathbf{s}^{-i}_{\tau|t}$ and $\mathbf{u}_{\tau|t}^{-i}$ represent, respectively, the predicted state of vehicle $i$, and the collections of predicted states and actions of the other vehicles $j \in \mathcal{M}$, $j \neq i$, i.e., $\mathbf{s}^{-i}_{\tau|t} = (s_{\tau|t}^j)_{j \in \mathcal{M}, j \neq i}$ and $\mathbf{u}^{-i}_{\tau|t} = (u_{\tau|t}^j)_{j \in \mathcal{M}, j \neq i}$; $R$ is a reward function depending on the states and actions of all interacting vehicles, which will be introduced in detail in the following section; and $\lambda \in (0,1]$ is a factor discounting future reward.

Once an optimal action sequence $(\mathbf{u}^i_t)^*$ is determined, vehicle $i$ applies the first element $(u_{0|t}^i)^*$ for one time step, i.e., $u^i_t = (u_{0|t}^i)^*$. After the states of all vehicles have been updated, vehicle $i$ repeats this procedure at $t+1$.

The fact that $R$ depends not only on the ego vehicle's state and action but also on those of the other vehicles determines the interactive nature of the drivers' decision-making processes in a multi-vehicle traffic scenario. Note that, due to the unknowns $\mathbf{u}_{\tau|t}^{-i}$ and $\mathbf{s}^{-i}_{\tau|t}$ for $\tau = 0, 1,\dots,N-1$, the problem \eqref{equ:general_rhc} has not been well-defined yet and cannot be solved. To be able to solve for $(\mathbf{u}^i_t)^*$, we will exploit a game-theoretic approach in Section~\ref{sec: game theorstic decision-making} to predict the values of $\mathbf{u}_{\tau|t}^{-i}$ and $\mathbf{s}^{-i}_{\tau|t}$.

\subsection{Reward function}
\label{sec: reward function}

We use the reward function $R$ in \eqref{equ:general_rhc} to represent vehicles' decision-making objectives in traffic. In this paper, we consider $R$ defined as follows:
\begin{equation}
\label{equ:reward}
    R(s^i_{\tau|t}, \mathbf{s}^{-i}_{\tau|t},u_{\tau|t}^i,\mathbf{u}^{-i}_{\tau|t}) = \mathbf{w}^\intercal \mathbf{\Phi}\big(s^i_{\tau+1|t}, (s_{\tau+1|t}^j)_{j \in \mathcal{M}, j \neq i}\big),
\end{equation}
where $\mathbf{\Phi} = [\phi_1,\phi_2,\dots,\phi_6]^\intercal$
is the feature vector and $\mathbf{w} \in \mathbb{R}_+^6$ is the weight vector. Note that $s^j_{\tau+1|t} = f(s^j_{\tau|t},u^j_{\tau|t})$ for all $j \in \mathcal{M}$ based on the kinematic vehicle model \eqref{equ:kinematics}.

The features $\phi_1,\phi_2,\dots,\phi_6$ are designed to encode common considerations in driving, such as safety, comfort, travel time, etc. They are defined as follows.

The feature $\phi_1$ characterizes the collision status of the vehicle. In particular, we over-bound the geometric contour of each vehicle by a rectangle, referred to as the collision-zone ($c$-zone). Then, $\phi_1 = -1$ if vehicle $i$'s $c$-zone at the predicted state $s^i_{\tau+1|t}$ overlaps with any of the other vehicles' $c$-zones at their predicted states $s^j_{\tau+1|t}$, which indicates a danger of collision; and $\phi_1 = 0$ otherwise. \di{The $c$-zone over-bounds the vehicle's geometric contour by a small margin to compensate for the effect of perception errors. Note that the size of these perception errors is small at low to medium vehicle speeds when compared to the resolution of the actions.} 

The feature $\phi_2$ characterizes the on-road status of the vehicle, taking $-1$ if vehicle $i$'s $c$-zone crosses any of the road boundaries, and $0$ otherwise. And similarly, $\phi_3$ characterizes the in-lane status of the vehicle. If vehicle $i$'s $c$-zone crosses a lane marking that separates the traffic of opposite directions or enters a lane different from its target lane when exiting an intersection, then $\phi_3 = -1$; $\phi_3 = 0$ otherwise.

To characterize the status of maintaining a safe and comfortable separation between vehicles, we further define a separation-zone ($s$-zone) for each vehicle, which over-bounds the vehicle's $c$-zone with a separation margin. The feature $\phi_4$ takes $-1$ if vehicle $i$'s $s$-zone overlaps with any of the other vehicles' $s$-zones at their predicted states, and takes $0$ otherwise. 

The features $\phi_5$ and $\phi_6$ characterize the vehicle's status of driving toward its target lane. They are defined as $\phi_5 = -\, |r^i_x - x^i| - |r^i_y - y^i|$ and $\phi_6 = v^i$, so that the vehicle is encouraged to approach the reference point $r^i$ in its target lane as quickly as it can.

The above reward function design reflects common driving objectives in traffic. The weight vector $\mathbf{w}$ can be tuned to achieve reasonable driving behavior, or be calibrated using traffic data and approaches such as inverse reinforcement learning \cite{ng2000algorithms,ziebart2008maximum}.

\section{Game-theoretic Decision-making and Explicit Realization via Imitation Learning}
\label{sec: game theorstic decision-making}

Game theory is a useful tool for modeling intelligent agents' strategic interactions. In this paper, we exploit the level-k game theory \cite{stahl1995players,costa2006cognition} to model vehicles' interactive decision-making.

\subsection{Level-k reasoning and decision-making}
\label{sec: level-k driver model}
%encoding

In level-k game theory, it is assumed that players make decisions based on finite depths of reasoning, called ``level,'' and different players may have different reasoning levels. In particular, a level-$0$ player makes non-strategic decisions -- decisions without regard to the other players' decisions. Then, a level-$k$, $k\ge 1$, player makes strategic decisions by assuming that all of the other players are level-($k-1$), predicting their decisions based on such an assumption, and optimally responding to their predicted decisions. It is verified by experimental results from cognitive science that such a level-k reasoning process can model human interactions with higher accuracy than traditional analytic methods in many cases \cite{costa2006cognition}.

To incorporate level-k reasoning in our decision-making model \eqref{equ:general_rhc}, we start with defining a level-$0$ decision rule. According to the non-strategic assumption on level-$0$ players, we let a level-$0$ decision of a vehicle $i$, $i \in \mathcal{M}$, depend only on the traffic state $\mathbf{s}_t$, including its own state $s^i_t$ and the other vehicles' states $\mathbf{s}^{-i}_t$, but not on the other vehicles' actions $\mathbf{u}^{-i}_t$. In this paper, a level-$0$ decision, $(\mathbf{u}^i_t)^0 = \big\{ (u_{0|t}^i)^0,(u_{1|t}^i)^0,\dots,(u_{N-1|t}^i)^0\big\}$, is a  sequence of predicted actions that maximizes the cumulative reward in \eqref{equ:general_rhc} with treating all of the other vehicles as stationary obstacles over the planning horizon, i.e., $v^j_{\tau|t} = 0$, $\omega^j_{\tau|t} = 0$ for all $j \neq i$, $\tau = 0, 1,\dots,N$. This way, a level-$0$ vehicle represents an aggressive vehicle which assumes that all of the other vehicles will yield the right of way to it.

On the basis of the formulated level-$0$ decision rule, the level-$k$ decisions of the vehicles are obtained based on
\begin{align}
   (\mathbf{u}^i_t)^k &= \big\{ (u_{0|t}^i)^k,(u_{1|t}^i)^k,\dots,(u_{N-1|t}^i)^k\big\} \label{equ:level-k_rhc} \\
   & \in \argmax_{\mathbf{u}^i_t \in \mathcal{U}^N} \sum_{\tau=0}^{N-1}\lambda^\tau R \big(s^i_{\tau|t},\mathbf{s}^{-i}_{\tau|t},u_{\tau|t}^i,(\mathbf{u}^{-i}_{\tau|t})^{k-1} \big),
    \nonumber
\end{align}
for every $i \in \mathcal{M}$, and for every $k = 1,2,\dots,k_{\max}$ through sequential, iterated computations, where $(\mathbf{u}^{-i}_{\tau|t})^{k-1}$ denotes the level-($k-1$) decisions of the other vehicles $j \neq i$, which have been determined either in the previous iteration or based on the level-$0$ decision rule (for $k=1$), and $k_{\max}$ is the highest reasoning level for computation.

Given a finite action set $\mathcal{U}$, the problem \eqref{equ:level-k_rhc} for every $i \in \mathcal{M}$ and $k = 1,2,\dots,k_{\max}$ can be solved with exhaustive search, e.g., based on a tree structure \cite{claussmann2015path}.

\subsection{Explicit level-k decision-making via imitation learning}
\label{sec: explicit method}

A level-$k$ vehicle drives in traffic by applying
$u^i_t = (u_{0|t}^i)^k$ at every time step, where $(u_{0|t}^i)^k$ is determined according to \eqref{equ:level-k_rhc} with the current state as the initial condition, i.e., $s_{0|t}^i = s_t^i$ and $\mathbf{s}^{-i}_{0|t} = \mathbf{s}^{-i}_t$.

\di{On the one hand, the problem \eqref{equ:level-k_rhc} needs to be numerically solved. The required computational effort to solve \eqref{equ:level-k_rhc} grows for larger $k$ and larger numbers of interacting vehicles, because in order to compute the level-$k$ decision of vehicle $i$, the level-($k-1$) decisions of all other vehicles $j \neq i$ need to be determined first, which, in turn, need the prerequisite determination of level-($k-2$) decisions for $k \ge 2$, and etc. On the other hand, for virtual testing of AV control systems, fast simulations are desired so that a large number of test scenarios can be covered within a short period of real time. To achieve fast simulations, we exploit machine learning techniques to move the computational tasks associated with \eqref{equ:level-k_rhc} offline and obtain explicit level-$k$ decision policies for online use.}

In particular, we define a policy as a map from a triple of the ego vehicle's state $s_t^i$, the other vehicles' states $\mathbf{s}_t^{-i}$, and the ego vehicle's reasoning level $k$ to the level-$k$ action of the ego vehicle, i.e.
\begin{equation}\label{equ:policy_K}
    \pi_{\text{k}}: (s_t^i,\mathbf{s}_t^{-i}, k) \mapsto (u^i_t)^k.
\end{equation}
This map is algorithmically determined by solving the problem \eqref{equ:level-k_rhc} and letting $(u^i_t)^k = (u^i_{0|t})^k$. In what follows, we pursue an explicit approximation of $\pi_{\text{k}}$, denoted by $\hat{\pi}_{\text{k}}$, exploiting the approach called ``imitation learning.''

Imitation learning is an approach for an autonomous agent to learn a control policy from expert demonstrations to imitate expert's behavior. The expert can be a human expert \cite{codevilla2018end} or a well-behaved artificial intelligence \cite{sun2018fast}. In this paper, we treat the algorithmically determined map $\pi_{\text{k}}$ as the expert.

Imitation learning can be formulated as a standard supervised learning problem, in which case it is also commonly referred to as ``behavioral cloning,'' where the learning objective is to obtain a policy from a pre-collected dataset of expert demonstrations that best approximates the expert's behavior at the states contained in the dataset. Such a procedure can be described as
\begin{equation}\label{equ:SL}
    \hat{\pi}_{\text{k}} \in \argmin_{\pi_\theta}\, \mathbb{E}_{\,\bar{\mathbf{s}}\sim \mathbb{P}(\bar{\mathbf{s}}|\pi_{\text{k}})}\big[ \mathcal{L}(\pi_{\text{k}}(\bar{\mathbf{s}}),\pi_\theta(\bar{\mathbf{s}}))\big],
\end{equation}
where $\bar{\mathbf{s}}$ denotes the triple $(s^i,\mathbf{s}^{-i}, k)$, $\pi_{\text{k}}$ denotes the expert policy \eqref{equ:policy_K}, $\pi_\theta$ denotes a policy parameterized by $\theta$ (e.g., the weights of a neural network) that is being evaluated and optimized, $\mathcal{L}$ is a loss function, and the notation $\mathbb{E}_{\,\bar{\mathbf{s}}\sim \mathbb{P}(\bar{\mathbf{s}}|\pi_{\text{k}})}(\cdot)$ is defined as
\begin{equation}
   \mathbb{E}_{\,\bar{\mathbf{s}}\sim \mathbb{P}(\bar{\mathbf{s}}|\pi_{\text{k}})}(\cdot) = \int (\cdot)\, \text{d}\mathbb{P}(\bar{\mathbf{s}}|\pi_{\text{k}}).
\end{equation}
We remark that a key feature of the procedure \eqref{equ:SL} is that the expectation is with respect to the probability distribution $\mathbb{P}(\bar{\mathbf{s}}|\pi_{\text{k}})$ of the data $\bar{\mathbf{s}}$ determined by the expert policy $\pi_{\text{k}}$, which is essentially the empirical distribution of $\bar{\mathbf{s}}$ in the pre-collected dataset.

In our previous work \cite{8619275_tian}, we have explored the procedure \eqref{equ:SL} to obtain an explicit policy that imitates level-$k$ decisions for an autonomous vehicle to drive through a roundabout intersection.

\di{Using \eqref{equ:SL} to train the policy $\hat{\pi}_{\text{k}}$ has a drawback in that only the states that can be reached by executing the expert policy $\pi_{\text{k}}$ will be included in the dataset, and such a sampling bias may cause the error between $\hat{\pi}_{\text{k}}$ and $\pi_{\text{k}}$ to propagate in time. In particular, a small error may cause the vehicle to reach a state that is not exactly included in the dataset and, consequently, a large error may occur at the next time step.

Therefore, in this paper we consider an alternative approach, based on the ``Dataset Aggregation'' (DAgger) algorithm, to train the policy $\hat{\pi}_{\text{k}}$. DAgger is an iterative algorithm that optimizes the policy under its induced state distribution \cite{ross2011reduction}. The learning objective of DAgger can be described as
\begin{align}
    & \hat{\pi}_{\text{k}} \in \argmin_{\pi_\theta}\, \mathbb{E}_{\,\bar{\mathbf{s}}\sim \mathbb{P}(\bar{\mathbf{s}}|\pi_\theta)}\big[ \mathcal{L}(\pi_{\text{k}}(\bar{\mathbf{s}}),\pi_\theta(\bar{\mathbf{s}}))\big], \label{equ:IL} \\
    & \mathbb{E}_{\,\bar{\mathbf{s}}\sim \mathbb{P}(\bar{\mathbf{s}}|\pi_\theta)}(\cdot) = \int (\cdot)\, \text{d}\mathbb{P}(\bar{\mathbf{s}}|\pi_\theta),
\end{align}
where the distinguishing feature from \eqref{equ:SL} is that the expectation is with respect to the probability distribution $\mathbb{P}(\bar{\mathbf{s}}|\pi_\theta)$ induced from the policy $\pi_\theta$ that is being evaluated and optimized. DAgger can effectively resolve the aforementioned issue with regard to the propagation of error in time, since there will be data points $(\bar{\mathbf{s}},\pi_{\text{k}}(\bar{\mathbf{s}}))$ for states $\bar{\mathbf{s}}$ reached by executing $\hat{\pi}_{\text{k}}$.}

The procedure to obtain explicit level-$k$ decision policies based on an improved version of DAgger algorithm \cite{sun2018fast} is presented as Algorithm~\ref{alg: learn level-k policy}. In Algorithm~\ref{alg: learn level-k policy}, $n_{\max}$ denotes the maximum number of simulation episodes and $t_{\max}$ represents the length of a simulation episode. By ``initialize the simulation environment,'' we mean constructing a traffic scene, including specifying the road layout and geometry as well as the number of vehicles. By ``initialize vehicle $i$,'' we mean putting the vehicle in a lane entering the scene while satisfying a minimum separation distance constraint from the other vehicles, and specifying a sequence of target lanes for the vehicle to traverse and finally leave the scene. By ``vehicle $i$ fails,'' we mean the occurrence of 1) vehicle $i$'s $c$-zone overlapping with any of the other vehicles' $c$-zones, 2) crossing any of the road boundaries, or 3) crossing a lane marking that separates the traffic of opposite directions. And, by ``vehicle $i$ succeeds,'' we mean vehicle $i$ gets to the last target lane in its target lane sequence so that it can leave the scene without further interactions with the other vehicles.

\begin{algorithm}
\small
\caption{Imitation learning algorithm to obtain explicit level-$k$ decision policies} \label{alg: learn level-k policy}
    Initialize $\hat{\pi}_{\text{k}}^0$ to an arbitrary policy; \\
    Initialize dataset $\mathcal{D} \leftarrow \emptyset$; \\
    \For{$n= 1:n_{\max}$}{
        Initialize the simulation environment; \\
        \For{$i \in \mathcal{M}$}{
         Initialize vehicle $i$;
         }
        \For{$t= 0:t_{\max}-1$}{
            \For{$i \in \mathcal{M}$}{
                \If{vehicle $i$ fails or succeeds}{
                    Re-initialize vehicle $i$;
                }
                \For{$k= 1:k_{\max}$}{ \If{$\hat{\pi}_{\text{\rm k}}^{n-1}(s_t^i, \mathbf{s}^{-i}_t,k) \neq \pi_{\text{\rm k}}(s_t^i, \mathbf{s}^{-i}_t,k)$}{
                    $\mathcal{D} \leftarrow \mathcal{D} \cup \big((s_t^i, \mathbf{s}^{-i}_t,k),\pi_{\text{k}}(s_t^i, \mathbf{s}^{-i}_t,k)\big)$
                }
                }
                Randomly generate $k_t \in \{1,\dots,k_{\max}\}$; \\
                $s^i_{t+1} = f \big(s^i_{t}, \hat{\pi}_{\text{k}}^{n-1}(s_t^i, \mathbf{s}^{-i}_t,k_t) \big)$;
            }
        }
        Train classifier $\hat{\pi}_{\text{k}}^{n}$ on $\mathcal{D}$; \\
    }
    Output $\hat{\pi}_{\text{k}} =  \hat{\pi}_{\text{k}}^{n_{\max}}$.
\end{algorithm}

\section{Traffic in Unsignalized Intersection Network}
\label{sec: traffic simulator}

We model urban traffic where the road system is composed of straight roads and three most common types of unsignalized intersections: four-way, T-shaped, and roundabout \cite{Fitzpatrick2005-02-01}. Such traffic models can be used as simulation environments for virtual testing of AV control systems, which will be introduced in Section~\ref{sec: VV}.

\begin{figure}[ht]
\begin{center}
\begin{picture}(200.0, 80.0)
%%%%%%%%%%%%%%%%%%%%%%%%%%%%%%
%%%%%%%%%%%%
\put(  -22,  0){\epsfig{file=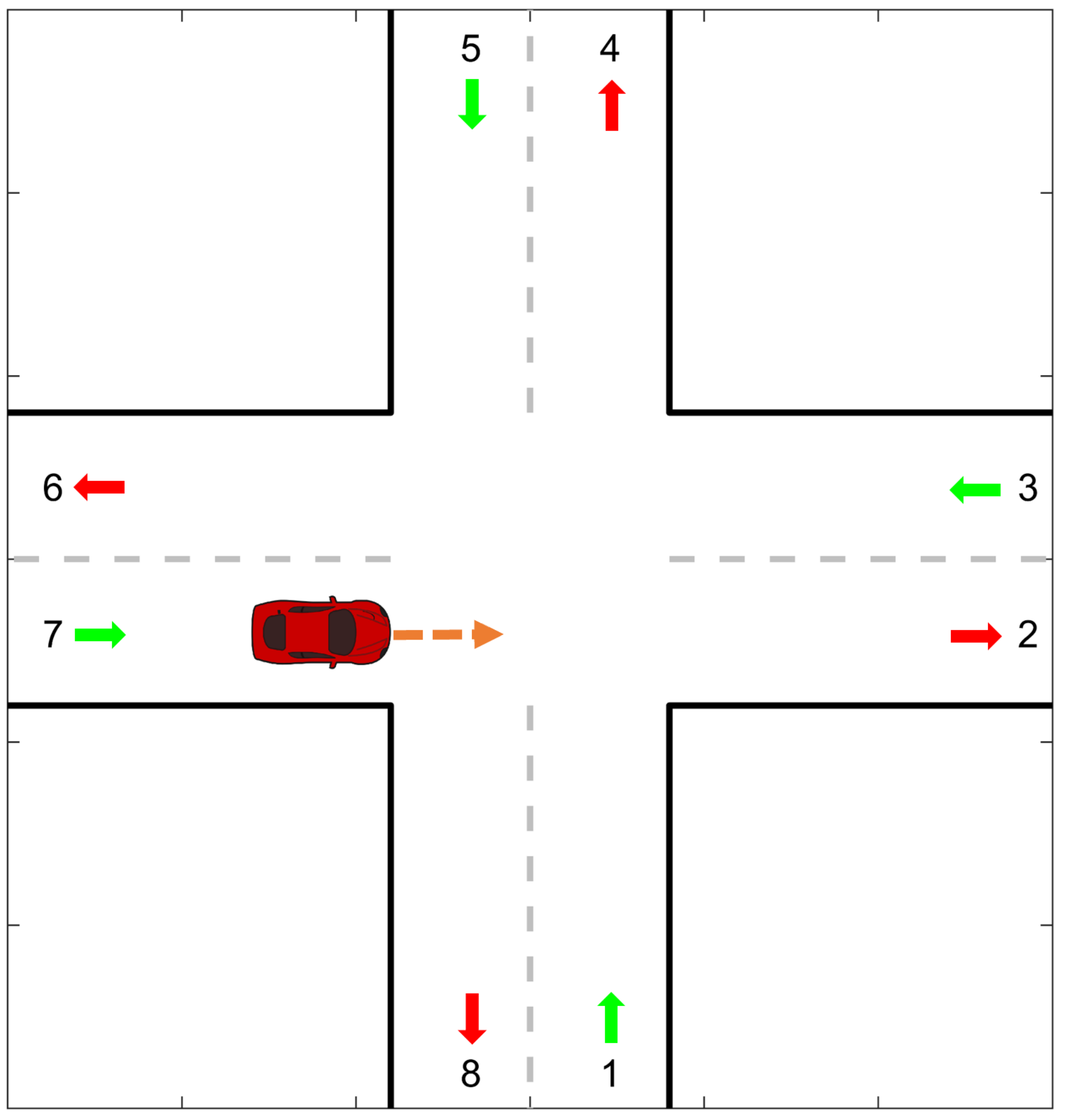,width = 0.31 \linewidth, trim=0.0cm 0.0cm 0cm 0cm,clip}}  %%%
\put(  60,  0){\epsfig{file=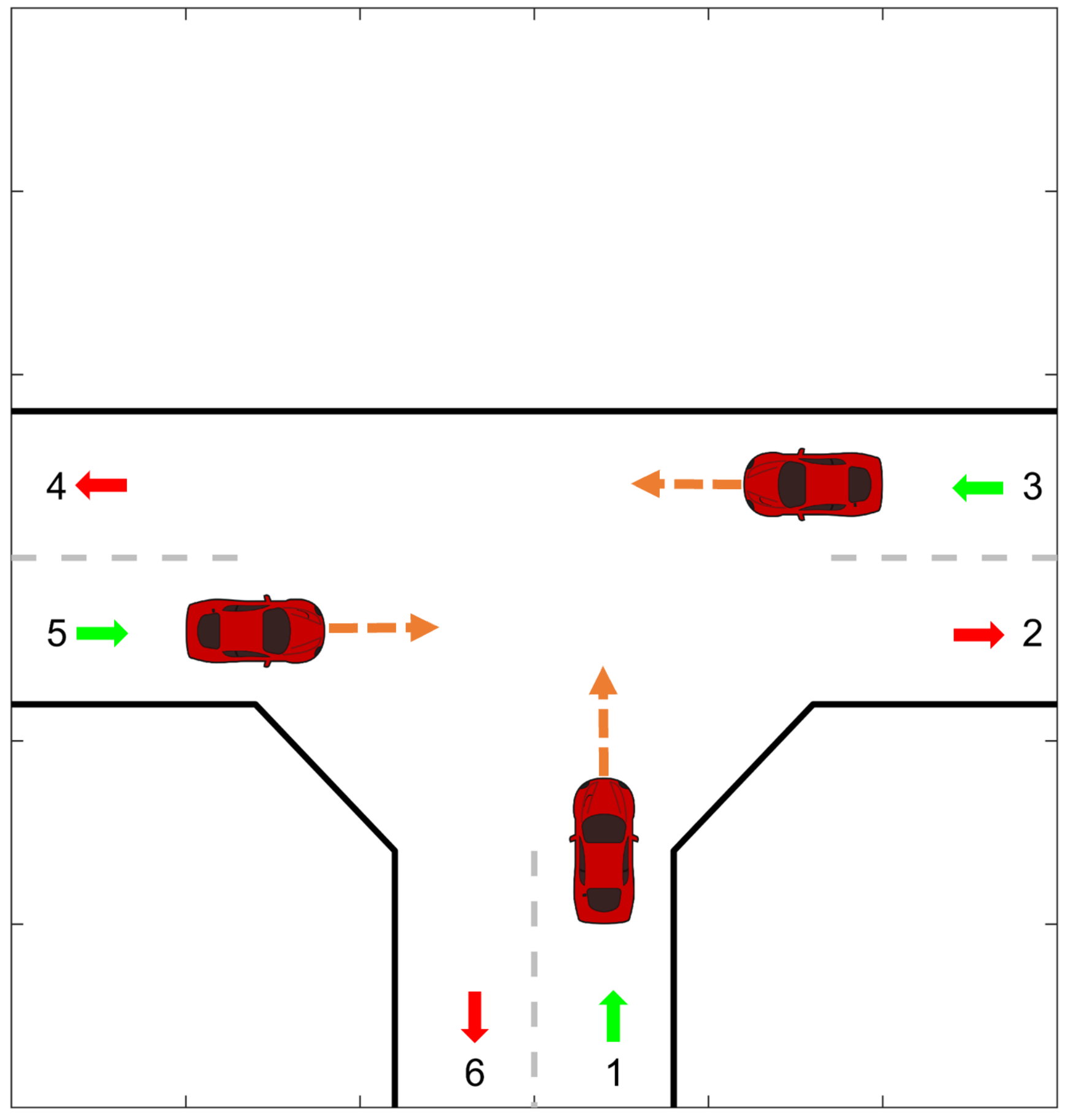,width = 0.31 \linewidth, trim=0.0cm 0.0cm 0cm 0cm,clip}}
\put(  142,  0.25){\epsfig{file=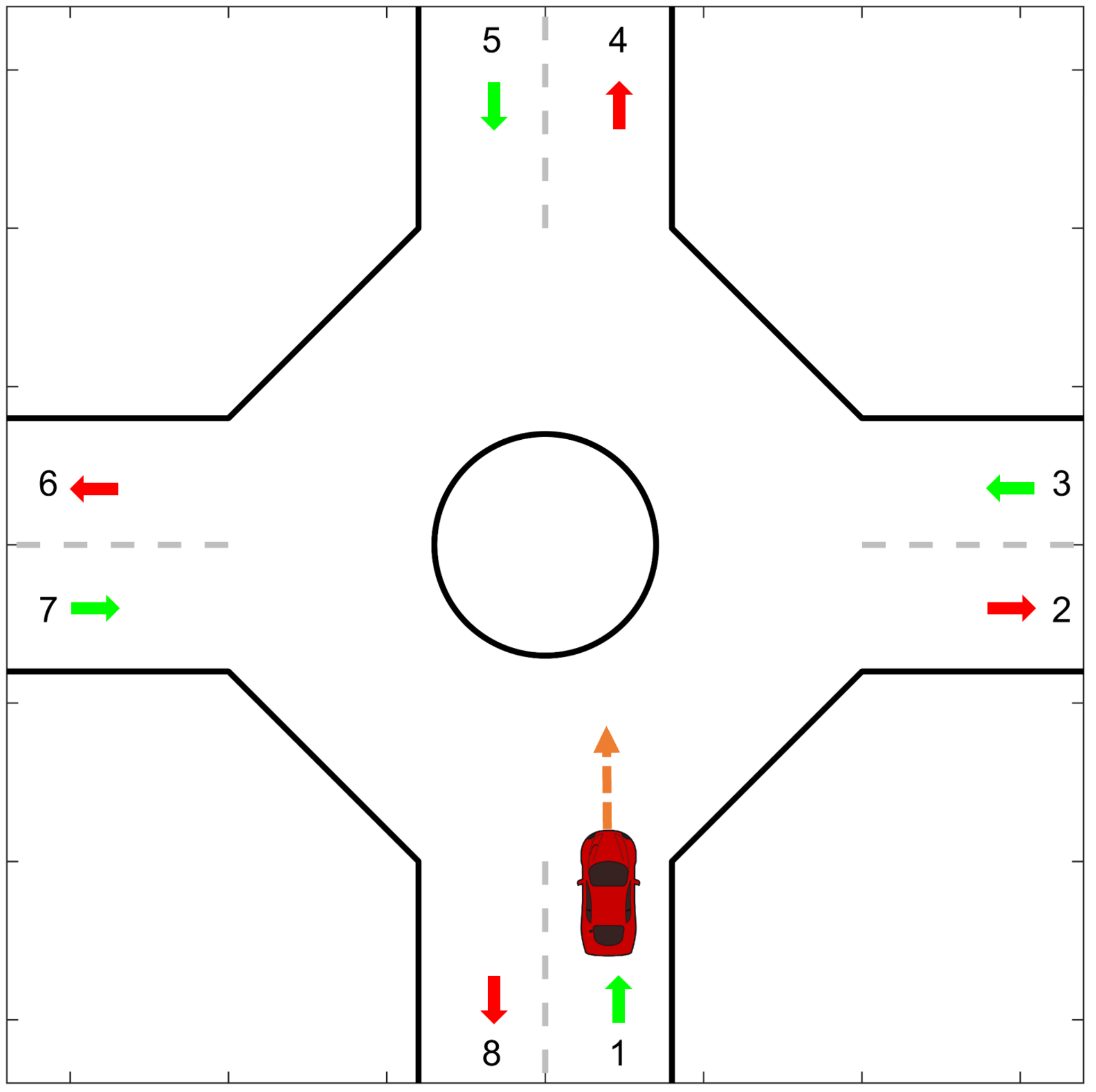,width = 0.325 \linewidth, trim=0.0cm 0.0cm 0cm 0cm,clip}}
%%%%%%%%%%%%%%%%%%%%%%
\small
\put(35,70){(a)}
\put(117,70){(b)}
\put(203,70){(c)}
\normalsize
\end{picture}
\end{center}
      \caption{Unsignalized intersections to be modeled: (a) four-way, (b) T-shaped, and (c) roundabout.}
      \label{fig: traffic}
\end{figure}
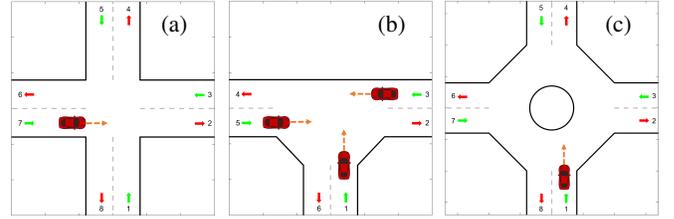

The three unsignalized intersections to be modeled are illustrated in Fig.~\ref{fig: traffic}. A vehicle can come from any of the entrance lanes (marked by green arrows) to enter an intersection and go to any of the exit lanes  (marked by red arrows) to leave it, except that U-turns are not allowed for four-way and T-shaped intersections.

When training the level-k policy $\hat{\pi}_{\text{k}}$ using Algorithm~\ref{alg: learn level-k policy}, we treat these three unsignalized intersections separately. Specifically, when initializing the simulation environment in step 4, we select one of these three unsignalized intersections as the traffic scene for the current simulation episode. In addition, since in this paper we only consider these three unsignalized intersections, their layout and geometry features can be characterized and distinguished using a label $\xi \in \{1,2,3\}$, i.e., the state $\xi^i$ of vehicle $i$ takes the value $1$ when vehicle $i$ operates in the area of the four-way intersection, $2$ for the T-shaped intersection, and $3$ for the roundabout. For more intersection types with various layout and geometry features, a higher dimensional vector $\xi$ may be used (e.g., see the intersection model in \cite{li_leader_follower}).

\begin{figure}[ht]
\centering
\includegraphics[width=0.48\textwidth]{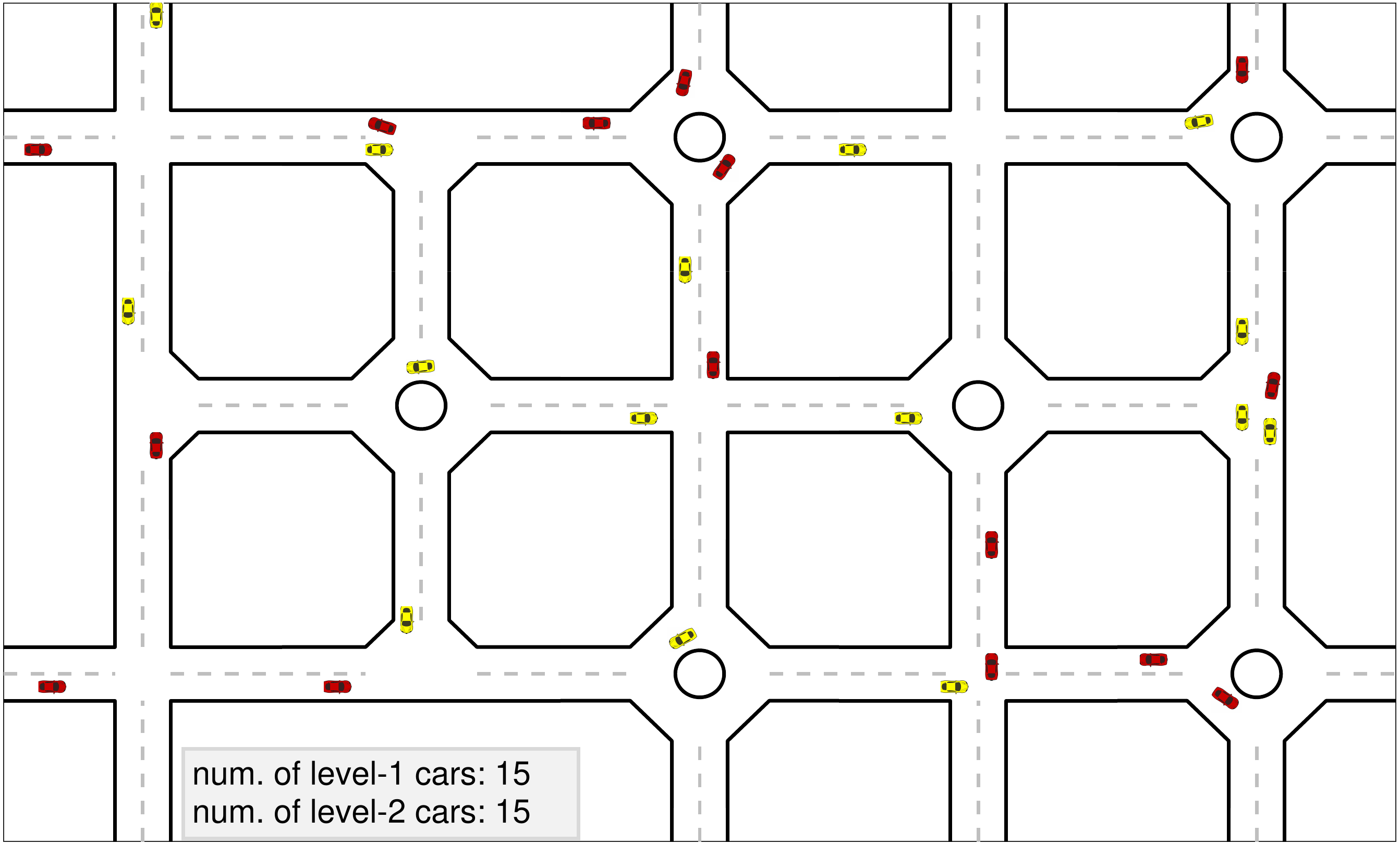}
\centering
\caption{An urban traffic scenario with $15$ level-$1$ cars (yellow) and $15$ level-$2$ cars (red).}
\label{fig: traffic_net}
\end{figure}

Once the policy $\hat{\pi}_{\text{k}}$ for each of these three unsignalized intersections has been obtained, we can model larger road systems using these three unsignalized intersections as modules and assembling them in arbitrary ways. Fig.~\ref{fig: traffic_net} shows an example of assembly. When a vehicle operates at/nearest to a specific intersection, it uses a local coordinate system, accounts for its interactions with only the vehicles in an immediate vicinity, and applies the $\hat{\pi}_{\text{k}}$ corresponding to this intersection.

To model the heterogeneity in driving styles of different drivers, we let different vehicles be of different reasoning levels. Specifically, a level-$k$ vehicle is controlled by the policy:
\begin{equation}\label{equ:policy_k}
    \hat{\pi}_{k} = \hat{\pi}_{\text{k}}(\cdot,\cdot, k): (s_t^i,\mathbf{s}_t^{-i}) \mapsto (u^i_t)^k.
\end{equation}
For instance, in Fig.~\ref{fig: traffic_net} the $15$ yellow cars are level-$1$ and the $15$ red cars are level-$2$.

\section{Autonomous Vehicle Control Approaches}\label{sec: VV}

In this section, we describe two AV control approaches for urban traffic environments with unsignalized intersections. These approaches will be tested and calibrated using our traffic model, thereby demonstrating its utility for verification and validation.

\subsection{Adaptive control based on level-$k$ models}

In this approach, the autonomous ego vehicle treats the other drivers as level-$k$ drivers. As different drivers may behave corresponding to different reasoning levels, the ego vehicle estimates their levels and adapts its own control strategy based on the estimation results.

The control strategy of the autonomous ego vehicle, $i$, can be described as: At each discrete-time instant $t$, vehicle $i$ solves for
\begin{align}
   (\mathbf{u}^i_t)^a &= \big\{ (u_{0|t}^i)^a,(u_{1|t}^i)^a,\dots,(u_{N-1|t}^i)^a\big\} \label{equ:adaptive_rhc} \\
   & \in \argmax_{\mathbf{u}^i_t \in \mathcal{U}^N }\sum_{\tau=0}^{N-1}\lambda^\tau R \big(s_{\tau|t}^i, \mathbf{s}^{-i}_{\tau|t},u_{\tau|t}^i,(\mathbf{u}_{\tau|t}^{-i})^{\tilde{k}} \big),
    \nonumber
\end{align}
where $(\mathbf{u}_{\tau|t}^{-i})^{\tilde{k}} = \big((u_{\tau|t}^j)^{\tilde{k}^j_t}\big)_{j \in \mathcal{M}, j \neq i}$ denotes the collection  of predicted actions of the other vehicles. In particular, the actions of vehicle $j$, $u_{\tau|t}^j$, $\tau = 0,1,\dots,N-1$, are predicted by modeling vehicle $j$ as level-$\tilde{k}^j_t$ and are solved for according to \eqref{equ:level-k_rhc}, where $\tilde{k}^j_t$ is determined according to the following maximum likelihood principle:
\begin{equation}
    \tilde{k}^j_t \in \argmax_{k \in \mathcal{K}}\, \mathbb{P}^i (k^j = k|t),
    \label{equ:level_estimate_1}
\end{equation}
in which $\mathbb{P}^i (k^j = k|t)$ represents vehicle $i$'s belief at time $t$ in that vehicle $j$ can be modeled as level-$k$, with $k$ taking values in a model set $\mathcal{K}$. The beliefs $\mathbb{P}^i (k^j = k|t)$ get updated after each time step according to the following algorithm: If there exist $k,k' \in \mathcal{K}$ such that $\pi_{\text{k}}(s_t^j,\mathbf{s}_t^{-j}, k) \neq \pi_{\text{k}}(s_t^j,\mathbf{s}_t^{-j}, k')$, then
\begin{align}
    & \mathbb{P}^i (k^j = k|t+1) = \frac{p^i (k^j = k|t+1)}{\sum_{k' \in \mathcal{K}} p^i (k^j = k'|t+1)},
    \label{equ:level_estimate_2} \\
    & p^i (k^j = k|t+1) =
        \begin{cases}
        (1-\beta) \mathbb{P}^i (k^j = k|t) + \beta
        & \text{if } k = \hat{k}^j_t, \\
        \mathbb{P}^i (k^j = k|t), & \text{otherwise},
        \end{cases} \nonumber
\end{align}
where $\beta \in [0,1]$ represents an update step size, and
\begin{align}
    \hat{k}^j_t &\in \argmin_{k \in \mathcal{K}}\, \text{dist}\big(u_t^j,(u_t^j)^k\big), \nonumber \\
    &= \sqrt{(a_t^j - (a_t^j)^k)^2 + (\omega_t^j - (\omega_t^j)^k)^2}\,; \label{equ:level_estimate_3}
\end{align}
if $\pi_{\text{k}}(s_t^j,\mathbf{s}_t^{-j}, k) = \pi_{\text{k}}(s_t^j,\mathbf{s}_t^{-j}, k')$ for all $k, k' \in \mathcal{K}$, then $\mathbb{P}^i (k^j = k|t+1) = \mathbb{P}^i (k^j = k|t)$ for all $k \in \mathcal{K}$.

The level estimation algorithm \eqref{equ:level_estimate_1}-\eqref{equ:level_estimate_3} has the following three features: 1) If the actions predicted by all of the models in $\mathcal{K}$ are the same, then the autonomous ego vehicle has no information to distinguish their relative accuracy and thus maintains its previous beliefs. 2) Otherwise, the ego vehicle identifies the model(s) in $\mathcal{K}$ whose prediction $(u_{t}^j)^k$ matches vehicle $j$'s actually applied action $u_t^j$ for time $t$ with the highest accuracy. 3) The ego vehicle improves its belief(s) in that model(s) from its previous beliefs. This way, it takes into account both its previous estimates and the current, latest estimate.

Similar to \eqref{equ:policy_K} defined by \eqref{equ:level-k_rhc}, we can define a policy to represent the control determined by \eqref{equ:adaptive_rhc} as follows:
\begin{equation}
    \pi_{\text{a}}: (s_t^i,\mathbf{s}^{-i}_t,\tilde{\mathbf{k}}^{-i}_t) \mapsto (u^i_t)^a,
\end{equation}
where $\tilde{\mathbf{k}}^{-i}_t = (\tilde{k}^j_t)_{j \in \mathcal{M}, j \neq i}$ denotes the collection of level estimates of the other vehicles and $(u^i_t)^a = (u^i_{0|t})^a$ is determined by \eqref{equ:adaptive_rhc}. Furthermore, we can train an explicit approximation $\hat{\pi}_{\text{a}}$ to $\pi_{\text{a}}$ using a similar imitation learning procedure as that for training the explicit approximation $\hat{\pi}_{\text{k}}$ to $\pi_{\text{k}}$. This way, together with replacing $\pi_{\text{k}}$ with $\hat{\pi}_{\text{k}}$ in the level estimation algorithm \eqref{equ:level_estimate_1}-\eqref{equ:level_estimate_3}, we can move the major computational tasks involved in this adaptive control approach \eqref{equ:adaptive_rhc}-\eqref{equ:level_estimate_3} offline, and thus, render its online computational feasibility.

The algorithm to train $\hat{\pi}_{\text{a}}$ using DAgger with $\pi_{\text{a}}$ as the expert policy is similar to Algorithm~\ref{alg: learn level-k policy} and is omitted.

\subsection{Rule-based control}

The second AV control approach that we consider is a rule-based solution. Compared to many other approaches, rule-based control has the advantage of interpretability and can often be calibrated by tuning a small number of parameters.

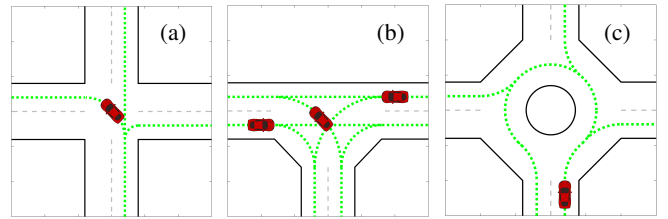
\begin{figure}[ht]
\begin{center}
\begin{picture}(200.0, 80.0)
%%%%%%%%%%%%%%%%%%%%%%%%%%%%%%
%%%%%%%%%%%%
\put(  -22,  0){\epsfig{file=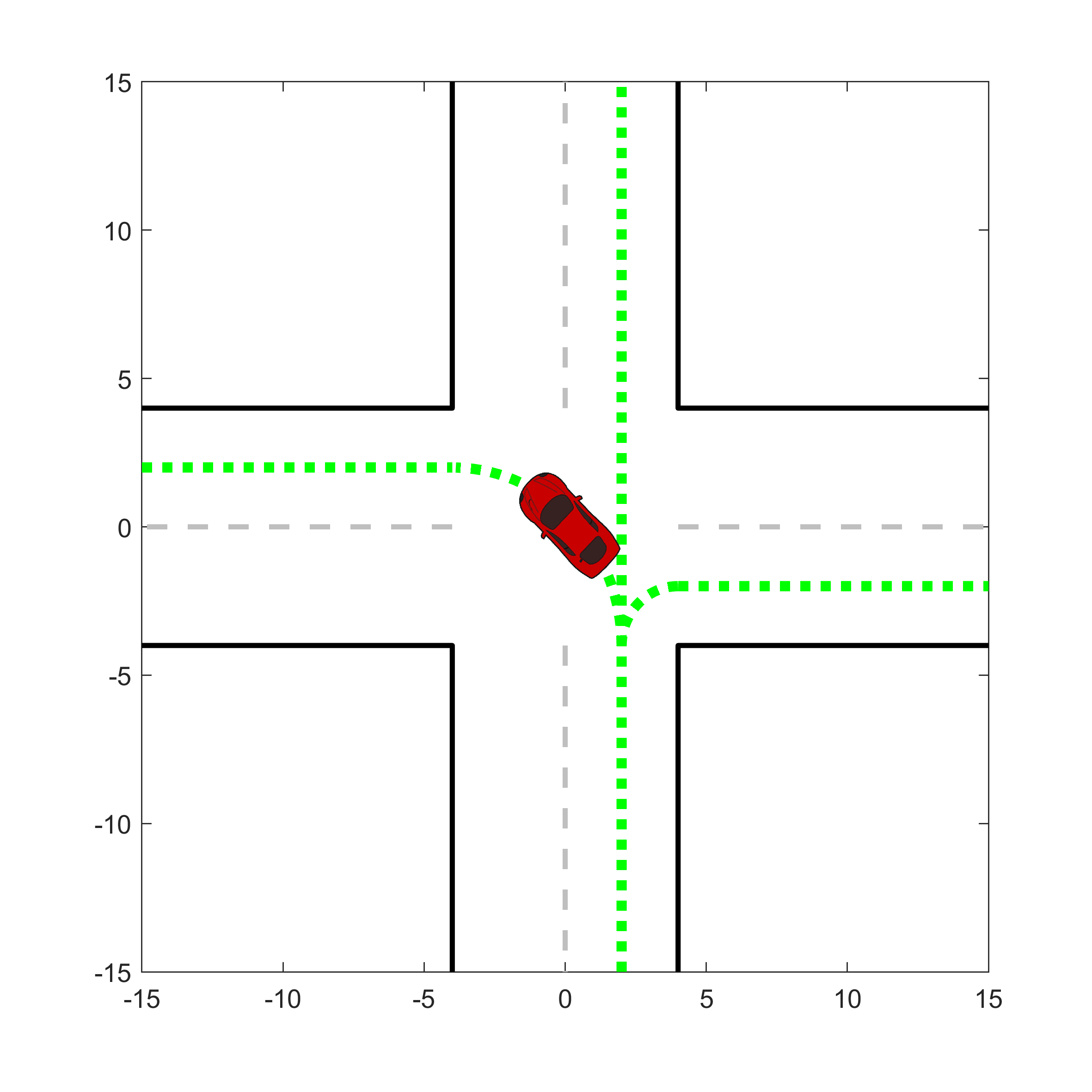,width = 0.34 \linewidth, trim=1.8cm 1.6cm 0cm 0cm,clip}}  %%%
\put(  60,  0){\epsfig{file=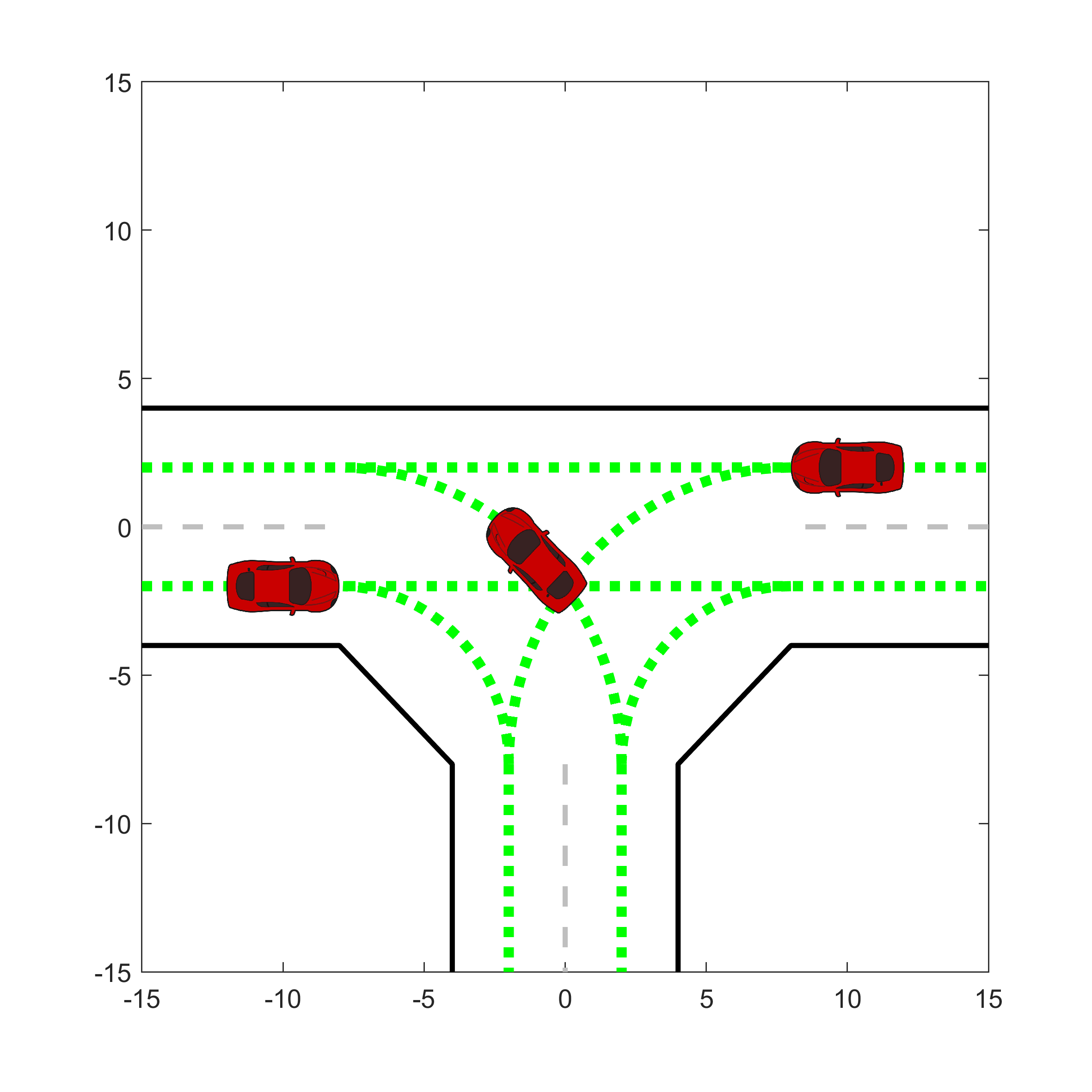,width = 0.34\linewidth, trim=1.82cm 1.6cm 0cm 0cm,clip}}
\put(  142,  -0.5){\epsfig{file=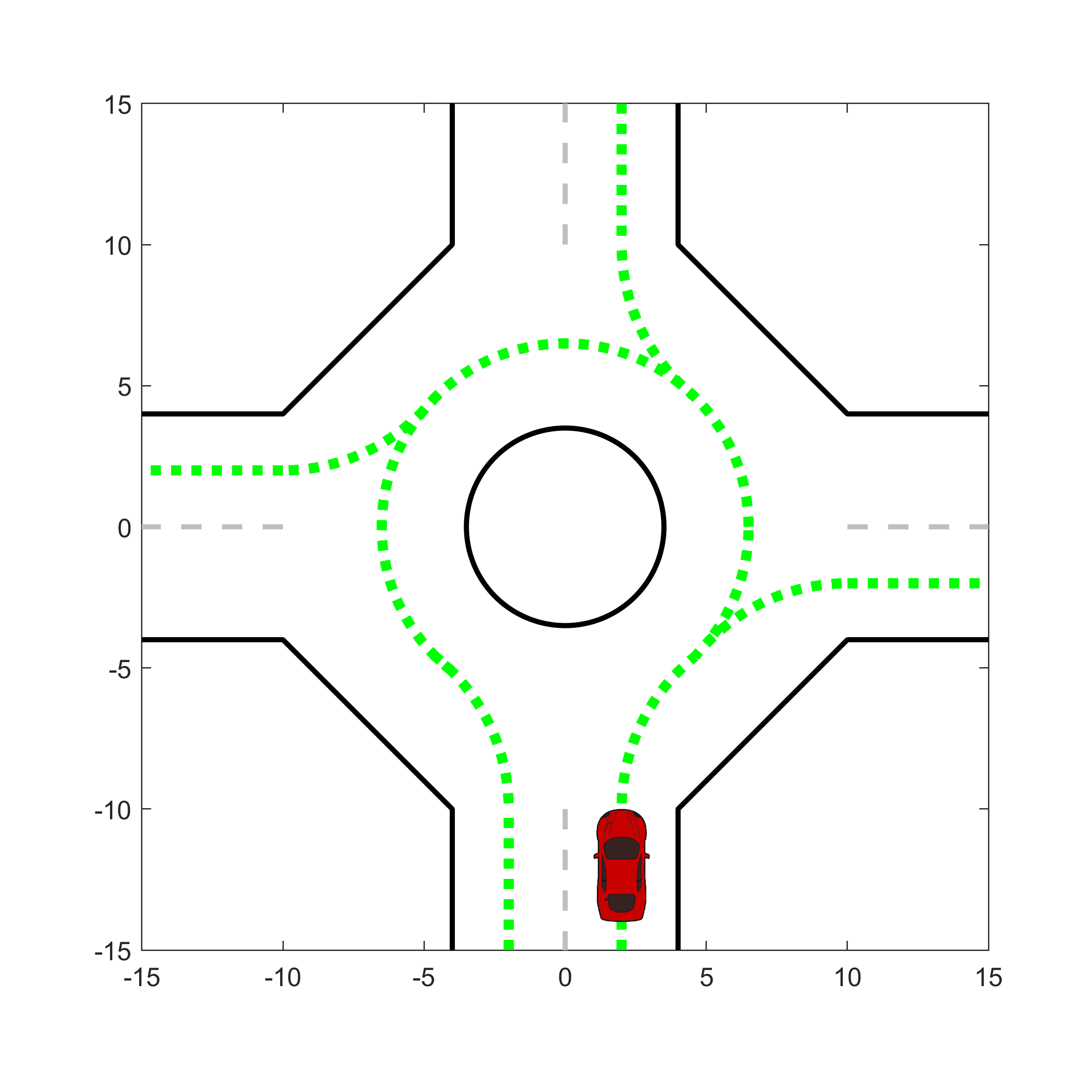,width = 0.358\linewidth, trim=1.82cm 1.8cm 0cm 0cm,clip}}
%%%%%%%%%%%%%%%%%%%%%%
\small
\put(35,67){(a)}
\put(117,67){(b)}
\put(203,67){(c)}
\normalsize
\end{picture}
\end{center}
      \caption{Reference paths for the autonomous ego vehicle to drive through (a) four-way, (b) T-shaped, and (c) roundabout intersections.}
      \label{fig: path}
\end{figure}

In this approach, the autonomous ego vehicle drives by following a pre-planned reference path and accounts for its interactions with other vehicles by adjusting its speed along the path correspondingly. Examples of reference paths for the autonomous ego vehicle to drive through intersections are illustrated by the green dotted curves in Fig.~\ref{fig: path}.

The basic control rules can be explained as follows: The autonomous ego vehicle pursues a higher speed along the reference path if there is no other vehicle in conflict with it. If there are other vehicles in conflict with it, then the autonomous ego vehicle yields to them by maximizing distances from them. Specifically, at each discrete-time instant $t$, the autonomous ego vehicle, $i$, selects and applies for one time step an acceleration value from a finite set of accelerations,  $\mathcal{A}$, according to Algorithm~\ref{alg: rule-based}.

\begin{algorithm}
\small
\caption{Rule-based autonomous vehicle control algorithm} \label{alg: rule-based}
Initialize $\mathcal{M}_c \leftarrow \emptyset$; \\
\For{$j \in \mathcal{M}, j \neq i$}{
    \If{the estimated future path of $j$ intersects with $i$'s future path and $\text{dist}\big((x_t^i,y_t^i),(x_t^j,y_t^j)\big) \leq R_c$}{
        $\mathcal{M}_c \leftarrow \mathcal{M}_c \cup \{j\}$;
    }
}
\eIf{$\mathcal{M}_c \neq \emptyset$}{
    $(a_t^i)^{r} = \argmax_{a \in \mathcal{A}} \big[ \min_{j \in \mathcal{M}_c} \text{dist}\big((x_{1|t}^i,y_{1|t}^i),(x_{1|t}^j,y_{1|t}^j)\big) \big]$;
}{
    $(a_t^i)^{r} = \max \{a \in \mathcal{A}\}$;
}
Output $(a_t^i)^{r}$.
\end{algorithm}

In Algorithm~\ref{alg: rule-based}, $\mathcal{M}_c$ represents the set of vehicles that are in conflict with the ego vehicle. In particular, the ego vehicle estimates each of the other vehicles' future paths based on their current positions and their target lanes and using the same path planning algorithm that is used by the ego vehicle to create its own path. If the estimated future path of a vehicle $j$ intersects with the ego vehicle's own future path and the current distance between these two vehicles is smaller than a threshold value $R_c$, then vehicle $j$ is identified as a vehicle in conflict, i.e., $j \in \mathcal{M}_c$. \di{In particular, the distance function $\text{dist}(\cdot,\cdot)$ measures the Euclidean distance.}

If there are vehicles in conflict, $\mathcal{M}_c \neq \emptyset$, then the ego vehicle maximizes the minimum among the predicted distances from these vehicles to improve safety. In step~8, $(x_{1|t}^i,y_{1|t}^i)$ represents the predicted position of the ego vehicle $i$ by driving along its reference path for one step with the speed after applying the acceleration $a$, and $(x_{1|t}^j,y_{1|t}^j)$ represents the predicted position of vehicle $j$ by driving along its current heading direction with its current speed. If there is no vehicle in conflict, $\mathcal{M}_c = \emptyset$, then the ego vehicle maximizes its speed.

Note that the key parameter for this rule-based control approach is the threshold value $R_c$. It determines both whether a vehicle will be identified as in conflict with the ego vehicle and the separation distance that the ego vehicle tries to keep from other vehicles. We will utilize our traffic model to calibrate $R_c$ in Section~\ref{sec: evaluation and calibration}.

\section{Results}
\label{sec: results}

In this section, we show simulation results of our level-k game theory-based vehicle interaction model, and illustrate its application to the verification, validation and calibration of AV control systems.

\subsection{Level-k vehicle models}
\label{sec: results_1}

We consider a sampling period $\Delta t = 0.25$[s] and an action set $\mathcal{U}$ consisting of $6$ actions representing common driving maneuvers in urban traffic, listed in Table~\ref{table: action-set}. The weight vector, the planning horizon, and the discount factor for the reward function \eqref{equ:reward} are $\mathbf{w} = [1000, 500, 50, 100, 5, 1]^{\intercal}$, $N = 4$, and $\lambda = 0.8$. When evaluating the features $\phi_1$ and $\phi_4$, we consider the $c$-zone of a vehicle as a $5 [\text{m}] \times 2 [\text{m}]$ rectangle centered at the vehicle's position $(x,y)$ and stretched along its heading direction $\theta$, and the $s$-zone of a vehicle as a rectangle concentric with its $c$-zone and $8 \text{[m]} \times 2.4 \text{[m]}$ in size. Furthermore, we consider a speed range $[v_{\min},v_{\max}] = [0,5]$[m/s]. When the speed calculated based on the model \eqref{equ:kinematics} gets outside $[v_{\min},v_{\max}]$, it is saturated to this range. \di{We note that $[v_{\min},v_{\max}] = [0,5]$[m/s] is a reasonable range to represent common speeds for vehicles to drive through unsignalized intersections. For instance, in California it is suggested to maintain the vehicle speed below $15$[mph] when traversing an uncontrolled highway intersection \cite{speed_limit}.}

\begin{table}[ht]
\centering
\caption{Action set $\mathcal{U}$.}
\label{table: action-set}
\begin{tabular}{ c  c  c}
 \hline
action $u$ & $a \ \text{[m}/\text{s}^2]$ & $\omega \ \text{[rad/s]}$  \\
 \hline
 maintain ($u_1$) & 0 & 0\\
 \hline
 accelerate ($u_2$) & 2.5 & 0\\
 \hline
 decelerate ($u_3$) & -2.5 & 0\\
 \hline
 hard brake ($u_4$) & -5 & 0\\
 \hline
 turn left ($u_5$) & 0 & $\pi/4$\\
 \hline
 turn right ($u_6$) & 0 & $-\pi/4$\\
 \hline
\end{tabular}
\end{table}

Experimental studies \cite{costa2006cognition,costa2009comparing} suggest that humans are most commonly level-$1$ and -$2$ reasoners in their interactions. Therefore, we model vehicles in traffic using level-$1$ and -$2$ policies in this paper. In particular, on the basis of our level-$0$ decision rule (see Section~\ref{sec: level-k driver model}), a level-$1$ vehicle represents a cautious/conservative vehicle and a level-$2$ vehicle represents an aggressive vehicle. Indeed, as level-$0$ and level-$2$ vehicles both represent aggressive vehicles, they behave similarly in many situations.

We use a neural network to represent a policy $\pi_{\theta}$ and train its weights $\theta$ using Algorithm~\ref{alg: learn level-k policy} to obtain an explicit approximation $\hat{\pi}_{\text{k}}$ to the level-k policy $\pi_{\text{k}}$, which is algorithmically determined based on \eqref{equ:level-k_rhc}. The accuracy of the obtained $\hat{\pi}_{\text{k}}$ in terms of matching $\pi_{\text{k}}$ on the training dataset is $98.3 \%$. Then, we generate $30 \%$ more data points of $\big((s_t^i, \mathbf{s}^{-i}_t,k),\pi_{\text{k}}(s_t^i, \mathbf{s}^{-i}_t,k)\big)$ for testing. The accuracy of $\hat{\pi}_{\text{k}}$ in terms of matching $\pi_{\text{k}}$ on the test dataset is $97.8 \%$. \di{As has been discussed at the beginning of Section~\ref{sec: explicit method}, the reason for generating $\hat{\pi}_{\text{k}}$ is to move the numerical computations for determining the level-$k$ decisions through \eqref{equ:level-k_rhc} offline. With $\hat{\pi}_{\text{k}}$, the interactive decision-making processes of vehicles are reduced to function evaluations (here, the function is expressed as a neural network). This way, the online simulations of traffic scenarios, used as environments for virtual testing of AV control systems, can be significantly accelerated.}

\begin{figure}[ht]
\begin{center}
\begin{picture}(200.0, 170.0)
%%%%%%%%%%%%%%%%%%%%%%%%%%%%%%
\put(  -22,  85){\epsfig{file=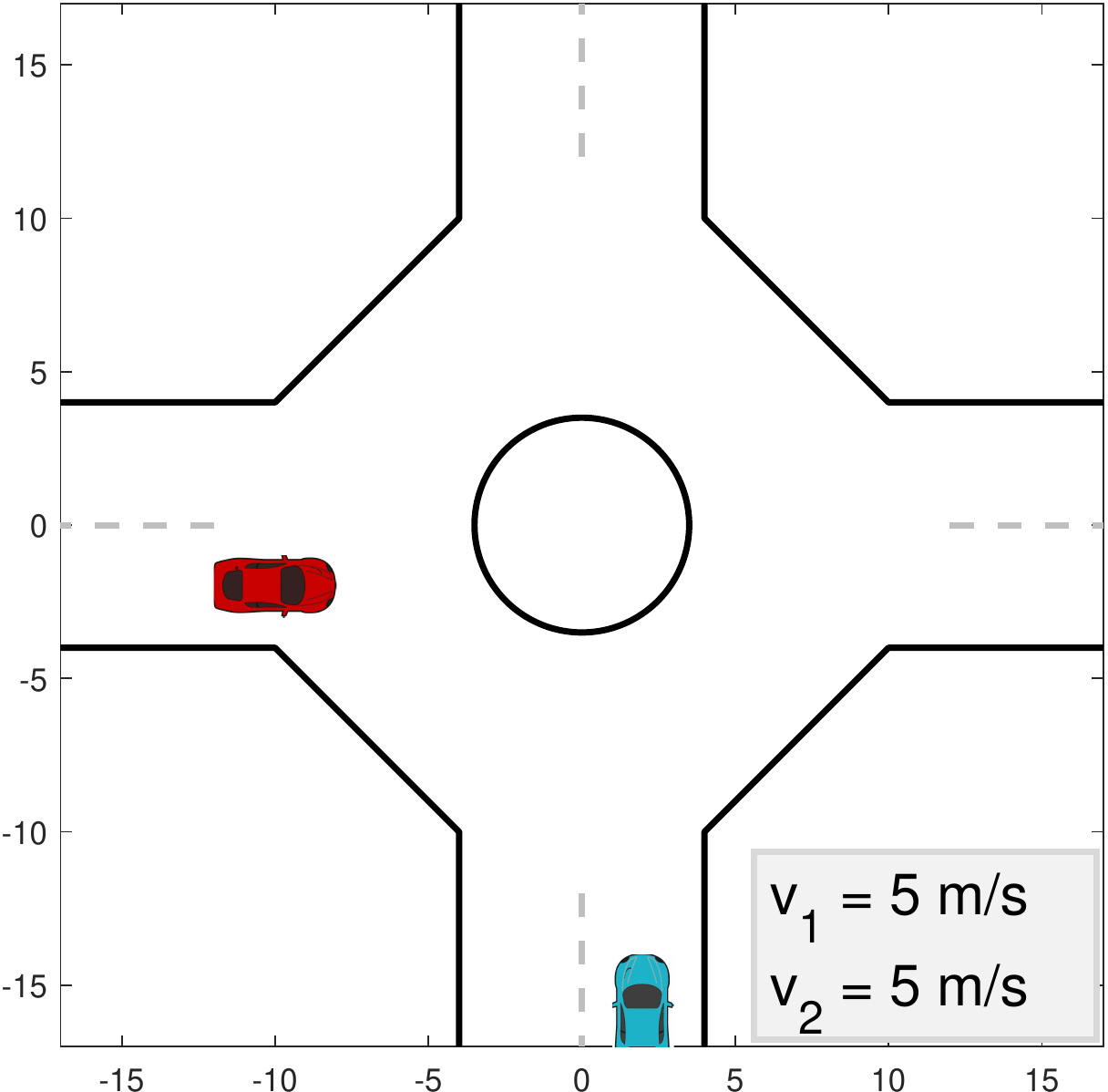,width = 0.33 \linewidth, trim=0.6cm 0.4cm 0cm 0cm,clip}}  %%%
\put(  60,  85){\epsfig{file=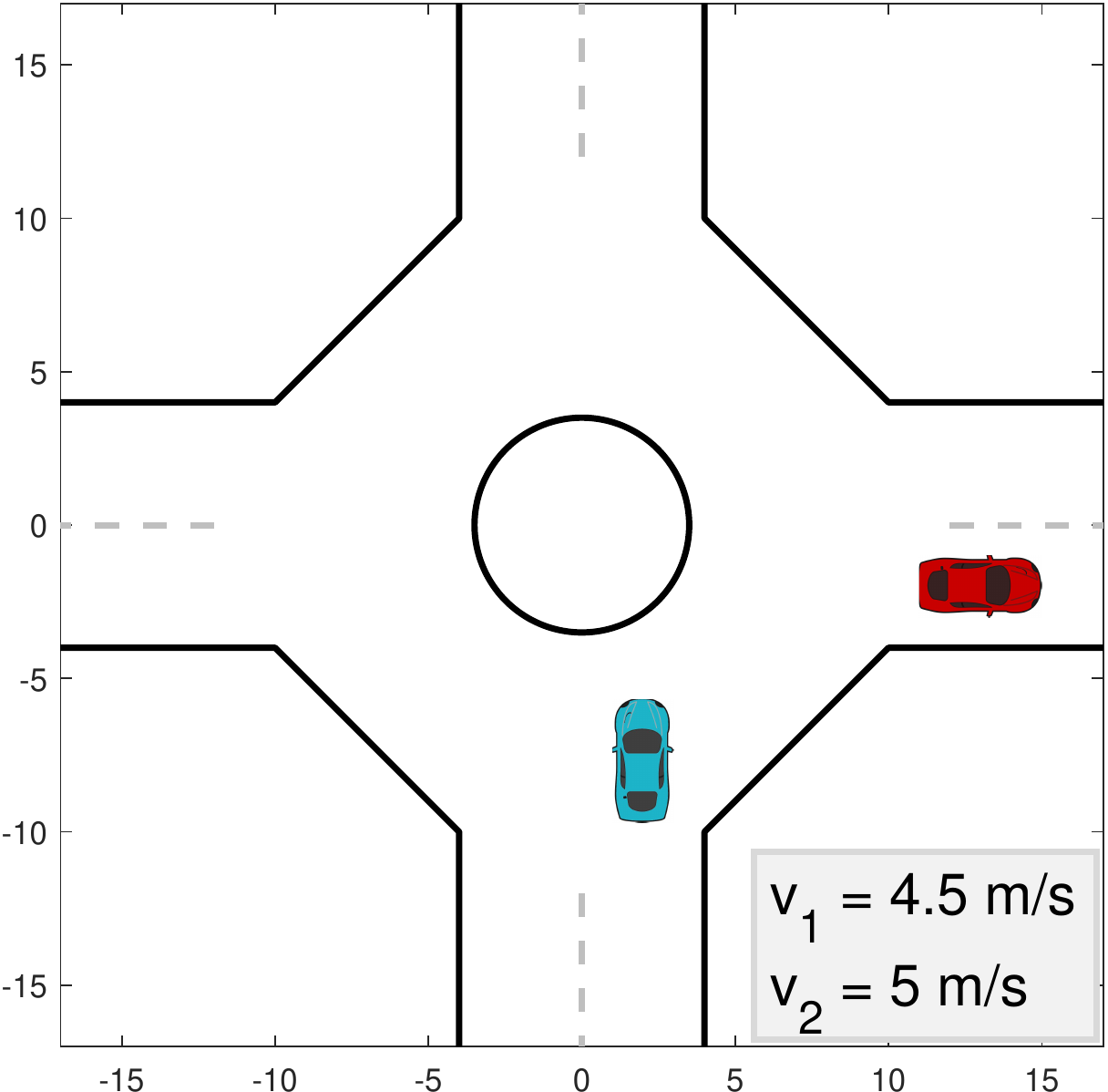, width = 0.33\linewidth, trim=0.6cm 0.4cm 0cm 0cm,clip}}

\put(  142,  85){\epsfig{file=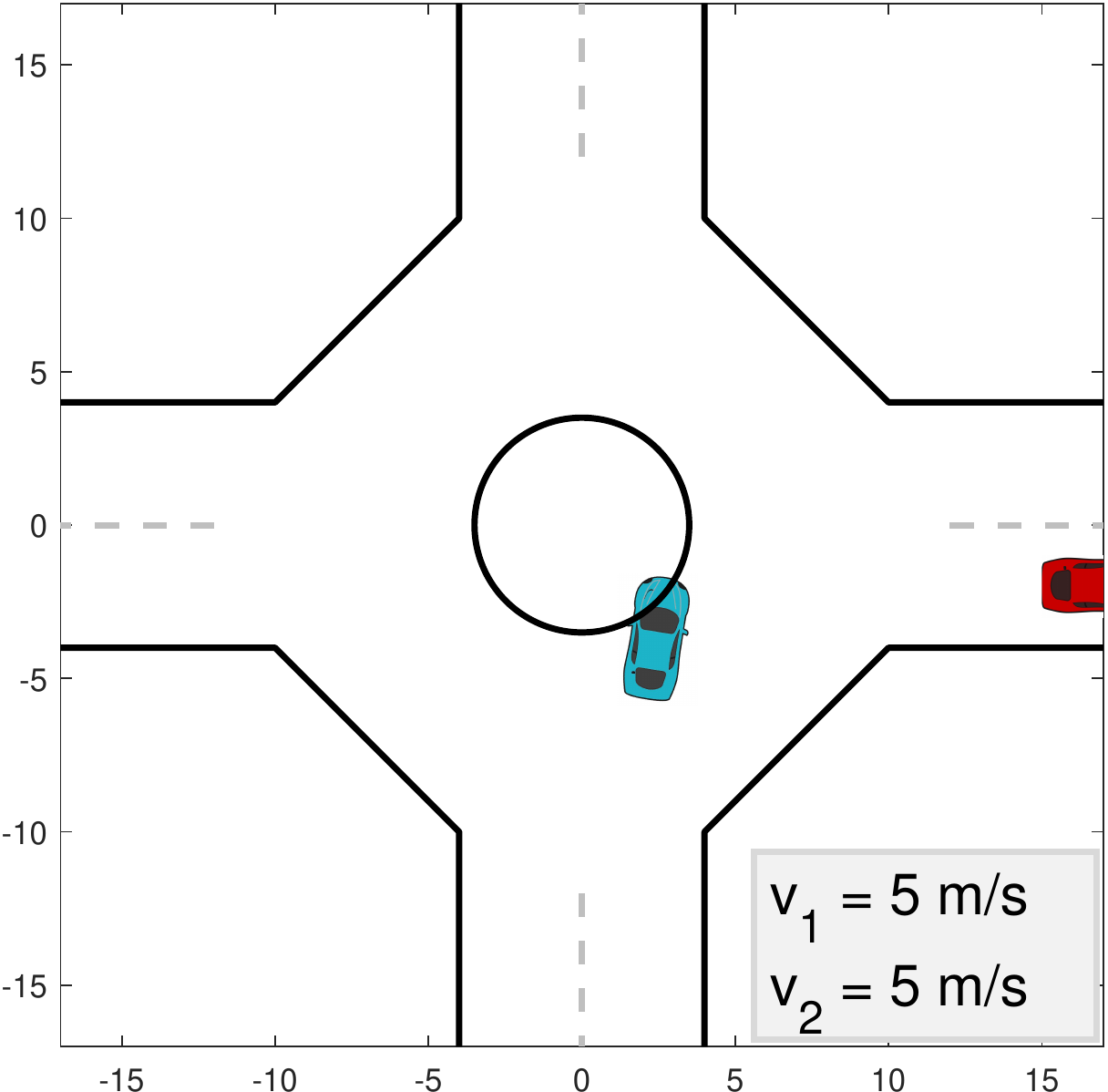, width = 0.33\linewidth, trim=0.6cm 0.4cm 0cm 0cm,clip}}
%%%%%%%%%%%%
\put(  -22,  0){\epsfig{file=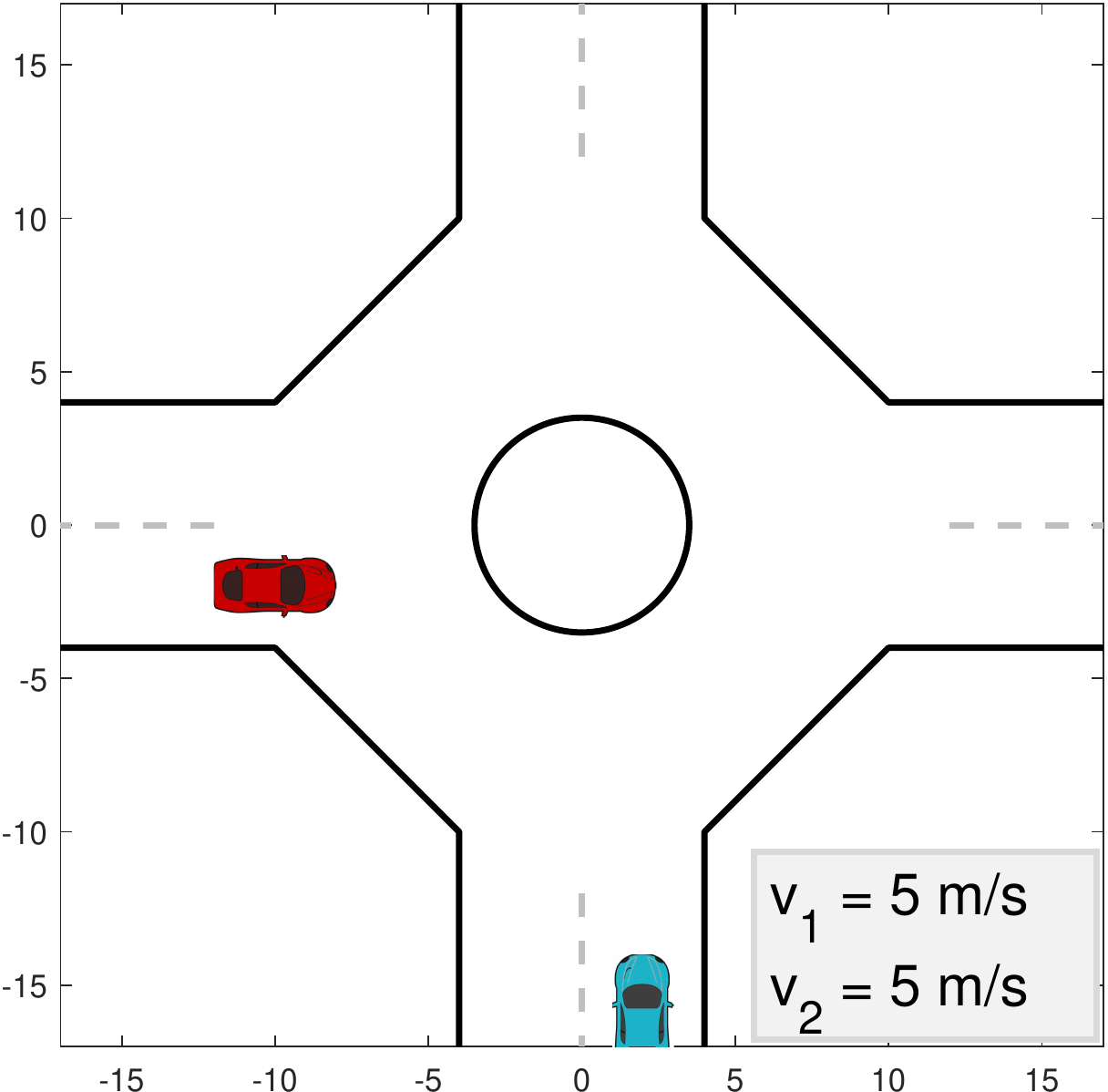,width = 0.33 \linewidth, trim=0.6cm 0.4cm 0cm 0cm,clip}}  %%%
\put(  60,  0){\epsfig{file=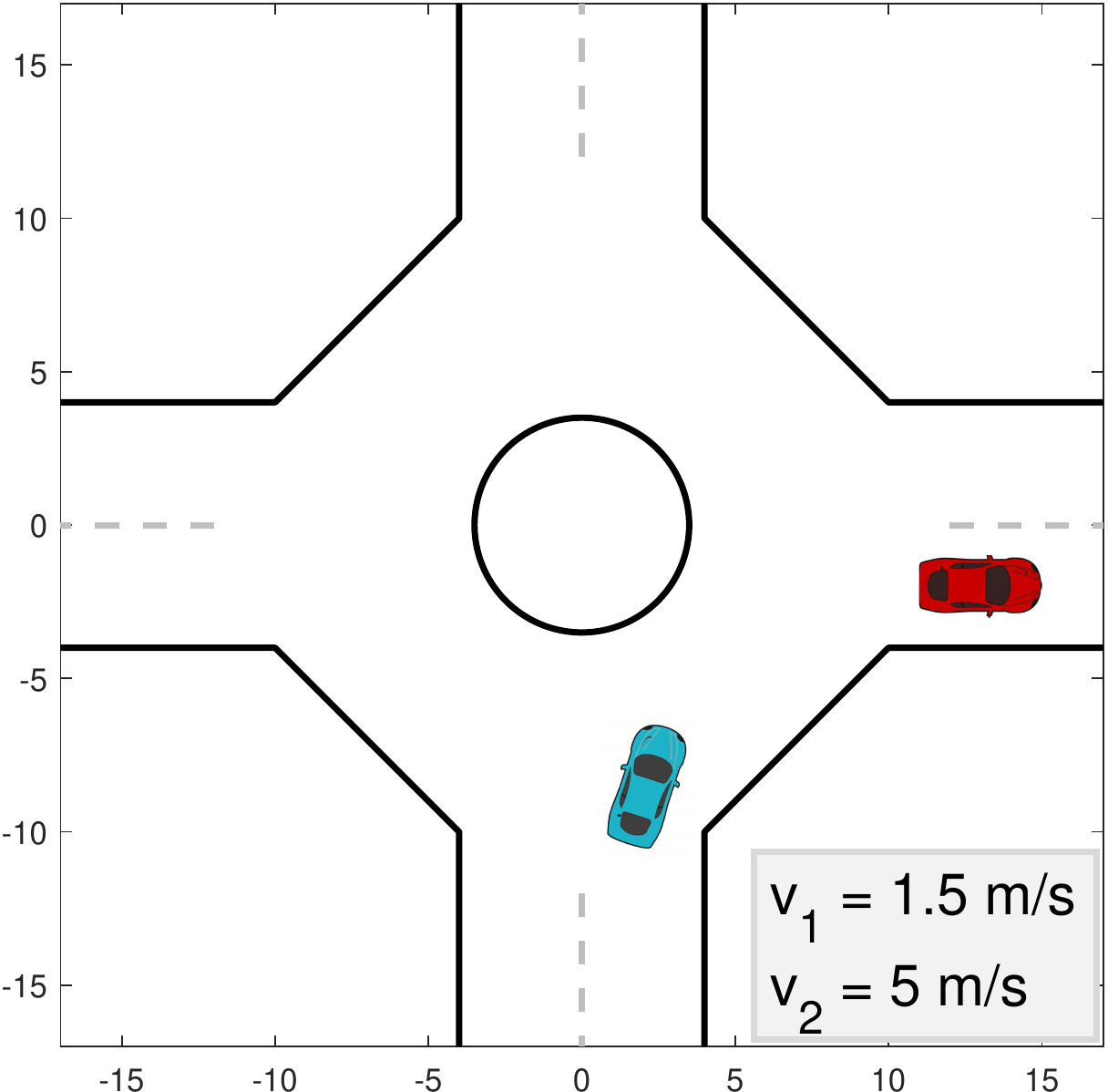, width = 0.33\linewidth, trim=0.6cm 0.4cm 0cm 0cm,clip}}

\put(  142,  0){\epsfig{file=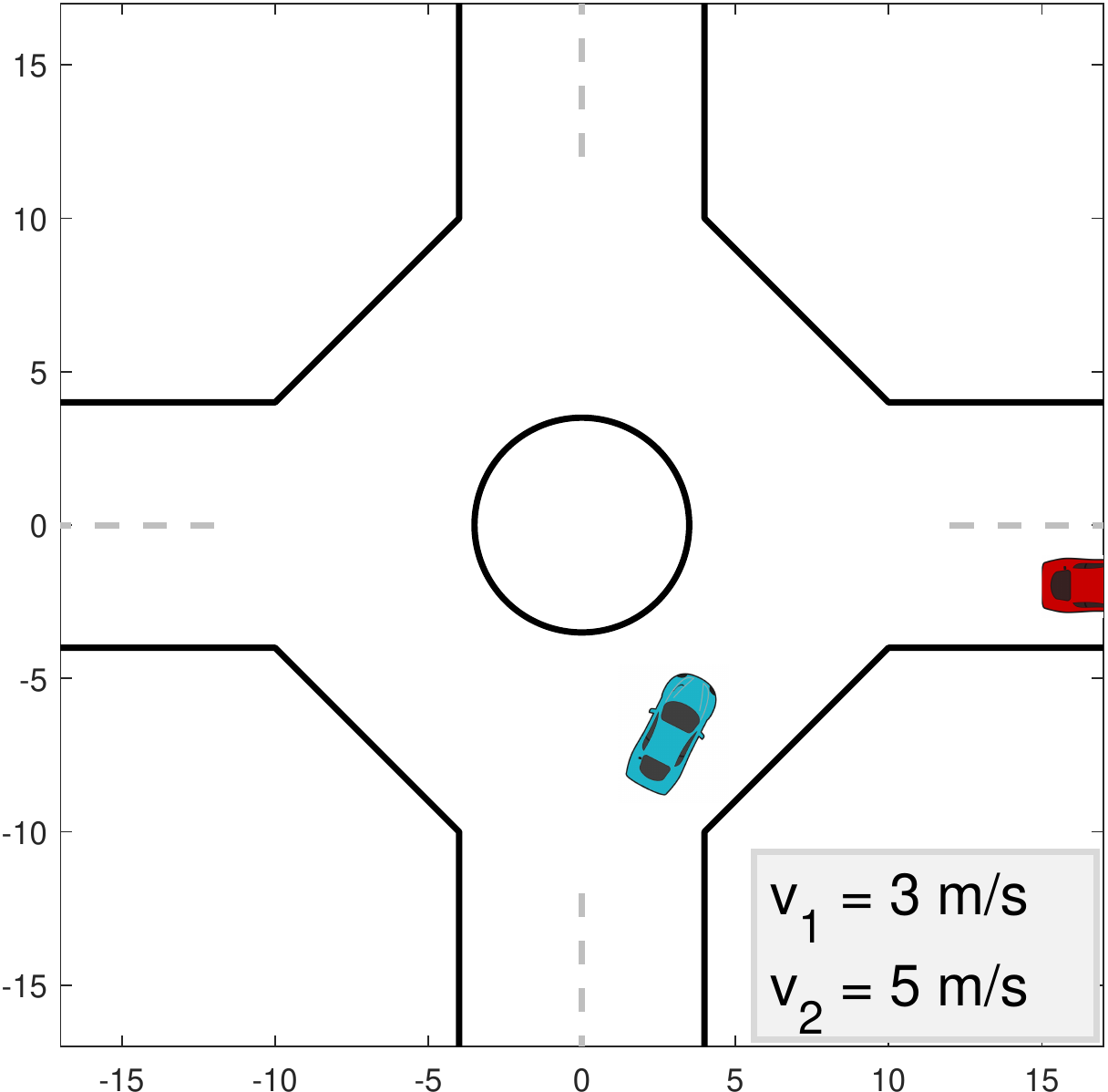, width = 0.33\linewidth, trim=0.6cm 0.4cm 0cm 0cm,clip}}
%%%%%%%%%%%%%%%%%%%%%%
\small
\put(35,155){(a-1)}
\put(117,155){(a-2)}
\put(199,155){(a-3)}
\put(35,70){(b-1)}
\put(117,70){(b-2)}
\put(199,70){(b-3)}
\normalsize
\end{picture}
\end{center}
      \caption{
      (a-1)-(a-3) show three sequential steps in a simulation where the blue vehicle controlled by $\hat{\pi}_{\text{k}}$ trained using standard supervised learning fails in making an adequate right turn to get around the central island of a roundabout; (b-1)-(b-3) show steps in a similar simulation where the blue vehicle controlled by $\hat{\pi}_{\text{k}}$ trained using DAgger succeeds in making a proper right turn.}
      \label{fig: dagger_compare}
\end{figure}
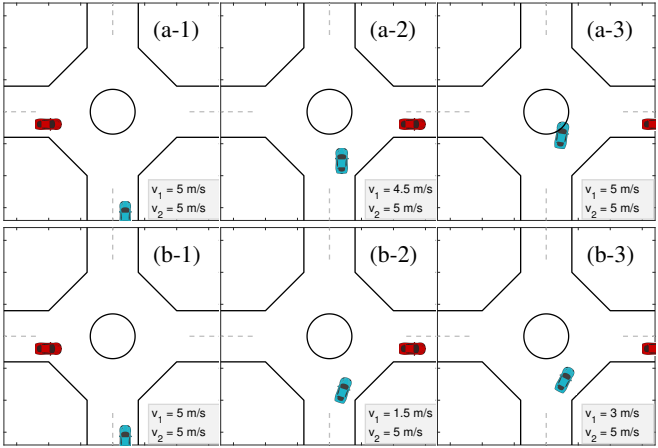

To show the advantage of using the DAgger algorithm \eqref{equ:IL} over using a standard supervised learning procedure \eqref{equ:SL} to obtain the policy $\hat{\pi}_{\text{k}}$, we show a case observed in our simulations where the policy trained using standard supervised learning fails but the one trained using DAgger succeeds. In Fig.~\ref{fig: dagger_compare}(a-3), the blue vehicle controlled by $\hat{\pi}_{\text{k}}$ trained using standard supervised learning fails in making an adequate right turn to get around the central island. This is due to a significant error of $\hat{\pi}_{\text{k}}$ from $\pi_{\text{k}}$ at certain states encountered by the blue vehicle when entering the roundabout, and the encounter with such states results from the issue of error propagation in time that has been discussed in Section~\ref{sec: explicit method}. In contrast, the blue vehicle in Fig.~\ref{fig: dagger_compare}(b-3) controlled by $\hat{\pi}_{\text{k}}$ trained using DAgger succeeds in making a proper right turn, illustrating the effectiveness of DAgger in avoiding such an issue.

In what follows we show the interactions between level-$k$ vehicles at the four-way, T-shaped, and roundabout intersections. In particular, we let three vehicles be controlled by different level-$k$ policies and show how the traffic scenarios evolve differently depending on the different combinations of level-$k$ policies.

\begin{figure}[ht]
\begin{center}
\begin{picture}(200.0, 260.0)
%%%%%%%%%%%%%%%%%%%%%%%%%%%%%%
\put(  -22,  174){\epsfig{file=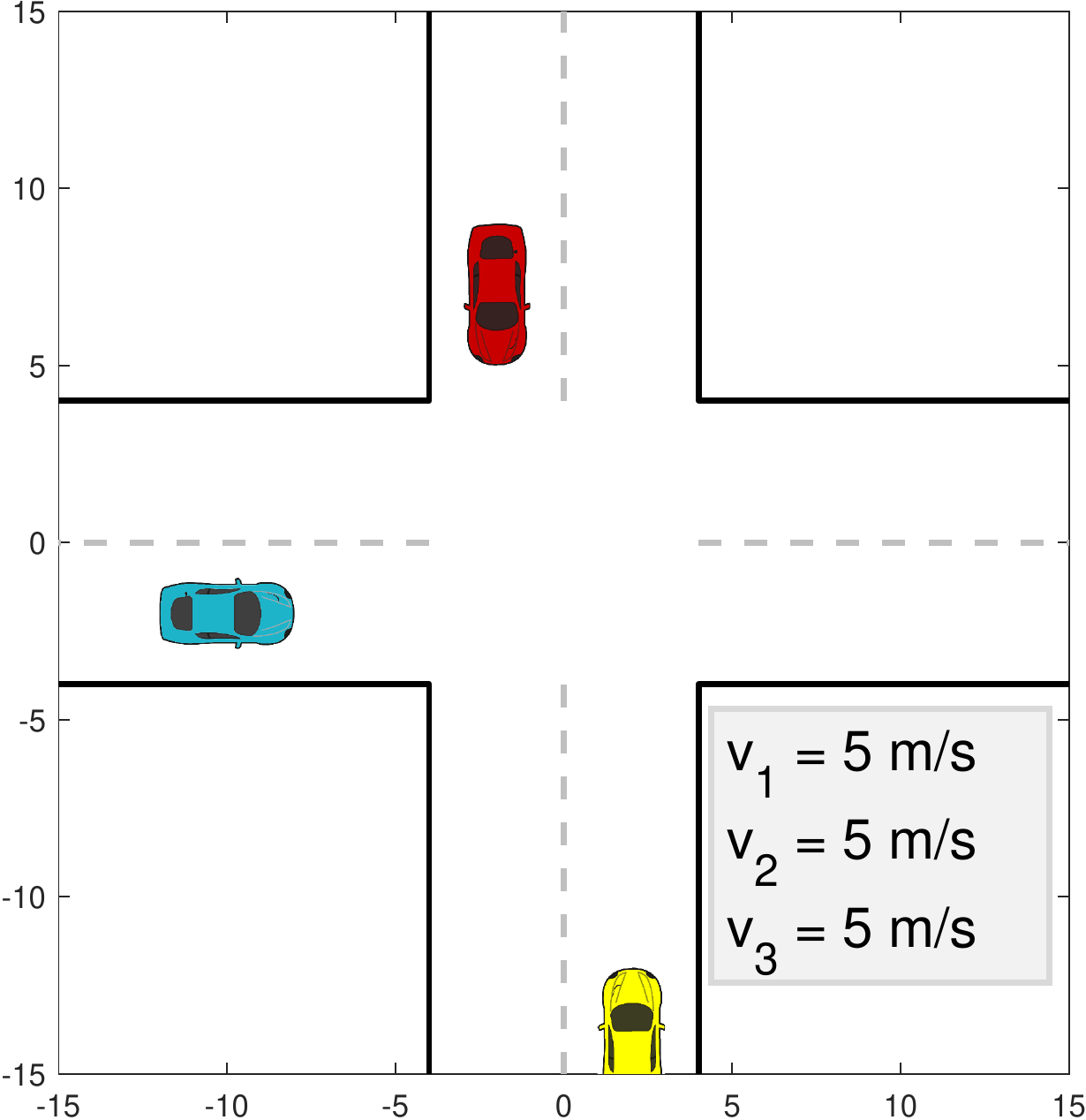,width = 0.33 \linewidth, trim=0.6cm 0.4cm 0cm 0cm,clip}}  %%%
\put(  60,  174){\epsfig{file=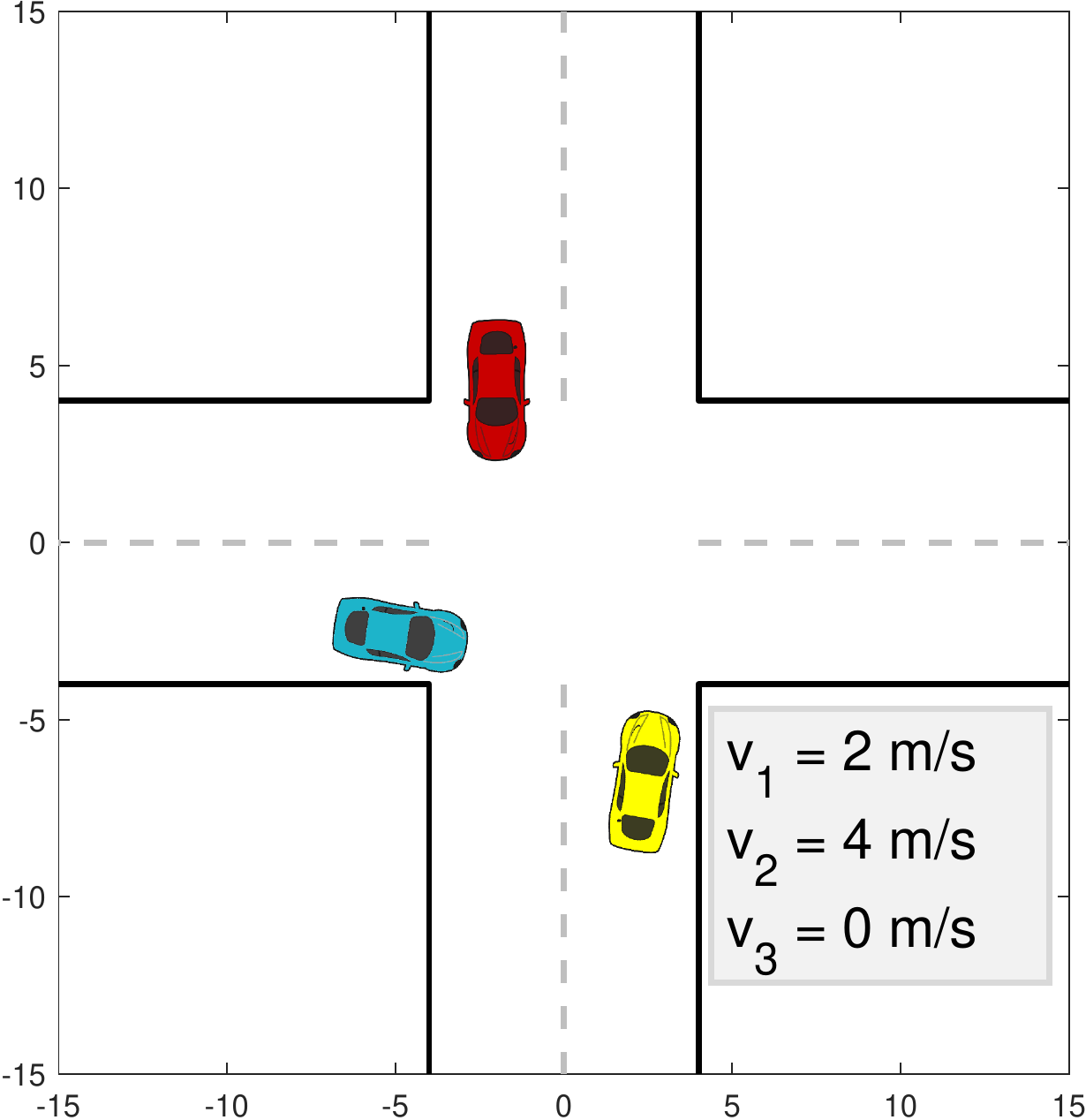, width = 0.33\linewidth, trim=0.6cm 0.4cm 0cm 0cm,clip}}

\put(  142,  174){\epsfig{file=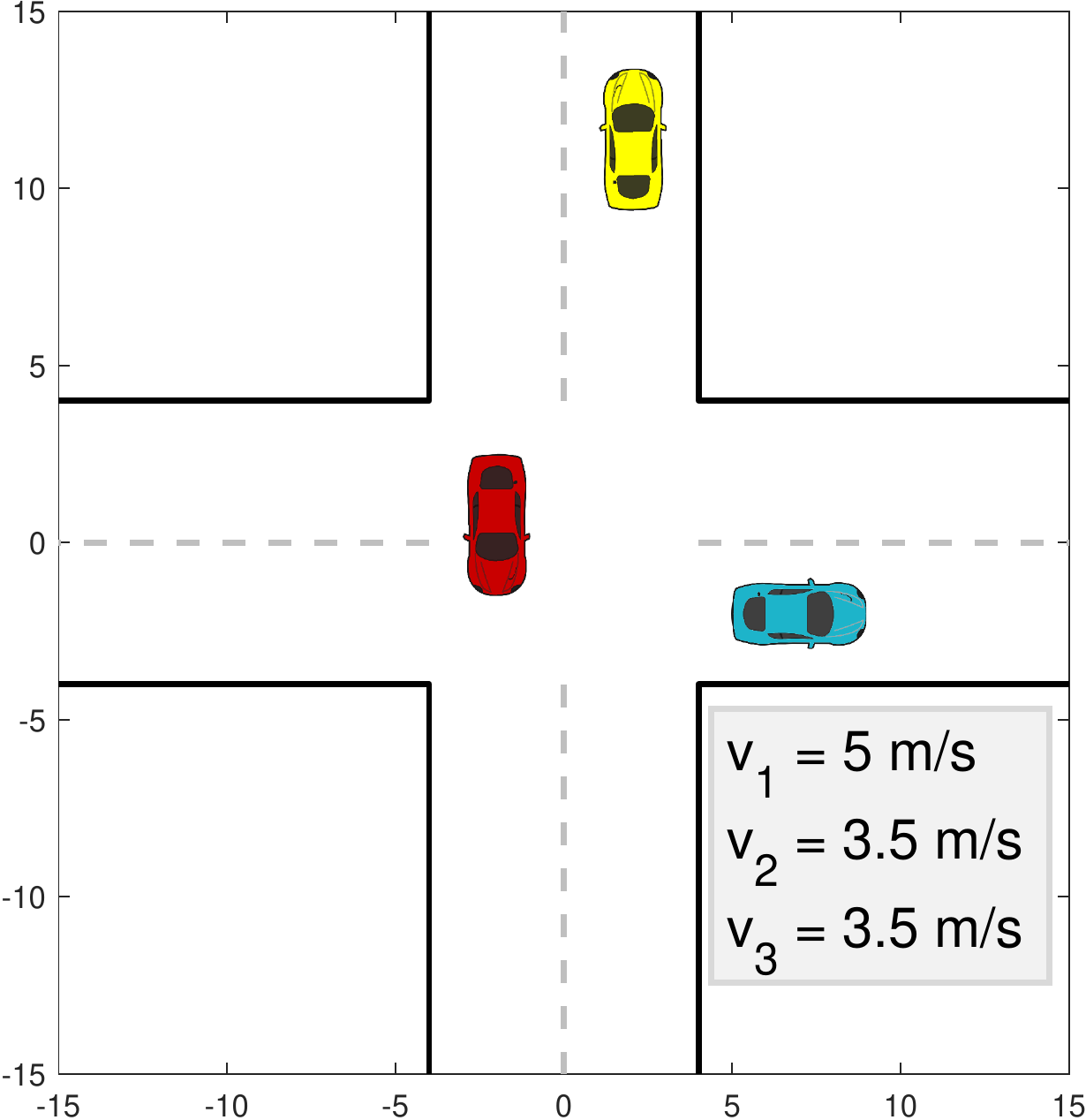, width = 0.33\linewidth, trim=0.6cm 0.4cm 0cm 0cm,clip}}
%%%%%%%%%%%%%%%%%%%%%%%%%%%%%%
\put(  -22, 87){\epsfig{file=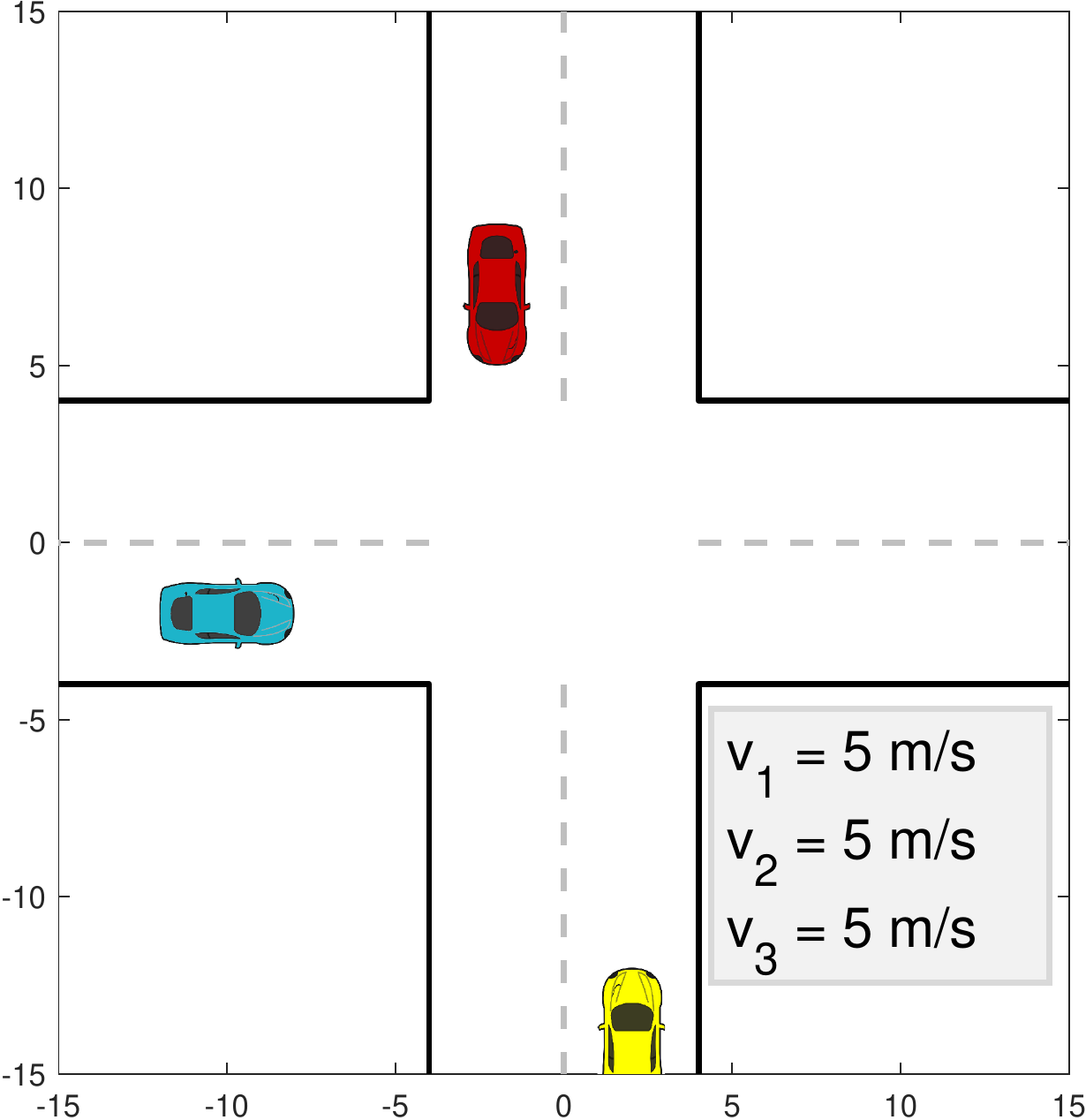,width = 0.33 \linewidth, trim=0.6cm 0.4cm 0cm 0cm,clip}}  %%%
\put(  60,  87){\epsfig{file=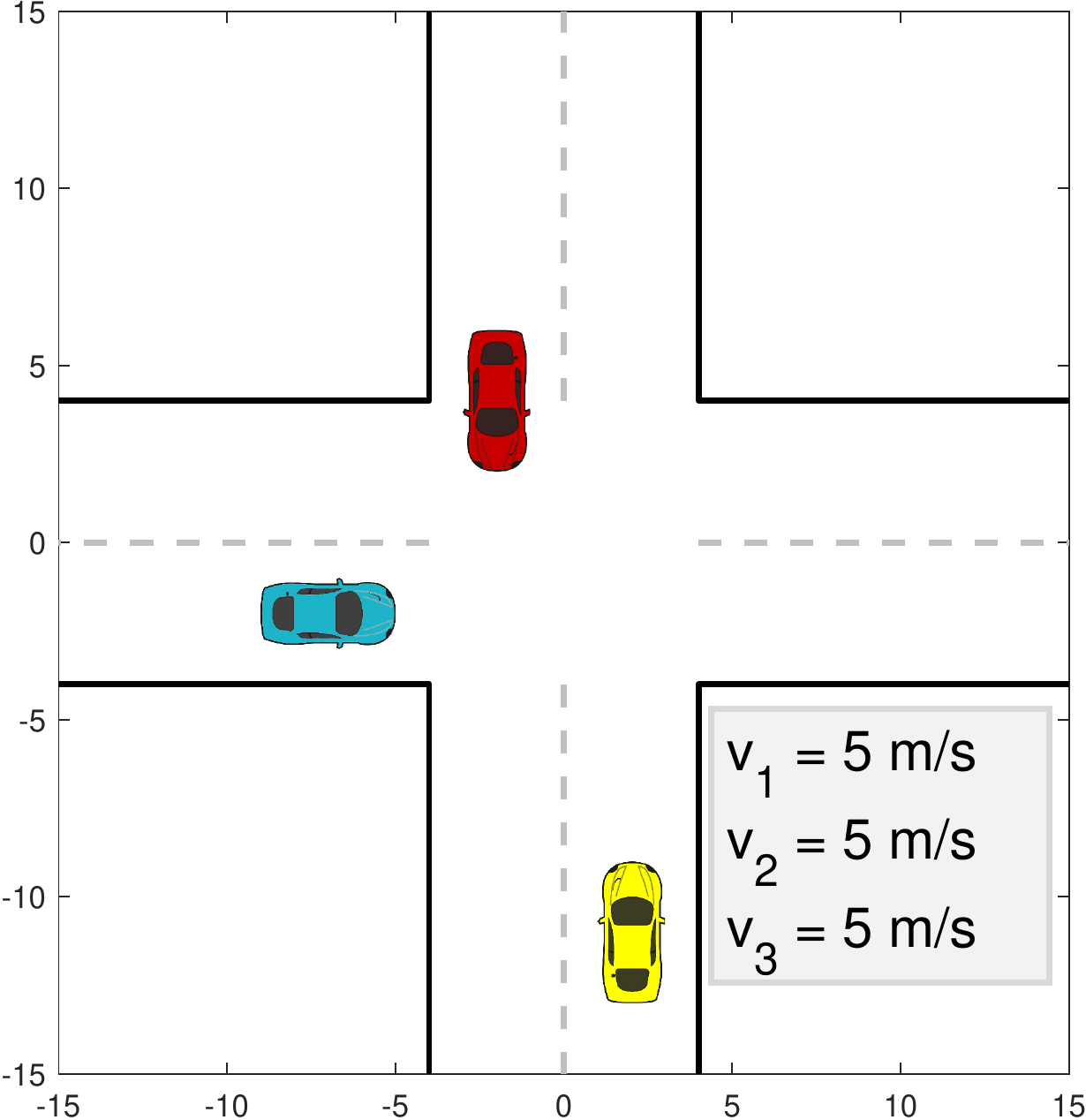, width = 0.33\linewidth, trim=0.6cm 0.4cm 0cm 0cm,clip}}

\put(  142, 87){\epsfig{file=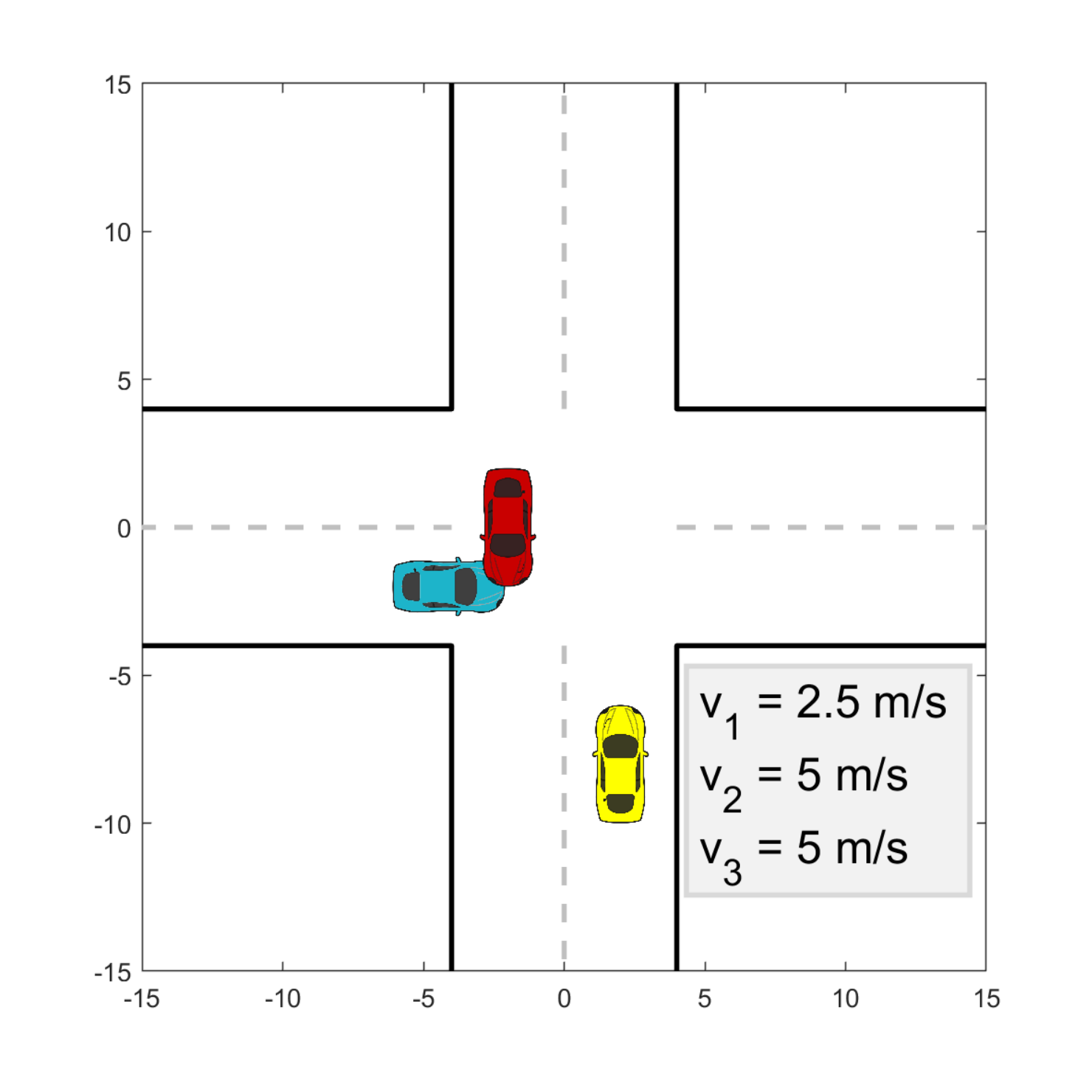, width = 0.338\linewidth, trim=1.8cm 1.5cm 1cm 1cm,clip}}

%%%%%%%%%%%%%%%%%%%%%%%%%%%%%%%%%%%%%%%%%%%%%
\put(  -22,  0){\epsfig{file=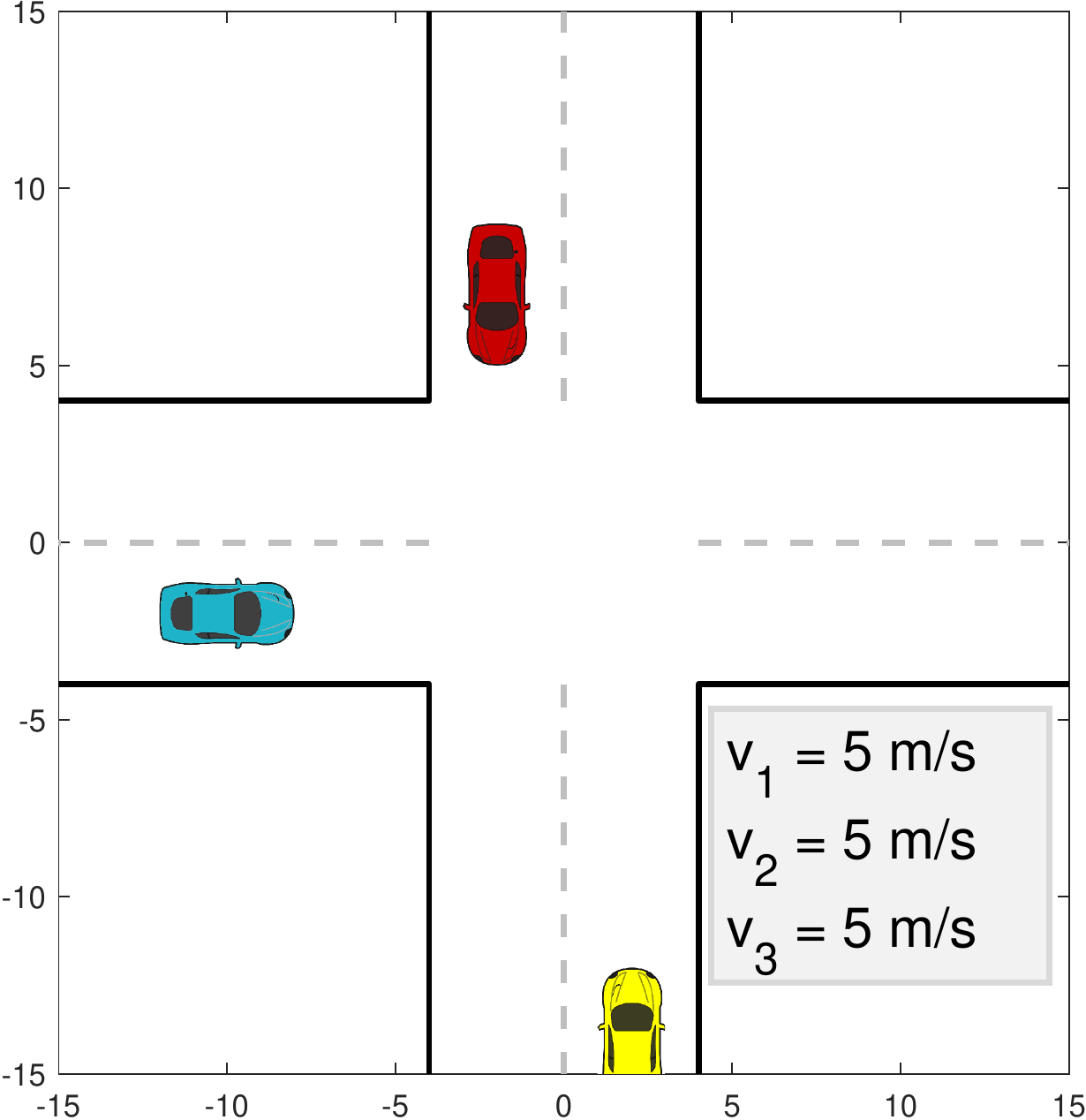,width = 0.33 \linewidth, trim=0.6cm 0.4cm 0cm 0cm,clip}}  %%%
\put(  60,  0){\epsfig{file=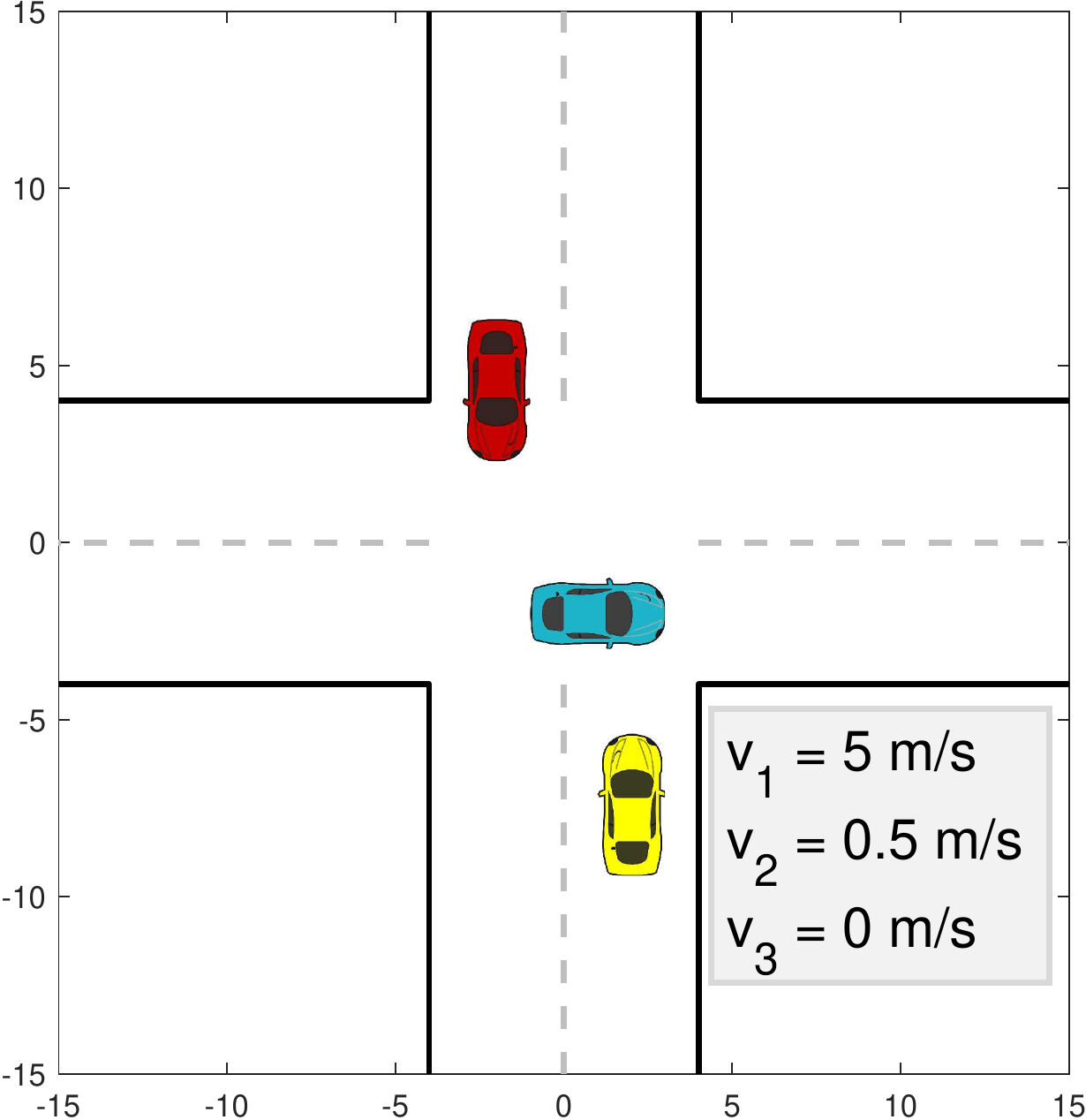, width = 0.33\linewidth, trim=0.6cm 0.4cm 0cm 0cm,clip}}
\put(  142,  0){\epsfig{file=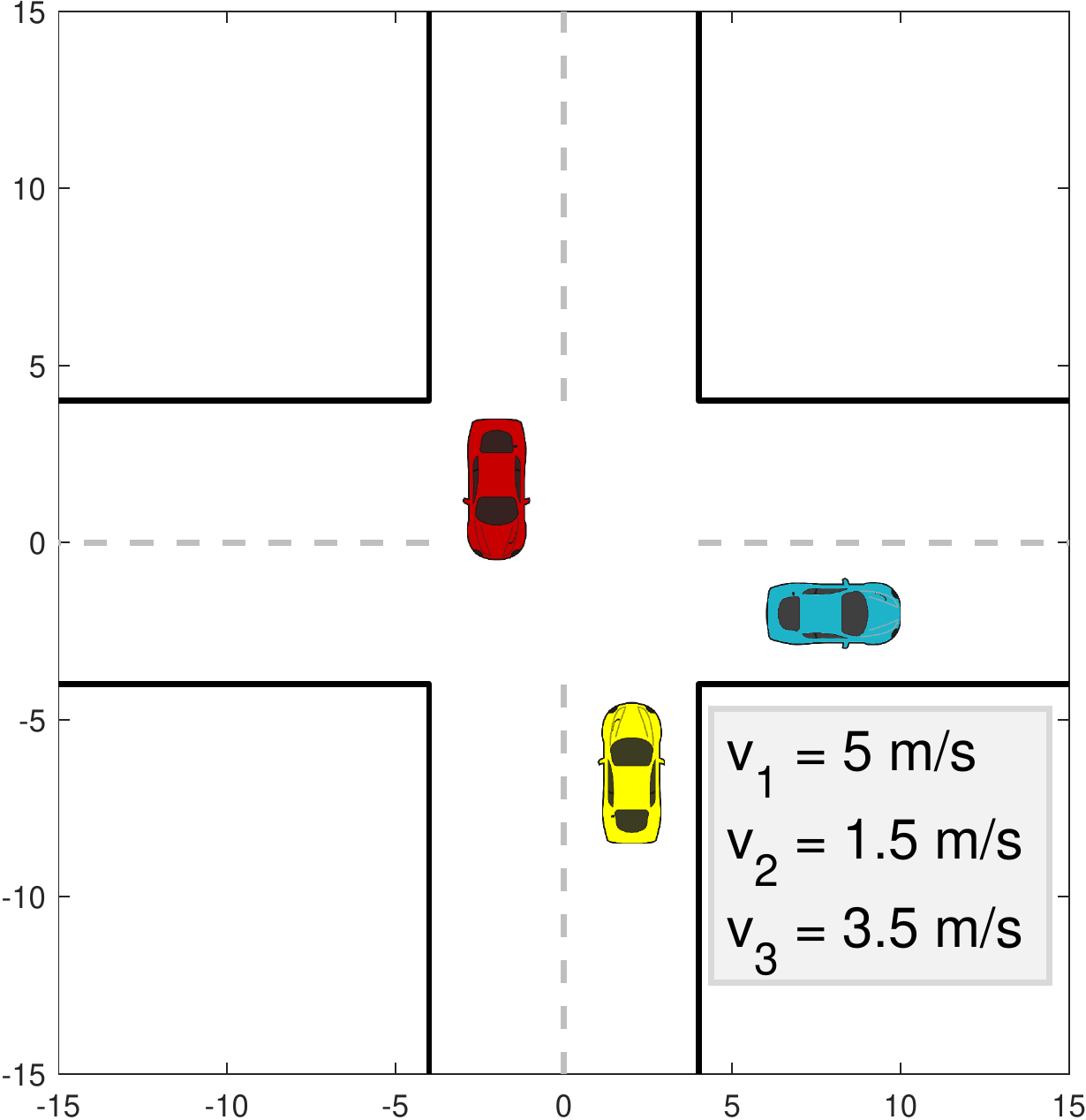, width = 0.33\linewidth, trim=0.6cm 0.4cm 0cm 0cm,clip}}
%%%%%%%%%%%%%%%%%%%%%%
\small
\put(35,246){(a-1)}
\put(117,246){(a-2)}
\put(199,246){(a-3)}
\put(35,159){(b-1)}
\put(117,159){(b-2)}
\put(199,159){(b-3)}
\put(35,72){(c-1)}
\put(117,72){(c-2)}
\put(199,72){(c-3)}
\normalsize
\end{picture}
\end{center}
       \caption{Interactions of level-$k$ vehicles at the four-way intersection. (a-1)-(a-3) show three sequential steps in a simulation where three level-$1$ vehicles interact with each other; (b-1)-(b-3) show steps of three level-$2$ vehicles interacting with each other; (c-1)-(c-3) show steps of a level-$2$ vehicle (blue) interacting with two level-$1$ vehicles (yellow and red); $v_1$, $v_2$ and $v_3$ are the speeds of the blue, yellow and red vehicles, respectively.}
      \label{fig: fourway-level-k}
\end{figure}
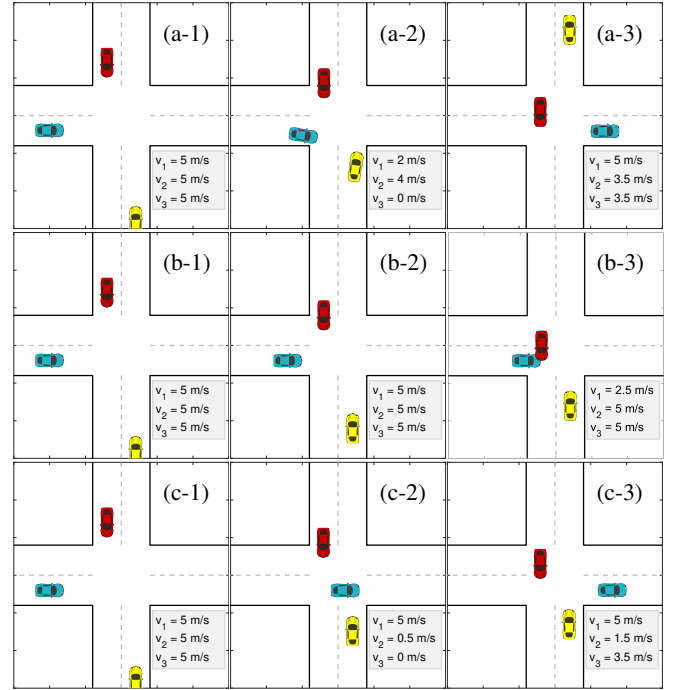

\begin{figure}[ht]
\begin{center}
\begin{picture}(200.0, 220.0)
%%%%%%%%%%%%%%%%%%%%%%%%%%%%%%
\put(  -22,  140){\epsfig{file=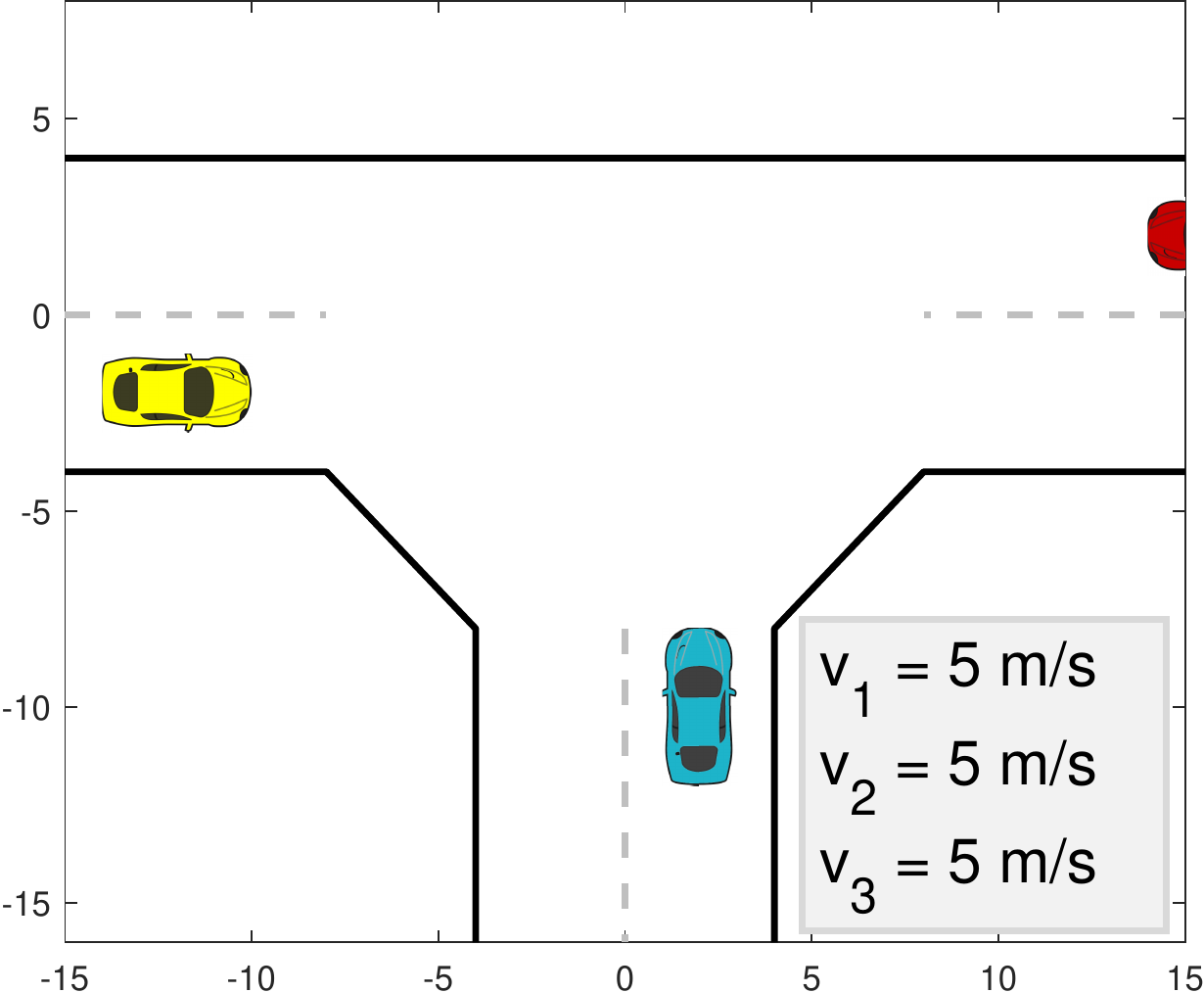,width = 0.33 \linewidth, trim=0.6cm 0.4cm 0cm 0cm,clip}}  %%%
\put(  60,  140){\epsfig{file=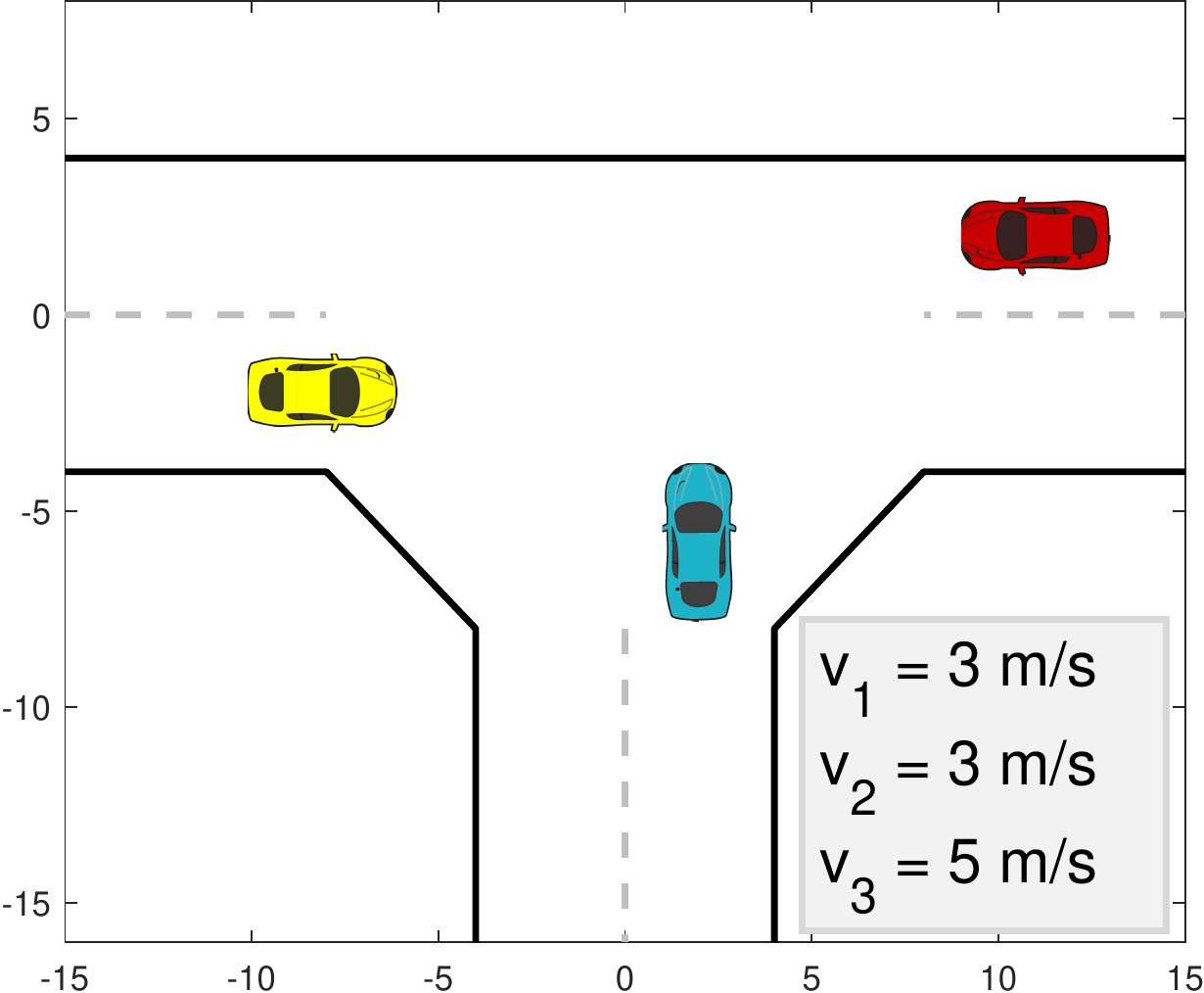, width = 0.33\linewidth, trim=0.6cm 0.4cm 0cm 0cm,clip}}

\put(  142,  140){\epsfig{file=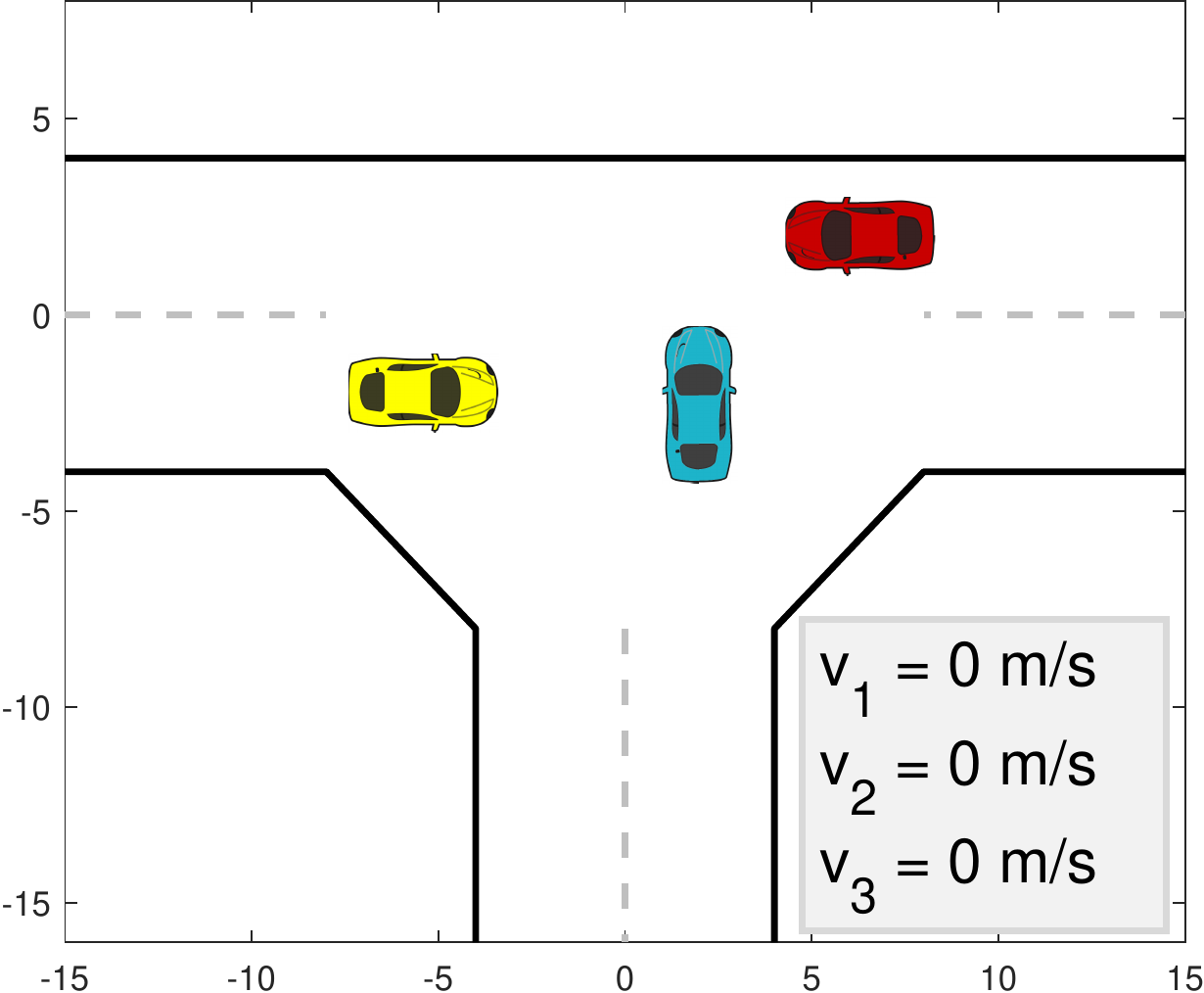, width = 0.33\linewidth, trim=0.6cm 0.4cm 0cm 0cm,clip}}

%%%%%%%%%%%%%%%%%%%%%%%%%%%%%%%%%%%%%%%%%%%%%
\put(  -22,  70){\epsfig{file=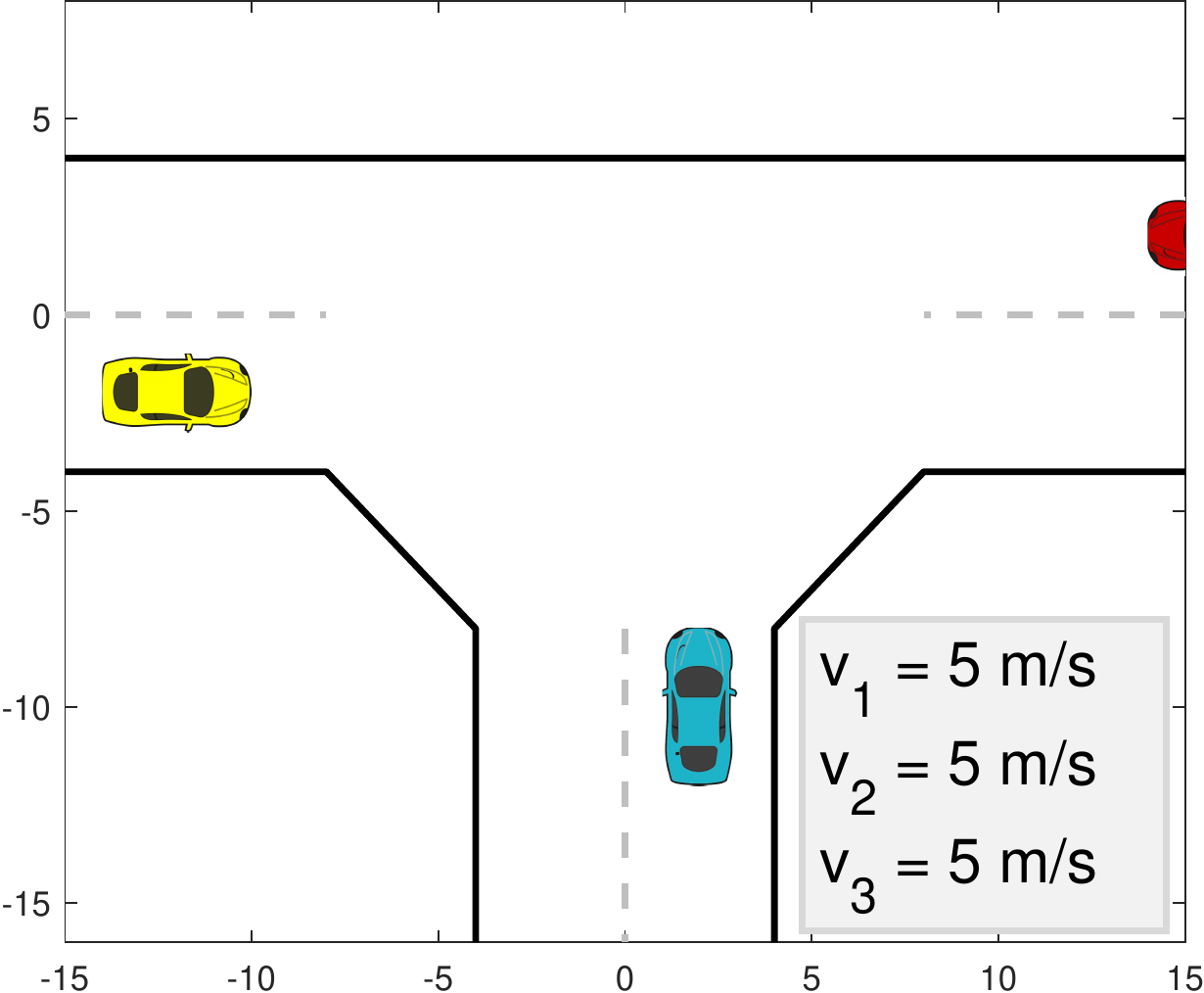,width = 0.33 \linewidth, trim=0.6cm 0.4cm 0cm 0cm,clip}}  %%%
\put(  60,  70){\epsfig{file=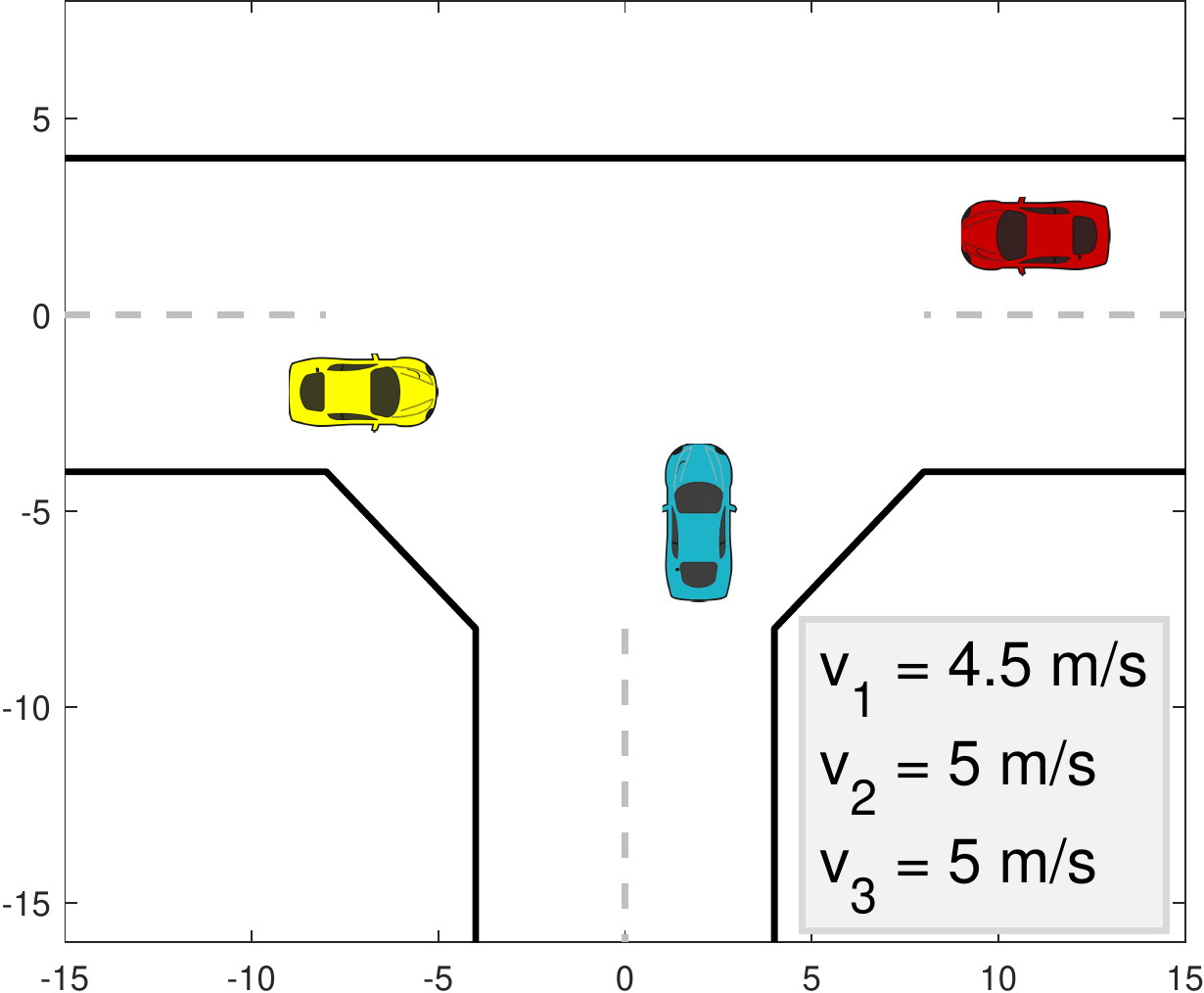, width = 0.33\linewidth, trim=0.6cm 0.4cm 0cm 0cm,clip}}
\put(  142,  71){\epsfig{file=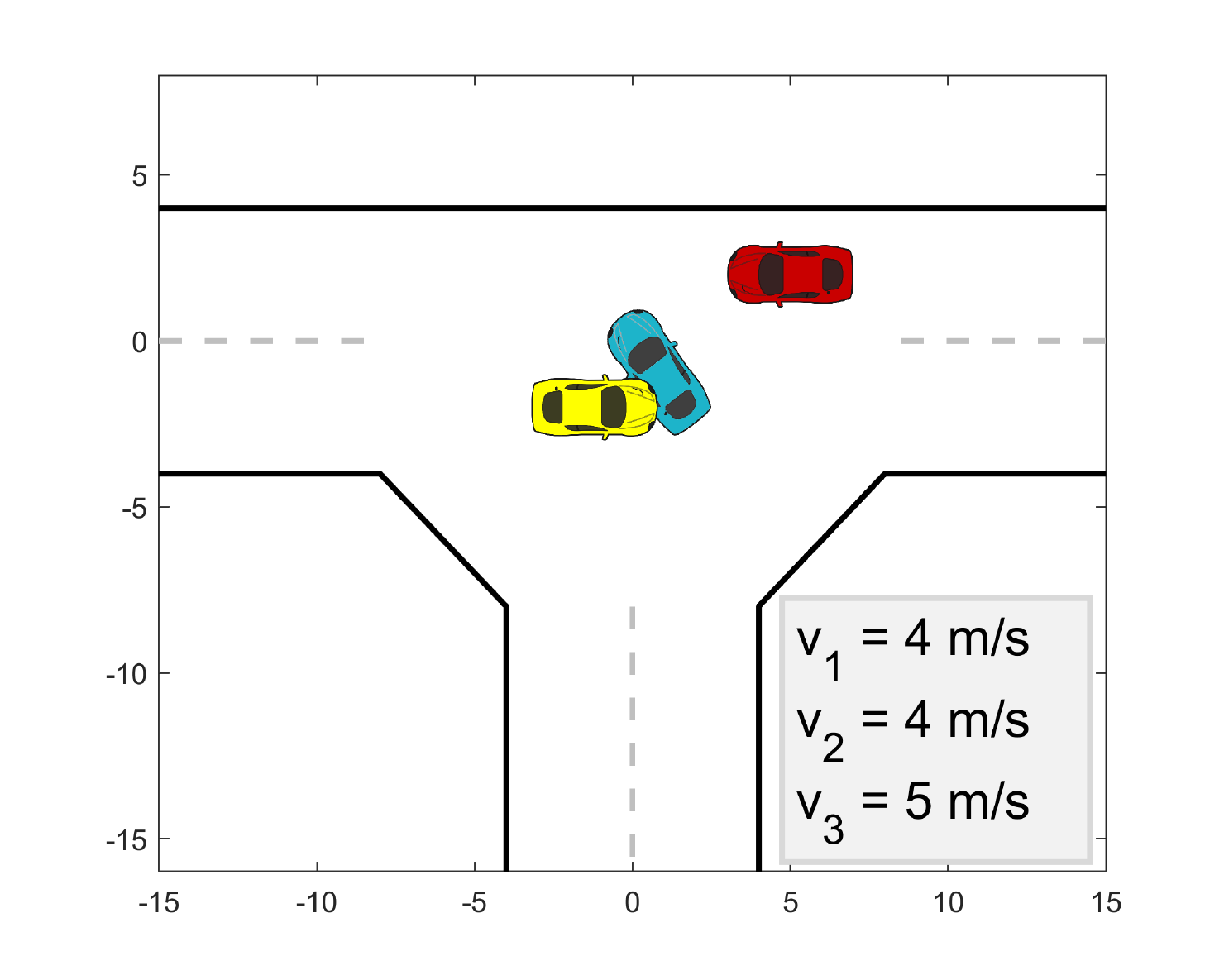, width = 0.337\linewidth, trim=1.8cm 1.3cm 1cm 0.8cm,clip}}
%%%%%%%%%%%%%%%%%%%%%%
\put(  -22,  0){\epsfig{file=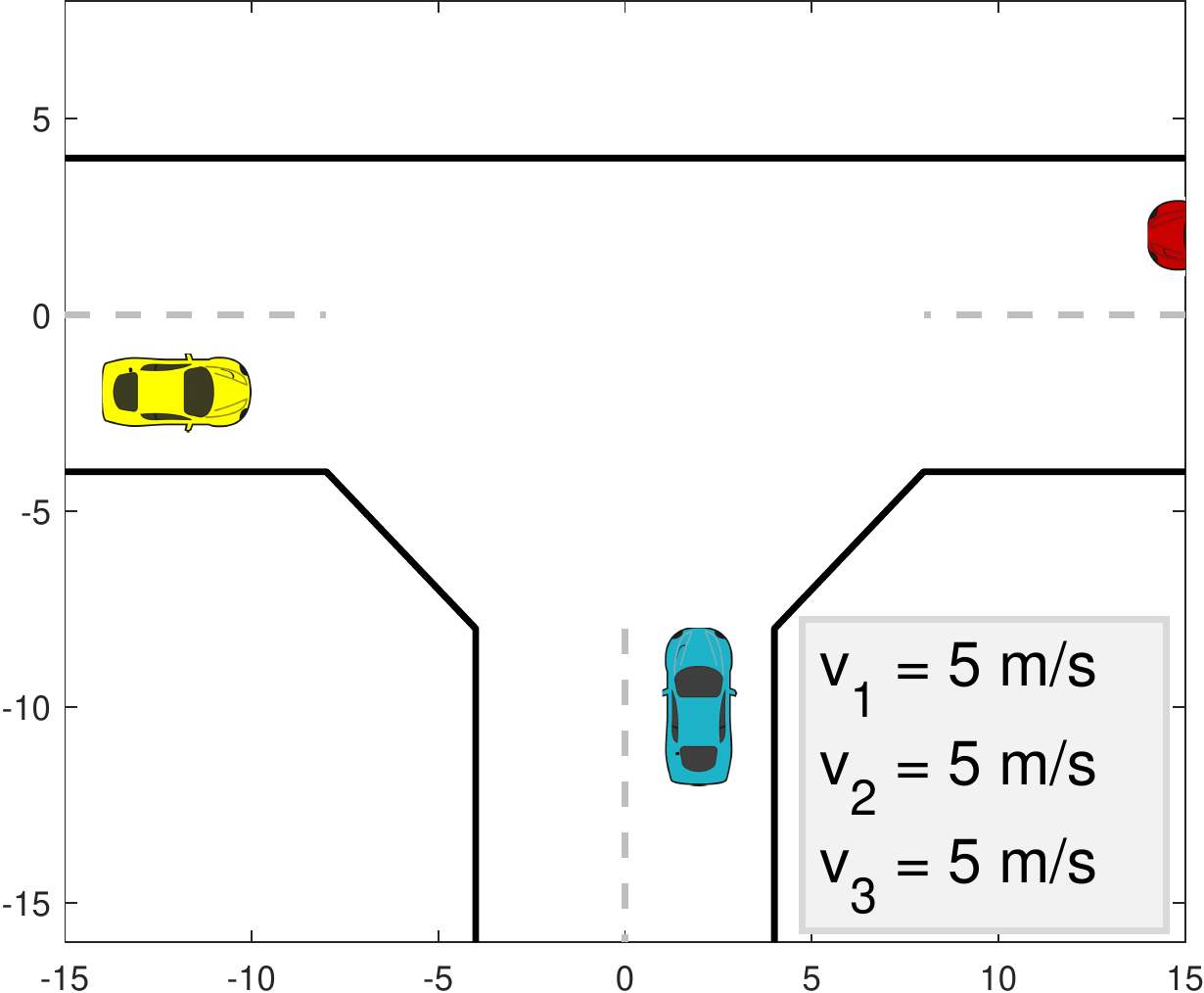,width = 0.33 \linewidth, trim=0.6cm 0.4cm 0cm 0cm,clip}}  %%%
\put(  60,  0){\epsfig{file=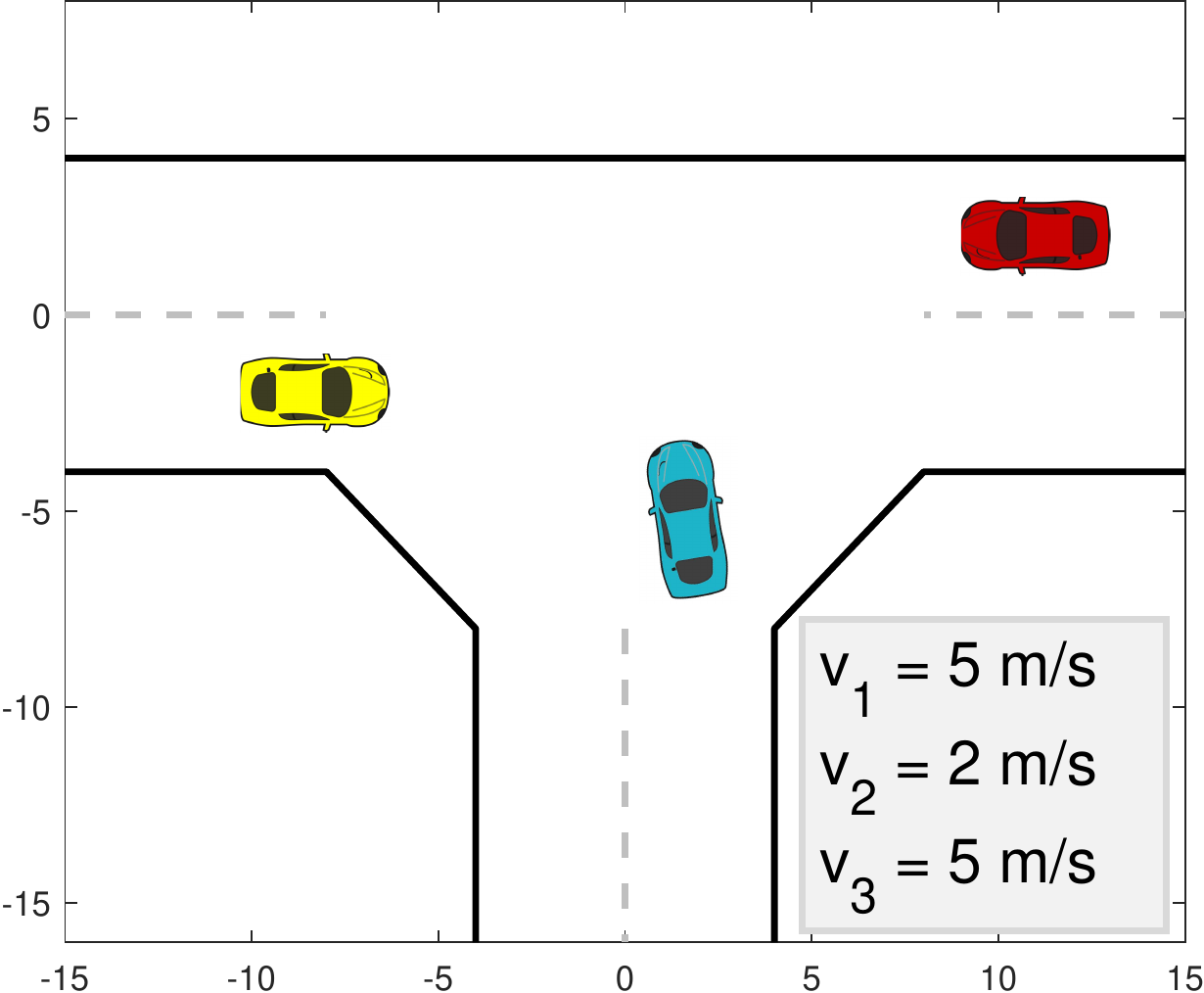, width = 0.33\linewidth, trim=0.6cm 0.4cm 0cm 0cm,clip}}
\put(  142,  0){\epsfig{file=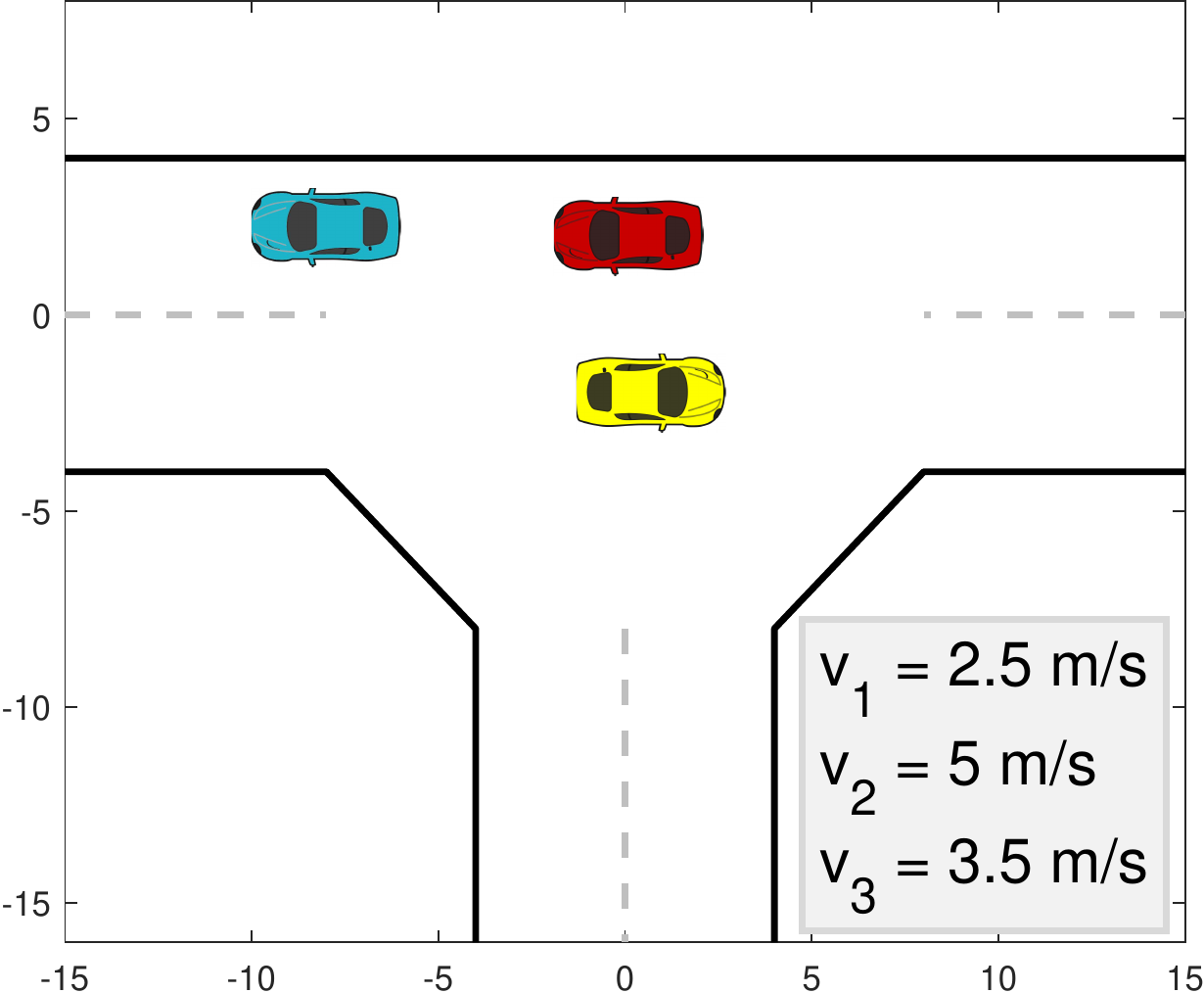, width = 0.33\linewidth, trim=0.6cm 0.4cm 0cm 0cm,clip}}
\small
\put(35,200){(a-1)}
\put(117,200){(a-2)}
\put(199,200){(a-3)}
\put(35,130){(b-1)}
\put(117,130){(b-2)}
\put(199,130){(b-3)}
\put(35,60){(c-1)}
\put(117,60){(c-2)}
\put(199,60){(c-3)}
\normalsize
\end{picture}
\end{center}
       \caption{Interactions of level-$k$ vehicles at the T-shaped intersection. (a-1)-(a-3) show three sequential steps in a simulation where three level-$1$ vehicles interact with each other; (b-1)-(b-3) show steps of three level-$2$ vehicles interacting with each other; (c-1)-(c-3) show steps of a level-$2$ vehicle (blue) interacting with two level-$1$ vehicles (yellow and red); $v_1$, $v_2$ and $v_3$ are the speeds of the blue, yellow and red vehicles, respectively.}
      \label{fig: threeway-level-k}
\end{figure}

\begin{figure}[ht]
\begin{center}
\begin{picture}(200.0, 264.0)
\put(  -22,  170){\epsfig{file=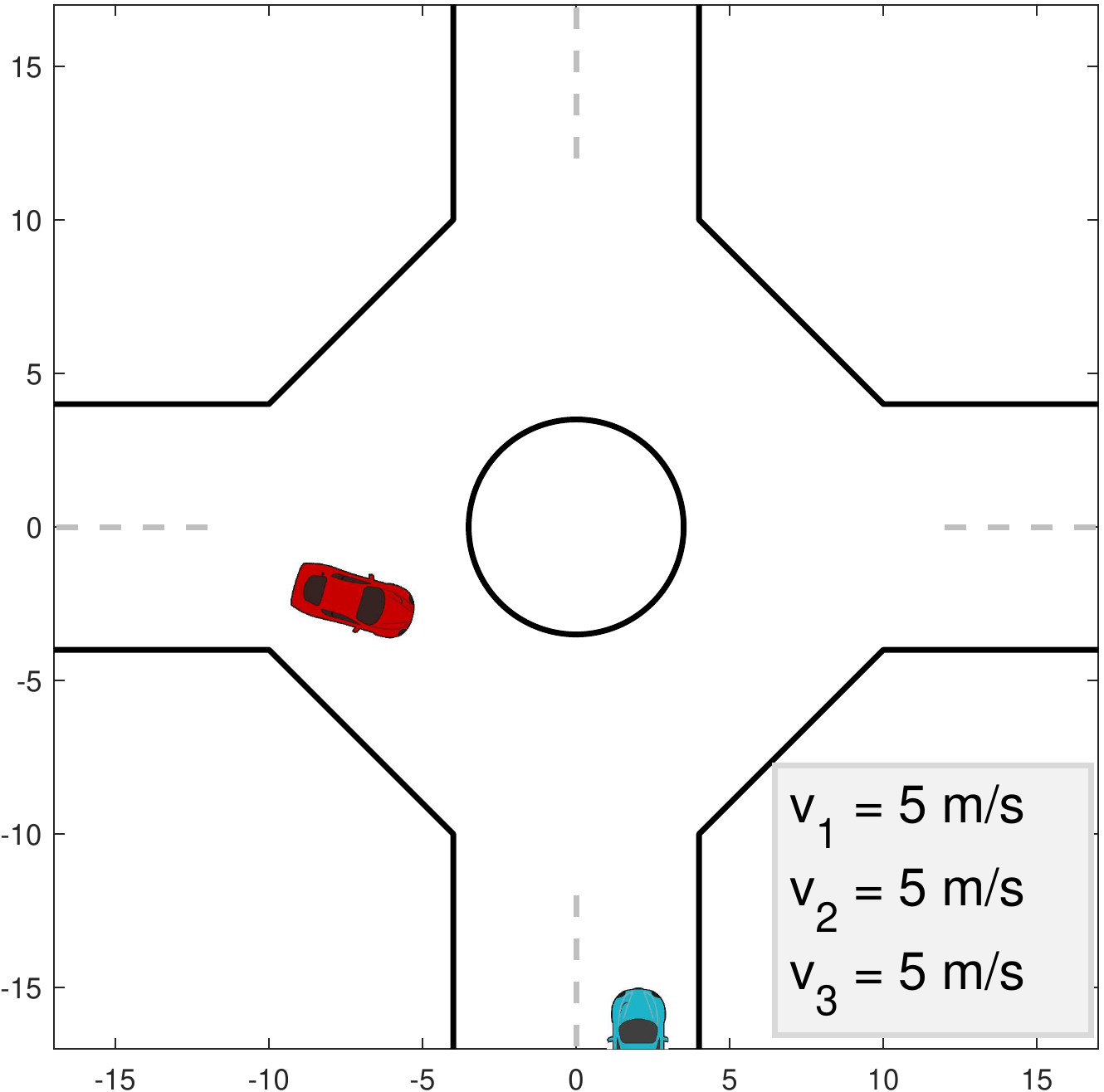,width = 0.33 \linewidth, trim=0.6cm 0.4cm 0cm 0cm,clip}}  %%%
\put(  60,  170){\epsfig{file=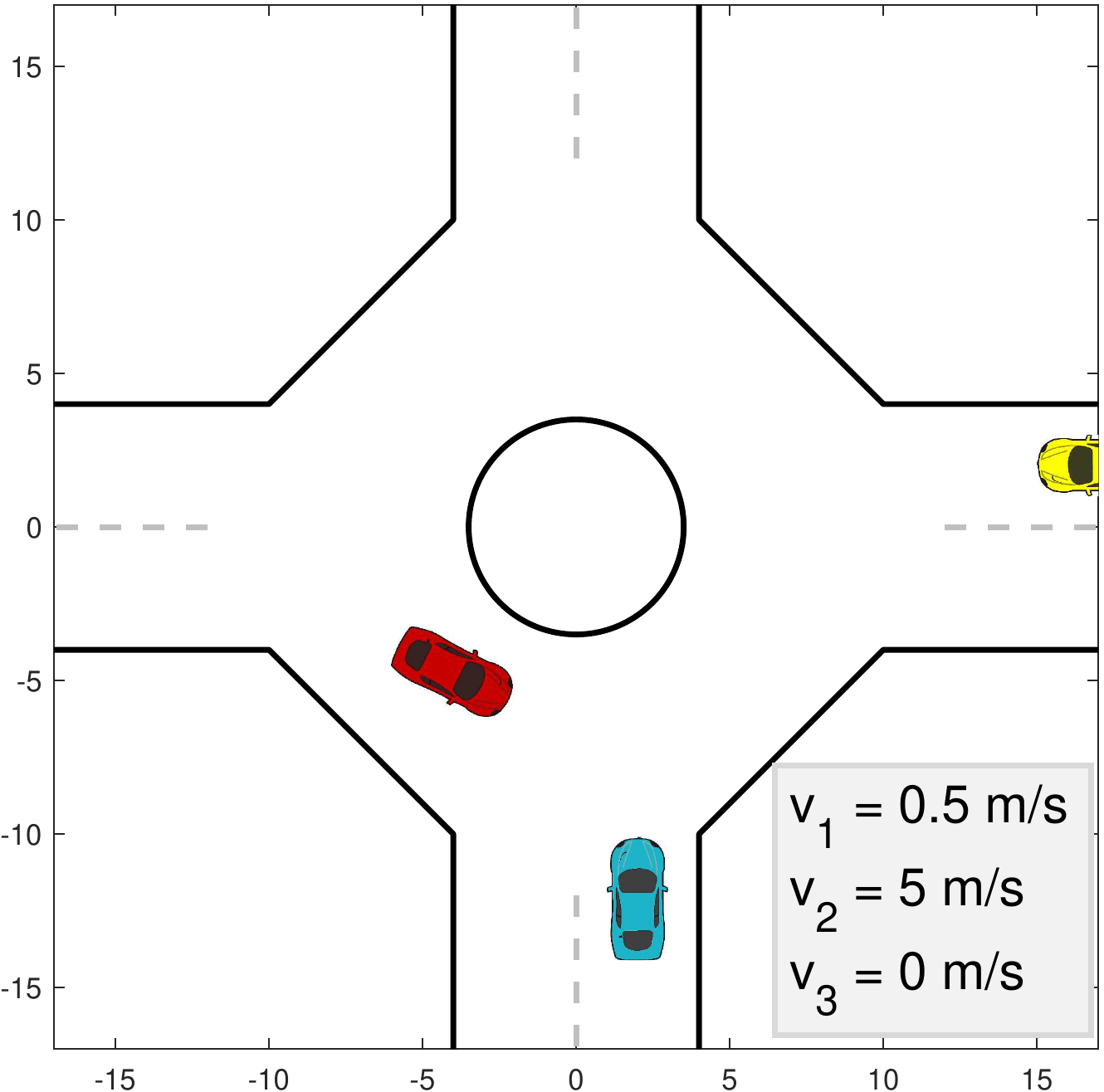, width = 0.33\linewidth, trim=0.6cm 0.4cm 0cm 0cm,clip}}
\put(  142,  169.97){\epsfig{file=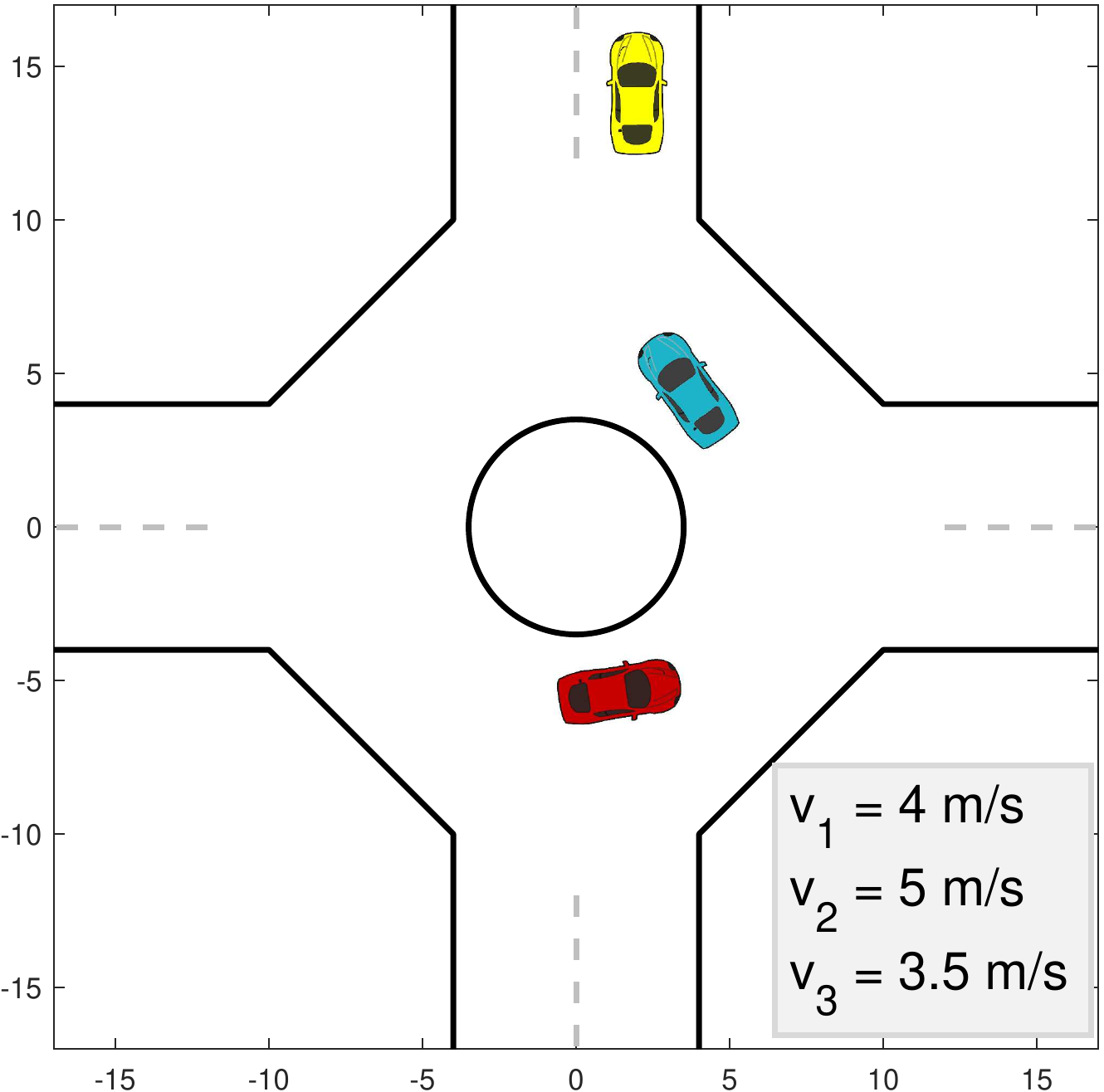, width = 0.33\linewidth, trim=0.6cm 0.4cm 0cm 0cm,clip}}
%%%%%%%%%%%%%%%%%%%%%%%%%%%%%%%%%%%%%%%%%
\put(  -22,  85){\epsfig{file=media/roundabout_2_2_2_step_9.pdf,width = 0.33 \linewidth, trim=0.6cm 0.4cm 0cm 0cm,clip}}  %%%
\put(  60,  85){\epsfig{file=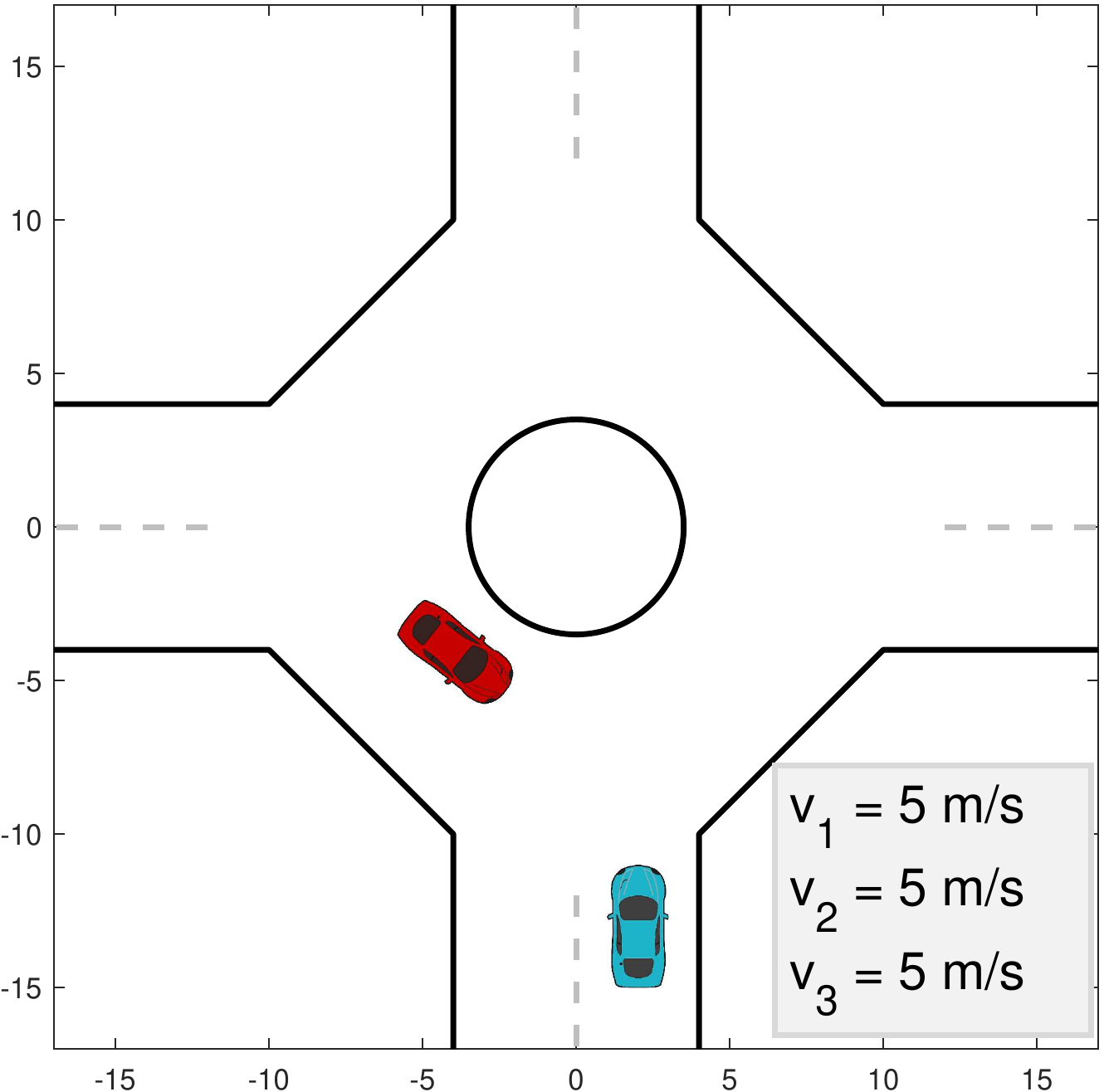, width = 0.33\linewidth, trim=0.6cm 0.4cm 0cm 0cm,clip}}
\put(  141.8,  85){\epsfig{file=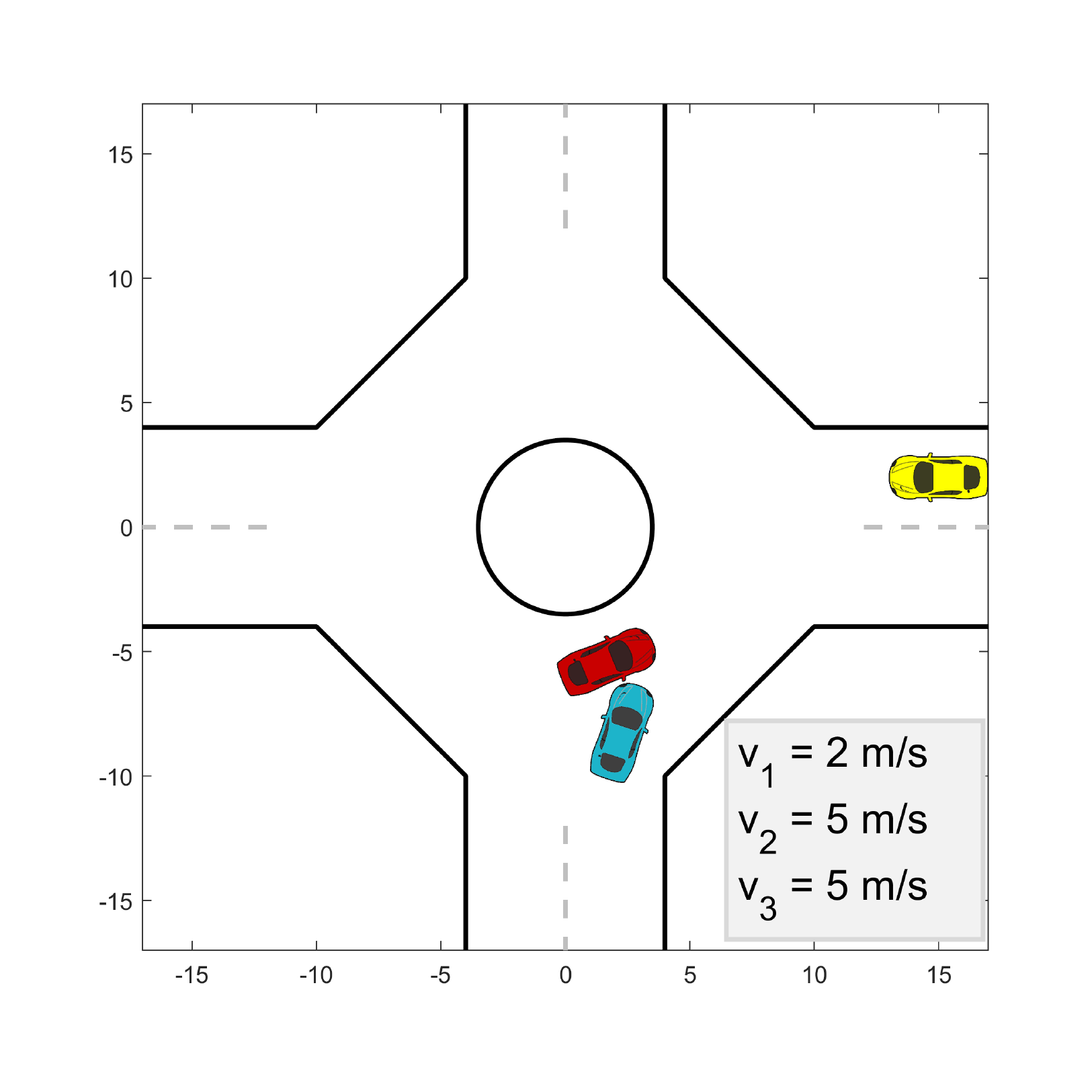, width = 0.344\linewidth, trim=2.05cm 2cm 1cm 1.2cm,clip}}
%%%%%%%%%%%%%%%%%%%%%%%%%%%%%%
\put(  -22,  0){\epsfig{file=media/roundabout_2_2_2_step_9.pdf,width = 0.33 \linewidth, trim=0.6cm 0.4cm 0cm 0cm,clip}}  %%%
\put(  60,  0){\epsfig{file=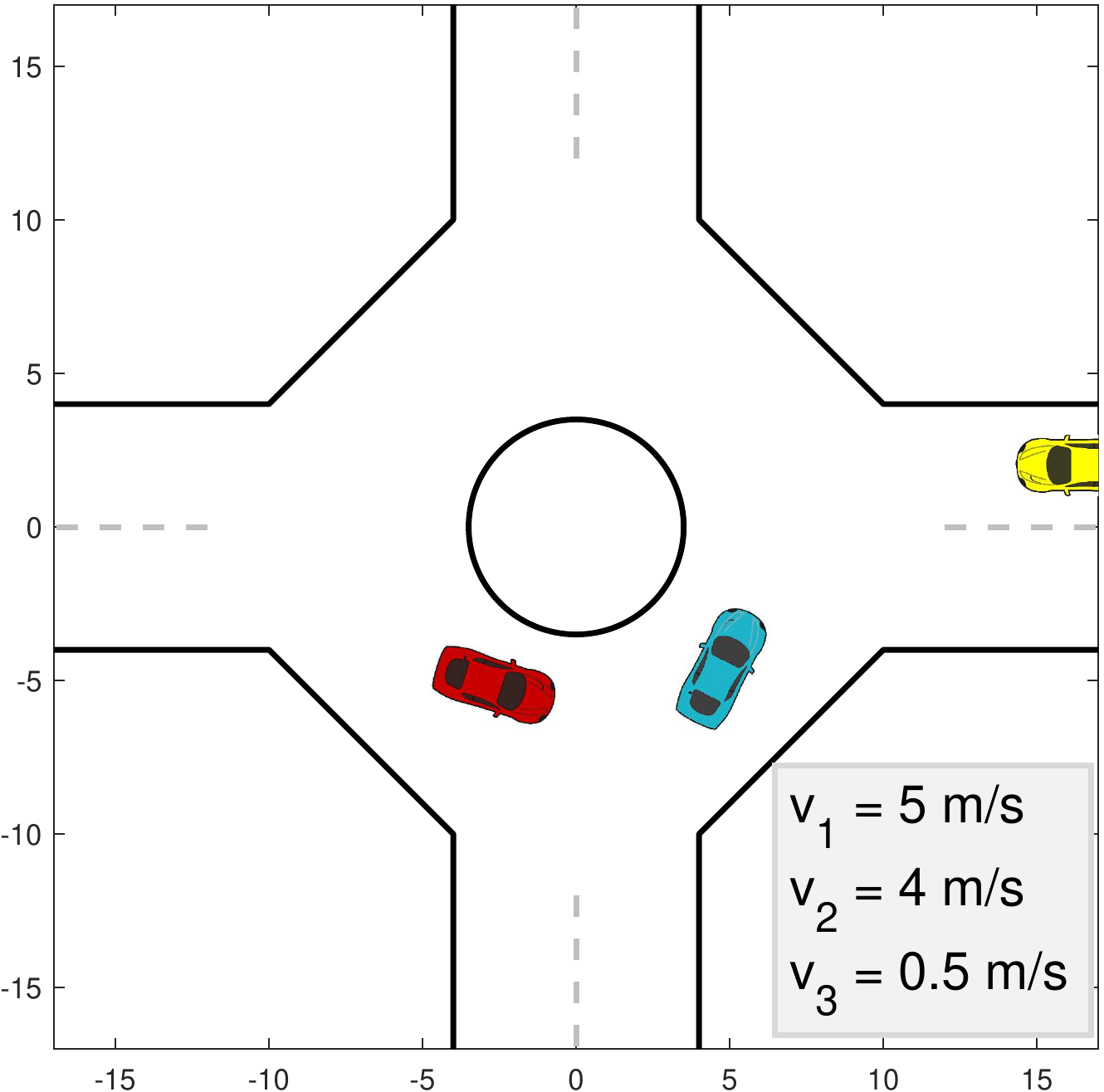, width = 0.33\linewidth, trim=0.6cm 0.4cm 0cm 0cm,clip}}

\put(  142,  0){\epsfig{file=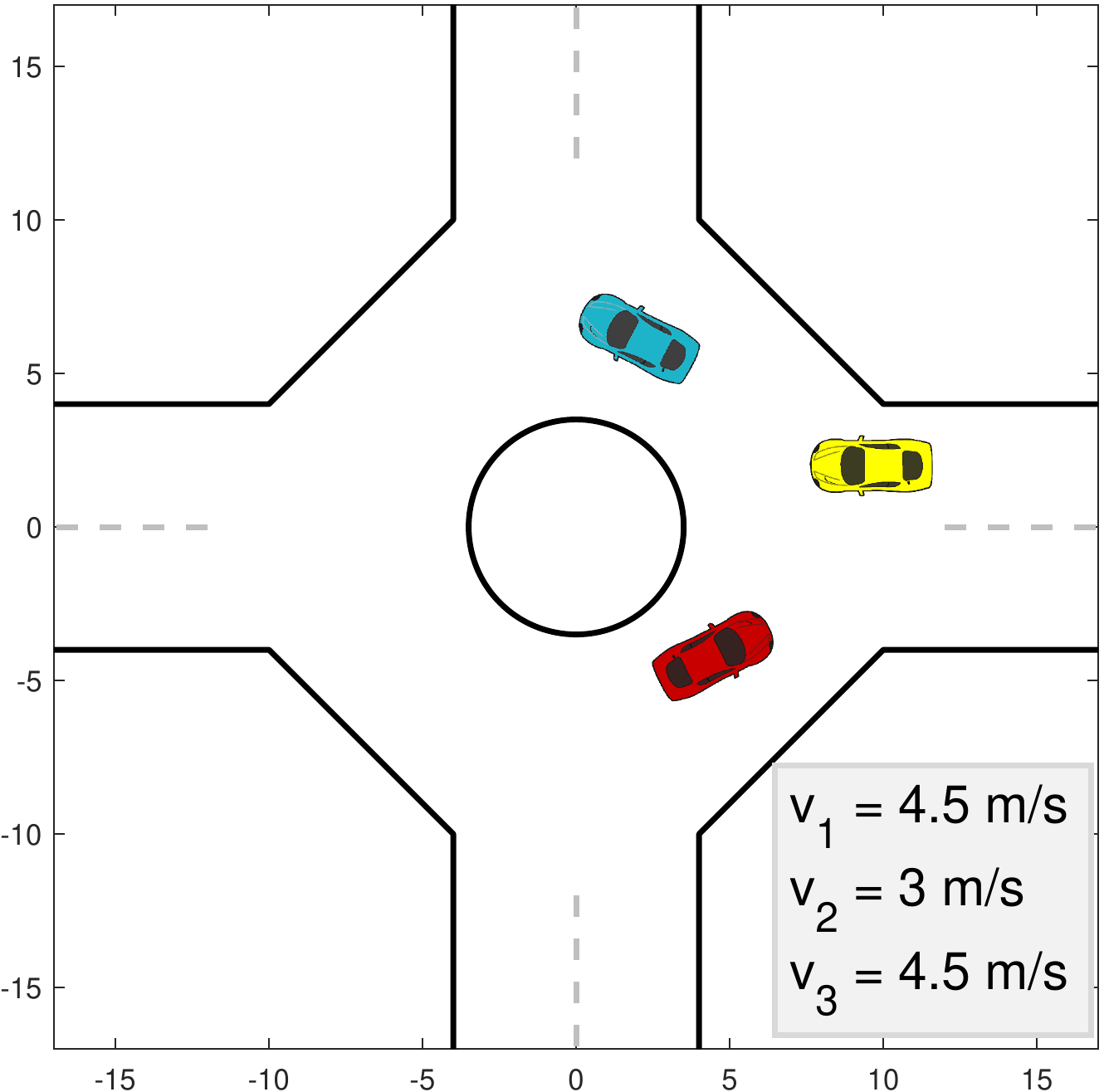, width = 0.33\linewidth, trim=0.6cm 0.4cm 0cm 0cm,clip}}

%%%%%%%%%%%%%%%%%%%%%%
\small
\put(35,240){(a-1)}
\put(117,240){(a-2)}
\put(199,240){(a-3)}
\put(35,155){(b-1)}
\put(117,155){(b-2)}
\put(199,155){(b-3)}
\put(35,70){(c-1)}
\put(117,70){(c-2)}
\put(199,70){(c-3)}
\normalsize
\end{picture}
\end{center}
       \caption{Interactions of level-$k$ vehicles at the roundabout intersection. (a-1)-(a-3) show three sequential steps in a simulation where three level-$1$ vehicles interact with each other; (b-1)-(b-3) show steps of three level-$2$ vehicles interacting with each other; (c-1)-(c-3) show steps of a level-$2$ vehicle (blue) interacting with two level-$1$ vehicles (yellow and red); $v_1$, $v_2$ and $v_3$ are the speeds of the blue, yellow and red vehicles, respectively.}
      \label{fig: roundabout-level-k}
\end{figure}

It can be observed from Figs.~\ref{fig: fourway-level-k}-\ref{fig: roundabout-level-k} that, in general, when level-$1$ and level-$2$ vehicles interact with each other, the conflicts between them can be resolved. This is expected since level-$1$ vehicles, representing cautious/conservative vehicles, will yield the right of way and level-$2$ vehicles, representing aggressive vehicles, will proceed ahead. In contrast, when level-$1$ vehicles interact with level-$1$ vehicles, deadlocks may occur, such as the one being observed in the T-shaped intersection in Fig.~\ref{fig: threeway-level-k}(a), because everyone yields to the others. When level-$2$ vehicles interact with level-$2$ vehicles, collisions may occur, such as the ones being observed in panel~(b) of Figs.~\ref{fig: fourway-level-k}-\ref{fig: roundabout-level-k}, because everyone assumes the others would yield.

\begin{figure}[ht]
\begin{center}
\begin{picture}(200.0, 310.0)
%%%%%%%%%%%%%%%%%%%%%%%%%%%%%%
\put(  -25,  204){\epsfig{file=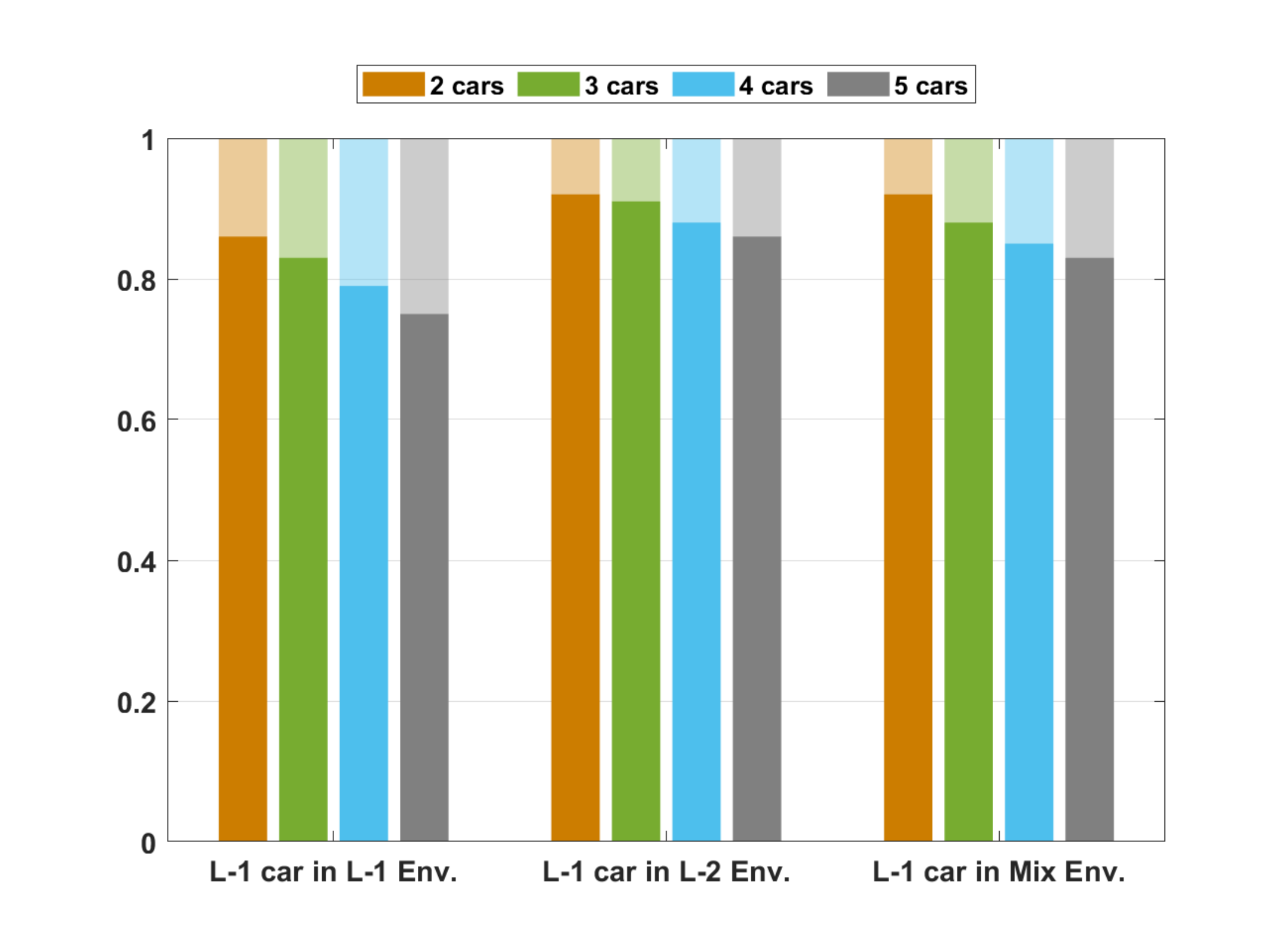,width = 0.55 \linewidth, trim=2.0cm 0.5cm 0cm 0.5cm,clip}}  %%%
%%%%%%%%%%%%
\put(  100,  204){\epsfig{file=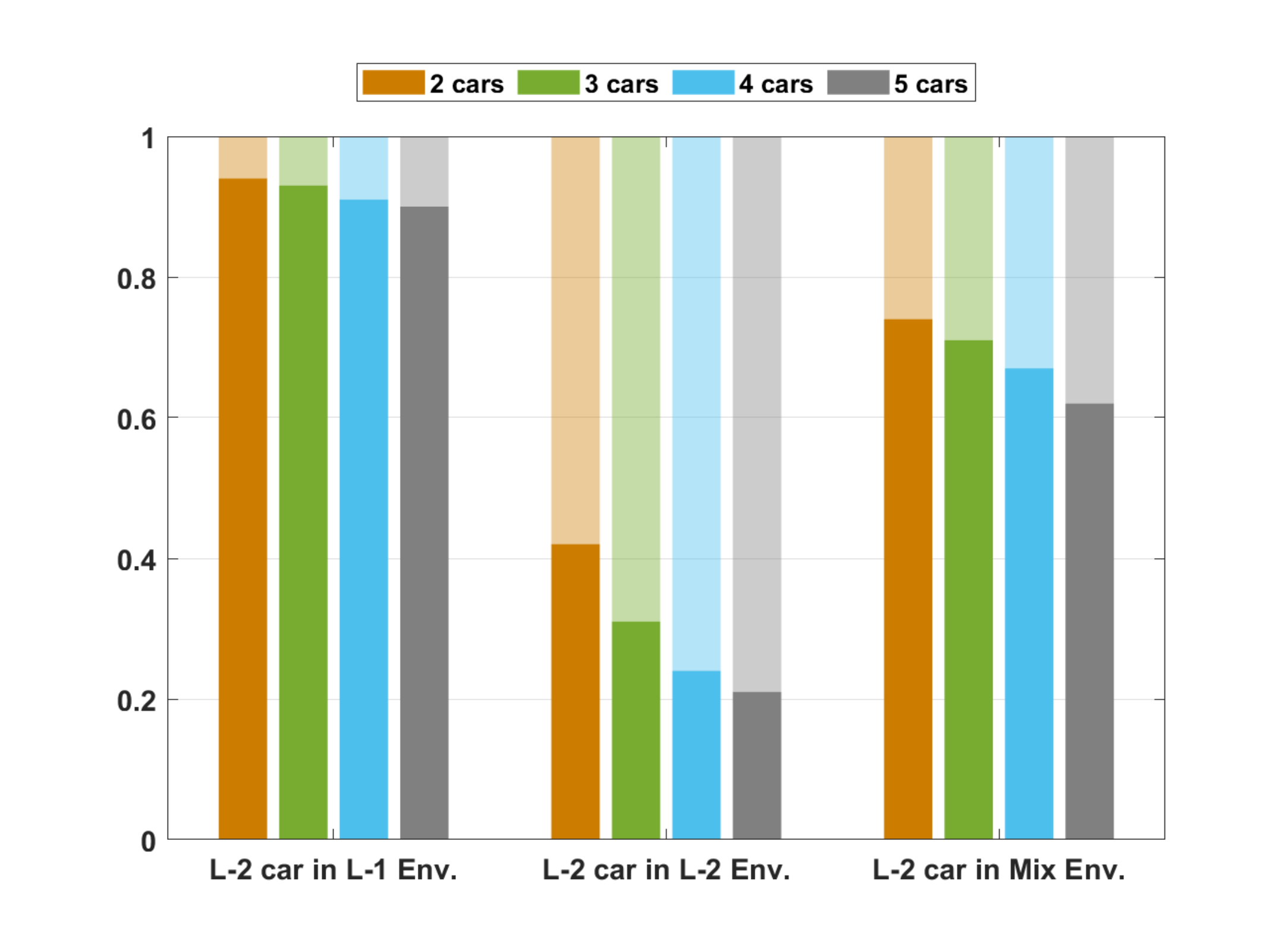,width = 0.55 \linewidth, trim=2.0cm 0.5cm 0cm 0.5cm,clip}}
%%%%%%%%%%%%%%%%%%%%%%%%%%%%%%%%%
\put(  -25,  102){\epsfig{file=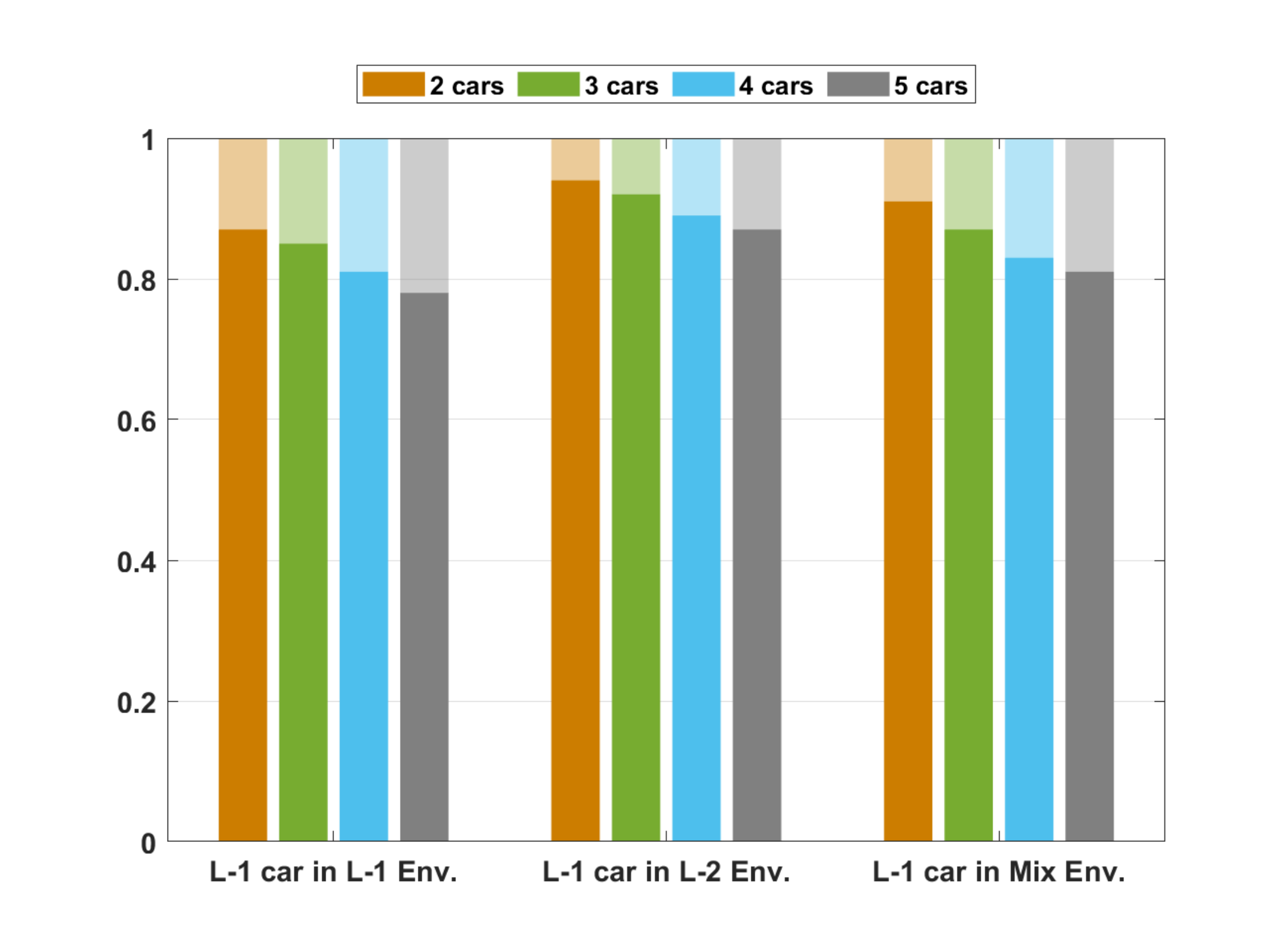,width = 0.55 \linewidth, trim=2.0cm 0.5cm 0cm 0.5cm,clip}}  %%%
%%%%%%%%%%%%
\put(  100,  102){\epsfig{file=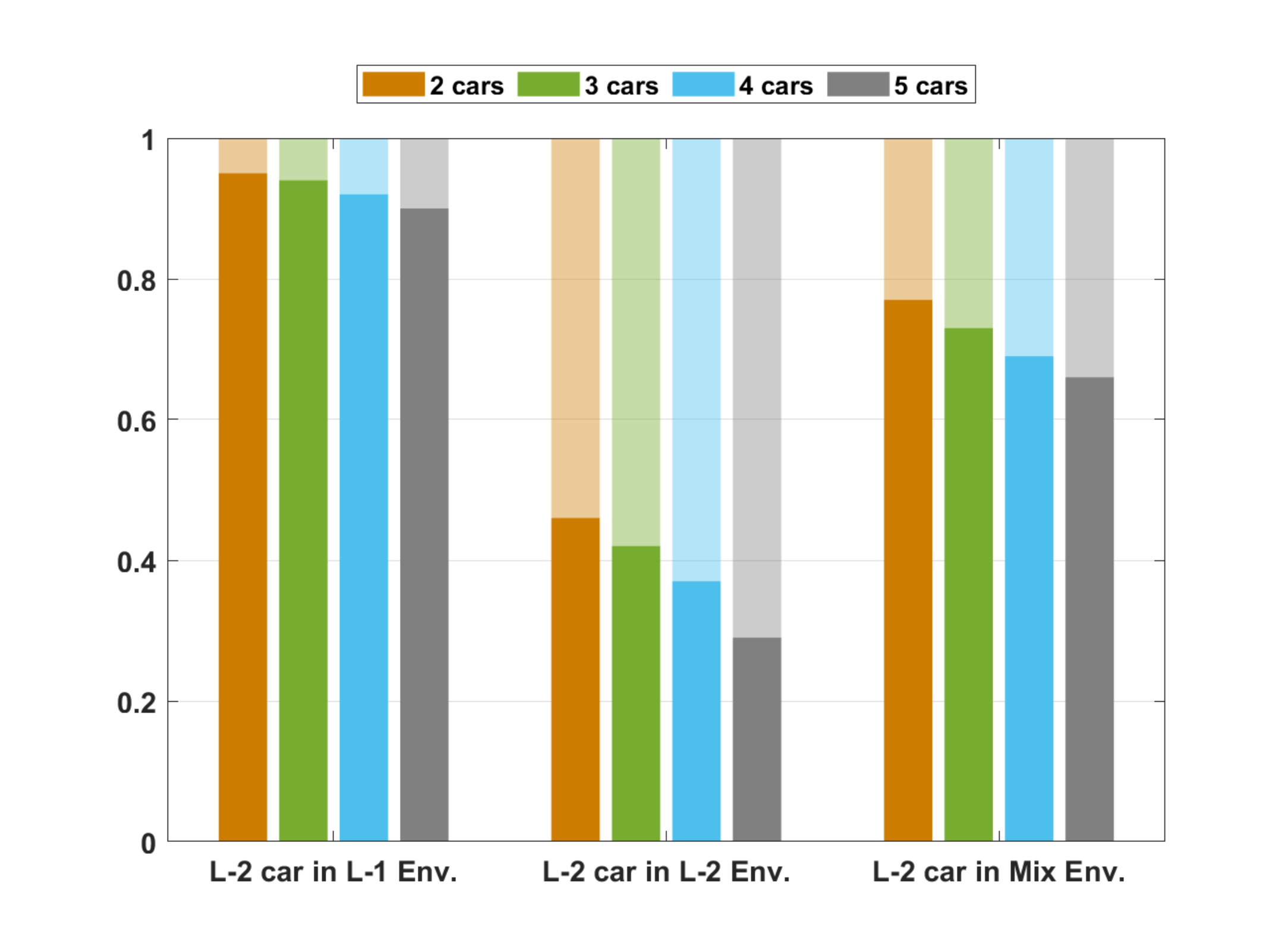,width = 0.55 \linewidth, trim=2.0cm 0.5cm 0cm 0.5cm,clip}}
%%%%%%%%%%%%%%%%%%%%%%%%%%%%%%%%
\put(  -25,  0){\epsfig{file=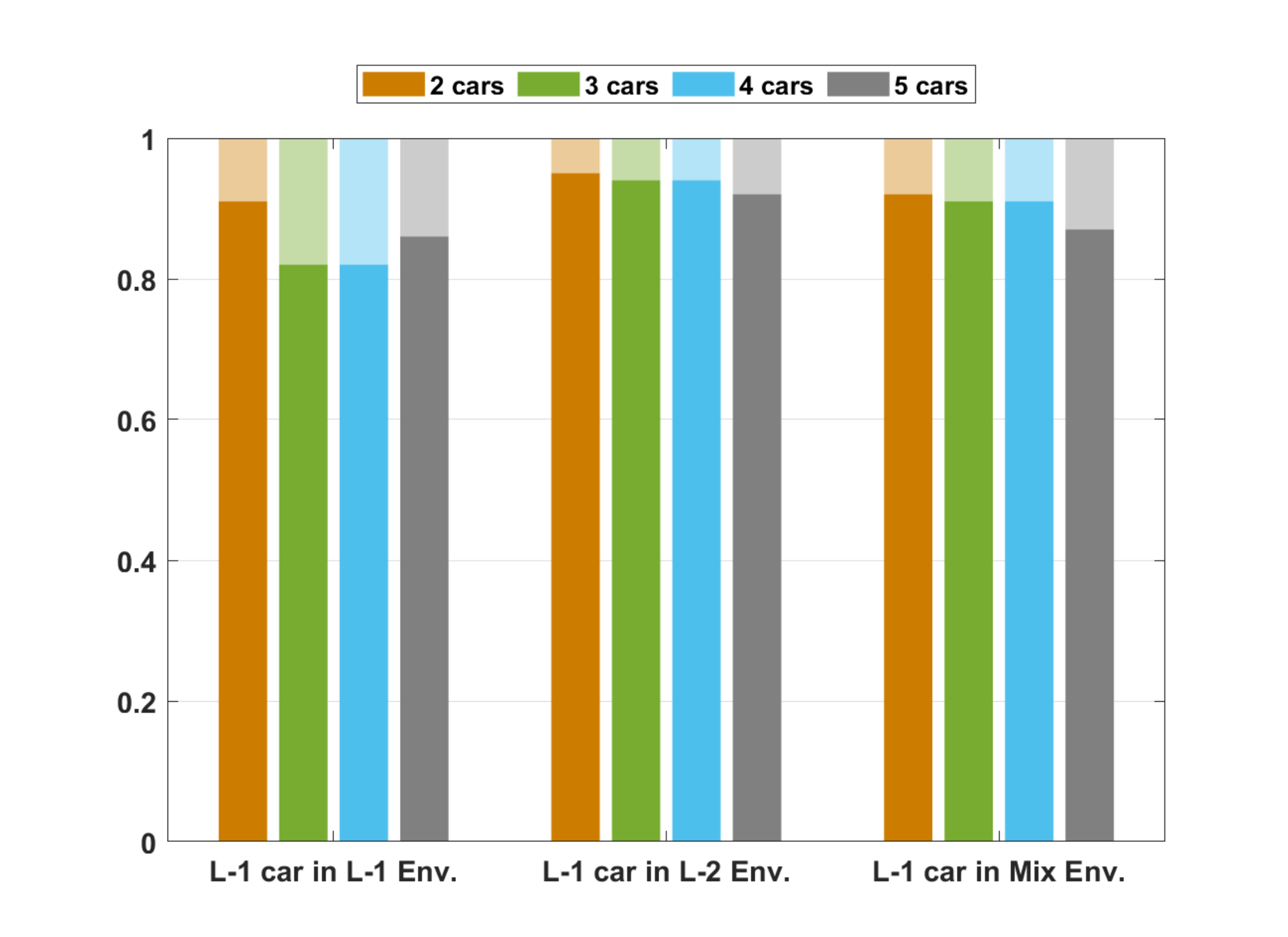,width = 0.55 \linewidth, trim=2.0cm 0.5cm 0cm 0.5cm,clip}}  %%%
%%%%%%%%%%%%
\put(  100,  0){\epsfig{file=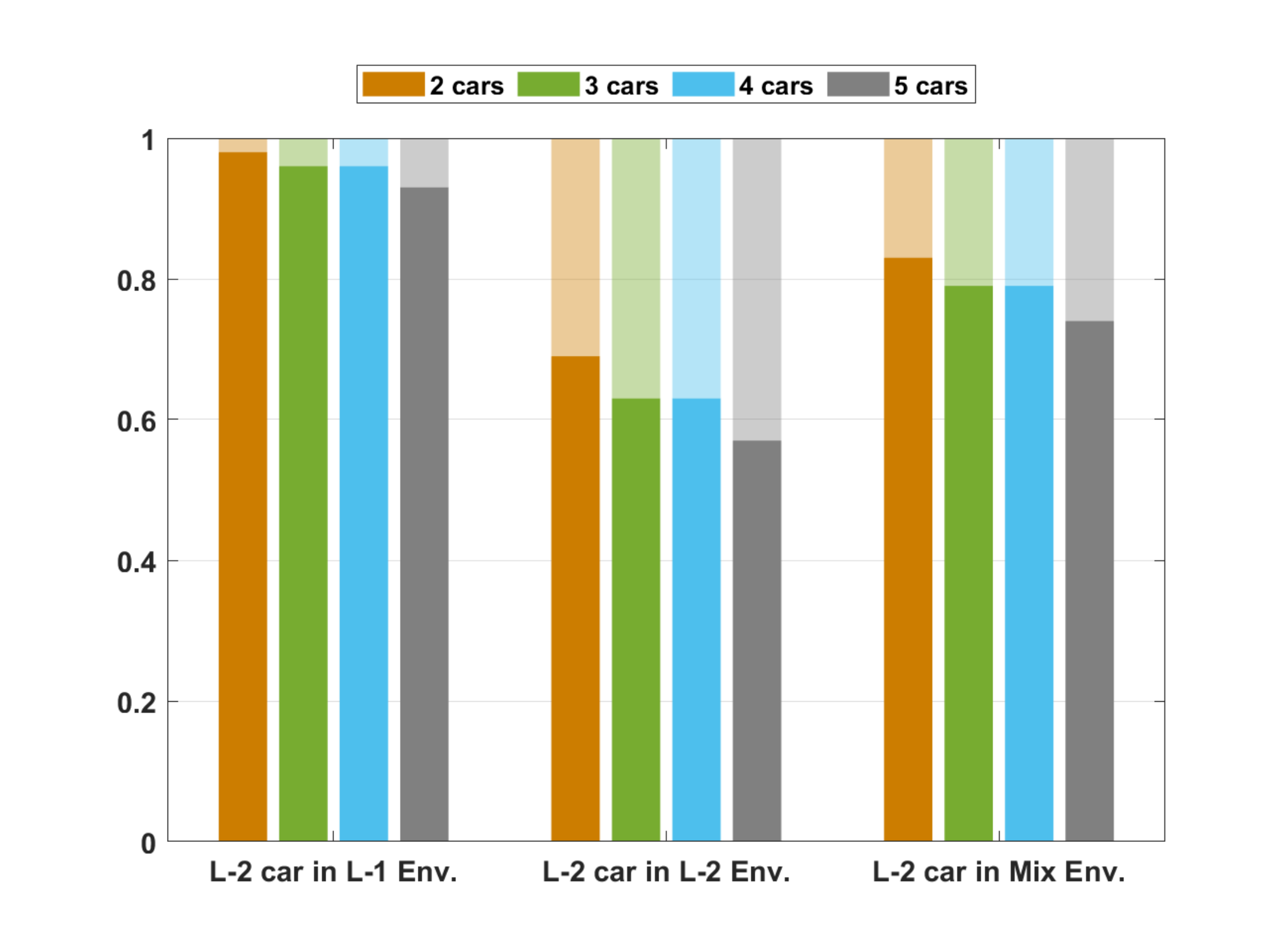,width = 0.55 \linewidth, trim=2.0cm 0.5cm 0cm 0.5cm,clip}}  %%%
%%%%%%%%%%%%%%%%%%%%%%
\small
\put(80,298){(a-1)}
\put(205,298){(b-1)}
\put(80,196){(a-2)}
\put(205,196){(b-2)}
\put(80,94){(a-3)}
\put(205,94){(b-3)}
\normalsize
\end{picture}
\end{center}
      \caption{The rates of success of level-$k$ policies. (a-1)-(a-3) show the rates of success of a level-$1$ ego vehicle operating in various traffic environments (various in the numbers and policies of interacting vehicles) at the four-way, T-shaped, and roundabout intersections; (b-1)-(b-3) show those of a level-$2$ ego vehicle; the bars in dark color represent the rates of success.}
      \label{fig: level-k-eval}
\end{figure}

We remark that deadlocks (collisions) do not always occur in level-$1$ (level-$2$) interactions. The initial conditions of Figs.~\ref{fig: fourway-level-k}-\ref{fig: roundabout-level-k} are chosen to show such situations. For randomized initial conditions, the rates of success, defined as the proportion of $2000$ simulation episodes where neither deadlocks nor collisions occur to the ego vehicle, for different numbers of interacting vehicles and different combinations of level-$k$ policies at the three intersections are shown in Fig.~\ref{fig: level-k-eval}. In Fig.~\ref{fig: level-k-eval}, ``L-$k$ car in L-$k'$ Env.'' shows the rate of success of a level-$k$ ego vehicle interacting with other vehicles that are all level-$k'$; ``L-$k$ car in Mix Env.'' shows the rate of success of a level-$k$ ego vehicle interacting with other vehicles whose control policies are randomly chosen between level-$1$ and level-$2$ with equal probability.

The following observations can be made: 1) As the number of interacting vehicles increases, the rate of success decreases for all the cases. This is reasonable since a larger number of interacting vehicles represents a more complex traffic scenario. 2) The rates of success of a level-$2$ ego vehicle interacting with other vehicles that are also level-$2$ are the lowest among the results of all combinations of level-$k$ policies. This is also reasonable since when all the vehicles are aggressive and assume the others would yield, traffic accidents are more likely to occur. 3) Among the results of the three intersection types, the rates of success for the roundabout intersection are the highest. This illustrates the effective functionality of roundabouts in reducing traffic conflicts.

We further remark that although the high rates of failure of ``level-$2$ versus level-$2$'' are not desired in real-world traffic, it is important for a simulation environment for AV control testing to include such cases that represent rational interactions between aggressive vehicles. Note that a level-$2$ vehicle is a rational decision maker that behaves aggressively, which is fundamentally different from a driver/vehicle model that acts aggressively but in an irrational way, e.g., taking actions randomly. \di{The cases of level-$2$ vehicle interactions provide challenging test scenarios for AV control systems, which can be more realistic than those provided by some worst-case (i.e., not necessarily rational) models \cite{chou2018using}.}

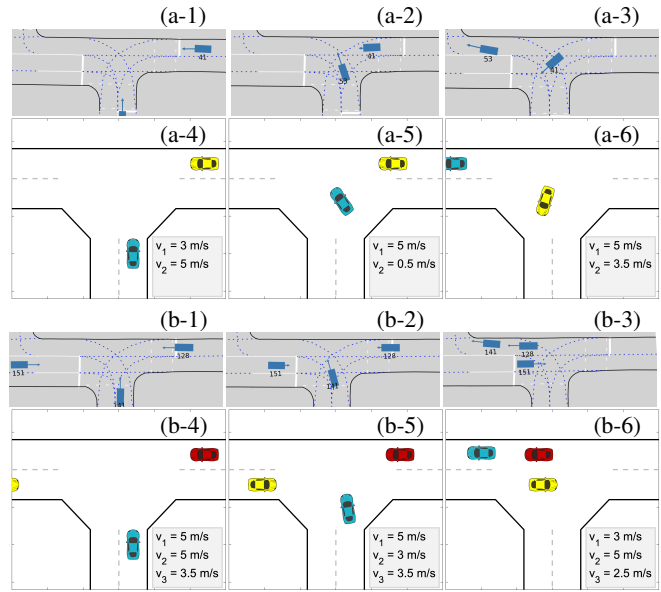
\begin{figure}[ht]
\begin{center}
\begin{picture}(200.0, 220.0)
%%%%%%%%%%%%%%%%%%%%%%%%%%%%%%
\put(  -21,  180){\epsfig{file=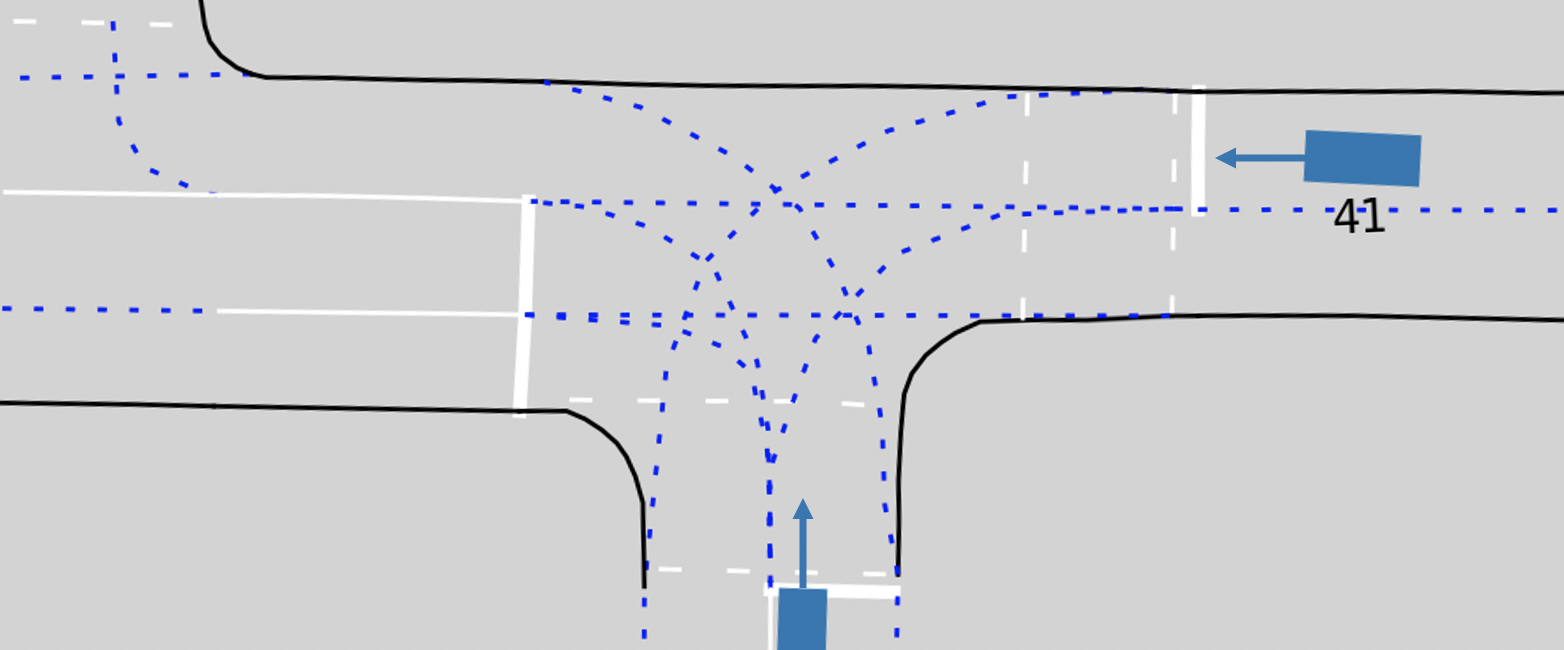,width = 0.32 \linewidth, angle=0, trim=0.6cm 0.5cm 0.8cm 0.4cm,clip}}  %%%
\put(  62,  180){\epsfig{file=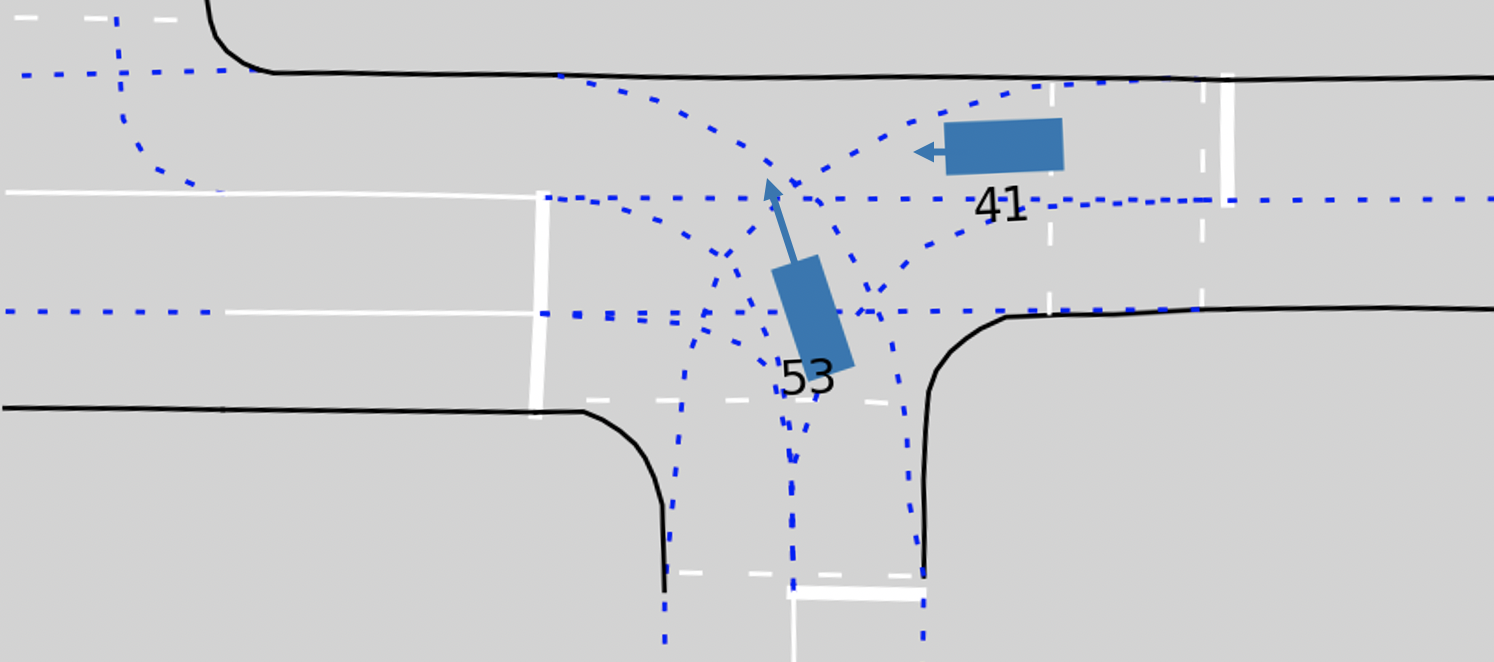, width = 0.31\linewidth, angle=0, trim=0.6cm 0.8cm 0.5cm 0.4cm,clip}}
\put(  143,  180){\epsfig{file=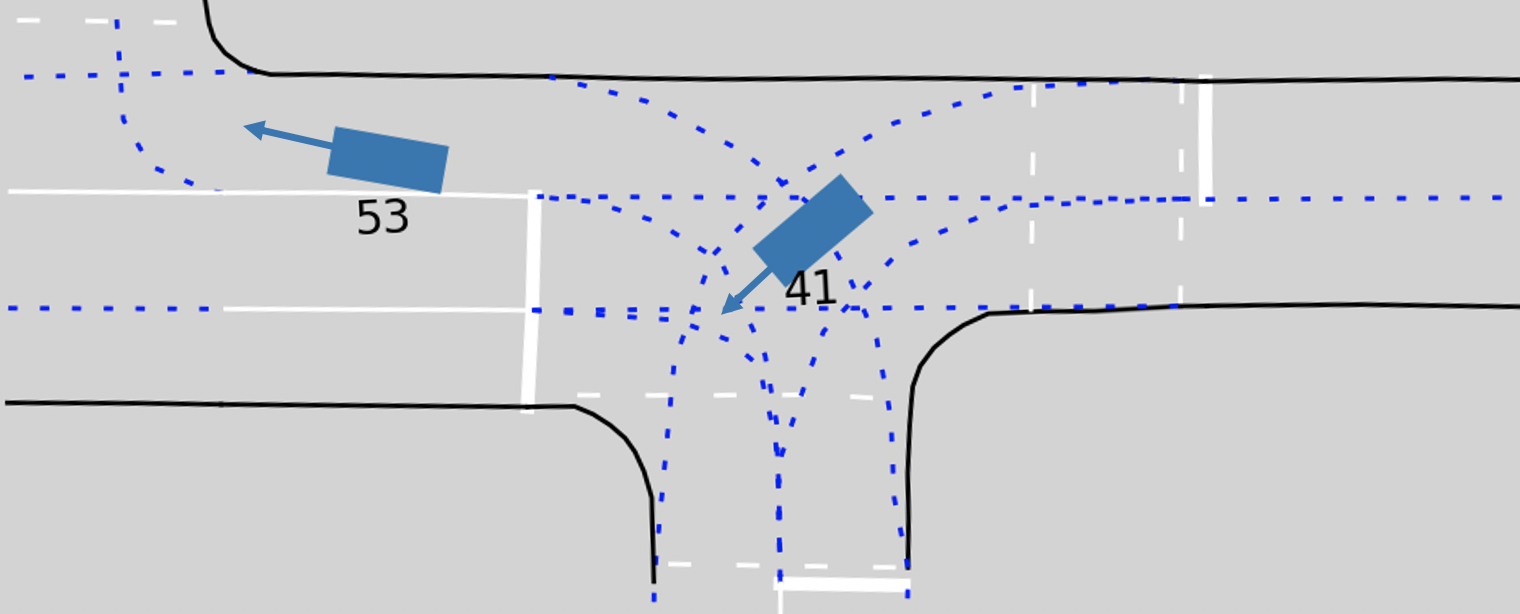, width = 0.32\linewidth, angle=0, trim=1.8cm 0.5cm 0.5cm 0.4cm,clip}}
%%%%%%%%%%%%%%%%%%%%%%%%%%%%%
\put(  -22,  110){\epsfig{file=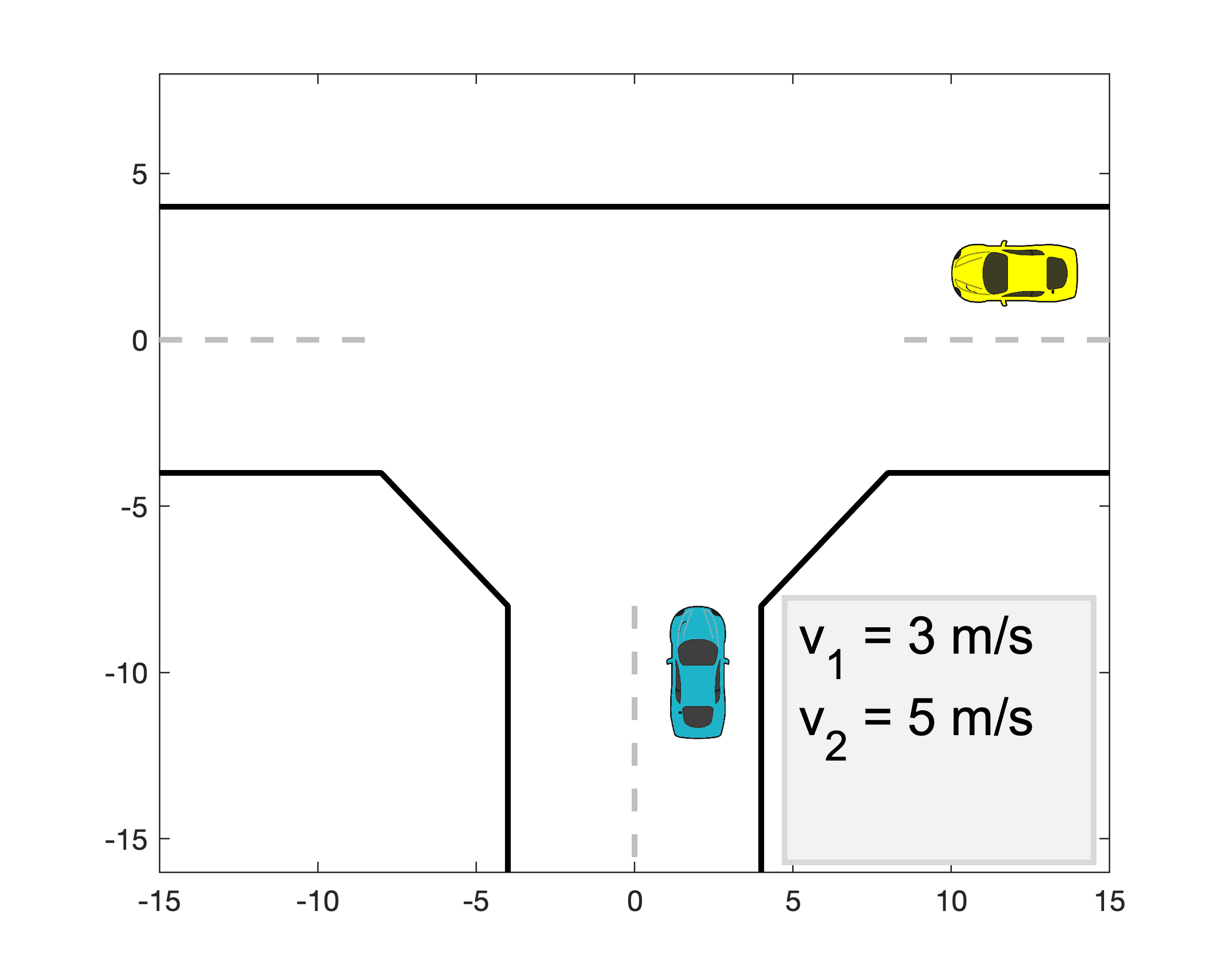,width = 0.33 \linewidth, trim=1.8cm 1.2cm 1.2cm 0.8cm,clip}}  %%%
\put(  60,  110){\epsfig{file=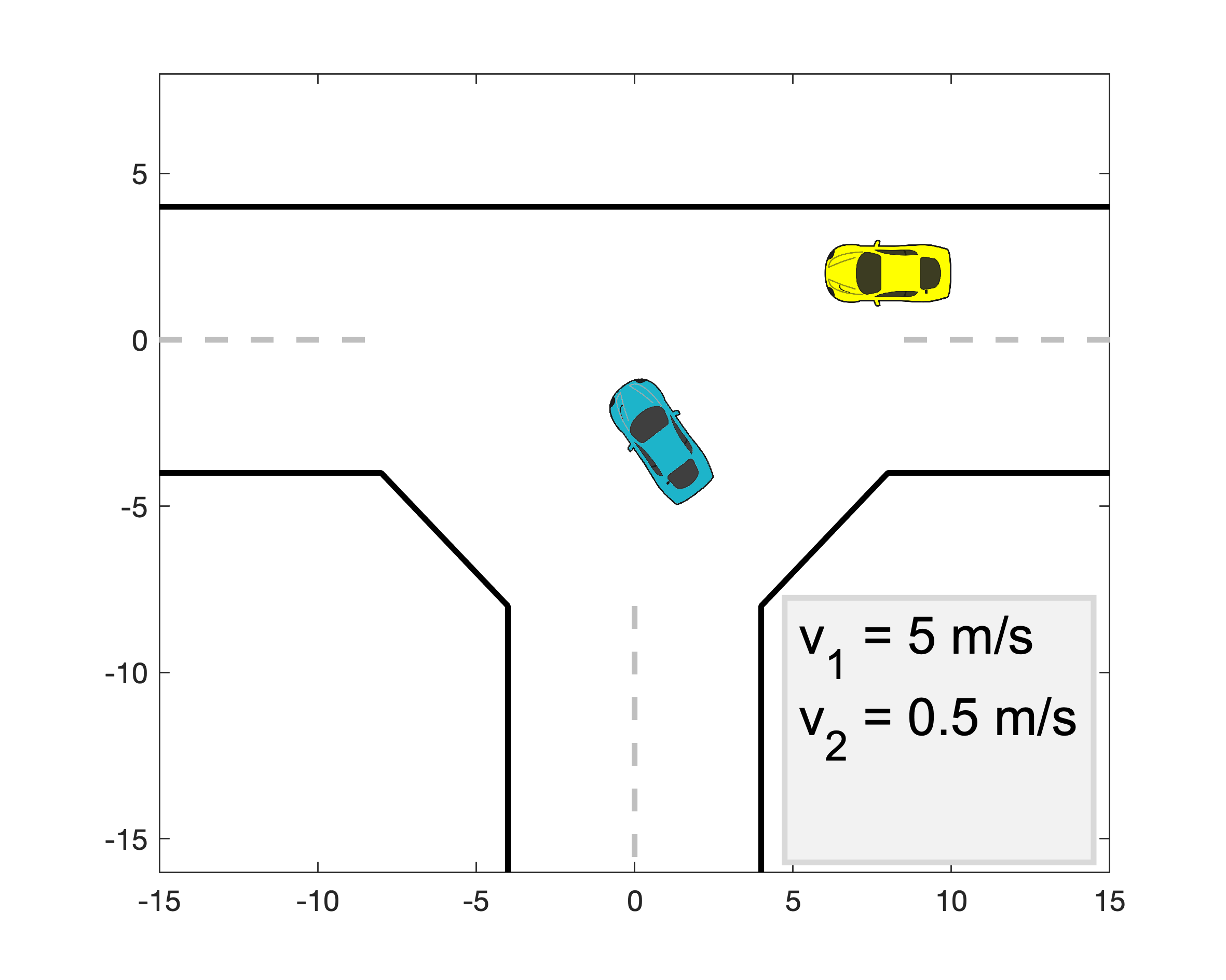, width = 0.33\linewidth, trim=1.8cm 1.2cm 1.2cm 0.8cm,clip}}
\put(  142,  110){\epsfig{file=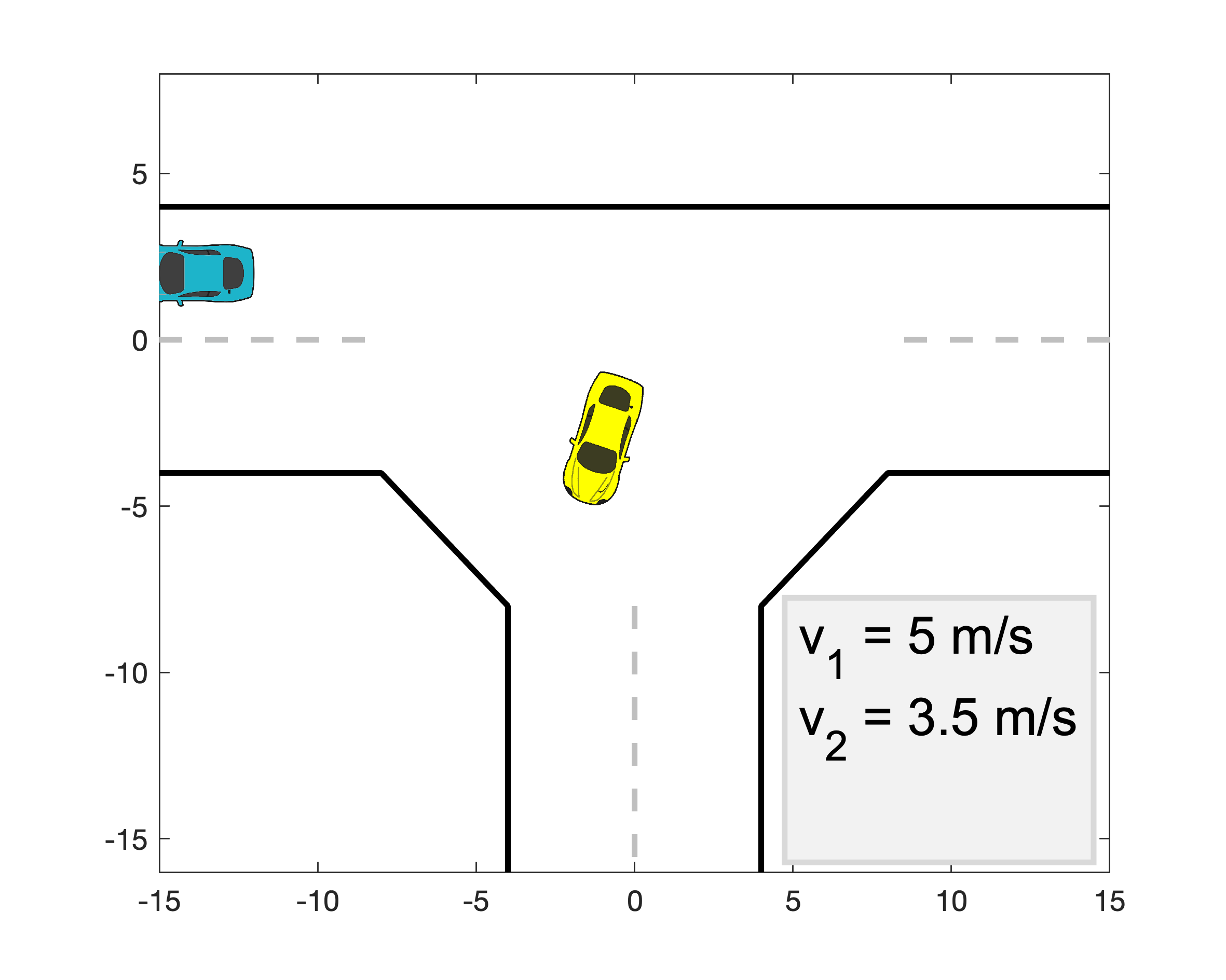, width = 0.33\linewidth, trim=1.8cm 1.2cm 1.2cm 0.8cm,clip}}
%%%%%%%%%%%%%%%%%%%%%%%%%%%%%%
\put(  -22,  70){\epsfig{file=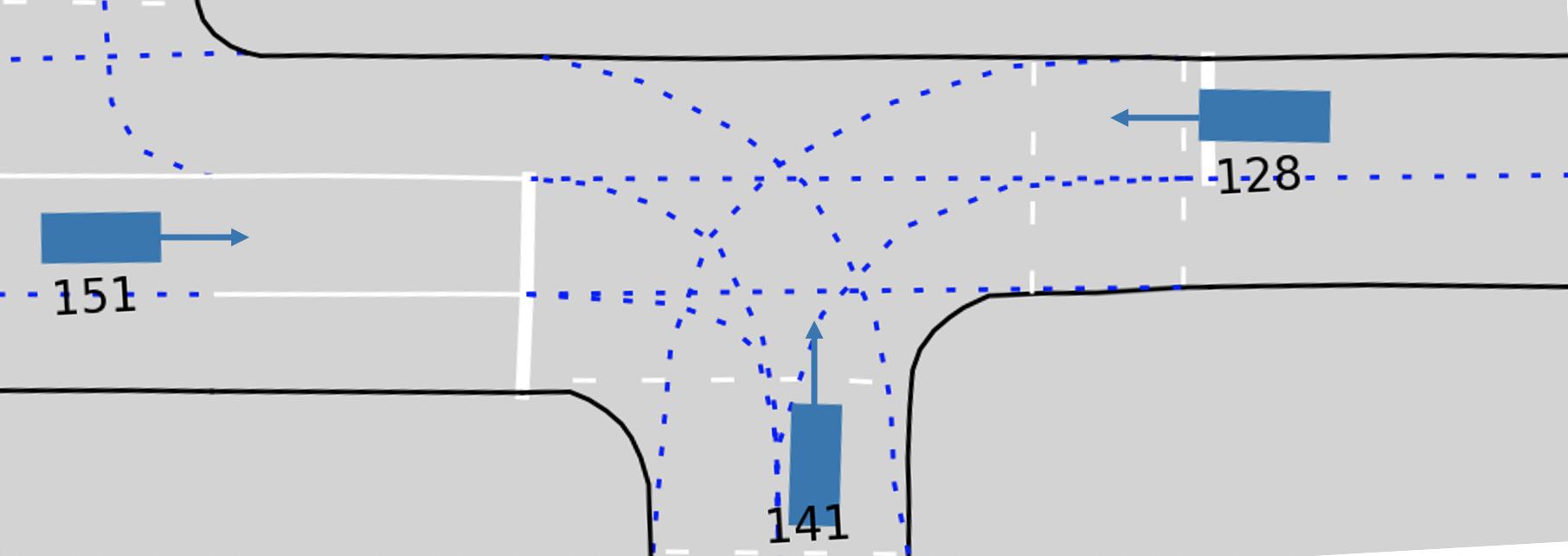,width = 0.32 \linewidth, angle=0, trim=0.6cm 0.5cm 0.5cm 0.2cm,clip}}  %%%
\put(  60,  70){\epsfig{file=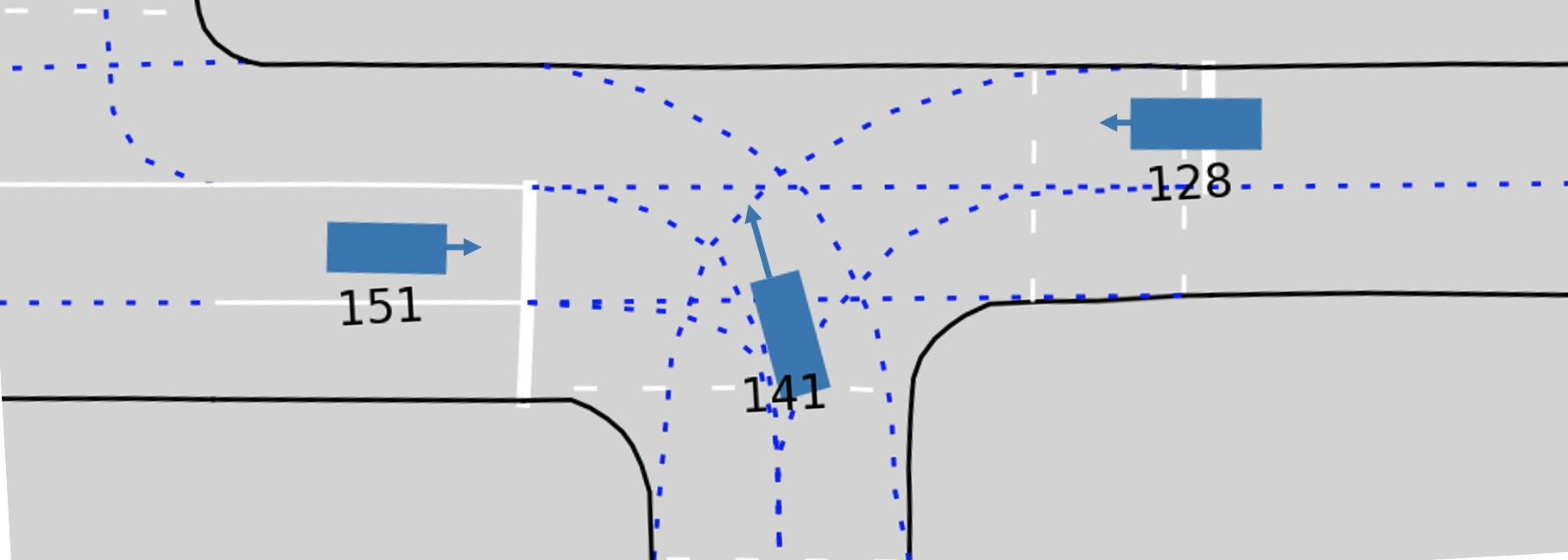, width = 0.32\linewidth, angle=0, trim=0.6cm 0.5cm 0.5cm 0.2cm,clip}}
\put(  142,  70){\epsfig{file=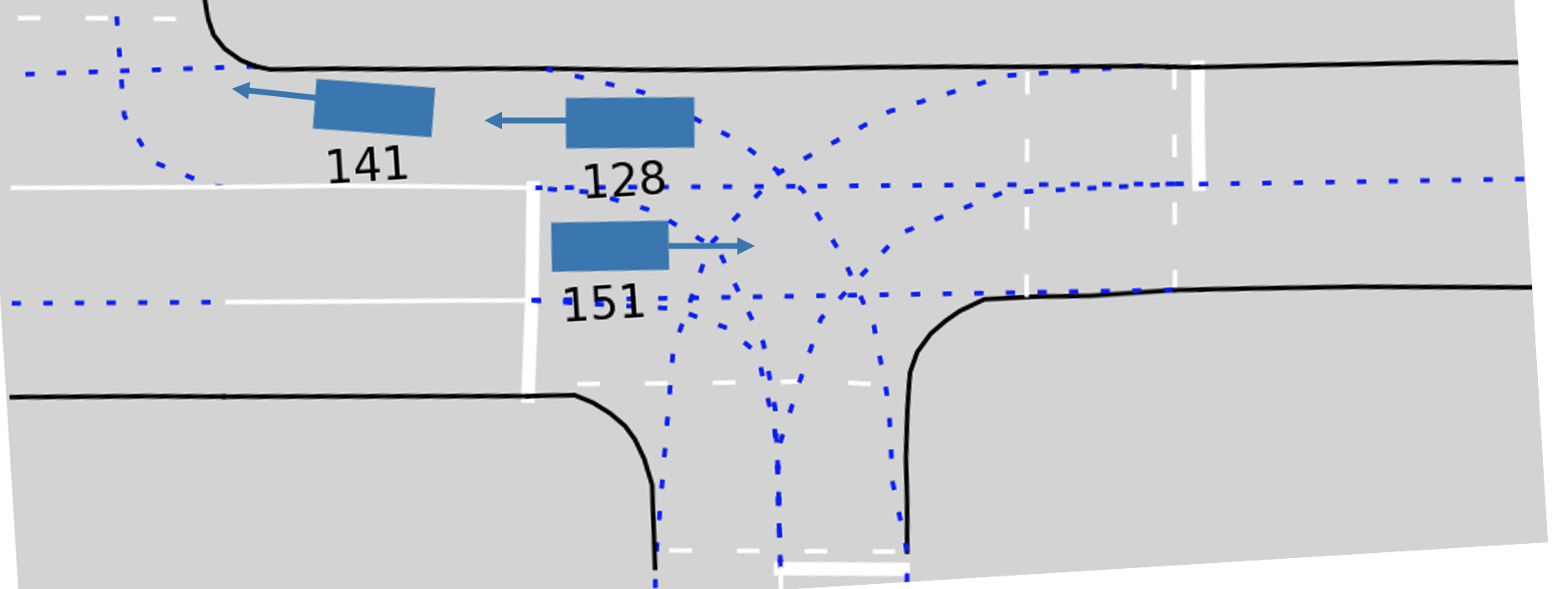, width = 0.32\linewidth, angle=0,, trim=0.6cm 1.0cm 1.0cm 0.5cm,clip}}
%%%%%%%%%%%%%%%%%%%%%%
\put(  -22,  0){\epsfig{file=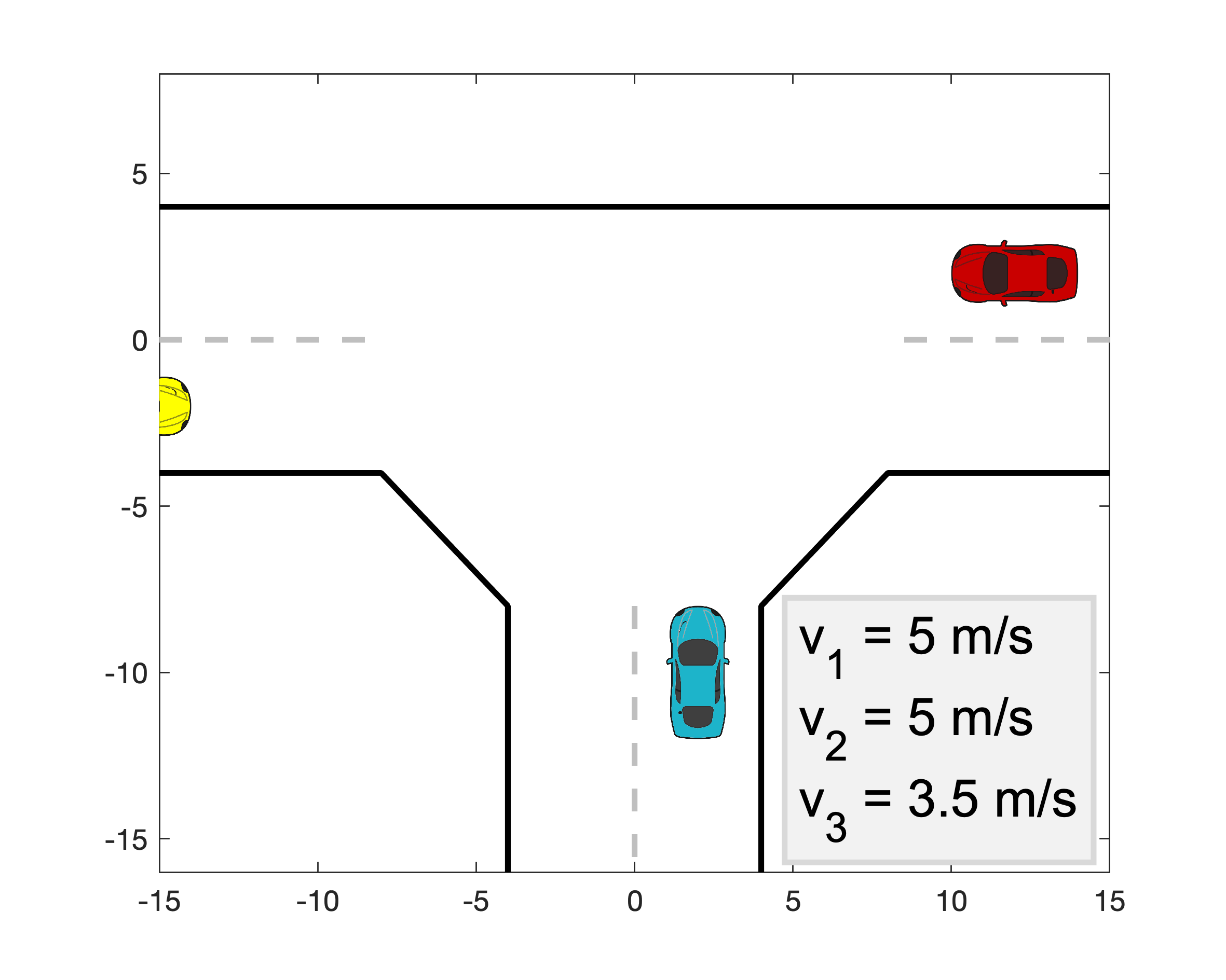,width = 0.33 \linewidth, trim=1.8cm 1.2cm 1.2cm 0.8cm,clip}}  %%%
\put(  60,  0){\epsfig{file=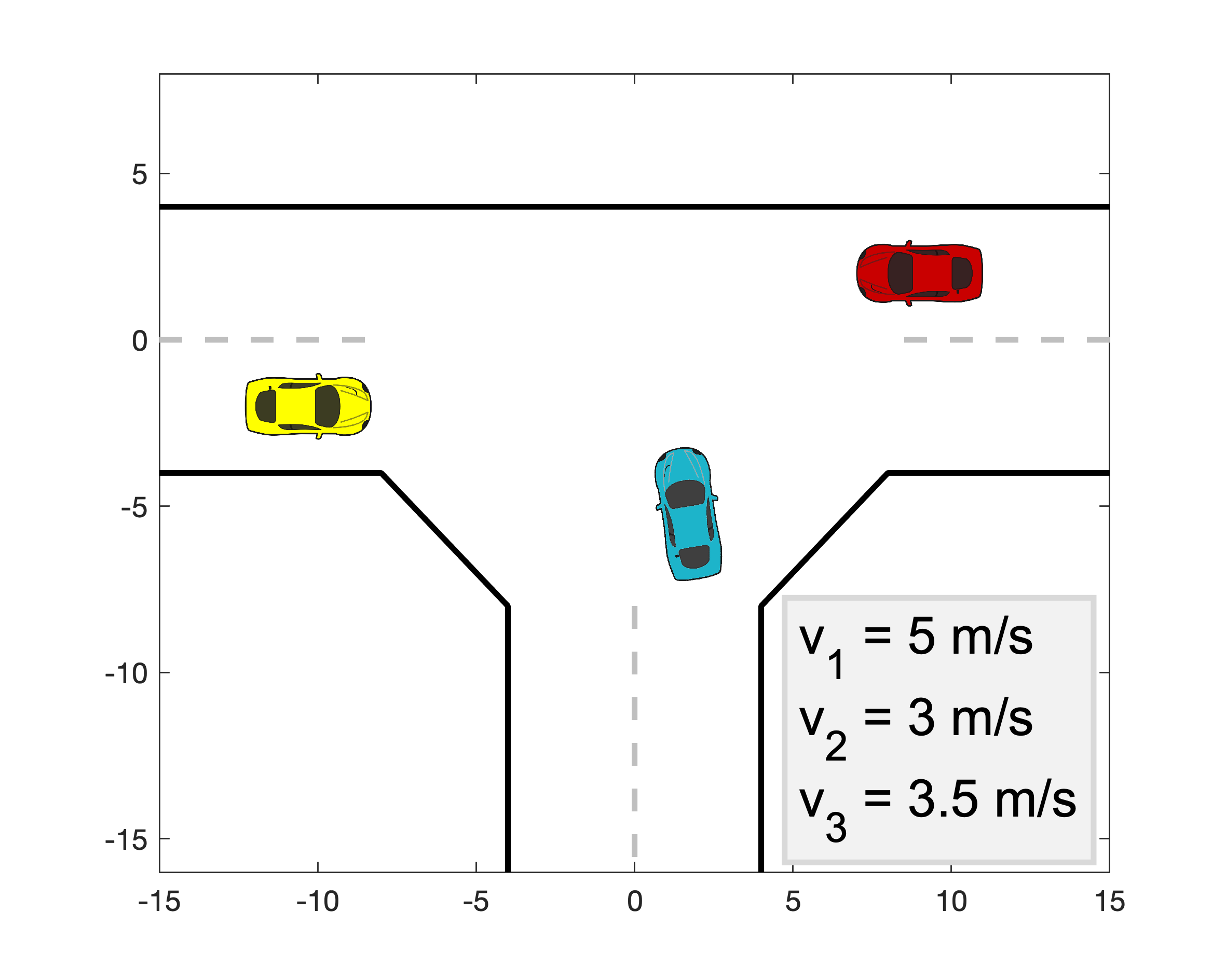, width = 0.33\linewidth, trim=1.8cm 1.2cm 1.2cm 0.8cm,clip}}
\put(  142,  0){\epsfig{file=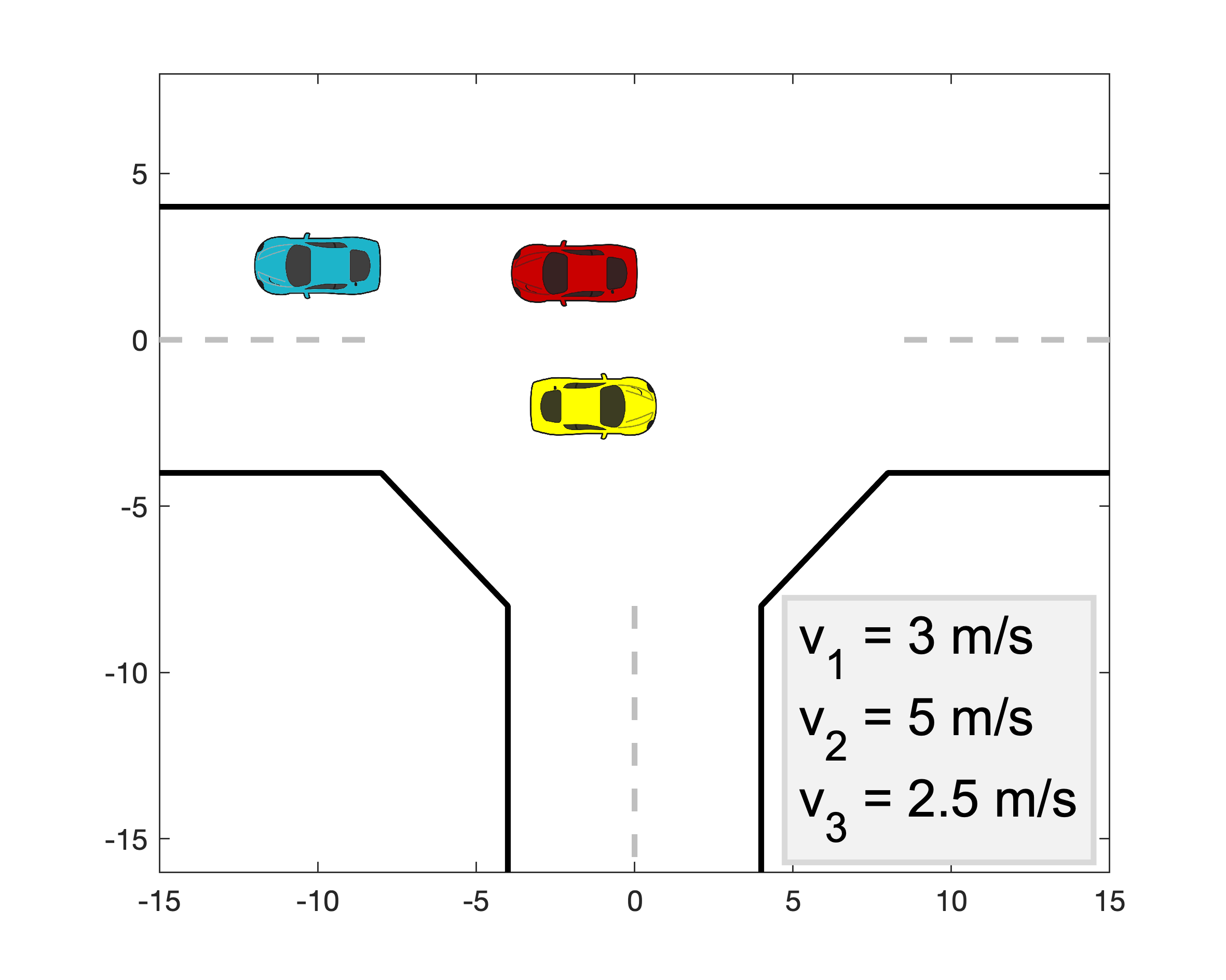, width = 0.33\linewidth, trim=1.8cm 1.2cm 1.2cm 0.8cm,clip}}
\small
\put(35,215){(a-1)}
\put(117,215){(a-2)}
\put(199,215){(a-3)}
\put(35,170){(a-4)}
\put(117,170){(a-5)}
\put(199,170){(a-6)}
\put(35,100){(b-1)}
\put(117,100){(b-2)}
\put(199,100){(b-3)}
\put(35,60){(b-4)}
\put(117,60){(b-5)}
\put(199,60){(b-6)}
\normalsize 
\end{picture}
\end{center}
       \caption{Reproduction of real-world traffic scenarios using our level-$k$ vehicle models. (a-1)-(a-3) visualize a traffic scenario with two interacting vehicles extracted from the dataset \cite{interactiondataset} at three sequential time instants; (a-4)-(a-6) show the simulation results of a level-$2$ vehicle (blue) interacting with a level-$1$ vehicle (yellow) in a similar scenario. (b-1)-(b-3) visualize a traffic scenario with three interacting vehicles extracted from the same dataset; (b-4)-(b-6) show the simulation results of a level-$2$ vehicle (blue) interacting with two level-$1$ vehicles (yellow and red) in a similar scenario.}   
      \label{fig: t_shape_traffic_compare}
\end{figure}

\begin{figure}[ht]
\centering
\includegraphics[width=0.45\textwidth]{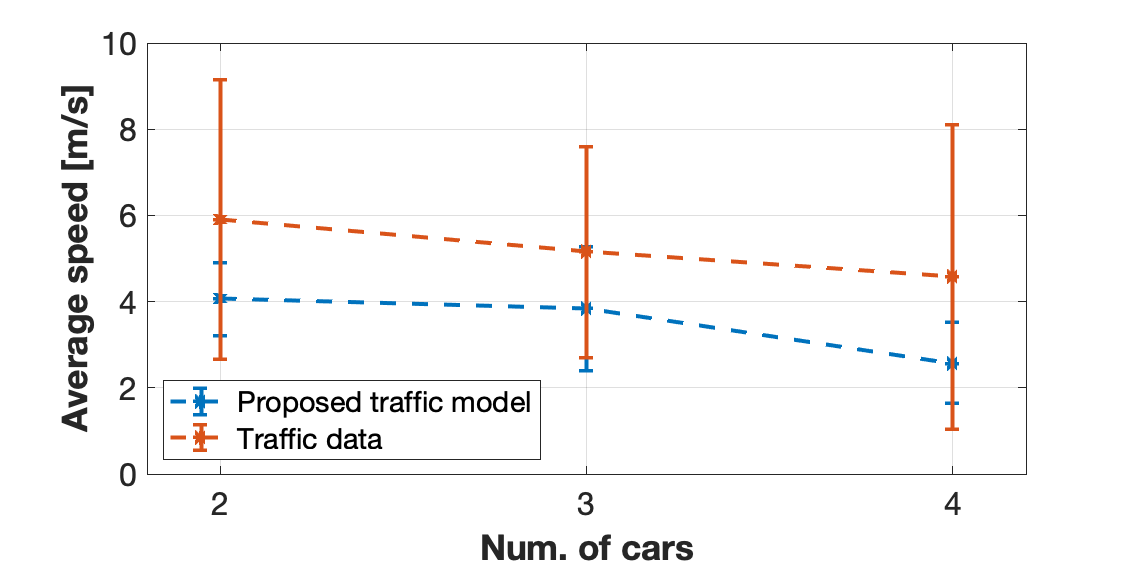}
\centering
\caption{Average speeds versus numbers of interacting vehicles for traversing T-shaped intersections of level-$k$ vehicle models and traffic data.}
\label{fig: traffic_data_speed_compare}
\end{figure}

\di{\subsection{Model validation}
\label{sec: results_2}

We validate our level-$k$ vehicle models before illustrating how to use them for AV control testing.

\subsubsection{Feasibility validation}

The unicycle model \eqref{equ:kinematics} has been used to represent vehicle kinematics and the action set $\mathcal{U}$ in Table~\ref{table: action-set} has been used to represent common  driving maneuvers. We now show that the trajectories generated by \eqref{equ:kinematics} with actions from $\mathcal{U}$ are feasible trajectories for vehicle systems. For this, we use a hybrid kinematic/dynamic bicycle model with the brush tire model \cite{kong2015kinematic} to represent high-fidelity vehicle dynamics, and we use a PID-based controller \cite{rajamani2011vehicle} to control the vehicle dynamics to execute the trajectory generated by \eqref{equ:kinematics} and $\mathcal{U}$. Specifically, at each discrete-time instant $t$
the level-$k$ decision policy selects an action from $\mathcal{U}$, which defines a desired state $s_{t+1}$ for the vehicle system through the unicycle model \eqref{equ:kinematics}. Then, over the continuous-time interval from $t$ to $t+1$, the PID-based controller controls the vehicle dynamics to track the desired state $s_{t+1}$. 

Two examples of tracking results for the T-shaped and the roundabout intersections are shown in Fig.~\ref{fig: motion control}, where the red solid curves represent the trajectories generated by \eqref{equ:kinematics} (referred to as ``reference trajectories'') and the black dotted curves represent the tracking trajectories. It can be observed that the tracking trajectories closely match the reference trajectories. This justifies the feasibility of trajectories generated by the unicycle model \eqref{equ:kinematics} and the action set $\mathcal{U}$.

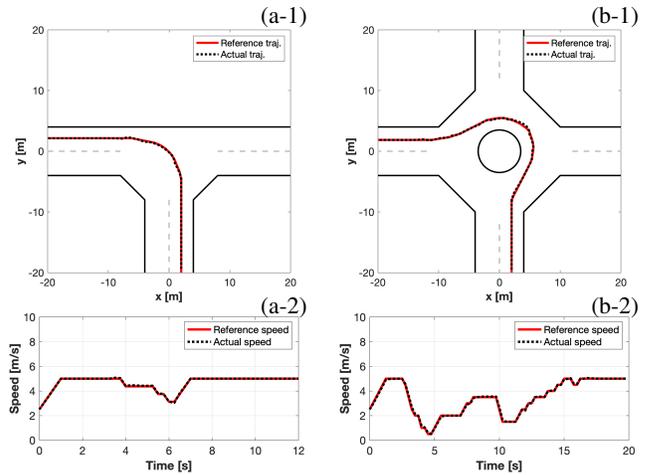
\begin{figure}[ht]
\begin{center}
\begin{picture}(200.0, 181.0)
\put(  104,  60){\epsfig{file=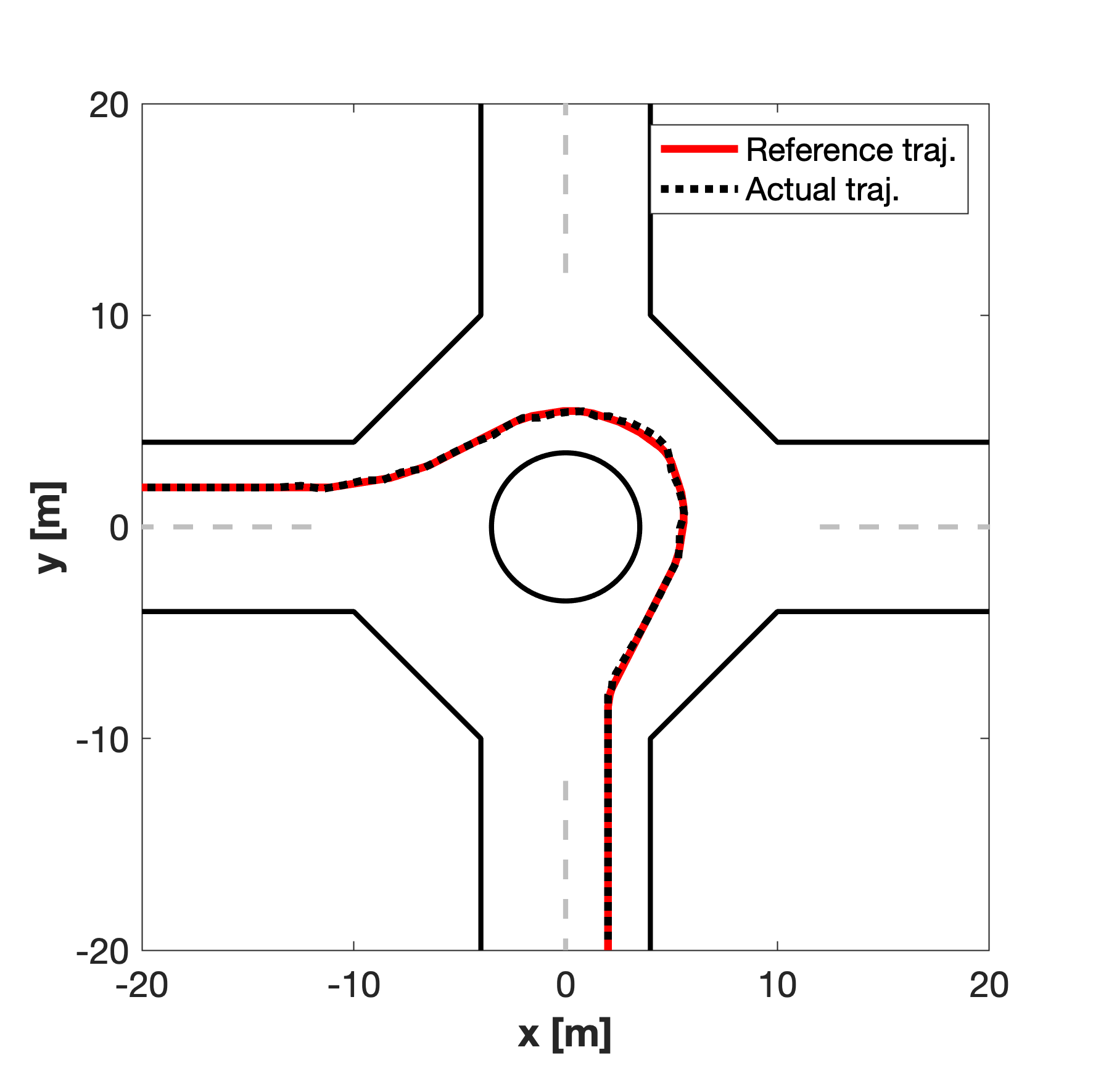,width = 0.47 \linewidth, trim=0.0cm 0.0cm 0cm 0.0cm,clip}}  %%%
%%%%%%%%%%%%
\put(  -21,  60){\epsfig{file=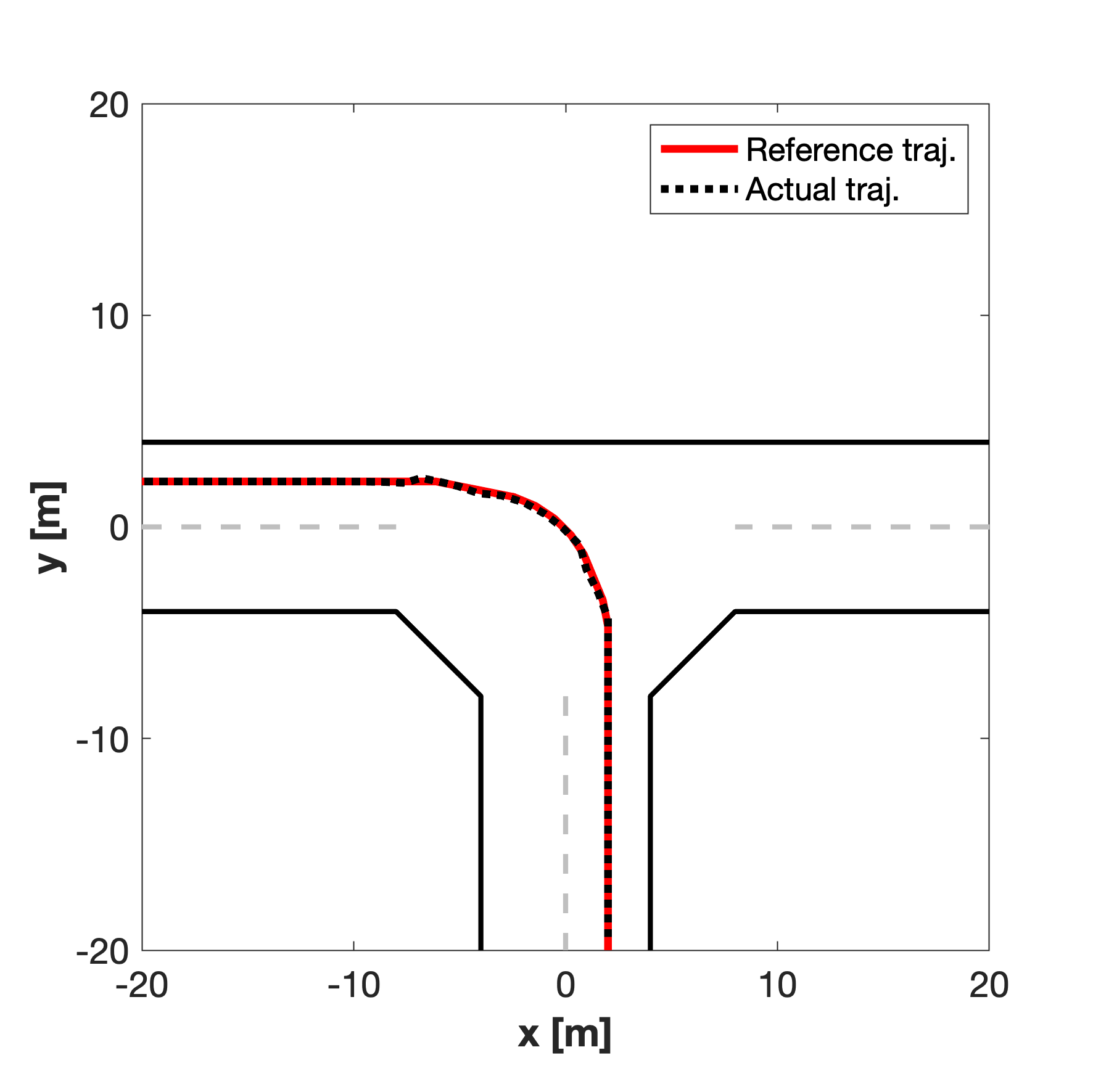,width = 0.47 \linewidth, trim=0.0cm 0.0cm 0cm 0.0cm,clip}}
%%%%%%%%%%%%%%%%%%%%%%%%%%%%%%%%
\put(  100,  0){\epsfig{file=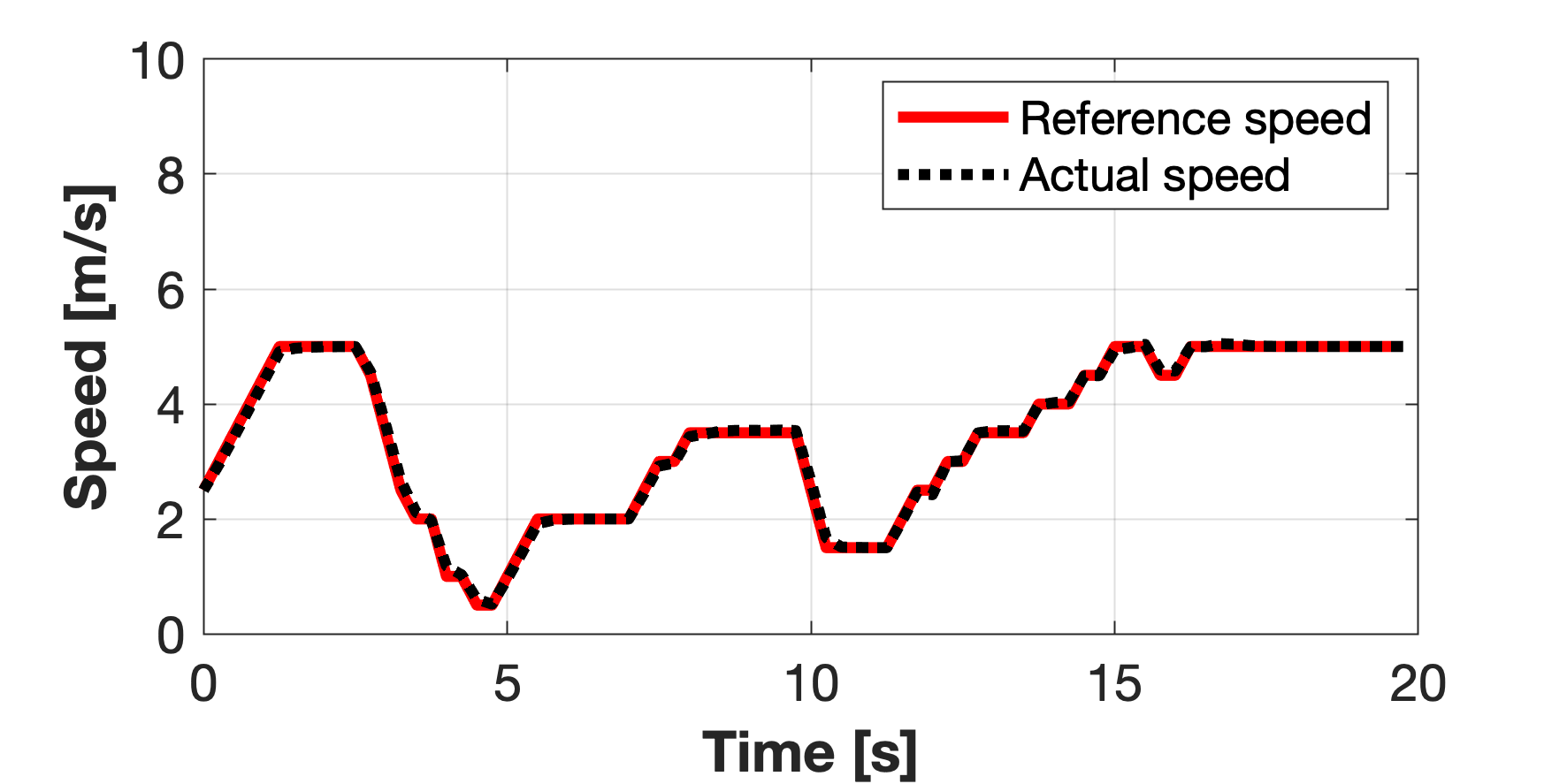,width = 0.5 \linewidth, trim=0.0cm 0.0cm 0cm 0.0cm,clip}}  %%%
%%%%%%%%%%%%
\put(  -25,  0){\epsfig{file=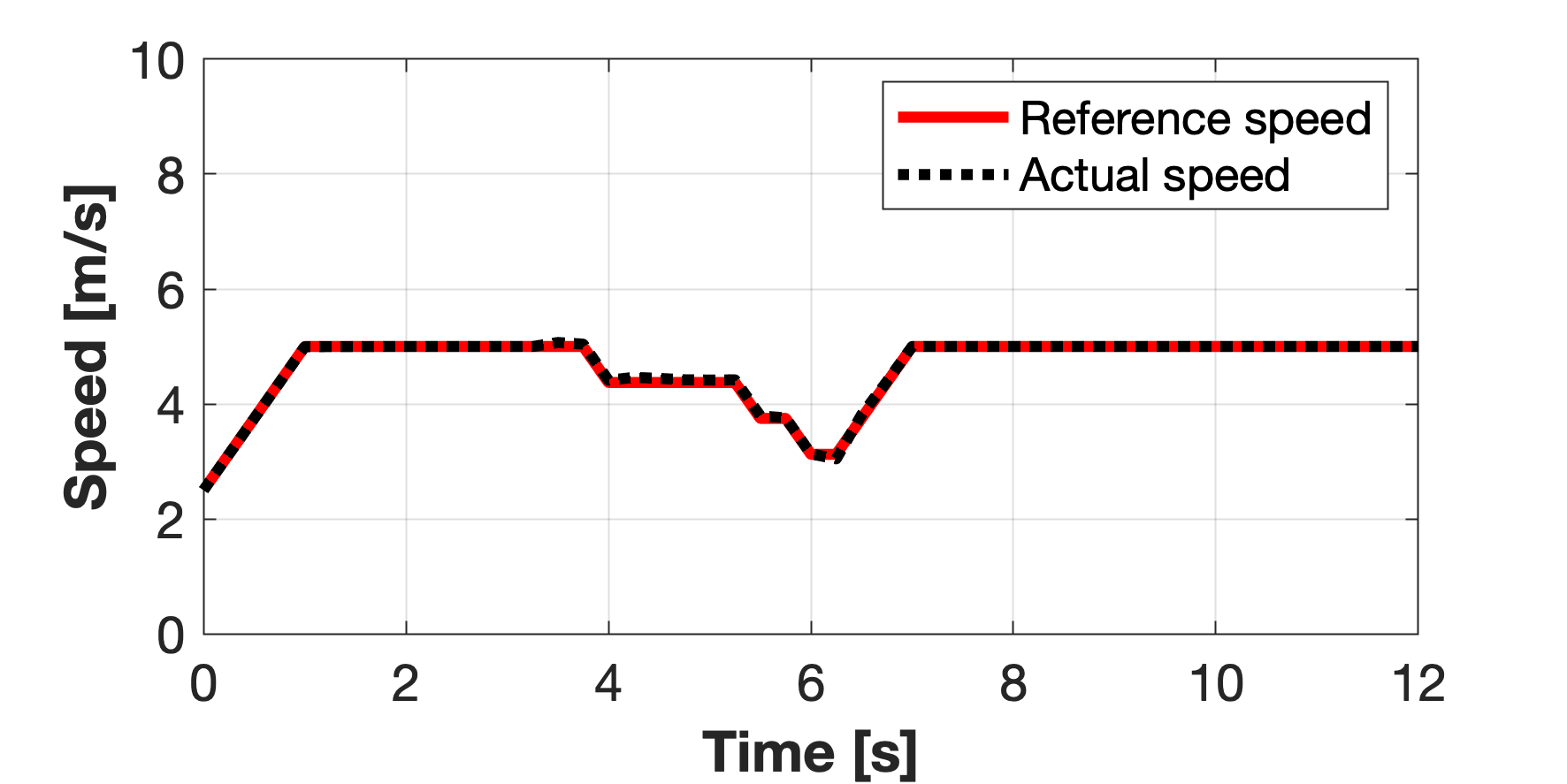,width = 0.5 \linewidth, trim=0.0cm 0.0cm 0cm 0.0cm,clip}}  %%%
%%%%%%%%%%%%%%%%%%%%%%
\small
\put(75, 170){(a-1)}
\put(200,170){(b-1)}
\put(75, 60){(a-2)}
\put(200,60){(b-2)}
\normalsize
\end{picture}
\end{center}
      \caption{Feasibility validation of the unicycle model \eqref{equ:kinematics} and the action set $\mathcal{U}$. (a-1)-(a-2) show an example of path and speed tracking result at a T-shaped intersection; (b-1)-(b-2) show that at a roundabout intersection.}
      \label{fig: motion control}
\end{figure}

\subsubsection{Comparison to traffic data}

We next validate our level-$k$ vehicle models using real-world traffic data. 

In Fig.~\ref{fig: t_shape_traffic_compare}, we show two traffic scenarios at a T-shaped intersection extracted from the INTERACTION dataset \cite{interactiondataset} and their reproduction by simulating our level-$k$ vehicle models. Specifically, we initialize the states of our level-$k$ vehicle models according to the initial scene of the scenario, and compare the evolution of the scenario simulated by our models to the actual one from data. It can be seen that the simulated evolution accurately matches the actual evolution for both cases.

We also compare the average speeds of our level-$k$ vehicle models and of actual vehicles in the dataset when traversing T-shaped intersections. The average speeds versus the numbers of interacting vehicles are plotted in Fig.~\ref{fig: traffic_data_speed_compare}, where the $95\%$ confidence intervals of data are indicated by the vertical error bars. It can be seen that the average speeds of our level-$k$ vehicle models are lower than the average speeds of actual vehicles. This is because some vehicles in the dataset drive at a speed that is higher than the speed upper bound $v_{\max} = 5$[m/s] of our models, and is also due to some differences in the road layout and geometry (e.g., three-lane versus two-lane on the left, see Fig.~\ref{fig: t_shape_traffic_compare}). Also, only $56$ scenarios at this T-shaped intersection are contained in the dataset and used to compute the average speed results of the red curve. This causes the relatively large error bars. In contrast, we run 2000 simulation episodes with randomized level-$k$ policy combinations and initial conditions to compute the average speed results of the blue curve. So the error bars are relatively small. In summary, similar trends of average speeds versus numbers of interacting vehicles are exhibited between the simulation results of our level-$k$ vehicle models and the traffic data. Also note that the $95\%$ confidence intervals of our models are contained in the $95\%$ confidence intervals of the traffic data.}

\subsection{Evaluation and calibration of autonomous vehicle control approaches}
\label{sec: evaluation and calibration}

We test the two AV control approaches described in Section~\ref{sec: VV} using a simulation environment constructed based on level-$k$ vehicle models.

For the first approach of adaptive control based on level-$k$ models, we use the same sampling period $\Delta t$, action set $\mathcal{U}$, reward function including the weight vector $\mathbf{w}$, planning horizon $N$, and discount factor $\lambda$ as those used for the level-$k$ vehicle models. In the level estimation algorithm \eqref{equ:level_estimate_1}-\eqref{equ:level_estimate_3}, we consider the model set $\mathcal{K} = \{1,2\}$ and the update step size $\beta = 0.6$.

When training the explicit approximation $\hat{\pi}_{\text{a}}$ to the policy $\pi_{\text{a}}$ that is algorithmically determined by \eqref{equ:adaptive_rhc}, we use the neural network. The accuracy of the obtained $\hat{\pi}_{\text{a}}$ in terms of matching $\pi_{\text{a}}$ is $98.8 \%$ on the training dataset and is $98.6 \%$ on a test dataset of $30 \%$ additional data points that are not used for training.

\begin{figure}[ht]
\begin{center}
\begin{picture}(200.0, 320.0)
%%%%%%%%%%%%%%%%%%%%%%%%%%%%%%
\put(  -22,  229){\epsfig{file=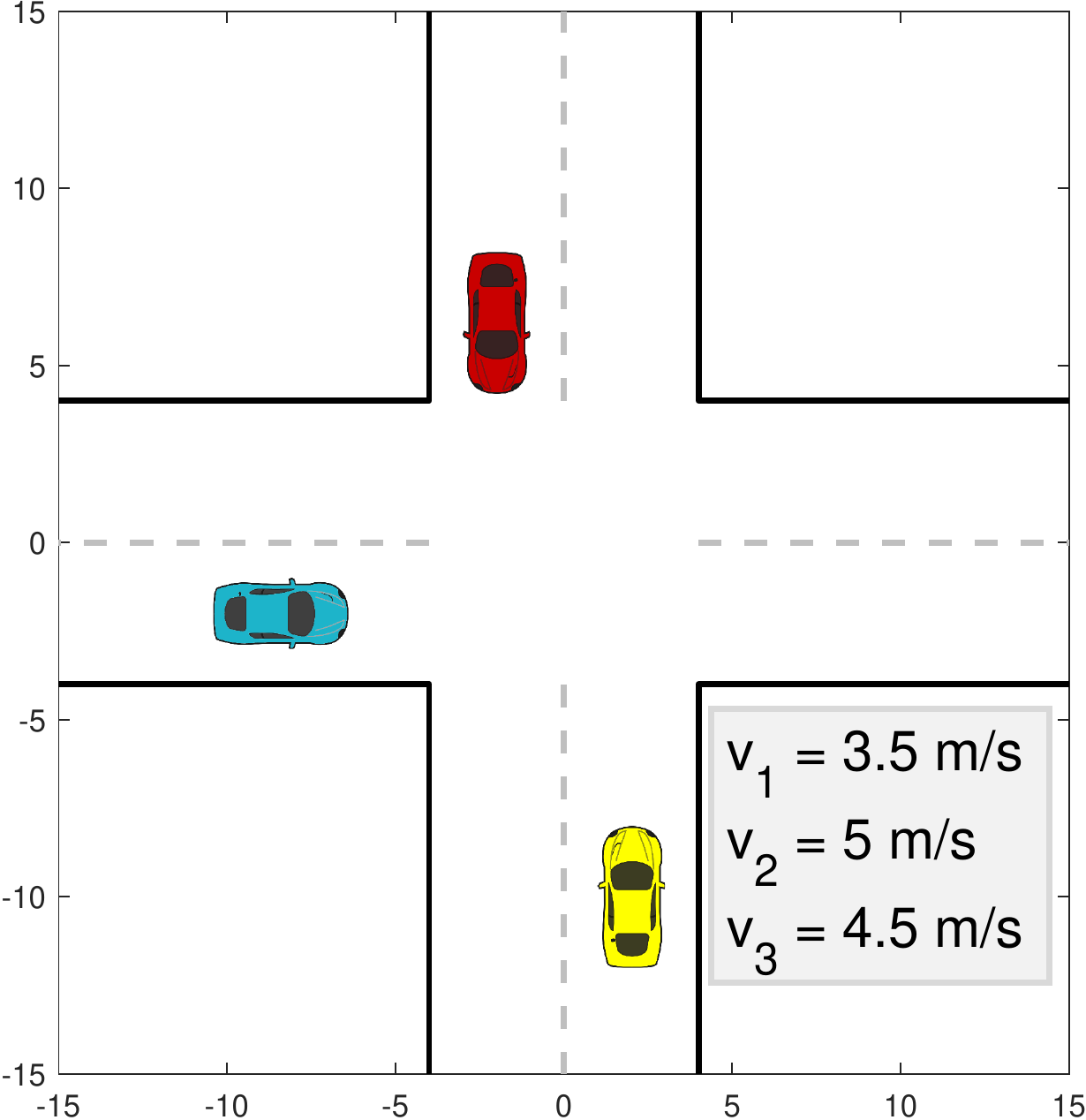,width = 0.33 \linewidth, trim=0.6cm 0.4cm 0cm 0cm,clip}}  %%%
\put(  60,  229){\epsfig{file=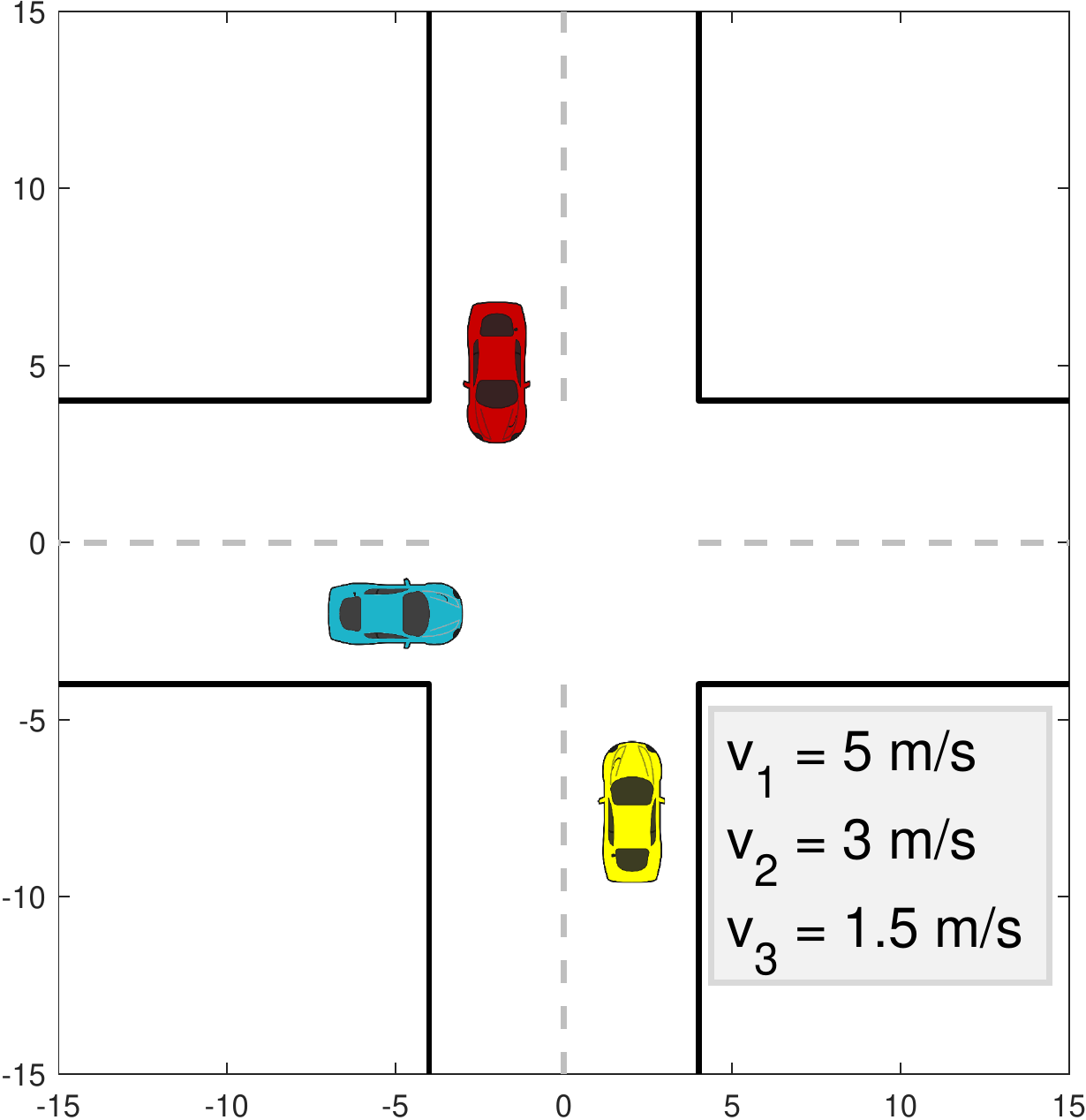, width = 0.33\linewidth, trim=0.6cm 0.4cm 0cm 0cm,clip}}

\put(  142,  229){\epsfig{file=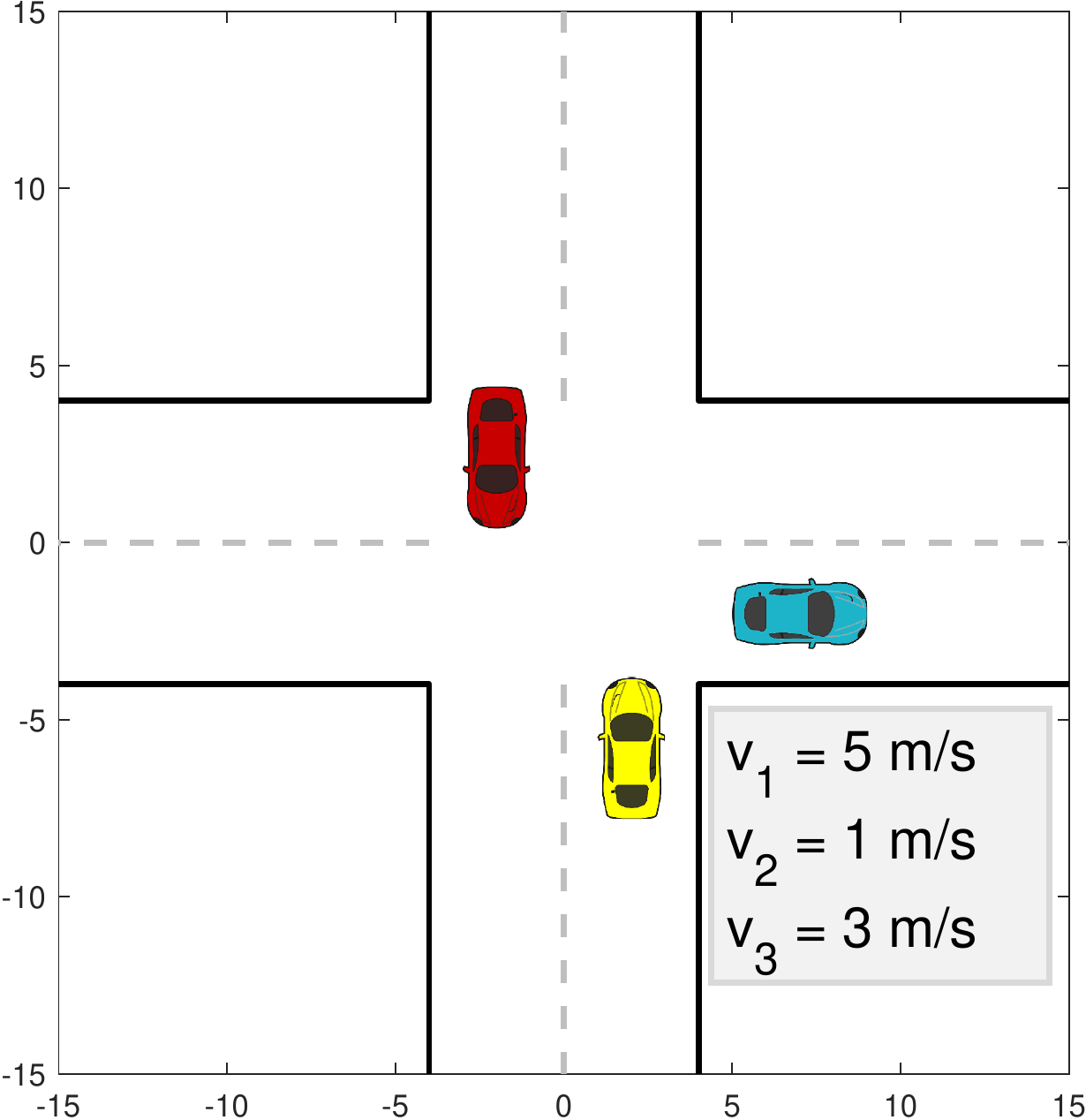, width = 0.33\linewidth, trim=0.6cm 0.4cm 0cm 0cm,clip}}
%%%%%%%%%%%%%%%%%%%%%%%%%%%%%%
\put(  -22,  142){\epsfig{file=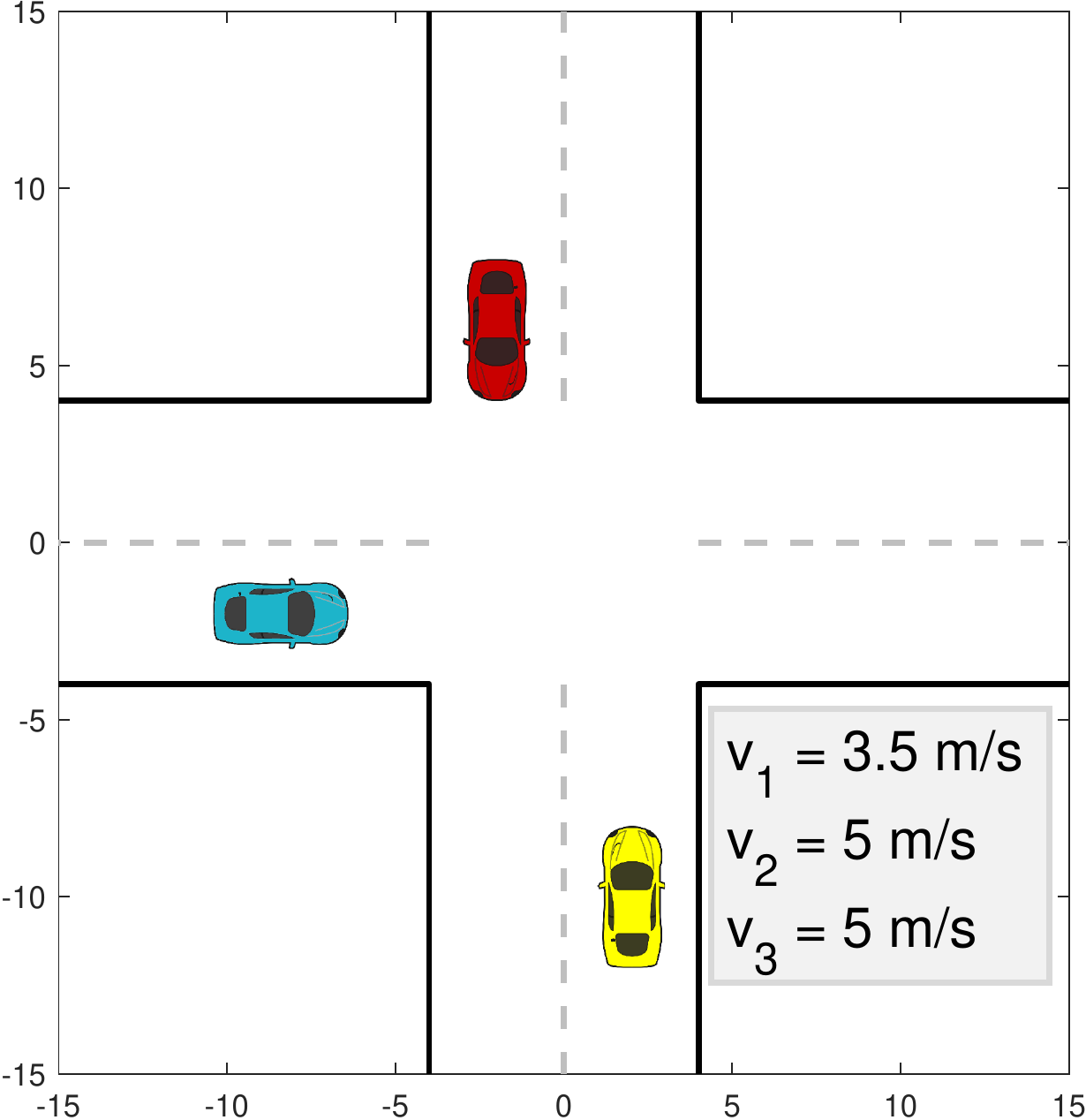,width = 0.33 \linewidth, trim=0.6cm 0.4cm 0cm 0cm,clip}}  %%%
\put(  60,  142){\epsfig{file=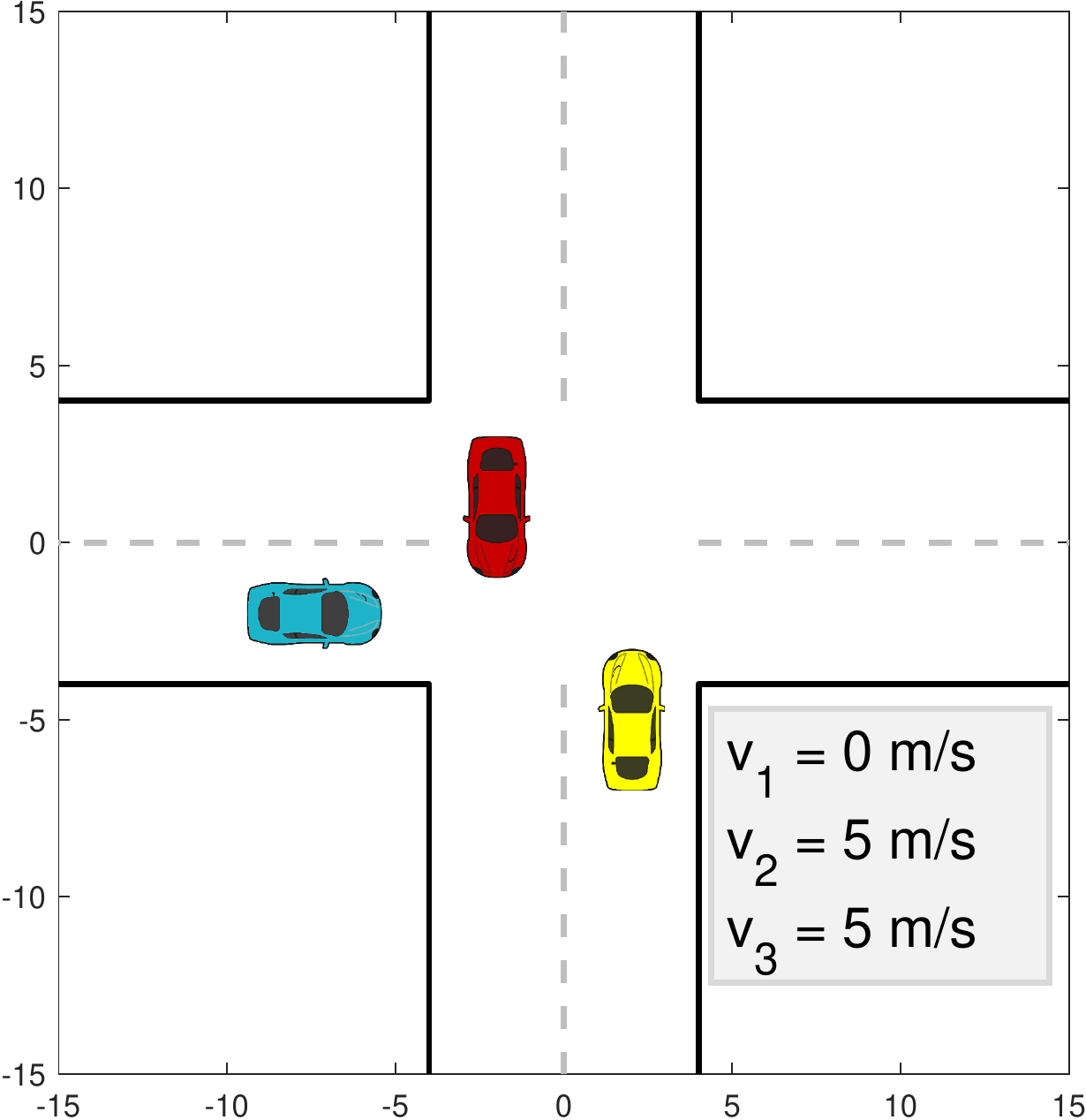, width = 0.33\linewidth, trim=0.6cm 0.4cm 0cm 0cm,clip}}

\put(  142,  142){\epsfig{file=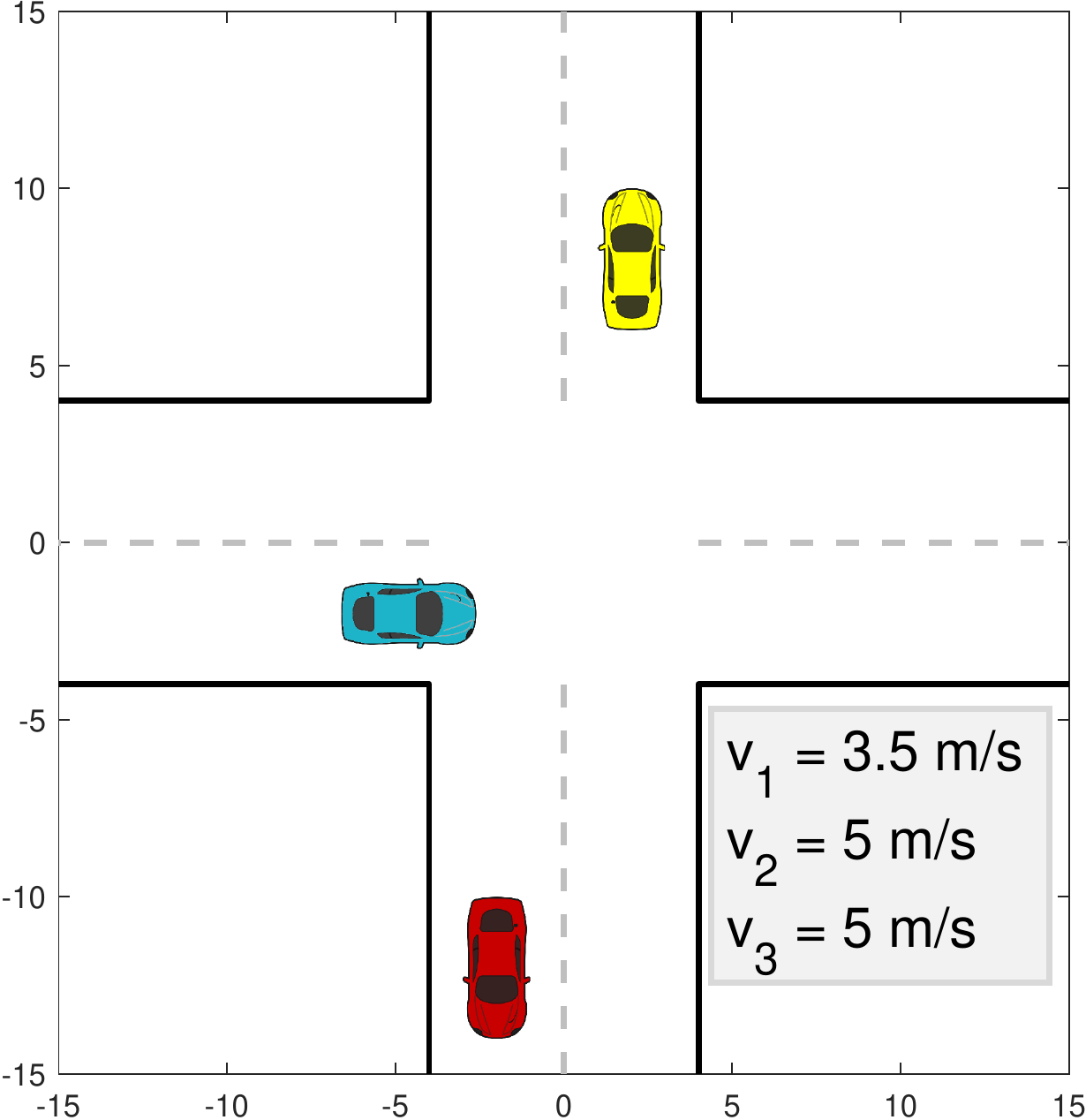, width = 0.33\linewidth, trim=0.6cm 0.4cm 0cm 0cm,clip}}

%%%%%%%%%%%%%%%%%%%%%%%%%%%%%%%%%%%%%%%%%%%%%
\put(  -22,  55){\epsfig{file=media/fourway_opp_1_2_step_12.pdf,width = 0.33 \linewidth, trim=0.6cm 0.4cm 0cm 0cm,clip}}  %%%
\put(  60,  55){\epsfig{file=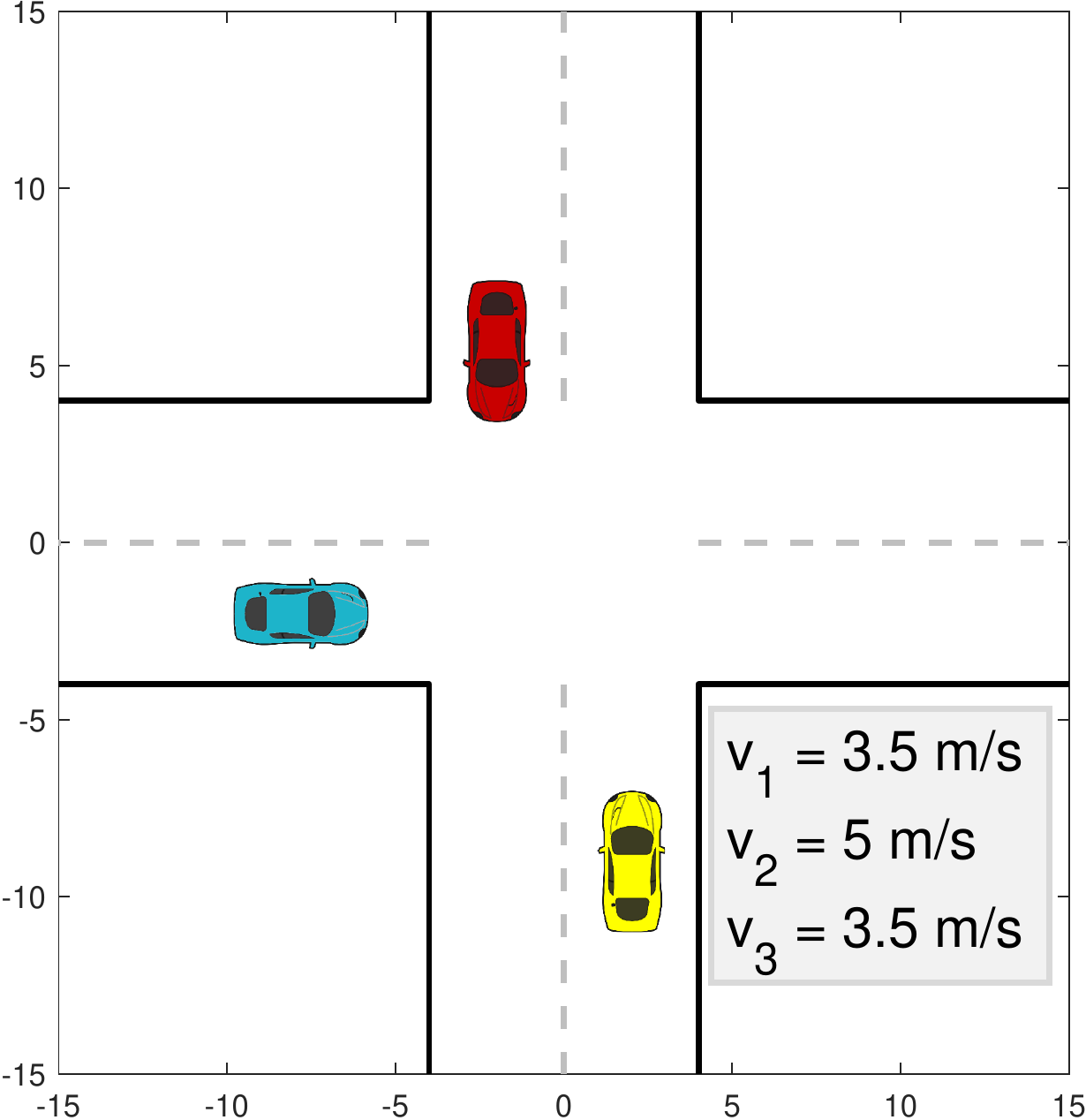, width = 0.33\linewidth, trim=0.6cm 0.4cm 0cm 0cm,clip}}
\put(  142,  55){\epsfig{file=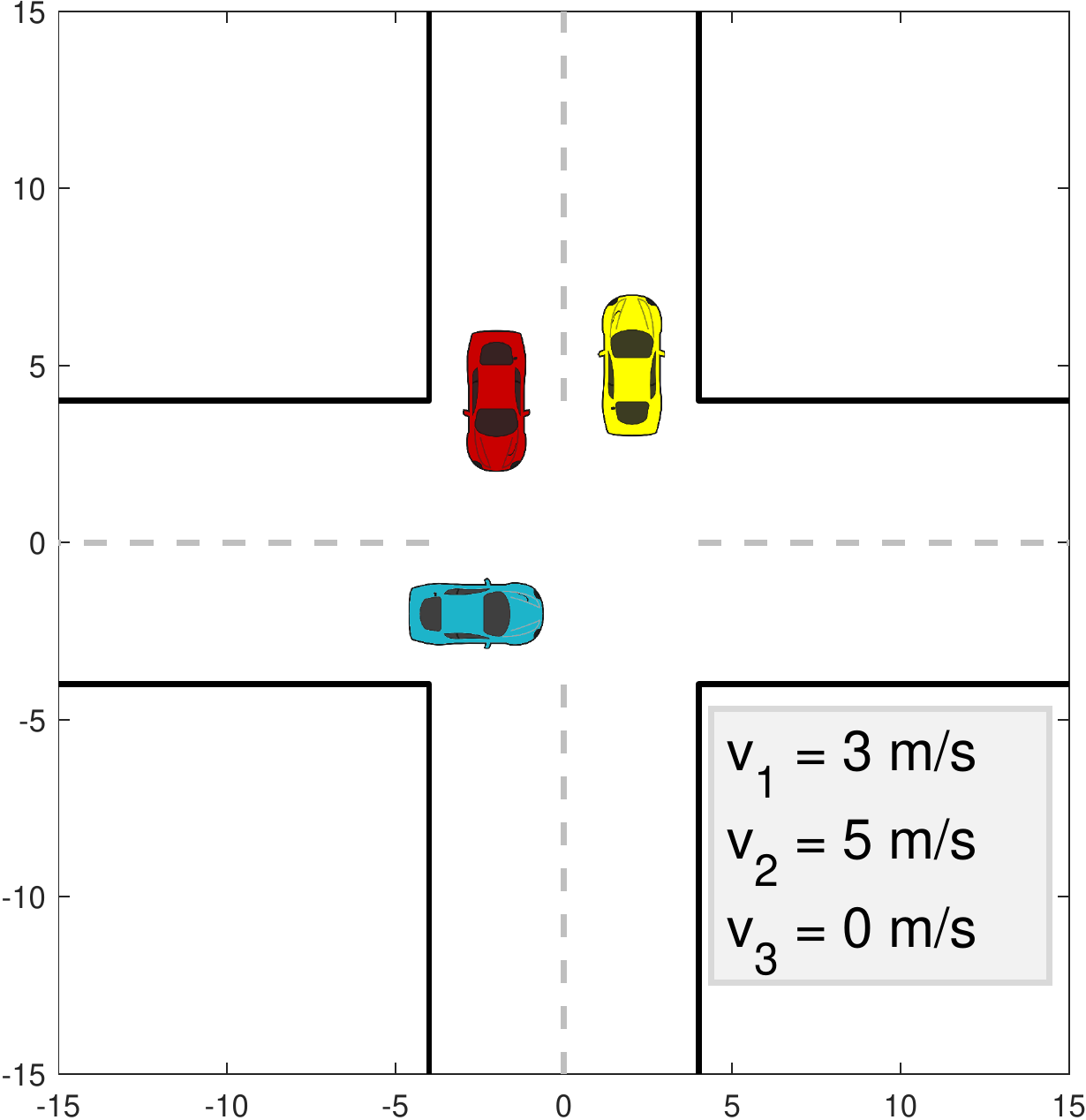, width = 0.33\linewidth, trim=0.6cm 0.4cm 0cm 0cm,clip}}
%%%%%%%%%%%%%%%%%%%%%%
\put(  -22,  -5){\epsfig{file=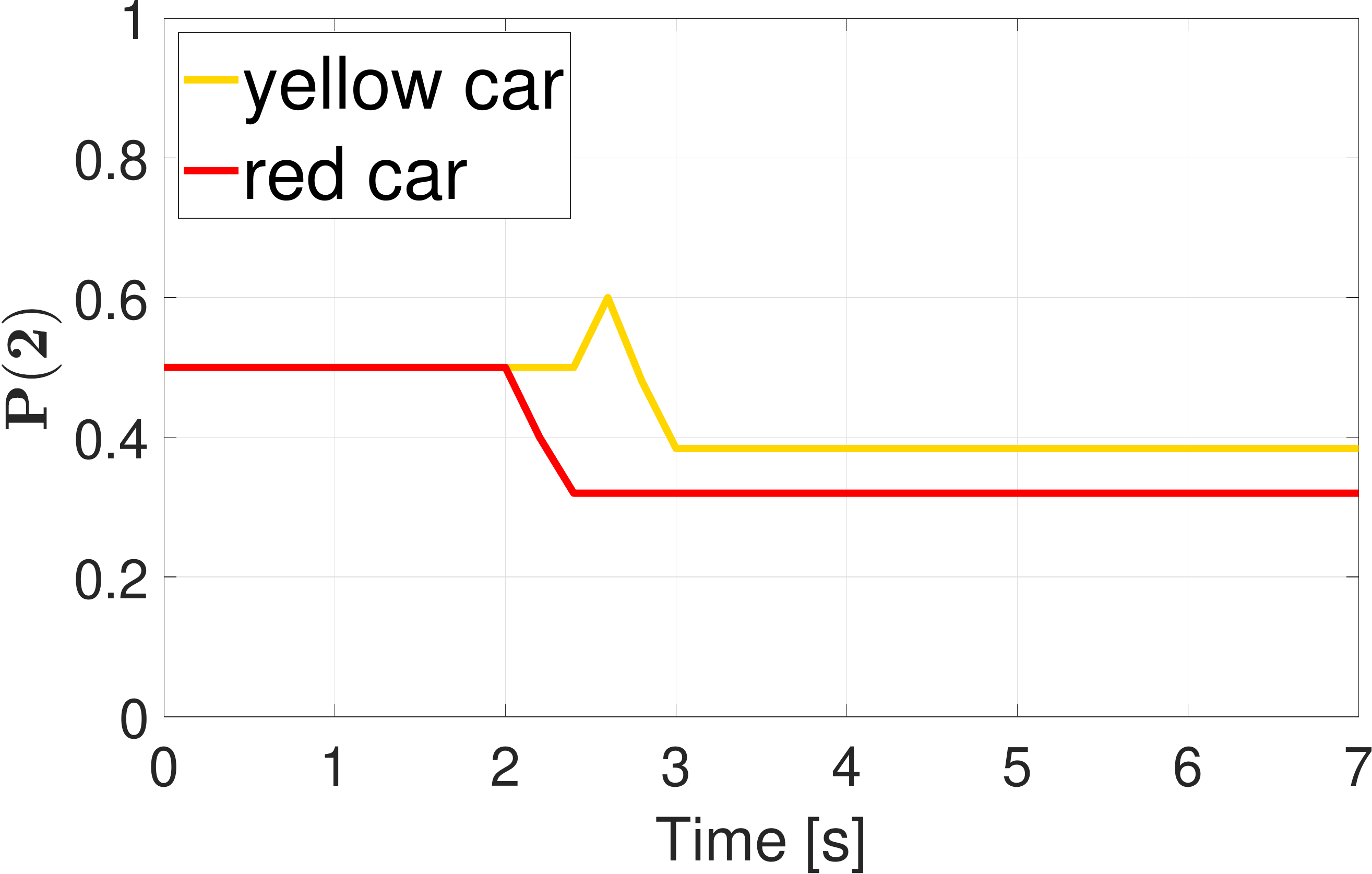,width = 0.32 \linewidth, trim=0.0cm 0.0cm 0cm 0cm,clip}}  %%%
\put(  60,  -5){\epsfig{file=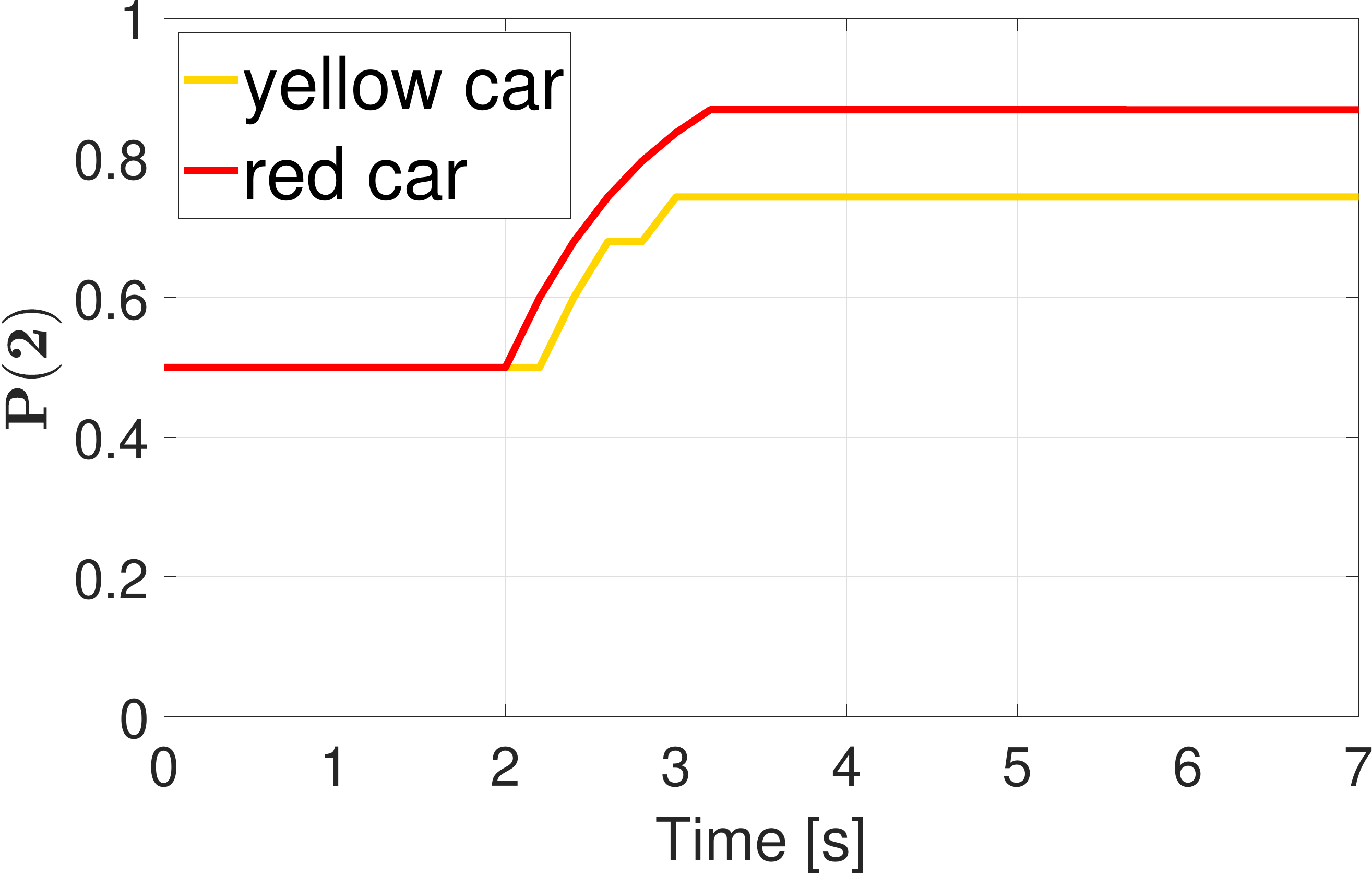, width = 0.32\linewidth, trim=0.0cm 0.0cm 0cm 0cm,clip}}
\put(  142,  -5){\epsfig{file=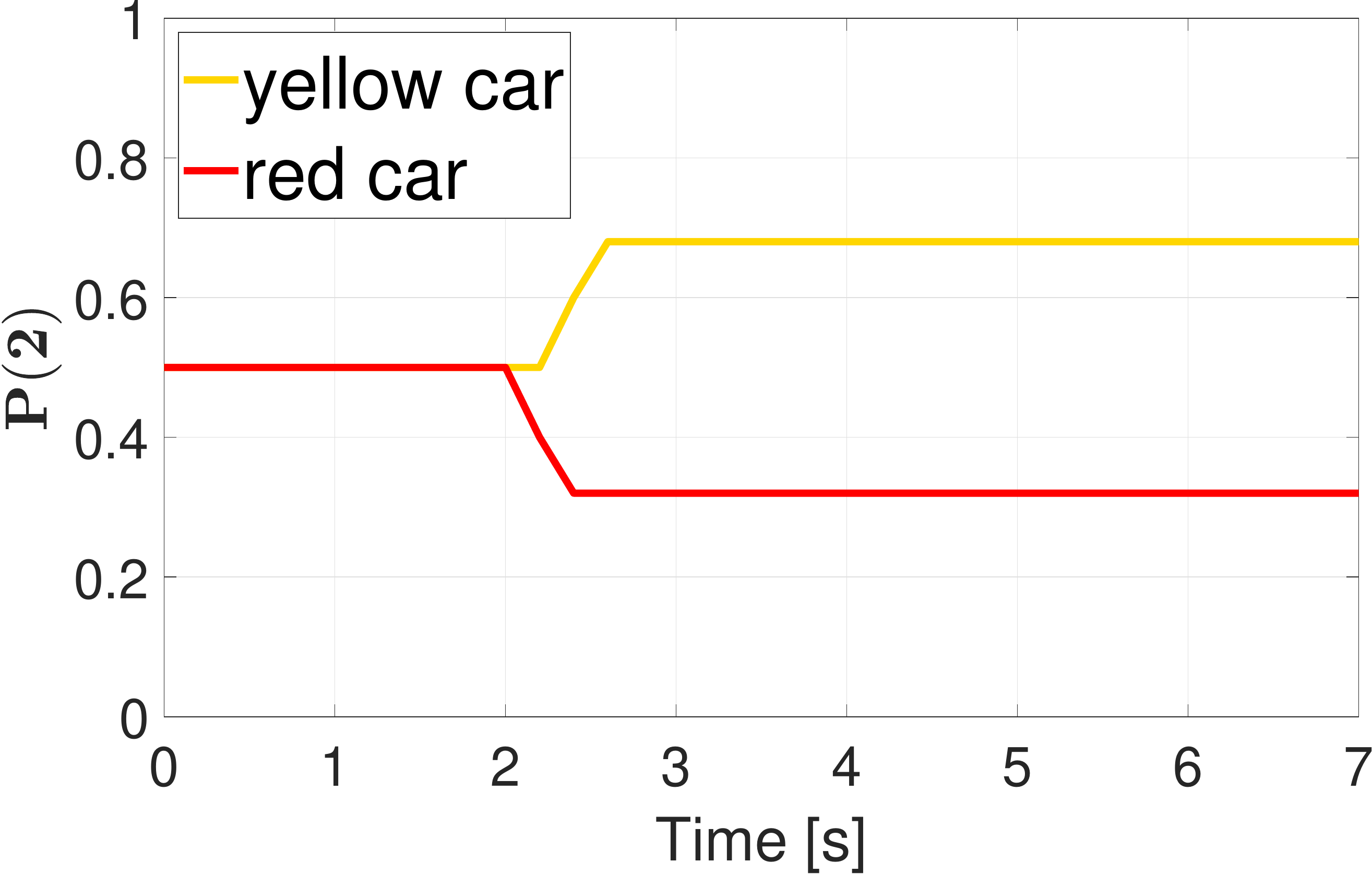, width = 0.32\linewidth, trim=0.0cm 0.0cm 0cm 0cm,clip}}
%%%%%%%%%%%%%%%%%%%%%%
\small
\put(35,301){(a-1)}
\put(117,301){(a-2)}
\put(199,301){(a-3)}
\put(35,214){(b-1)}
\put(117,214){(b-2)}
\put(199,214){(b-3)}
\put(35,127){(c-1)}
\put(117,127){(c-2)}
\put(199,127){(c-3)}
\put(35,48){(a-4)}
\put(117,48){(b-4)}
\put(199,48){(c-4)}
\normalsize
\end{picture}
\end{center}
       \caption{Interactions of the autonomous ego vehicle (blue) controlled by the adaptive control approach with level-$k$ vehicles at the four-way intersection. (a-1)-(a-3) show three sequential steps in a simulation where the autonomous ego vehicle interacts with two level-$1$ vehicles, and (a-4) shows the time histories of the two vehicles' level estimates where $\mathbb{P}(2) = \mathbb{P}(k = 2)$ denotes the ego vehicle's belief in the level-$2$ model; (b-1)-(b-4) show those of the autonomous ego vehicle interacting with two level-$2$ vehicles; (c-1)-(c-4) show those of the autonomous ego vehicle interacting with a level-$1$ vehicle (red) and a
       level-$2$ vehicle (yellow); $v_1$, $v_2$ and $v_3$ are the speeds of the blue, yellow and red vehicles, respectively.}
      \label{fig: adaptive-fourway}
\end{figure}

\begin{figure}[ht]
\begin{center}
\begin{picture}(200.0, 270.0)
%%%%%%%%%%%%%%%%%%%%%%%%%%%%%%
\put(  -22,  195){\epsfig{file=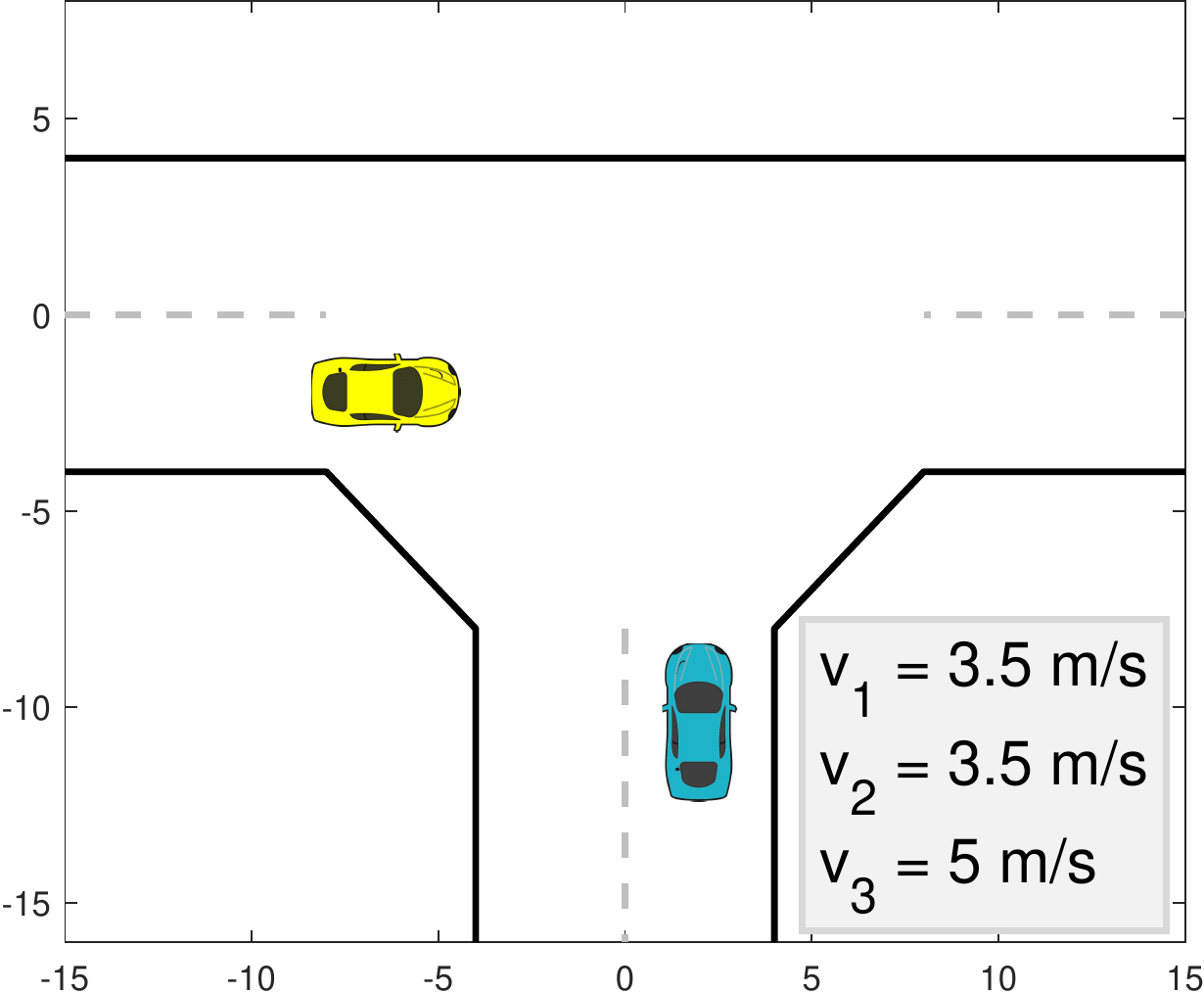,width = 0.33 \linewidth, trim=0.6cm 0.4cm 0cm 0cm,clip}}  %%%
\put(  60,  195){\epsfig{file=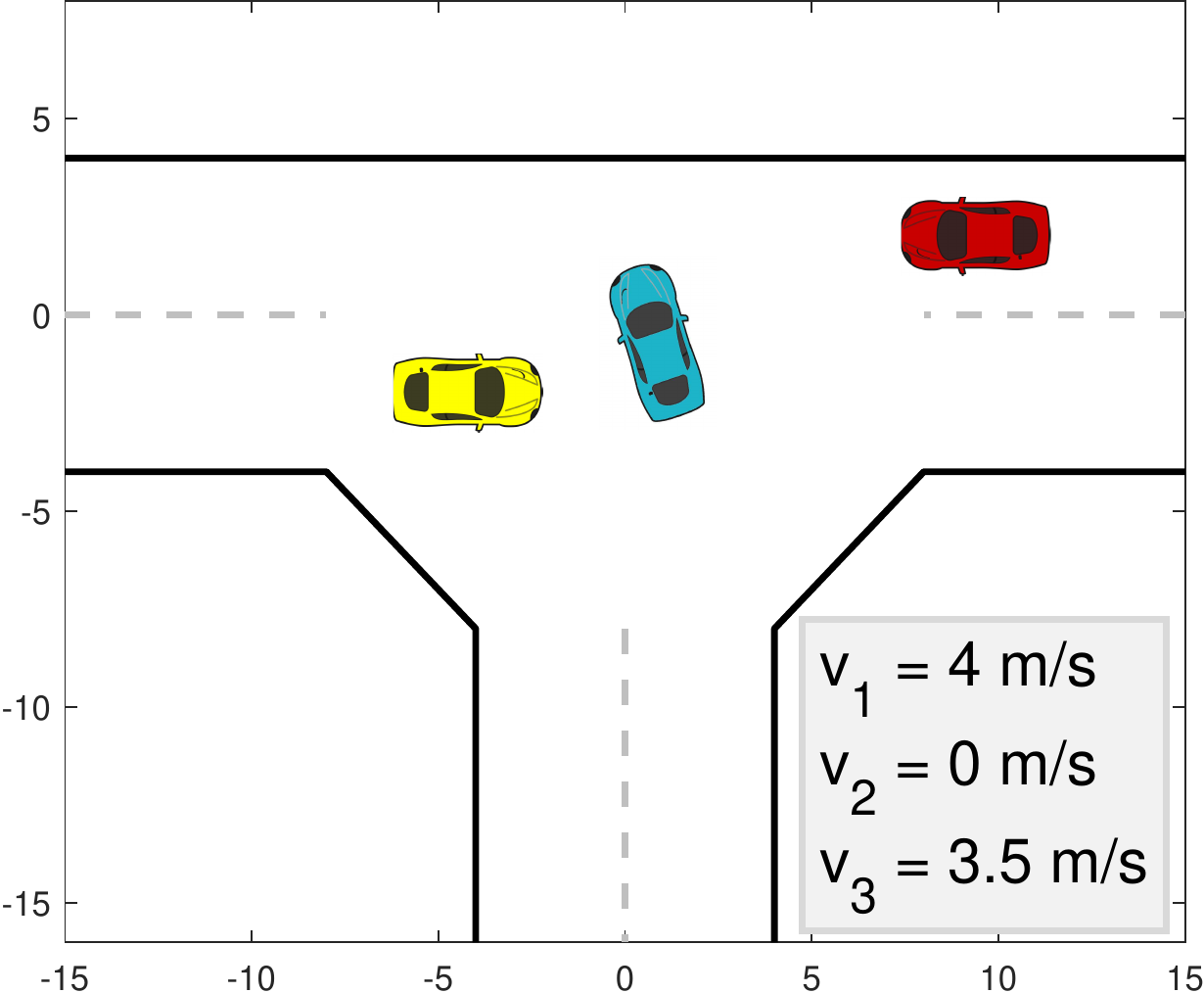, width = 0.33\linewidth, trim=0.6cm 0.4cm 0cm 0cm,clip}}

\put(  142,  195){\epsfig{file=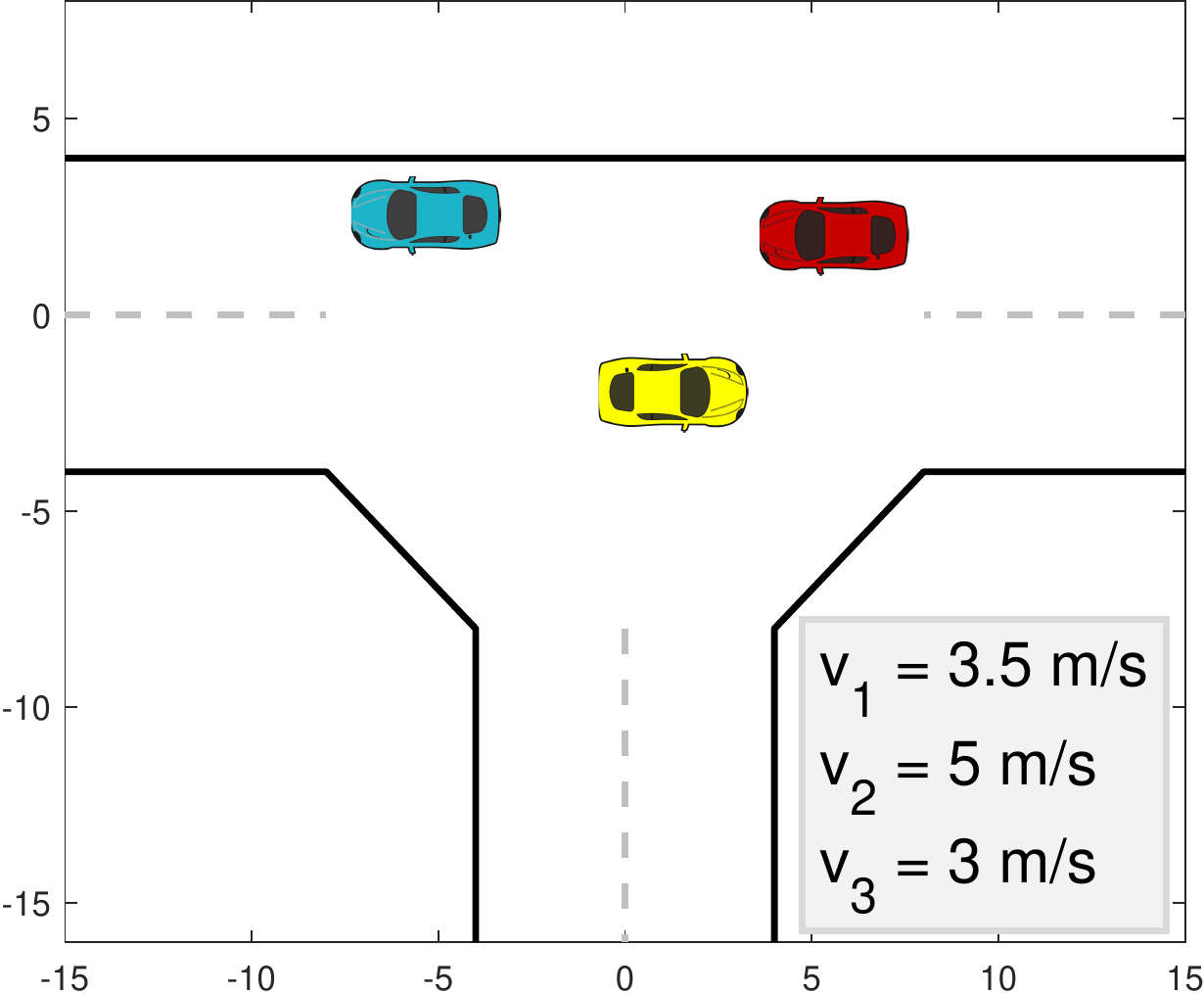, width = 0.33\linewidth, trim=0.6cm 0.4cm 0cm 0cm,clip}}
%%%%%%%%%%%%%%%%%%%%%%%%%%%%%%
\put(  -22,  125){\epsfig{file=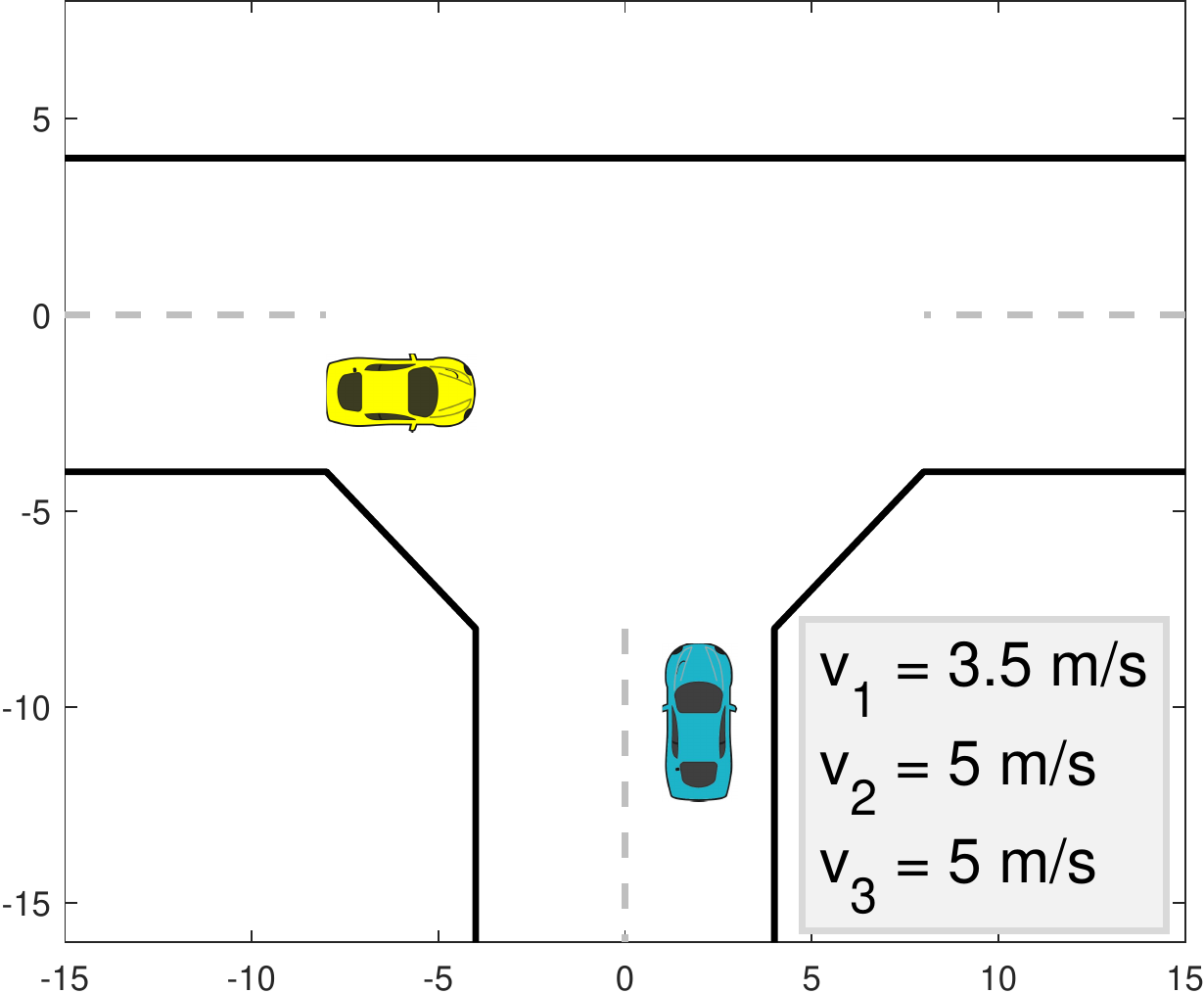,width = 0.33 \linewidth, trim=0.6cm 0.4cm 0cm 0cm,clip}}  %%%
\put(  60,  125){\epsfig{file=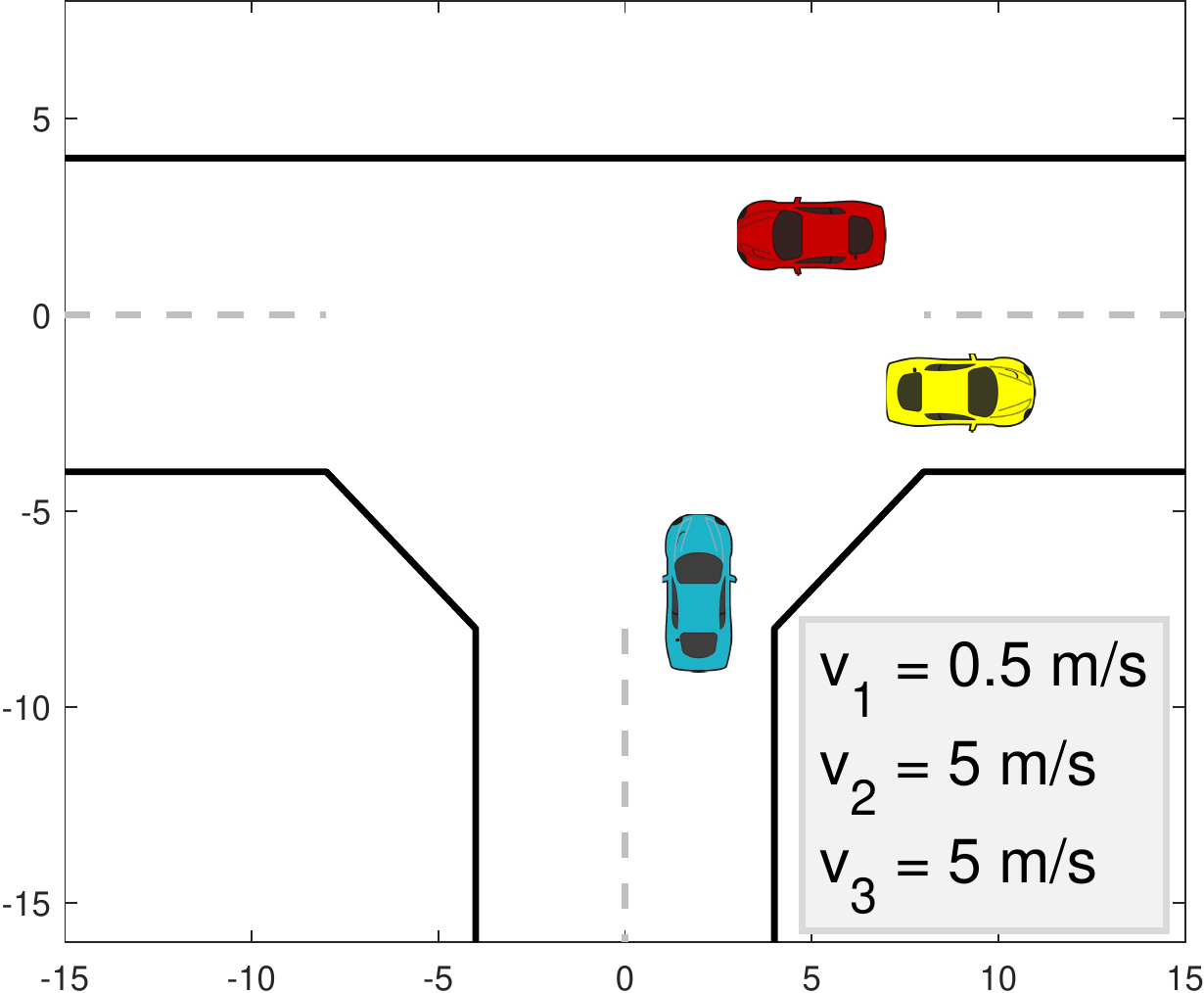, width = 0.33\linewidth, trim=0.6cm 0.4cm 0cm 0cm,clip}}

\put(  142,  125){\epsfig{file=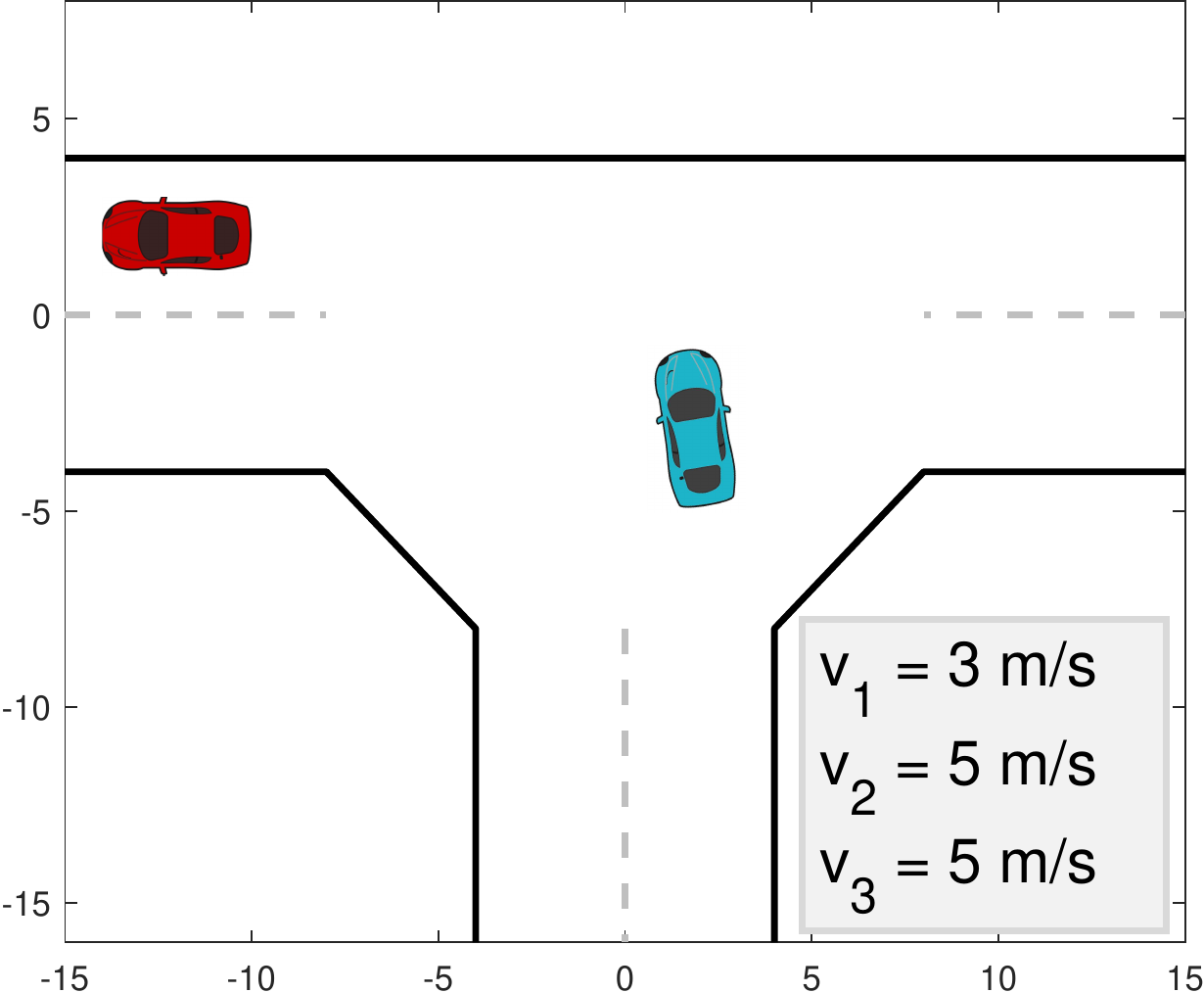, width = 0.33\linewidth, trim=0.6cm 0.4cm 0cm 0cm,clip}}

%%%%%%%%%%%%%%%%%%%%%%%%%%%%%%%%%%%%%%%%%%%%%
\put(  -22,  55){\epsfig{file=media/t_shape_ada_2_2_step_14.pdf,width = 0.33 \linewidth, trim=0.6cm 0.4cm 0cm 0cm,clip}}  %%%
\put(  60,  55){\epsfig{file=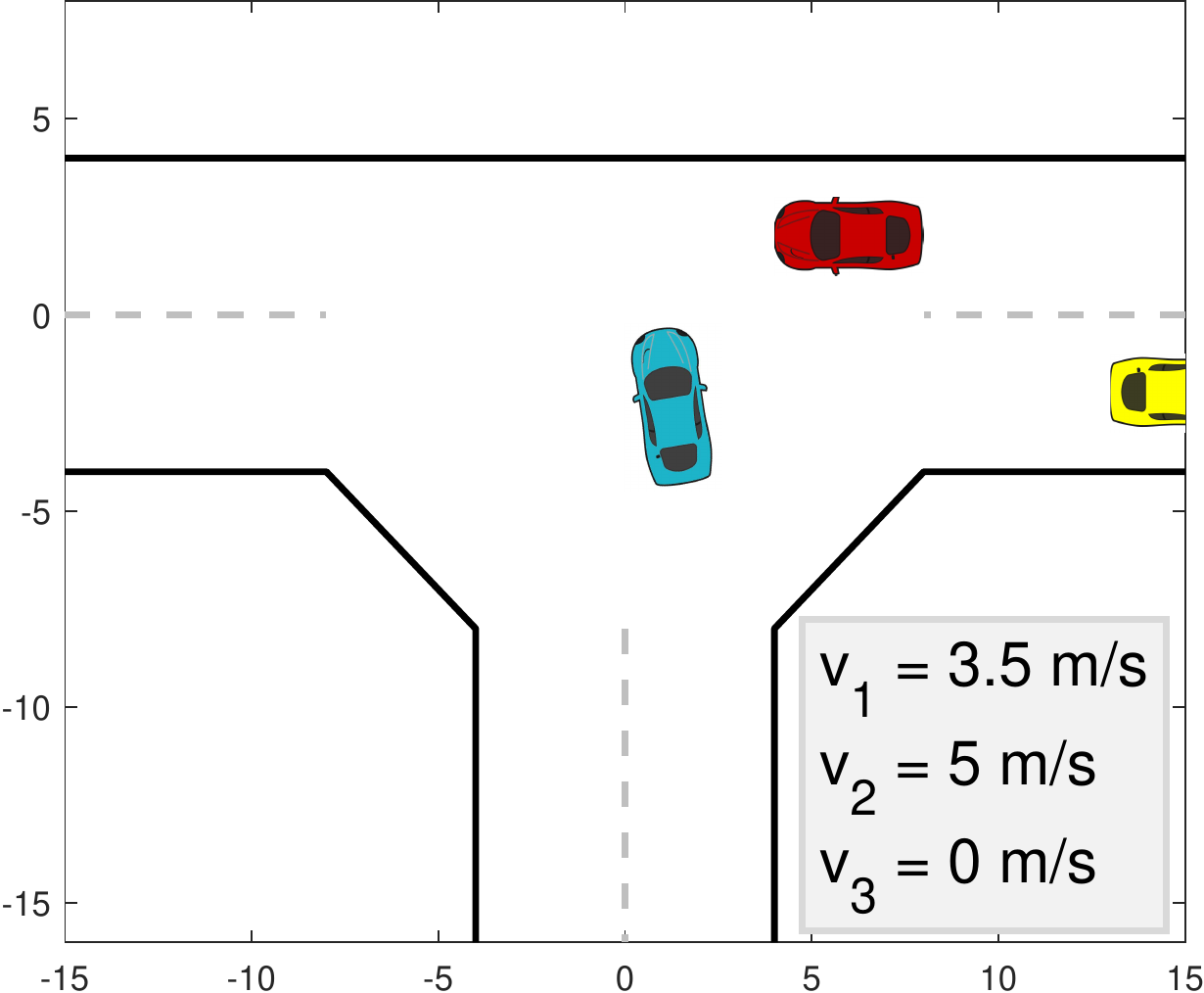, width = 0.33\linewidth, trim=0.6cm 0.4cm 0cm 0cm,clip}}
\put(  142,  55){\epsfig{file=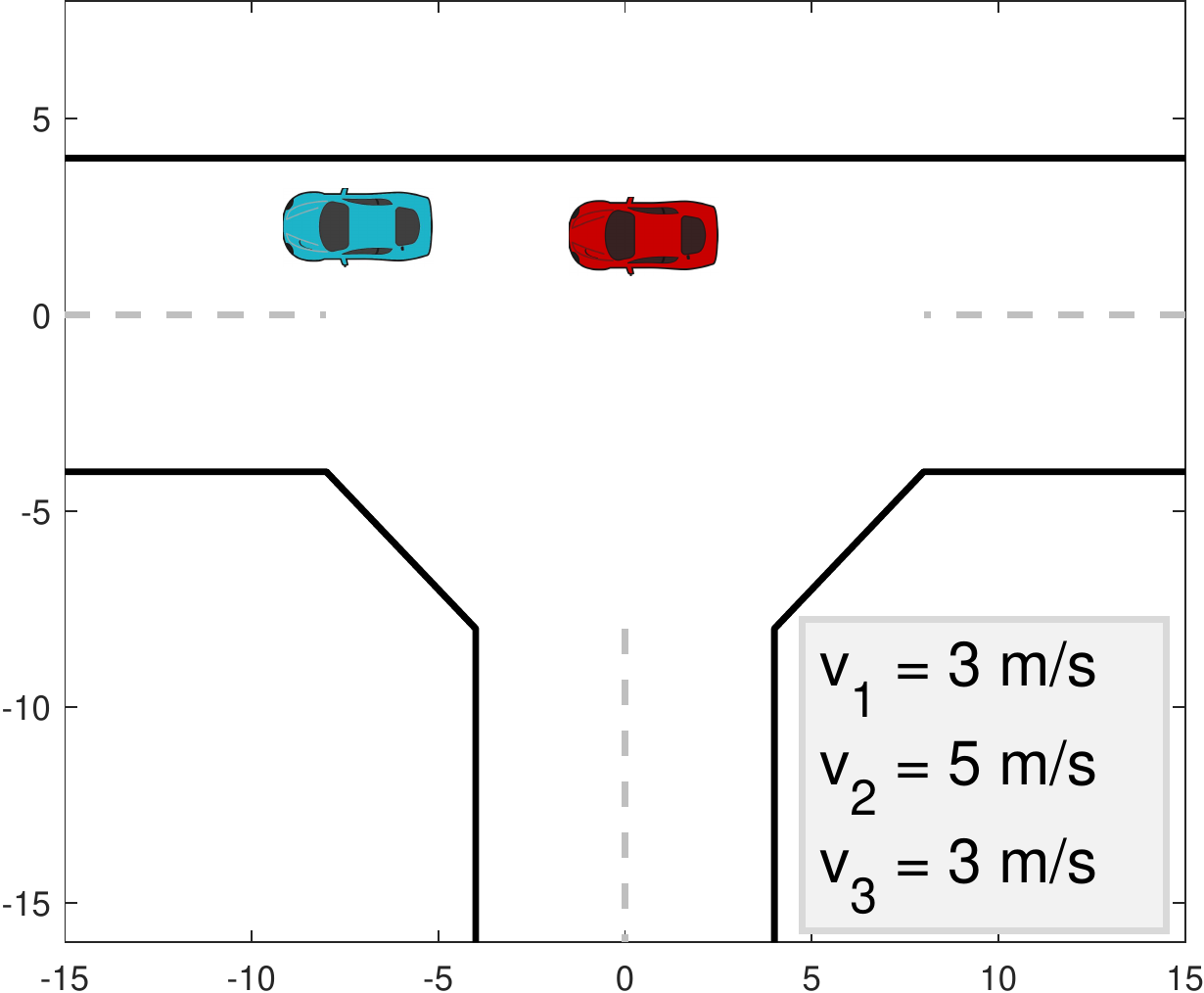, width = 0.33\linewidth, trim=0.6cm 0.4cm 0cm 0cm,clip}}
%%%%%%%%%%%%%%%%%%%%%%
\put(  -22,  -5){\epsfig{file=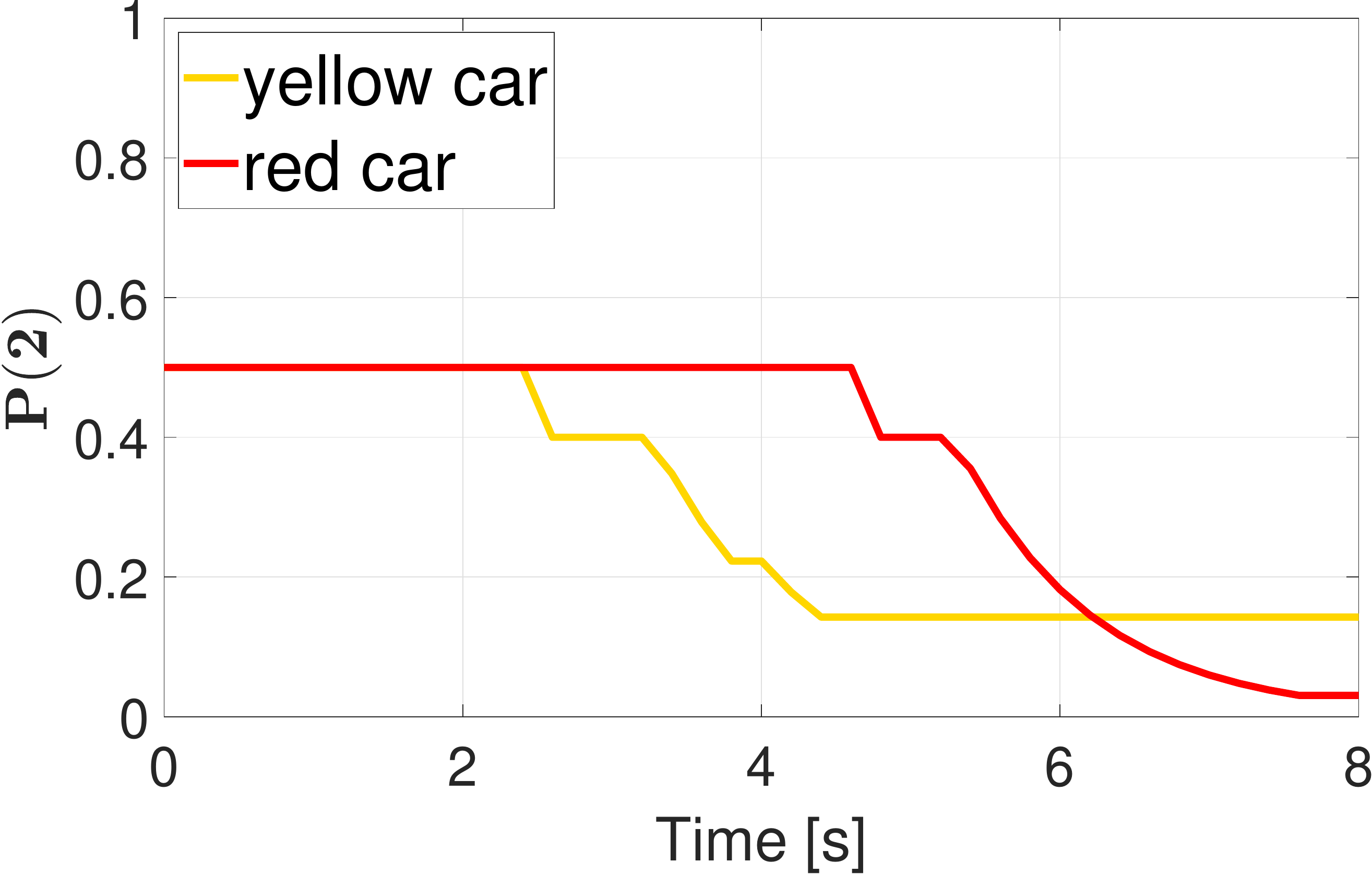,width = 0.32 \linewidth, trim=0.0cm 0.3cm 0cm 0cm,clip}}  %%%
\put(  60,  -5){\epsfig{file=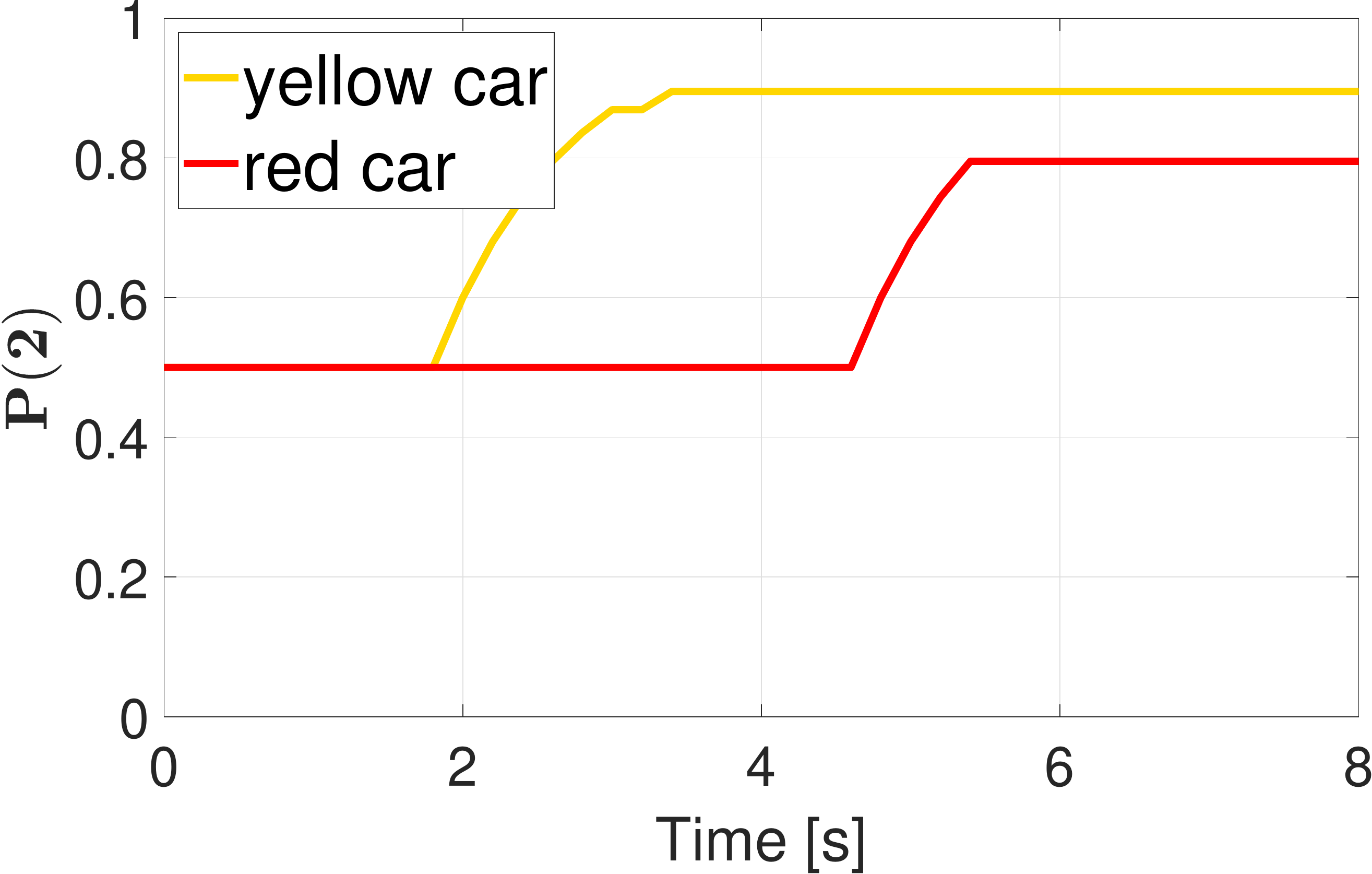, width = 0.32\linewidth, trim=0.0cm 0.3cm 0cm 0cm,clip}}
\put(  142,  -5){\epsfig{file=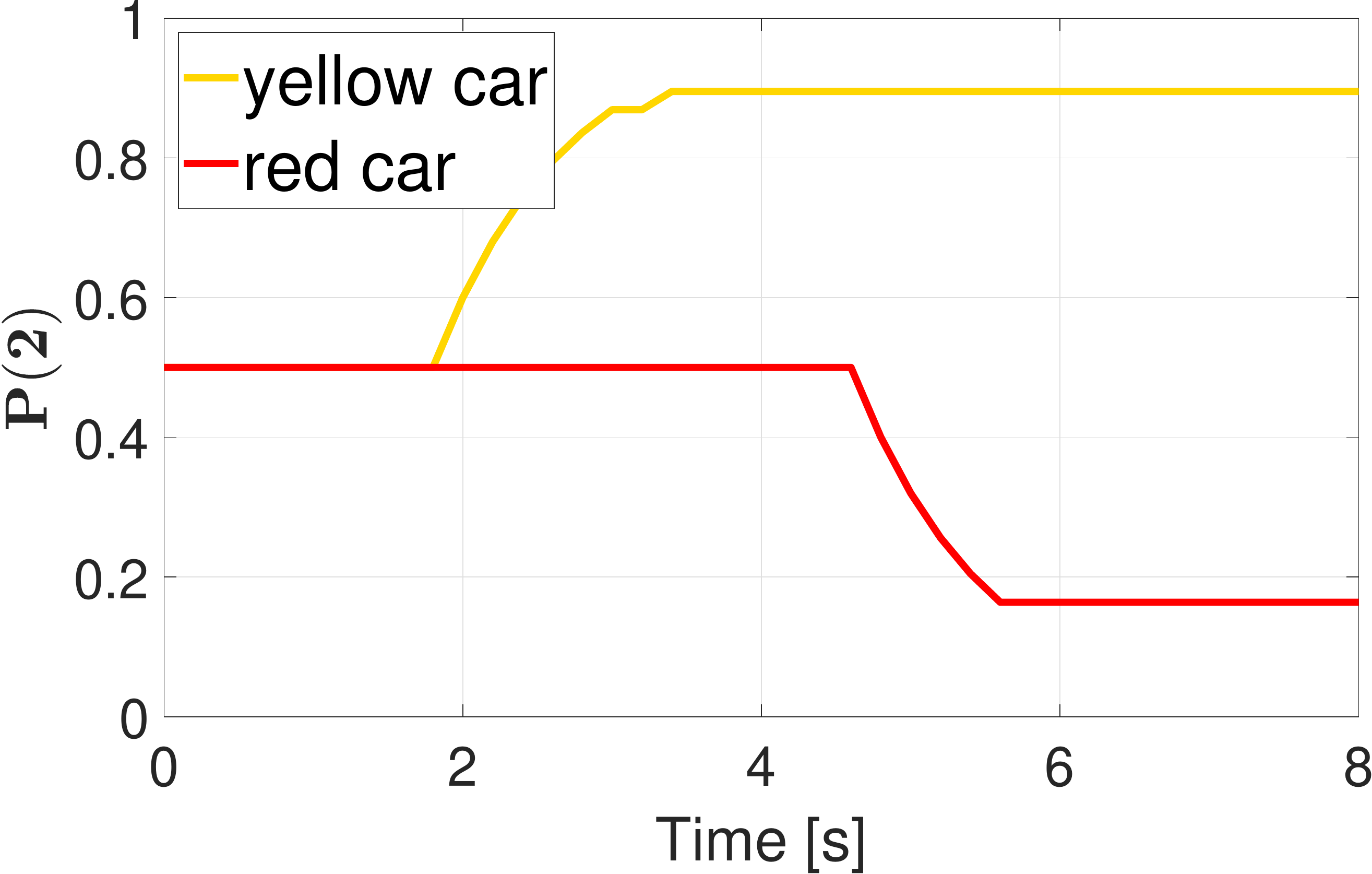, width = 0.32\linewidth, trim=0.0cm 0.3cm 0cm 0cm,clip}}
\small
\put(35,255){(a-1)}
\put(117,255){(a-2)}
\put(199,255){(a-3)}
\put(35,185){(b-1)}
\put(117,185){(b-2)}
\put(199,185){(b-3)}
\put(35,115){(c-1)}
\put(117,115){(c-2)}
\put(199,115){(c-3)}
\put(35,47){(a-4)}
\put(117,47){(b-4)}
\put(199,47){(c-4)}
\normalsize
\end{picture}
\end{center}
       \caption{Interactions of the autonomous ego vehicle (blue) controlled by the adaptive control approach with level-$k$ vehicles at the T-shaped intersection. (a-1)-(a-3) show three sequential steps in a simulation where the autonomous ego vehicle interacts with two level-$1$ vehicles, and (a-4) shows the time histories of the two vehicles' level estimates where $\mathbb{P}(2) = \mathbb{P}(k = 2)$ denotes the ego vehicle's belief in the level-$2$ model; (b-1)-(b-4) show those of the autonomous ego vehicle interacting with two level-$2$ vehicles; (c-1)-(c-4) show those of the autonomous ego vehicle interacting with a level-$1$ vehicle (red) and a
       level-$2$ vehicle (yellow); $v_1$, $v_2$ and $v_3$ are the speeds of the blue, yellow and red vehicles, respectively.}
      \label{fig:  adaptive-threeway}
\end{figure}

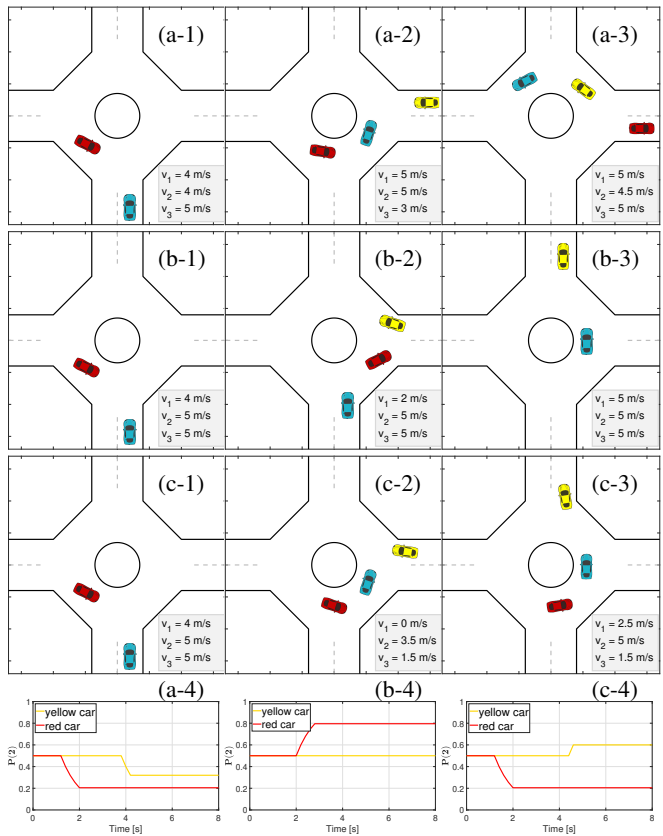
\begin{figure}[ht]
\begin{center}
\begin{picture}(200.0, 314.0)
\put(  -22,  225){\epsfig{file=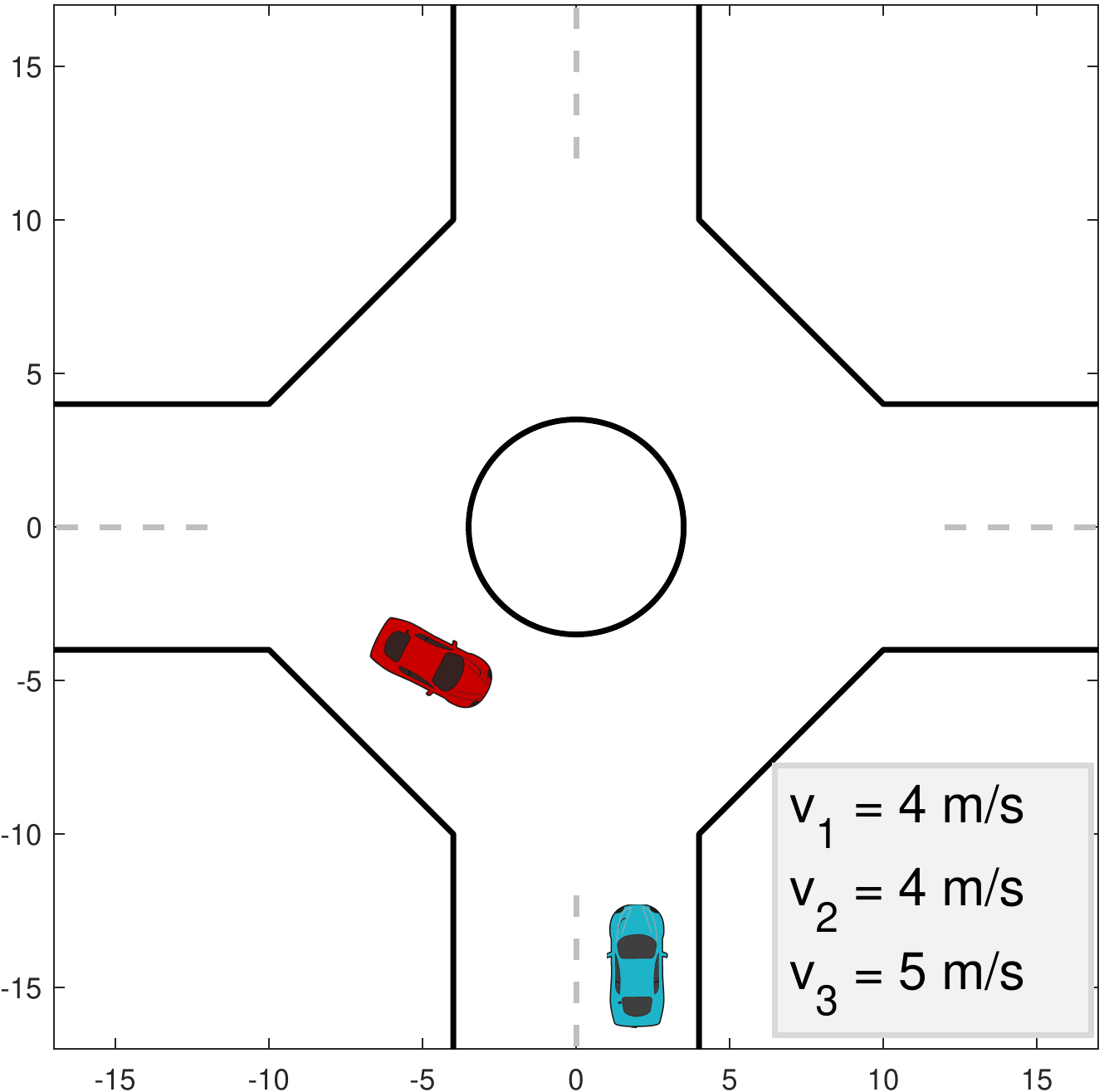,width = 0.33 \linewidth, trim=0.6cm 0.4cm 0cm 0cm,clip}}  %%%
\put(  60,  225){\epsfig{file=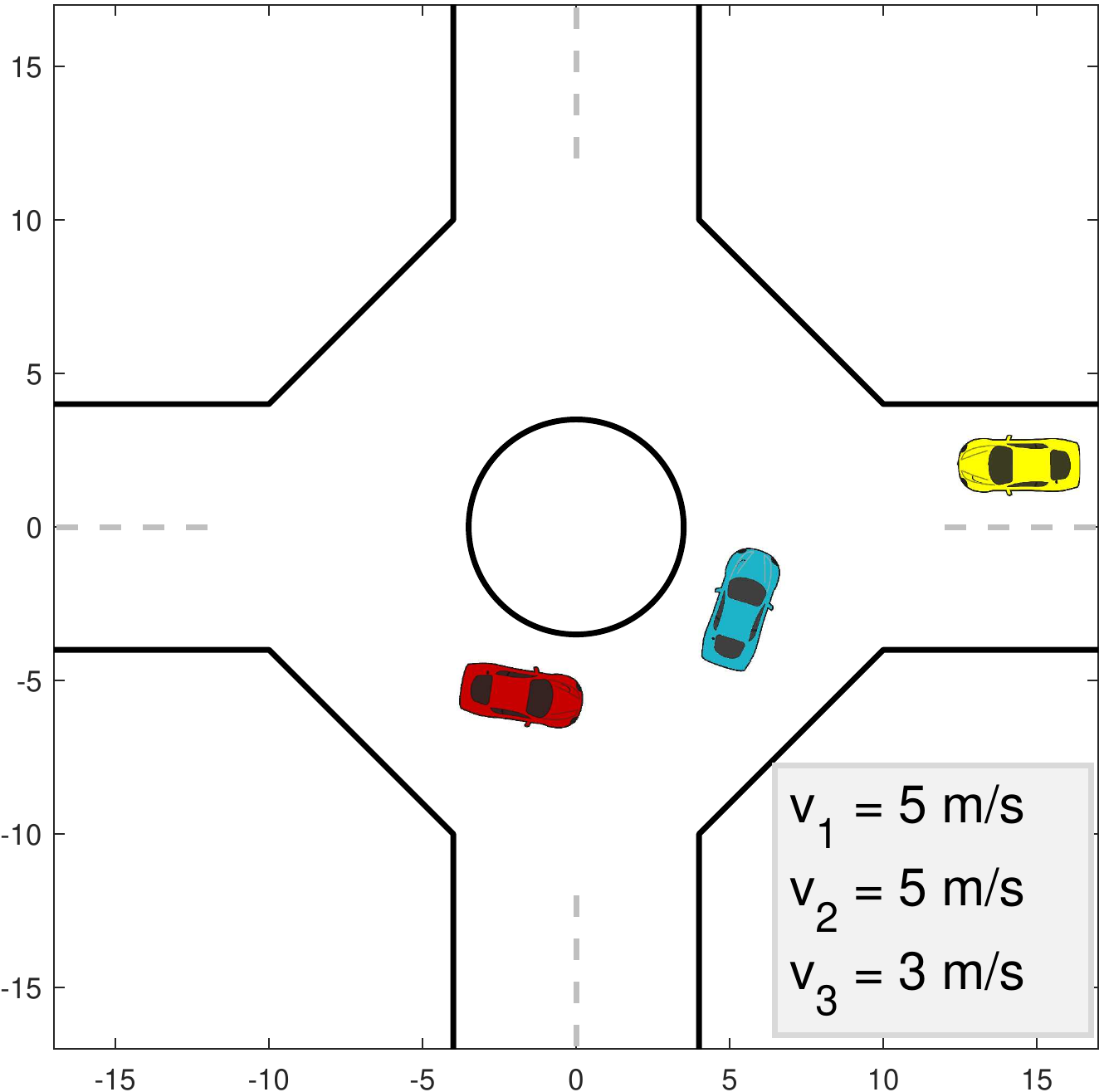, width = 0.33\linewidth, trim=0.6cm 0.4cm 0cm 0cm,clip}}

\put(  142,  225){\epsfig{file=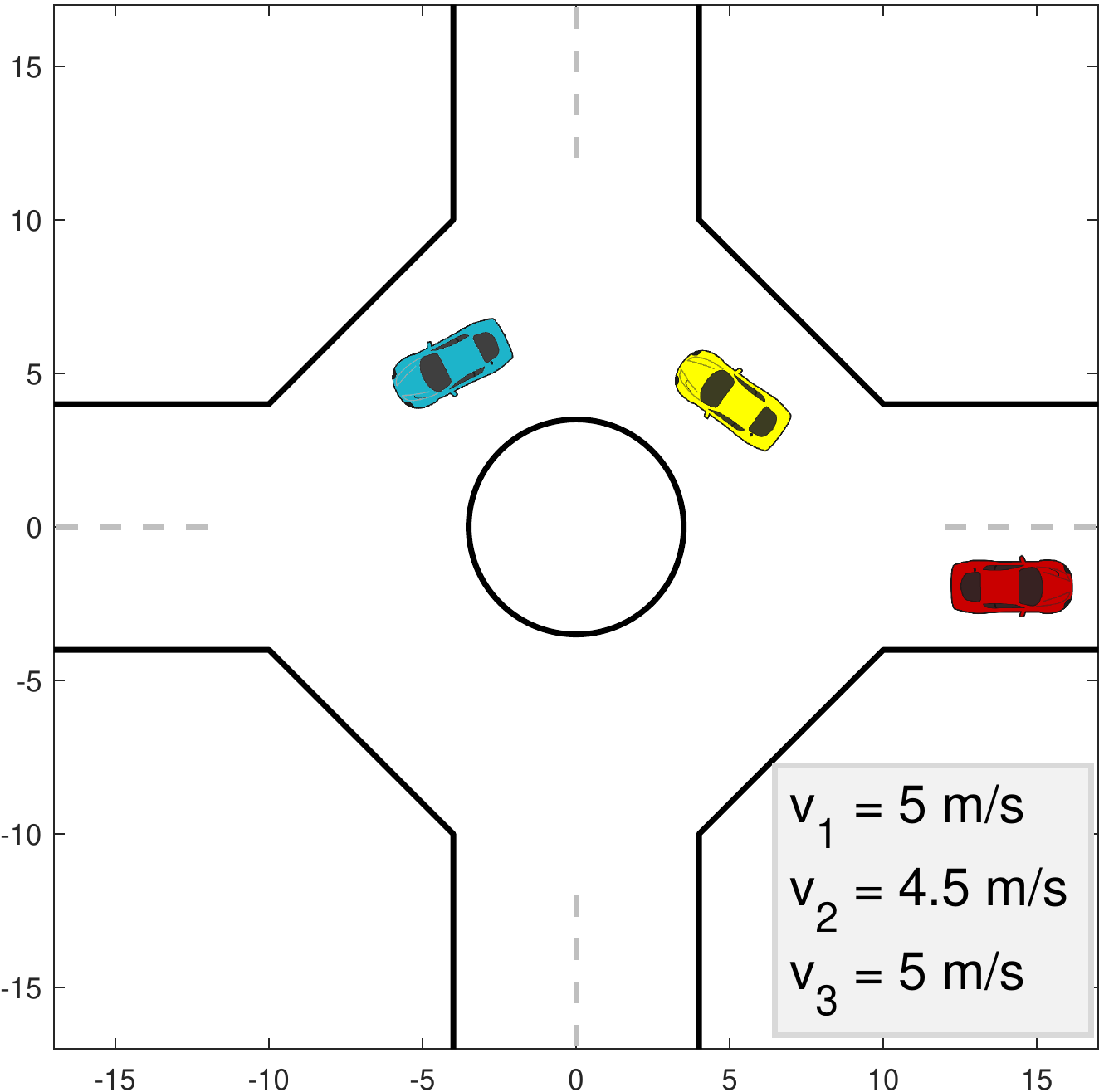, width = 0.33\linewidth, trim=0.6cm 0.4cm 0cm 0cm,clip}}
%%%%%%%%%%%%%%%%%%%%%%%%%%%%%%
\put(  -22,  140){\epsfig{file=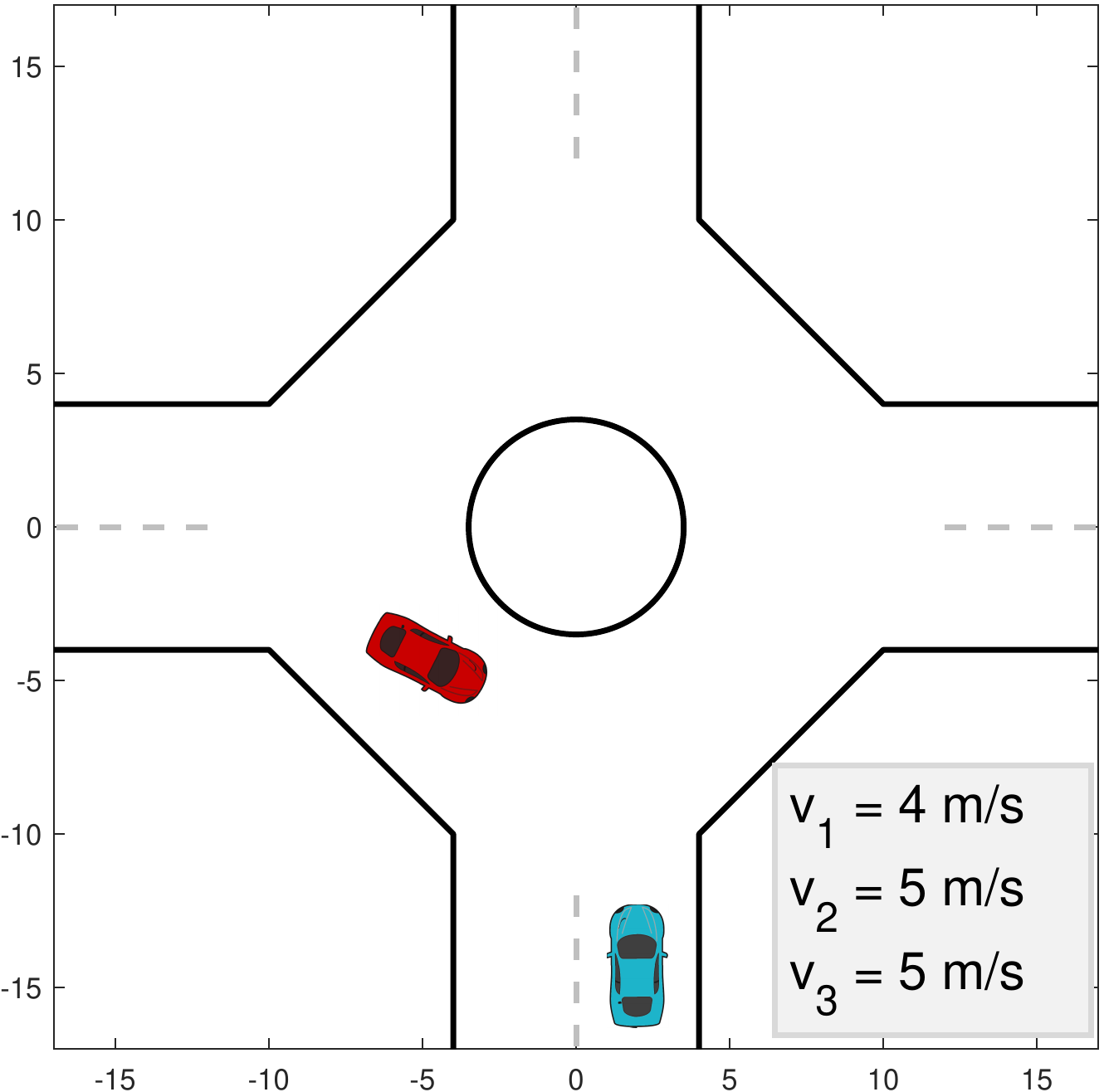,width = 0.33 \linewidth, trim=0.6cm 0.4cm 0cm 0cm,clip}}  %%%
\put(  60,  140){\epsfig{file=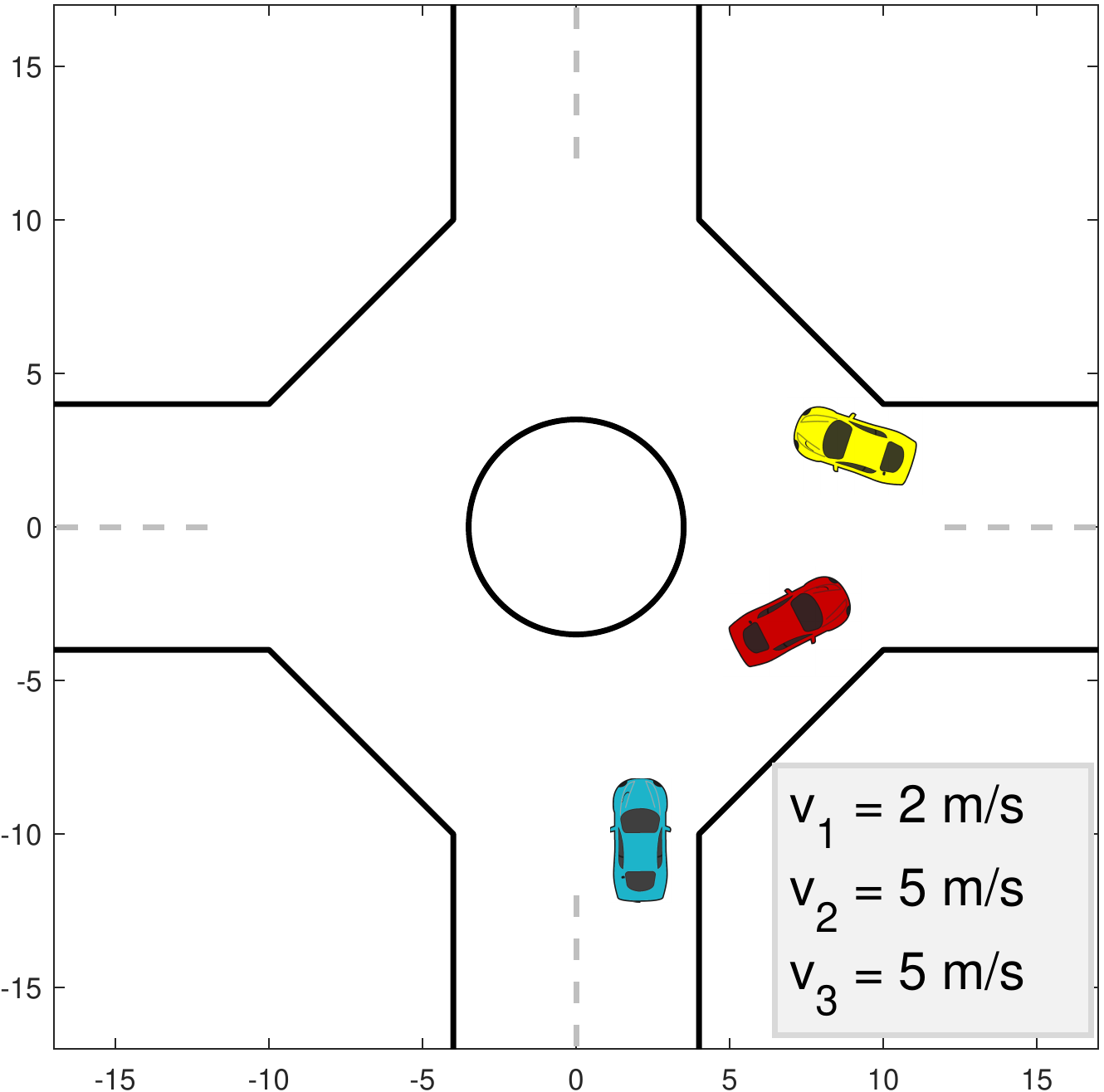, width = 0.33\linewidth, trim=0.6cm 0.4cm 0cm 0cm,clip}}

\put(  142,  140){\epsfig{file=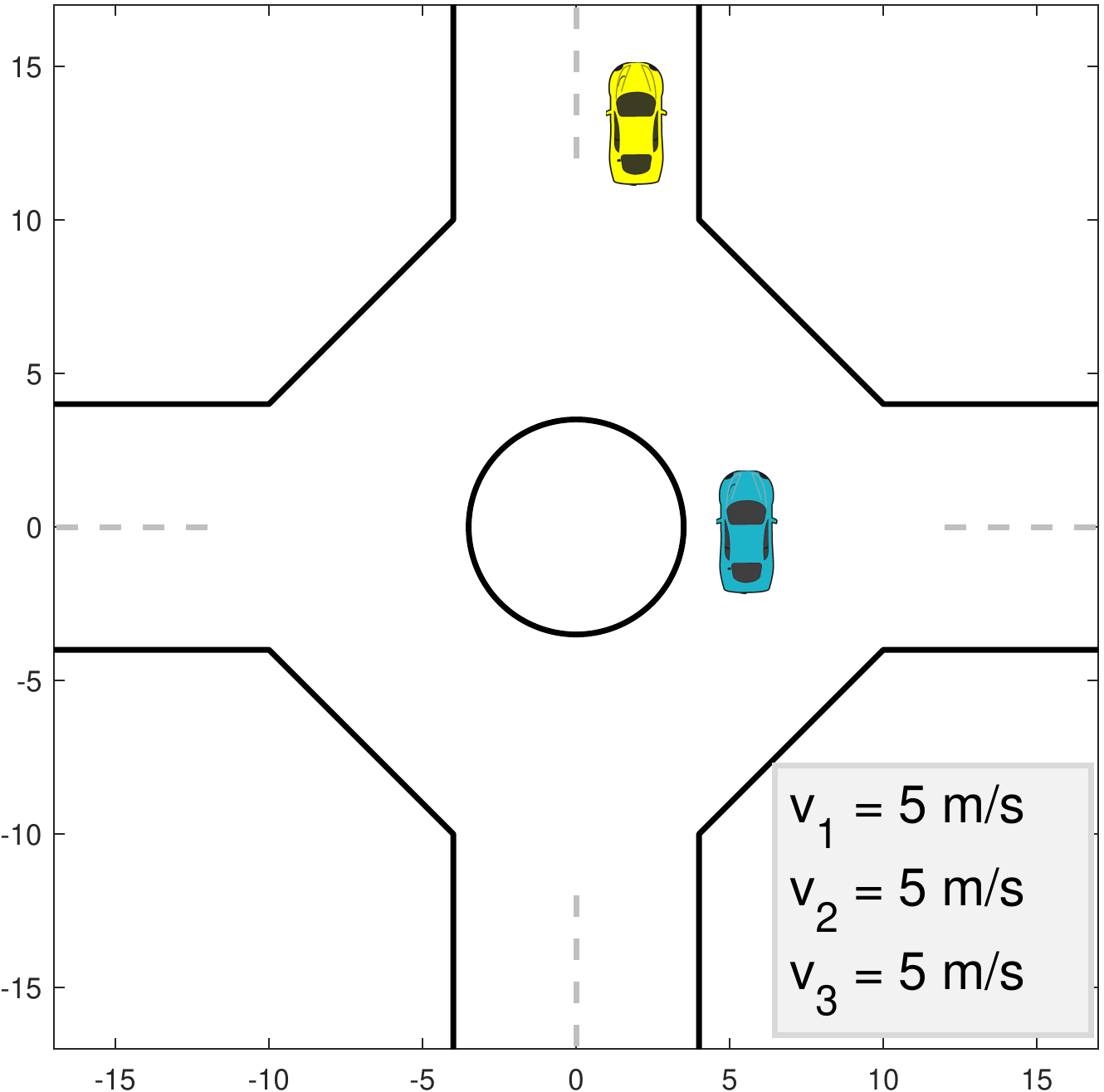, width = 0.33\linewidth, trim=0.6cm 0.4cm 0cm 0cm,clip}}
%%%%%%%%%%%%%%%%%%%%%%%%%%%%%%
\put(  -22, 55){\epsfig{file=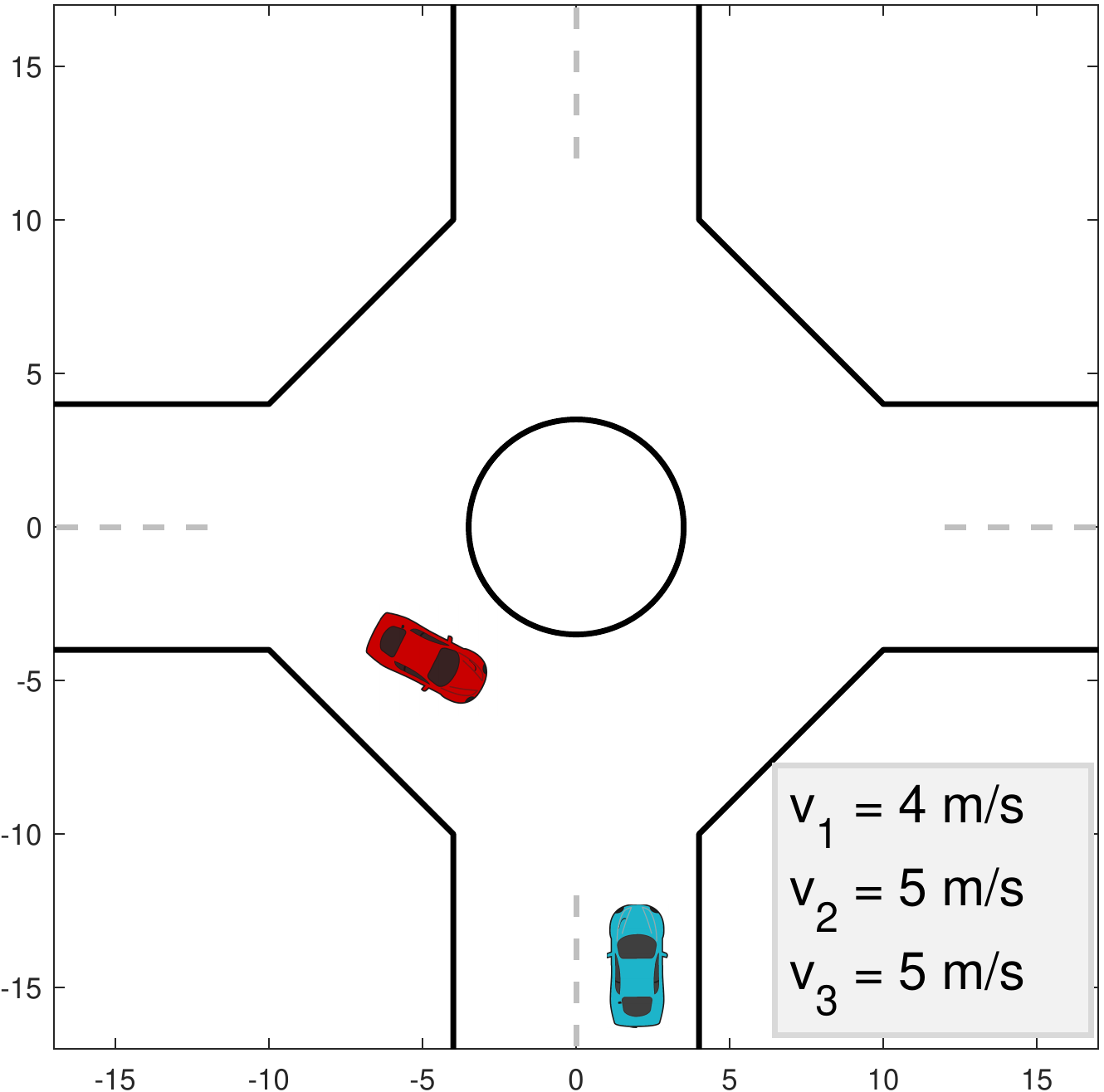,width = 0.33 \linewidth, trim=0.6cm 0.4cm 0cm 0cm,clip}}  %%%
\put(  60,  55){\epsfig{file=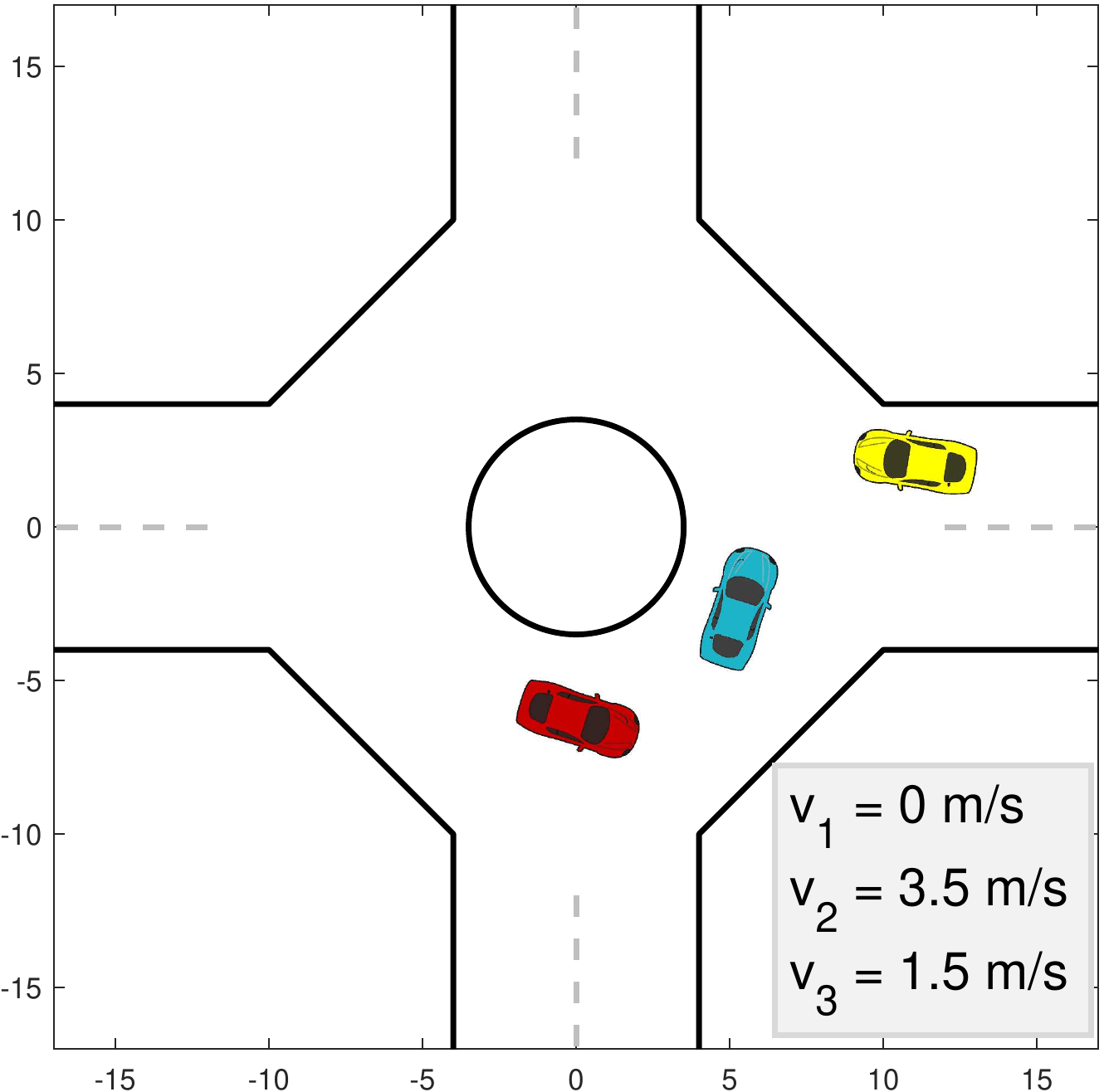, width = 0.33\linewidth, trim=0.6cm 0.4cm 0cm 0cm,clip}}

\put(  142, 55){\epsfig{file=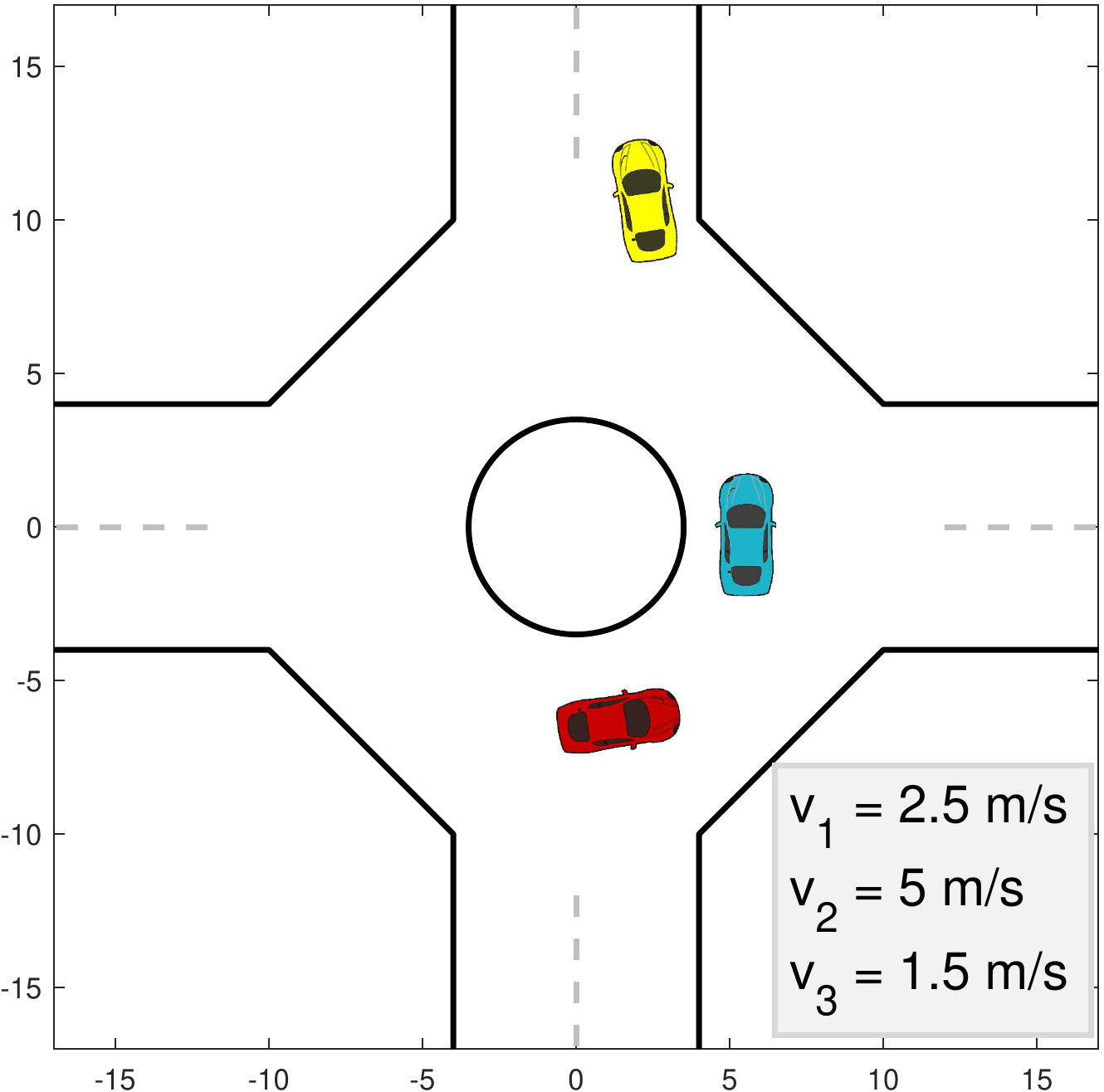, width = 0.33\linewidth, trim=0.6cm 0.4cm 0cm 0cm,clip}}

%%%%%%%%%%%%%%%%%%%%%%%%%%%%%%%%%%%%%%%%%%%%%
\put(  -22,  -5){\epsfig{file=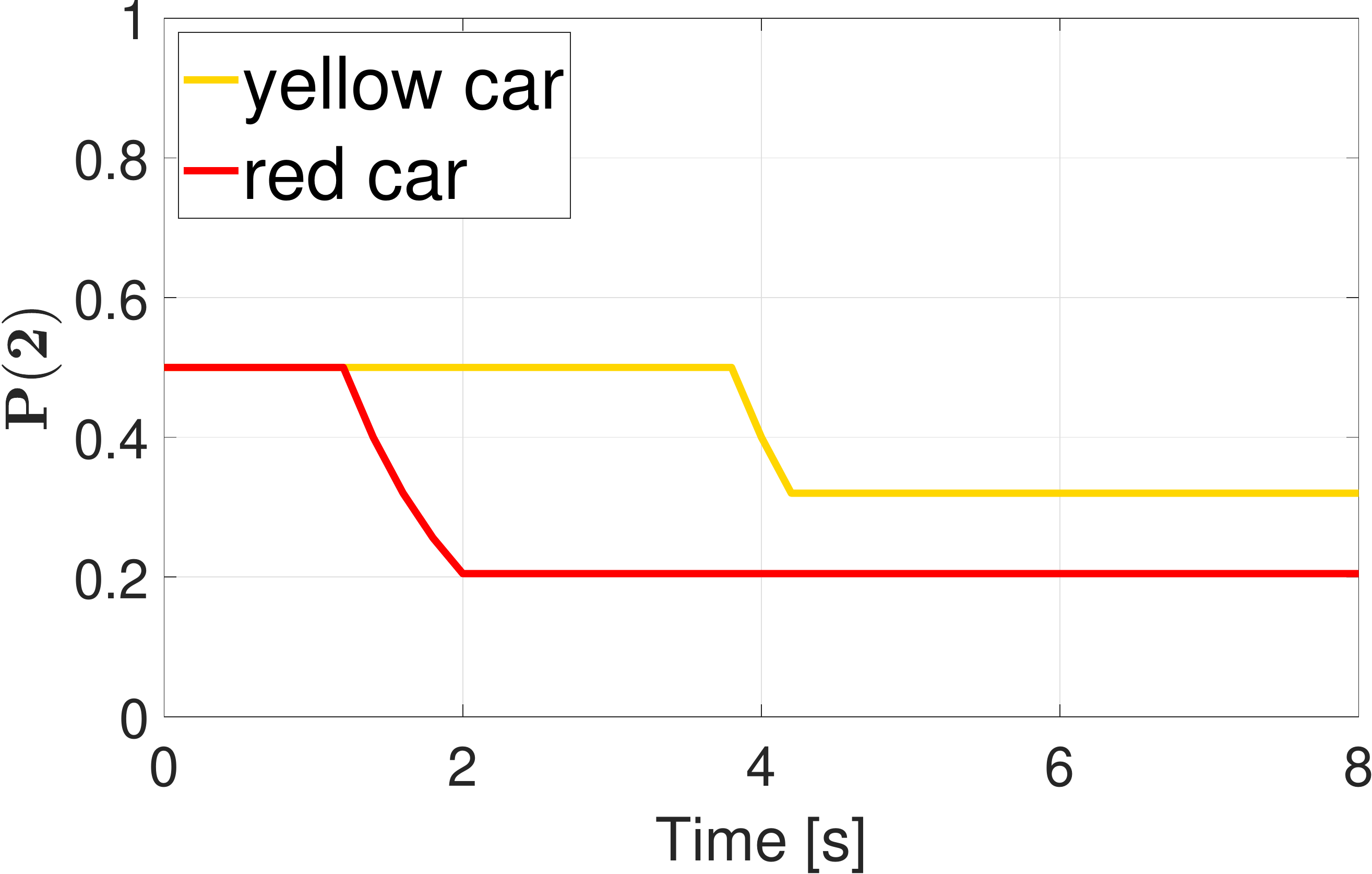,width = 0.32 \linewidth, trim=0.0cm 0.0cm 0cm 0cm,clip}}  %%%
\put(  60,  -5){\epsfig{file=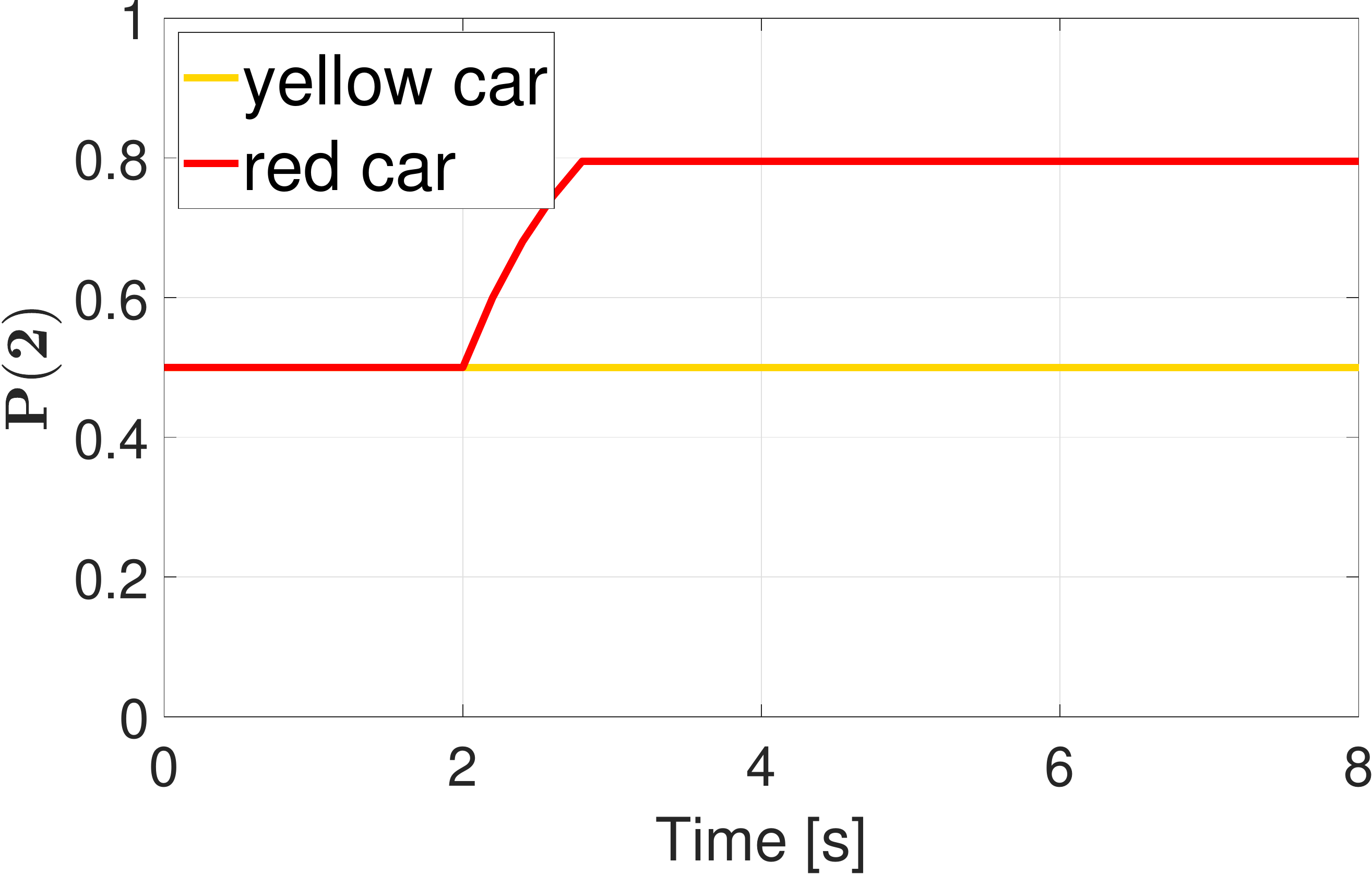, width = 0.32\linewidth, trim=0.0cm 0.0cm 0cm 0cm,clip}}
\put(  142,  -5){\epsfig{file=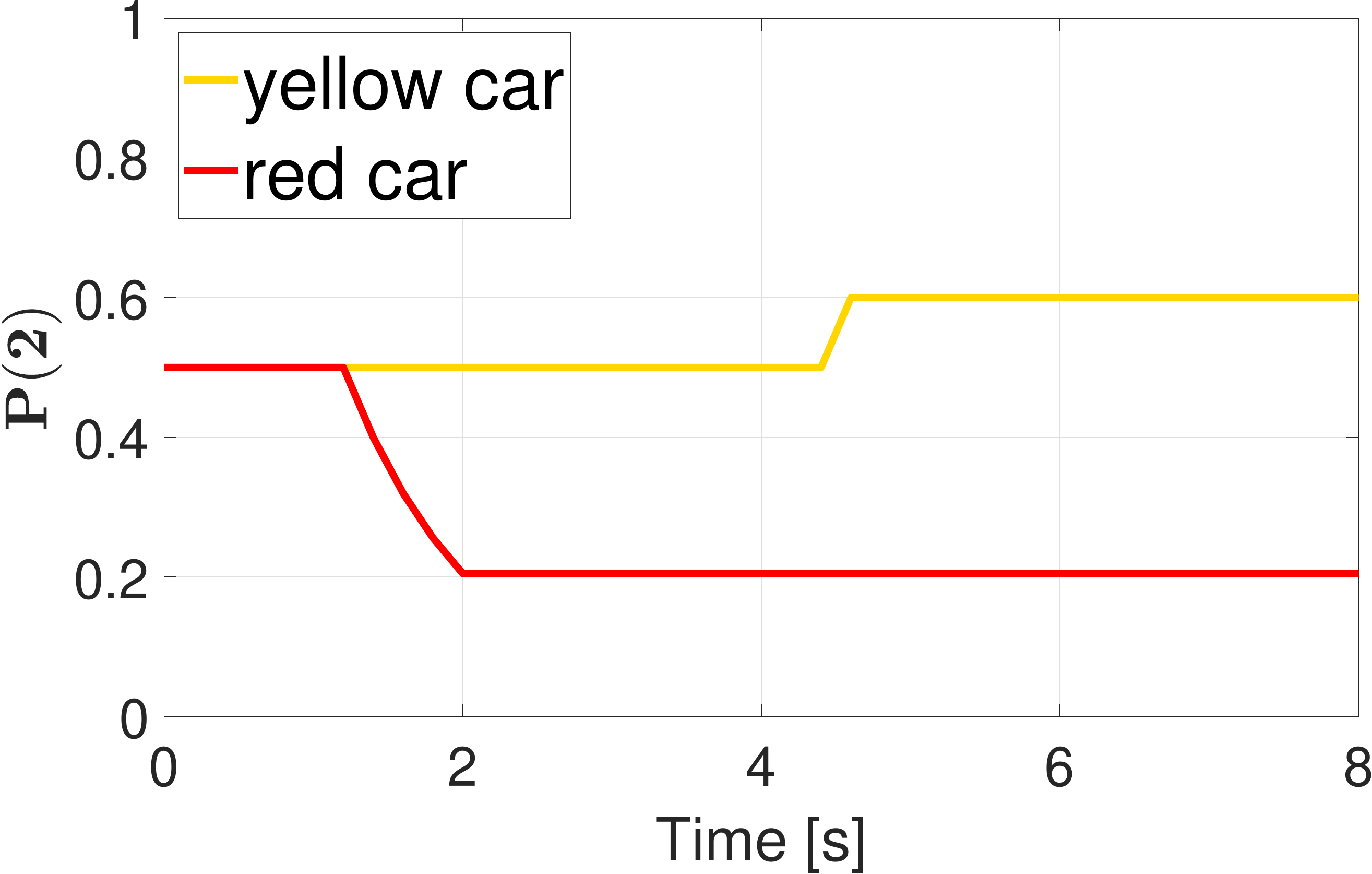, width = 0.32\linewidth, trim=0.0cm 0.0cm 0cm 0cm,clip}}
%%%%%%%%%%%%%%%%%%%%%%
\small
\put(35,295){(a-1)}
\put(117,295){(a-2)}
\put(199,295){(a-3)}
\put(35,210){(b-1)}
\put(117,210){(b-2)}
\put(199,210){(b-3)}
\put(35,125){(c-1)}
\put(117,125){(c-2)}
\put(199,125){(c-3)}
\put(35,47){(a-4)}
\put(117,47){(b-4)}
\put(199,47){(c-4)}
\normalsize
\end{picture}
\end{center}
       \caption{Interactions of the autonomous ego vehicle (blue) controlled by the adaptive control approach with level-$k$ vehicles at the roundabout intersection. (a-1)-(a-3) show three sequential steps in a simulation where the autonomous ego vehicle interacts with two level-$1$ vehicles, and (a-4) shows the time histories of the two vehicles' level estimates where $\mathbb{P}(2) = \mathbb{P}(k = 2)$ denotes the ego vehicle's belief in the level-$2$ model; (b-1)-(b-4) show those of the autonomous ego vehicle interacting with two level-$2$ vehicles; (c-1)-(c-4) show those of the autonomous ego vehicle interacting with a level-$1$ vehicle (red) and a
       level-$2$ vehicle (yellow); $v_1$, $v_2$ and $v_3$ are the speeds of the blue, yellow and red vehicles, respectively.}
      \label{fig: adaptive-roundabout}
\end{figure}

Firstly, we simulate similar scenarios as those shown in Figs.~\ref{fig: fourway-level-k}-\ref{fig: roundabout-level-k}, but let the autonomous ego vehicle (blue) be controlled by the adaptive control approach instead of level-$k$ policies. Figs.~\ref{fig: adaptive-fourway}-\ref{fig: adaptive-roundabout} show snapshots of the simulations. It can be observed that the autonomous ego vehicle can resolve the conflicts with the other two vehicles and safely drive through the intersections although the other two vehicles are controlled by varying policies. The bottom panels show the level estimation histories of the simulations. It can be observed that the autonomous ego vehicle can resolve the conflicts because it successfully identifies the level-$k$ models of the other two vehicles. Recall that vehicle $j$ is identified as level-$1$ (level-$2$) when $\mathbb{P}(k^j=2) < 0.5$ ($\mathbb{P}(k^j=2) \ge 0.5$).

The success of the adaptive control approach in situations where level-$k$ control policies with fixed $k$ fail suggests the significance in AV control of intention recognition and action prediction for the other vehicles. Note that these two steps are achieved in our adaptive control approach through the level estimates and the level-$k$ models of the other vehicles.

We then statistically evaluate and compare the two AV control approaches. For the second approach of rule-based control, we consider an acceleration set $\mathcal{A} = \{-5,-2.5,0,2.5\}$[m/s$^2$] and an initial design of the threshold value $R_c = 14$[m].

\begin{figure}[ht]
\begin{center}
\begin{picture}(200.0, 138.0)
%%%%%%%%%%%%%%%%%%%%%%%%%%%%%%
%%%%%%%%%%%%
\put(  -22,  0){\epsfig{file=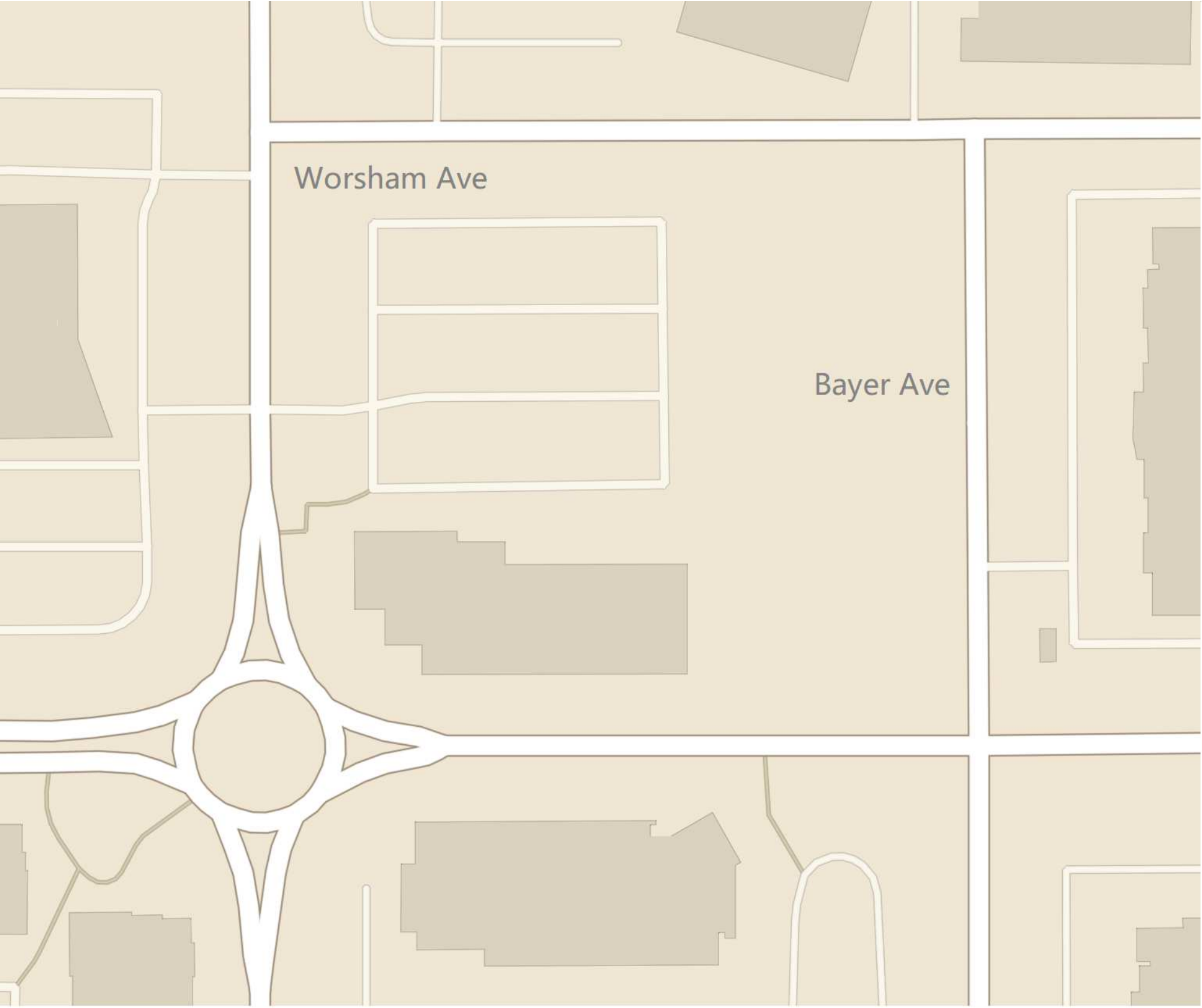,width = 0.48 \linewidth,height = 0.48 \linewidth, trim=0cm 0cm 0cm 0cm,clip}}  %%%
\put(  102,  0){\epsfig{file=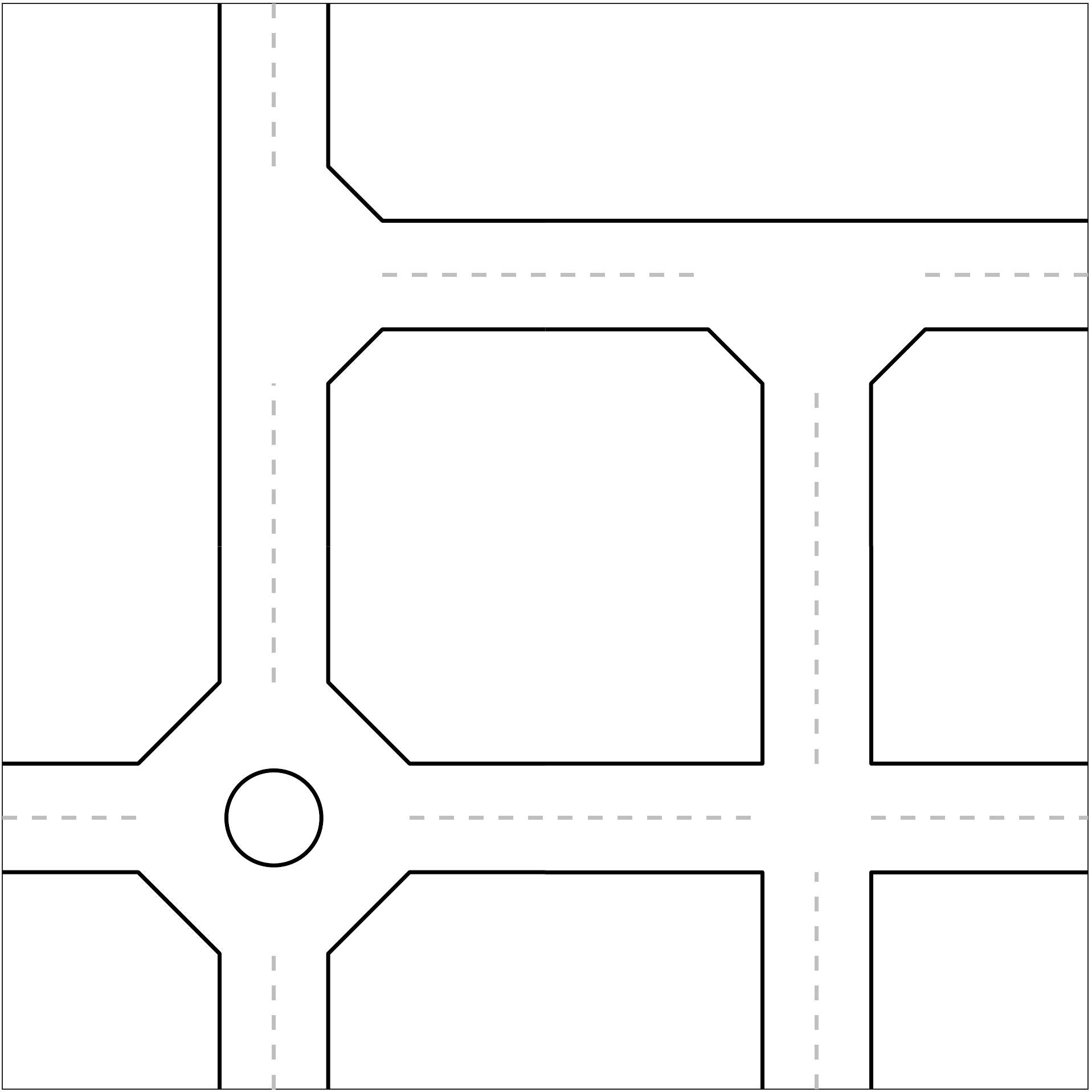,width = 0.48 \linewidth, trim=0.0cm 0.0cm 0cm 0cm,clip}}
%%%%%%%%%%%%%%%%%%%%%%
\small
\put(72,124){(a)}
\put(194,124){(b)}
\normalsize
\end{picture}
\end{center}
      \caption{Traffic scene for evaluating autonomous vehicle  control approaches. (a) shows an urban area in Los Angeles (provided by Google Maps) and (b) shows the model of the road system in (a).}
      \label{fig: traffic scene}
\end{figure}

In order to cover a rich set of scenarios, we construct a larger traffic scene shown in Fig.~\ref{fig: traffic scene}, which models the road system of an urban area in Los Angeles and consists of one four-way intersection, one roundabout, and two T-shaped intersections. We let an autonomous ego vehicle controlled by the adaptive control approach or the rule-based control approach drive through this traffic scene. Apart from the autonomous ego vehicle, we also put multiple other vehicles controlled by level-$k$ policies in the scene and let them drive through the scene repeatedly. Their initial positions, lanes entering the scene, and sequences of target lanes to traverse the scene are all randomly chosen.

We evaluate the two control approaches based on two statistical metrics: the rate of collision (CR) and the rate of deadlock (DR). The rate of collision is defined as the proportion of $2000$ simulation episodes where the autonomous ego vehicle collides with another vehicle or with the road boundaries. The rate of deadlock is defined as the proportion of $2000$ simulation episodes where no collision occurs to the autonomous ego vehicle but it fails to drive through the scene in $300$[s] of simulation time. We consider three traffic models: 1) all of the other vehicles are level-$1$, called a ``level-1 environment,'' 2) all of the other vehicles are level-$2$, called a ``level-2 environment,'' and 3) the control policy of each of the other vehicles is randomly chosen between level-$1$ and level-$2$ with equal probability, called a ``mixed environment.''

\begin{figure}[ht]
\begin{center}
\begin{picture}(200.0, 80.0)
%%%%%%%%%%%%%%%%%%%%%%%%%%%%%%
\put(  -28,  0){\epsfig{file=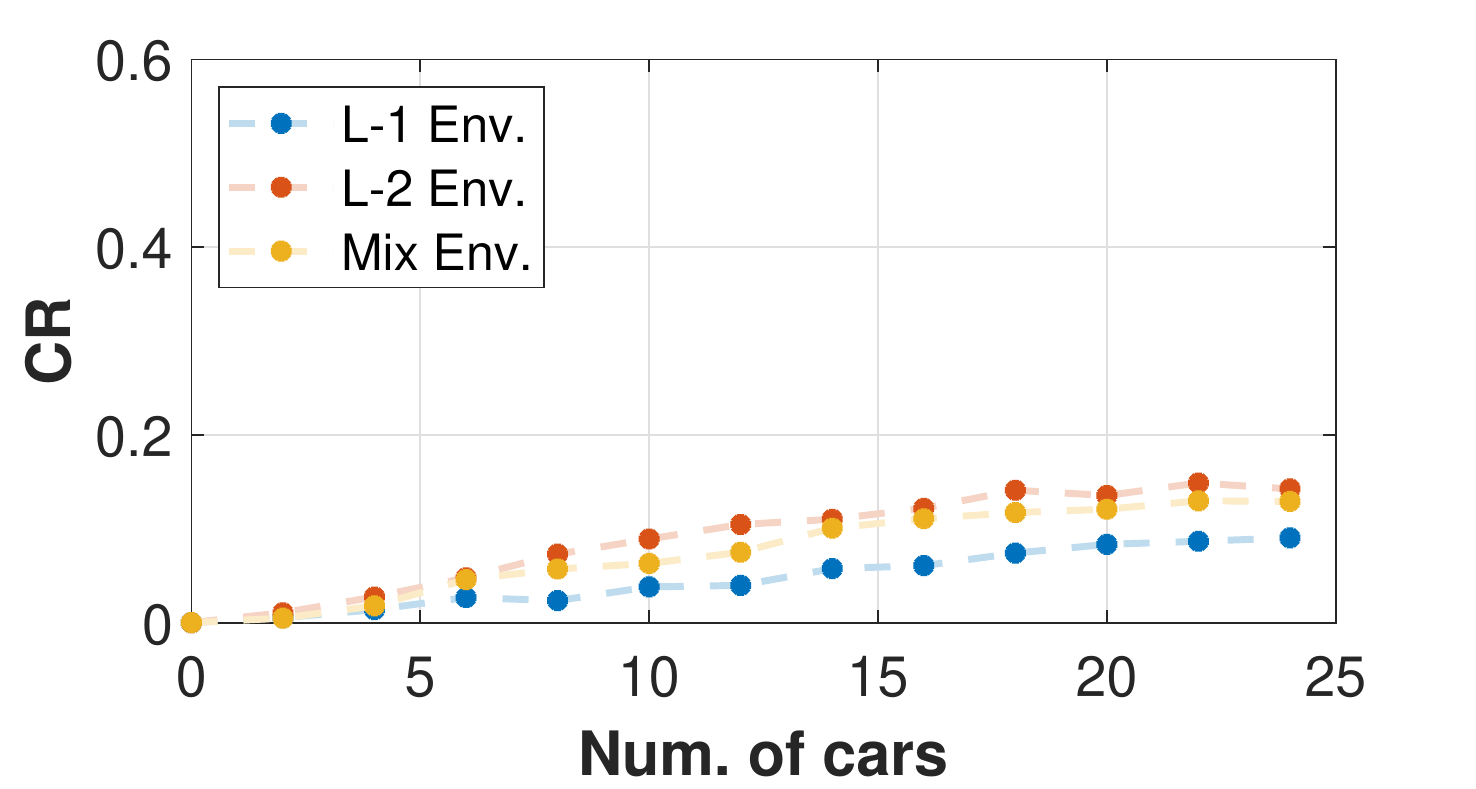,width = 0.54 \linewidth, trim=0.0cm 0.0cm 0cm 0cm,clip}}  %%%
%%%%%%%%%%%%
\put(  98,  0){\epsfig{file=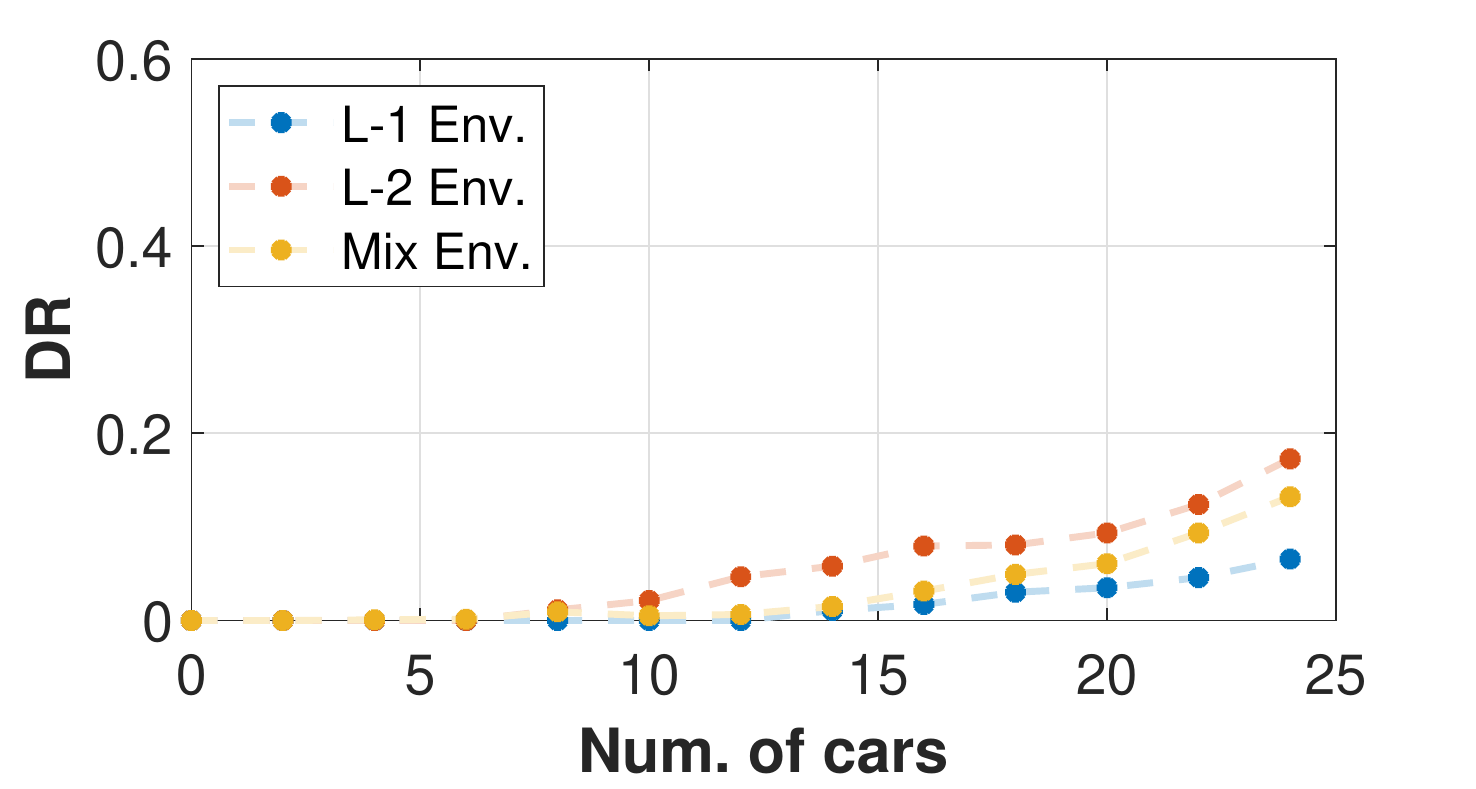,width = 0.54 \linewidth, trim=0.0cm 0.0cm 0cm 0cm,clip}}  %%%
%%%%%%%%%%%%%%%%%%%%%%
\small
\put(75,70){(a)}
\put(200,70){(b)}
\normalsize
\end{picture}
\end{center}
      \caption{Evaluation results of the adaptive control approach: (a) the rate of collision (CR) and (b) the rate of deadlock (DR) versus different numbers of environmental vehicles and different traffic models.}
      \label{fig: eval_ada}
\end{figure}

\begin{figure}[ht]
\begin{center}
\begin{picture}(200.0, 85.0)
%%%%%%%%%%%%%%%%%%%%%%%%%%%%%%
\put(  -27,  0){\epsfig{file=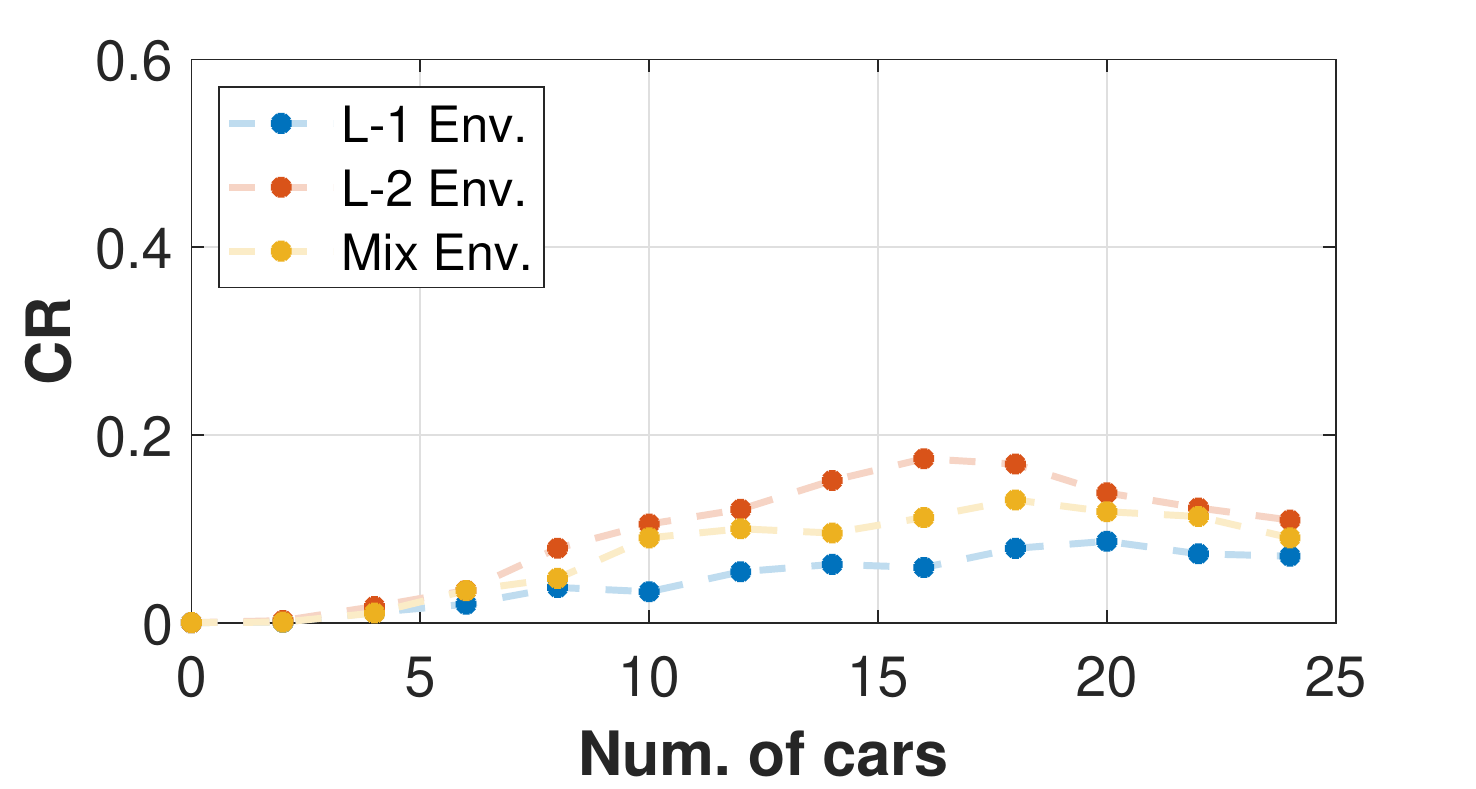,width = 0.54 \linewidth, trim=0.0cm 0.0cm 0cm 0cm,clip}}  %%%
%%%%%%%%%%%%
\put(  99,  0){\epsfig{file=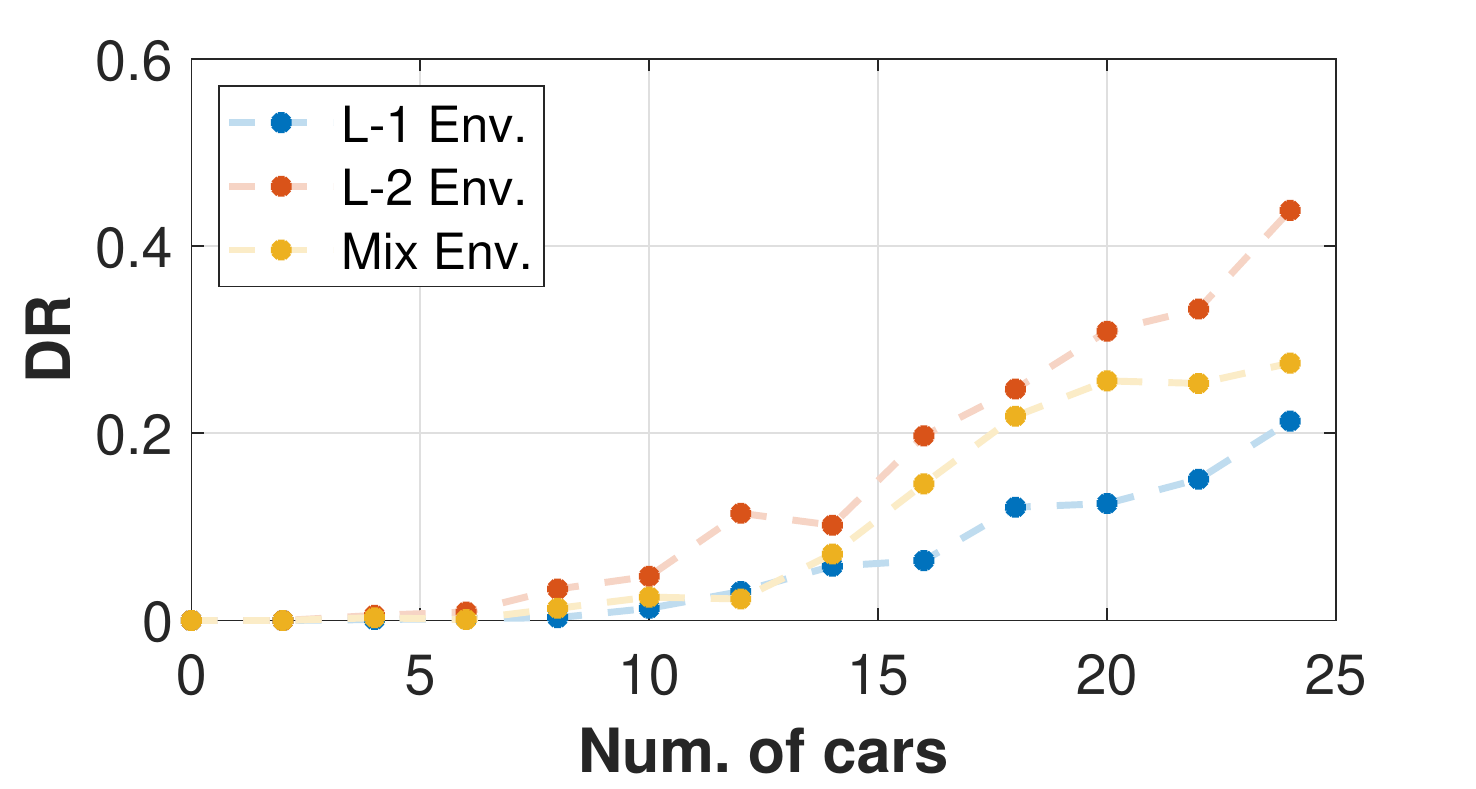,width = 0.54 \linewidth, trim=0.0cm 0.0cm 0cm 0cm,clip}}  %%%
%%%%%%%%%%%%%%%%%%%%%%
\small
\put(75,70){(a)}
\put(200,70){(b)}
\normalsize
\end{picture}
\end{center}
      \caption{Evaluation results of the rule-based control approach with $R_c = 14\,$[m]: (a) the rate of collision (CR) and (b) the rate of deadlock (DR) versus different numbers of environmental vehicles and different traffic models.}
      \label{fig: eval_rule_R2}
\end{figure}

The CR and DR results of the adaptive control approach and the rule-based control approach for different numbers of other vehicles in the scene are shown in Figs.~\ref{fig: eval_ada} and~\ref{fig: eval_rule_R2}. The number of other vehicles, $n_v$, represents traffic density, roughly, $2.87 n_v\,$[vehicles/mile] (the total length of the roads is about $560\,$[m]).

From Fig.~\ref{fig: eval_ada} it can be observed that, for the adaptive control approach, the CR and DR increase as the traffic density increases, which is reasonable. In particular, the increase in CR slows down as the number of other vehicles goes beyond $20$. Among the results for different traffic models, the CR and DR for the level-1 environment are the lowest and those for the level-2 environment are the highest. This is also reasonable since the level-1 environment, composed of level-$1$ vehicles, represents a cautious/conservative traffic model, the level-2 environment represents an aggressive traffic model and is thus most challenging for the autonomous ego vehicle, while the mixed environment lies in between. Furthermore, the results for the adaptive control approach are less sensitive to changes in traffic models than those for level-$k$ policies with fixed $k$ shown in Fig.~\ref{fig: level-k-eval}. This shows again the significance of adaptation of AV control strategy to other vehicles' intentions and actions. Note that the rate of success for a single intersection of the adaptive control approach, if estimated as $1-\frac{CR+DR}{4}$, is close to that of ``L-$1$ car in L-$2$ Env.'' and that of ``L-$2$ car in L-$1$ Env.,'' which represent the best performance of level-$k$ policies.

For the rule-based control approach, it can be observed from Fig.~\ref{fig: eval_rule_R2} that as the traffic density increases, the CR first increases and then decreases, while the DR keeps increasing. The decrease in CR when the traffic becomes very dense is due to the constant yielding of the autonomous ego vehicle to other vehicles, which causes the dramatic increase in DR.

Comparing the results of the two approaches, the adaptive control approach performs better than the rule-based control approach in the above experiments. This is attributed to the more sophisticated and complicated algorithm behind the adaptive control approach. However, the rule-based control is more interpretable (e.g., the reason for the decrease in CR is easily understood), and is easier to calibrate.

\begin{figure}[ht]
\begin{center}
\begin{picture}(200.0, 120.0)
%%%%%%%%%%%%%%%%%%%%%%%%%%%%%%
%%%%%%%%%%%%
\put(  -15,  0){\epsfig{file=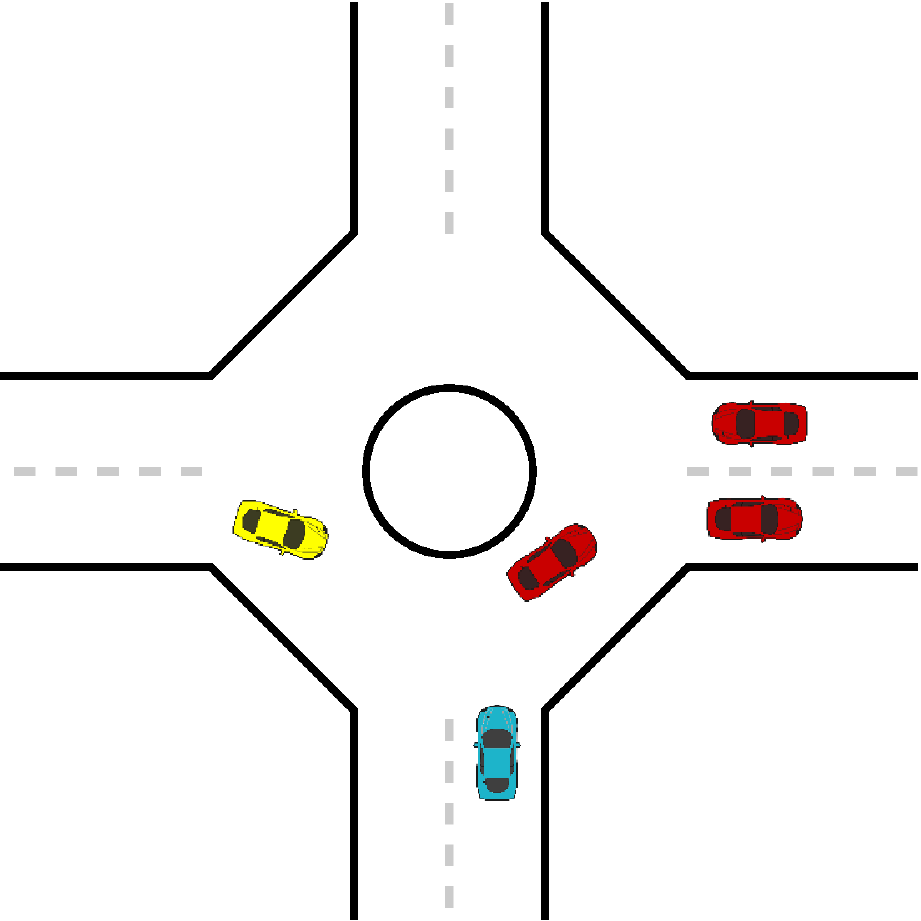,width = 0.46 \linewidth, trim=0cm 0cm 0cm 0cm,clip}}  %%%
\put(  105,  0.5){\epsfig{file=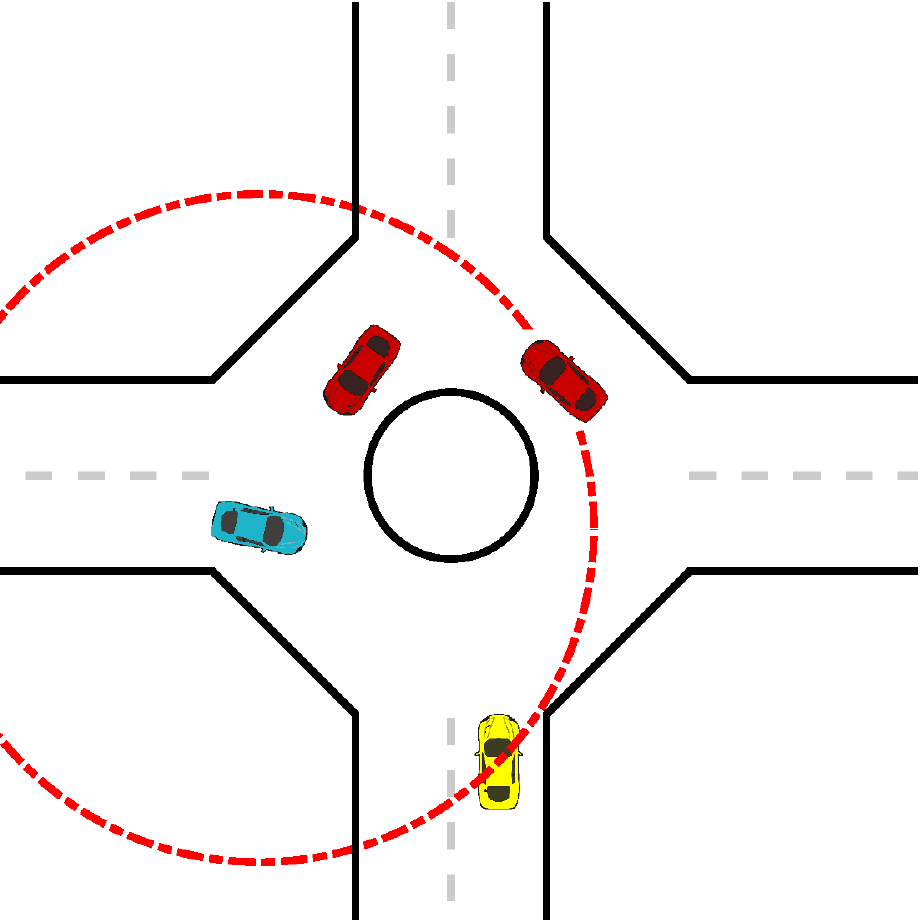,width = 0.46 \linewidth, trim=0.0cm 0.0cm 0cm 0cm,clip}}
%%%%%%%%%%%%%%%%%%%%%%
\small
\put(70,107.5){(a)}
\put(190,107.5){(b)}
\normalsize
\end{picture}
\end{center}
      \caption{Failure cases. (a) shows a scenario where the autonomous ego vehicle (blue) controlled by the adaptive control approach gets stuck at the entrance of the roundabout due to the level-$1$ vehicle (yellow) on its left. (b) shows a scenario where the autonomous ego vehicle (blue) controlled by the rule-based control approach gets hit by the level-$2$ vehicle (red) on its left.}
      \label{fig: failure case}
\end{figure}

In Fig.~\ref{fig: failure case}, we show two informative cases observed in our simulations. In the first case in Fig.~\ref{fig: failure case}(a), the autonomous ego vehicle (blue) controlled by the adaptive control approach and the level-$1$ vehicle (yellow) on its left both yield to the other and cause a deadlock. Note that a level-$1$ vehicle represents a vehicle with a cautious/conservative driver, and accordingly, yields to the autonomous ego vehicle. Although the autonomous ego vehicle eventually decides to proceed ahead and successfully drives through the roundabout, it takes too long for such a conflict to be resolved, and thus this scenario falls into our DR category. To avoid such deadlock scenarios, the autonomous ego vehicle may need to identify the driving style of the opponent vehicle faster, which may be achieved through a larger update step size $\beta$. In the second case in Fig.~\ref{fig: failure case}(b), the autonomous ego vehicle controlled by the rule-based control approach stops in the roundabout to yield to the yellow vehicle on its right and within the critical distance $R_c$ (marked by the red dashed circle). However, because the gap between the autonomous ego vehicle and the yellow vehicle is still quite large, the red vehicle on the left of the autonomous ego vehicle expects it to proceed and thus does not slow down, which causes a collision. This scenario shows that a larger critical distance $R_c$ may not always correspond to a safer driving behavior. \di{We remark that failure/corner cases identified by our simulations, such as the above two cases, can also inform the design of specific test trajectories for AV control systems.}

We now optimize the threshold value $R_c$ in the rule-based control approach to achieve better performance defined by a performance index as follows:
\begin{equation}\label{equ:Index}
    J = \frac{1}{n_{\max}}\sum_{n = 1}^{n_{\max}} \Big(w_c\phi_c(\mathcal{S}_n) + w_d\phi_d(\mathcal{S}_n) + w_v\frac{\phi_s(\mathcal{S}_n)}{\bar{v}(\mathcal{S}_n)+\epsilon}\Big),
\end{equation}
where $\mathcal{S}_n$ denotes the $n$th simulation episode; the $\phi_c(\mathcal{S}_n)$, $\phi_d(\mathcal{S}_n)$, and $\phi_s(\mathcal{S}_n)$ are indicator functions, taking $1$ if, respectively, a collision occurs to the autonomous ego vehicle, no collision but a deadlock occurs to the autonomous ego vehicle, and neither collision nor deadlock occur and the autonomous ego vehicle successfully drives through the scene in $300$[s] of simulation time in the $n$th simulation episode, and taking $0$ otherwise; $\bar{v}(\mathcal{S}_n)$ is the average speed of the autonomous ego vehicle in the $n$th simulation episode; $w_c, w_d, w_v \ge 0$ are weighting factors, and $\epsilon > 0$ is a constant to adjust the shape of the function with respect to the average speed $\bar{v}(\mathcal{S}_n)$ and to avoid the denominator being $0$.

The performance index function \eqref{equ:Index} imposes penalties for collisions and deadlocks through the first two terms, and rewards higher average speeds through the last term. Note that the last term is designed in such a way that the penalty increases fast for decrease in speed values that are already very low, and decreases slowly for increase in speed values that are already very high. In obtaining the following results, we run simulations in the same scene shown in Fig.~\ref{fig: traffic scene} with $15$ other vehicles, and we use $w_c = 10$, $w_d = 5$, $w_v = 1$, and $\epsilon = 0.1$.

\begin{figure}[ht]
\centering
\includegraphics[width=0.35\textwidth]{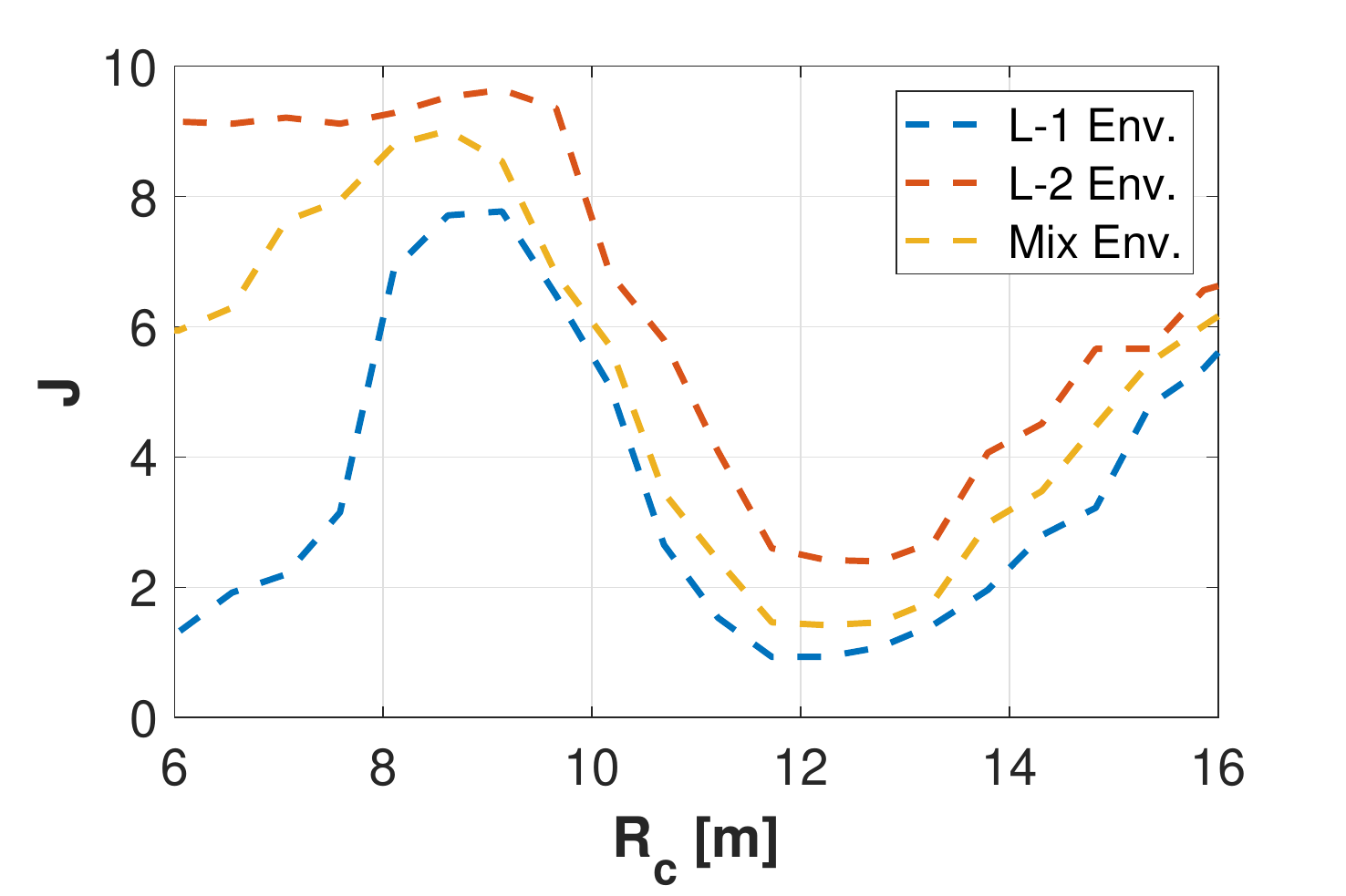}
\centering
\caption{Performance index $J$ as function of $R_c$ of the rule-based control approach with different traffic models.}
\label{fig: calibration}
\end{figure}

We plot the values of \eqref{equ:Index} for different values of $R_c$ in Fig.~\ref{fig: calibration}. Specifically, for each value of $R_c$, we run $n_{\max} = 2000$ simulation episodes and calculate the value of \eqref{equ:Index} based on the simulation results. Lower values of \eqref{equ:Index} represent better performance in terms of having less collisions, less deadlocks, and higher average travel speeds.

In Fig.~\ref{fig: calibration}, the blue curve represents the result when the autonomous ego vehicle operates in the level-$1$ environment. It can be observed that the performance is good when $R_c$ takes very small values, i.e., in the range of $[6,7.5]$[m]. This is because small $R_c$ corresponds to aggressive behavior and the level-$1$ environment represents a conservative traffic model, thus, the other vehicles almost always yield to the autonomous ego vehicle when there is a conflict. Since the autonomous ego vehicle proceeds ahead while the other vehicles yield, collisions and deadlocks are avoided. However, when operating in the level-$2$ or mixed environment, small $R_c$ leads to poor performance. This is because both the autonomous ego vehicle and the other vehicles behave aggressively and cause many collisions. When $R_c$ takes values in the range of $[7.5,11]$[m], the performance is the worst for all of the three traffic models. This is because such $R_c$ values correspond to behaviors between aggressive and conservative, which can cause collisions with both aggressive and conservative interacting vehicles. The range $[11.5,13]$[m] is suitable for choosing the value of $R_c$, where the performance is good and insensitive to changes in the traffic models. For larger $R_c$ values, the autonomous ego vehicle becomes overly conservative and almost always yields to the other vehicles, which causes it difficulties to enter the intersections and leads to many deadlocks.

\section{Conclusion}
\label{sec: conclusion}

In this paper, we described a framework based on level-k game theory for modeling traffic consisting of heterogeneous (in terms of their driving styles) and interactive vehicles in urban environments with unsignalized intersections. An algorithm integrating the level-k decision-making formalism, receding-horizon optimization, and imitation learning was proposed and used to solve for level-$k$ control policies.

The developed traffic models are useful as simulation environments for verification and validation of autonomous vehicle control systems. In particular, we considered two autonomous vehicle control approaches as case studies: an adaptive control approach based on level-$k$ vehicle models and a rule-based control approach. We analyzed their characteristics and evaluated their performance based on their testing results with our traffic models, and then optimized the parameters of the rule-based approach based on a performance index.

We envision that traffic models developed using the framework proposed in this paper can also be integrated with urban traffic/driving simulators with higher-fidelity car dynamics and environmental representations, such as CARLA \cite{Dosovitskiy17}, using an approach similar to that of \cite{su2019traffic}, to create more realistic urban traffic simulations and support autonomous driving system development.

\bibliographystyle{IEEEtran}

%%\bibliography{ref} C:\Users\hp\Documents

\bibliography{Ref}
\begin{IEEEbiography}[{\includegraphics[width=1.05in,height=1.05in]{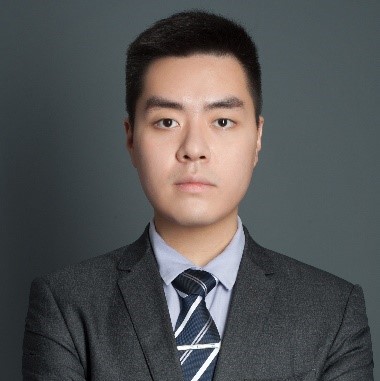}}]
{\textbf{Ran Tian}} received his B.S. degree in aerospace engineering from the University of Michigan, Ann Arbor, MI, USA, in 2016, and his B.S. degree in mechanical engineering from the Shanghai Jiao Tong University, Shanghai, China, in 2017. He received his M.S. degree in Robotics from the University of Michigan, Ann Arbor, MI, USA, in 2019. His current research interests include decision-making under uncertainty and human-robot interaction.
\end{IEEEbiography}

\begin{IEEEbiography}[{\includegraphics[width=1in,height=1.25in]{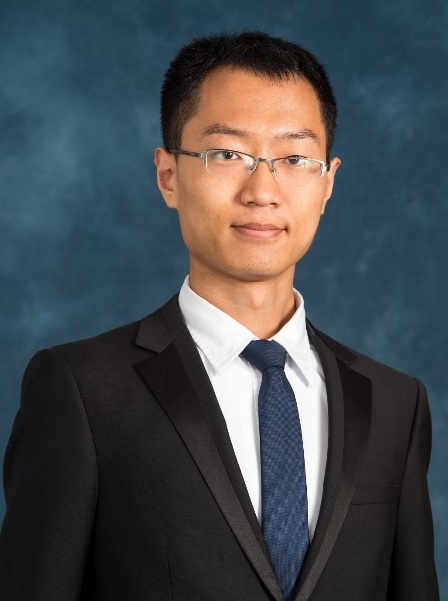}}]
{\textbf{Nan Li}} received the B.S. degree in automotive engineering from Tongji University, Shanghai, China, in 2014, and the M.S. degrees in mechanical engineering and in mathematics from the University of Michigan, Ann Arbor, MI, USA, in 2016 and in 2020, respectively, where he is currently pursuing the Ph.D. degree in aerospace
engineering. His research interests are stochastic control and multi-agent systems.
\end{IEEEbiography}

\begin{IEEEbiography}[{\includegraphics[width=1.05in,height=1.05in,keepaspectratio]{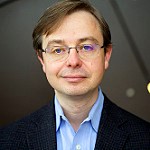}}]
{\textbf{Ilya Kolmanovsky}} is a professor in the department of aerospace engineering at the University of Michigan, with research interests in control theory for systems with state and control constraints,
and in control applications to aerospace and automotive systems. He received his Ph.D. degree
from the University of Michigan in 1995.
\end{IEEEbiography}

\begin{IEEEbiography}[{\includegraphics[width=1in,height=1.25in,keepaspectratio]{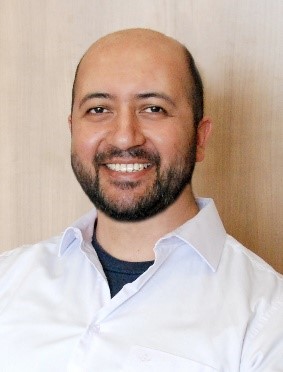}}]
{\textbf{Yildiray Yildiz}} is an assistant professor at Bilkent University, Ankara. He received his B.S. degree (valedictorian) in mechanical engineering from Middle East Technical
University, Ankara in 2002; M.S. degree in mechatronics from Sabanci University,
Istanbul, in 2004; and Ph.D. degree in mechanical engineering with a mathematics
minor from MIT in 2009. He held postdoctoral associate and associate scientist
positions with NASA Ames Research Center, California, employed by the University of
California, Santa Cruz, through its University Affiliated Research Center, from 2009 to
2010 and 2010 to 2014, respectively. He is the recipient of NASA Honor Award, Young
Scientist Award from Science Academy of Turkey, Young Scientist Award from Turkish
Academy of Sciences, Research Incentive Award from Prof. Mustafa Parlar Education
and Research Foundation, and best student conference paper award from ASME. He is
an IEEE Senior Member and currently serving as an associate editor for IEEE Control
Systems Magazine and European Journal of Control. His research interests include
control, machine learning, game theory, and applications of these fields for modeling
and control of automotive and aerospace systems.
\end{IEEEbiography}

\begin{IEEEbiography}[{\includegraphics[width=1.05in,height=1.05in]{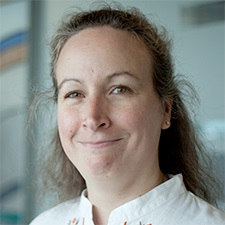}}]
{\textbf{Anouck R. Girard}} received the Ph.D. degree in ocean engineering from the University of
California, Berkeley, CA, USA, in 2002. She has been with the University of Michigan, Ann Arbor, MI, USA, since 2006, where she is currently an Associate Professor of Aerospace Engineering. She has co-authored the book Fundamentals of Aerospace Navigation and Guidance (Cambridge University Press, 2014). Her current research interests include flight dynamics and control systems. Dr. Girard was a recipient of the Silver Shaft Teaching Award from the University of Michigan and a Best Student Paper
Award from the American Society of Mechanical Engineers.
\end{IEEEbiography}

\end{document}